\documentclass[10pt,journal,c]{IEEEtran}
\usepackage{times}
\usepackage{epsfig}
\usepackage{graphicx}
\usepackage{subcaption}
\usepackage{amsmath}
\usepackage{amssymb}
\usepackage{calrsfs}
\usepackage{algorithm}
\usepackage{algorithmic}
\usepackage{microtype}
\usepackage{float}
\usepackage{alphalph}
\usepackage{multicol}
\usepackage{multirow}
\usepackage{mathrsfs}

\usepackage[top=0.75in, bottom=1.6in, left=0.516in, right=0.516in]{geometry}

\usepackage[noadjust]{cite}

\def\eg{\emph{e.g.}}
\def\ie{\emph{i.e.}}
\def\etal{\emph{et al.}}

\DeclareMathAlphabet{\pazocal}{OMS}{zplm}{m}{n}

\usepackage[pagebackref=true,breaklinks=true,colorlinks,bookmarks=false]{hyperref}

\begin{document}
%
\title{Partial Membership Latent Dirichlet Allocation}

\author{Chao~Chen,~\IEEEmembership{Member,~IEEE,}
        Alina~Zare,~\IEEEmembership{Senior Member,~IEEE,}
        Huy N.~Trinh, 
        Gbenga O.~Omotara,
        ~J. Tory ~Cobb,~\IEEEmembership{Senior Member,~IEEE,}
        and Timotius A.~Lagaunne
\IEEEcompsocitemizethanks{\IEEEcompsocthanksitem C. Chen, H. Trinh, G. O. Omotara, and T. Lagaunne are with the Department
of Electrical and Computer Engineering, University of Missouri, Columbia,
MO, 65203. 
E-mail: ccwwf@mail.missouri.edu
\IEEEcompsocthanksitem A. Zare is with the Department of Electrical and Computer Engineering, University of Florida, Gainesville, FL, 32611. 
E-mail: azare@ece.ufl.edu
\IEEEcompsocthanksitem J. Tory Cobb is with Naval Surface Warfare Center, Panama City, FL.

The authors graciously thank the Office of Naval Research, Code 321, for funding this research. Any opinions, findings, and conclusions or recommendations expressed in this material are those of the author(s) and do not necessarily reflect the views of the Office of Naval Research.

}
\thanks{Manuscript received December 28, 2016.}}

\markboth{Chen \MakeLowercase{\textit{et al.}}: Partial Membership Latent Dirichlet Allocation}{}
\IEEEtitleabstractindextext{%
\begin{abstract}
	
Topic models (e.g., pLSA, LDA, sLDA) have been widely used for segmenting imagery. However, these models are confined to crisp segmentation, forcing a visual word (\ie, an image patch) to belong to one and only one topic.  Yet, there are many images in which some regions cannot be assigned a crisp categorical label (\eg, transition regions between a foggy sky and the ground or between sand and water at a beach).  In these cases, a visual word is best represented with partial memberships across multiple topics.  To address this, we present a partial membership latent Dirichlet allocation (PM-LDA) model and an associated  parameter estimation algorithm. This model can be useful for imagery where a visual word  may be a mixture of multiple topics. Experimental results on visual and sonar imagery show that PM-LDA can produce both crisp and soft semantic image segmentations; a capability previous topic modeling methods do not have. 
	
\end{abstract}

}

\maketitle

\IEEEdisplaynontitleabstractindextext

%
\IEEEpeerreviewmaketitle

\section{Introduction}\label{sec:introduction}

\begin{figure}[!htb]
	\centering
	\begin{subfigure}[b]{0.15\textwidth}
		\includegraphics[width=1\linewidth,height=0.8\linewidth]{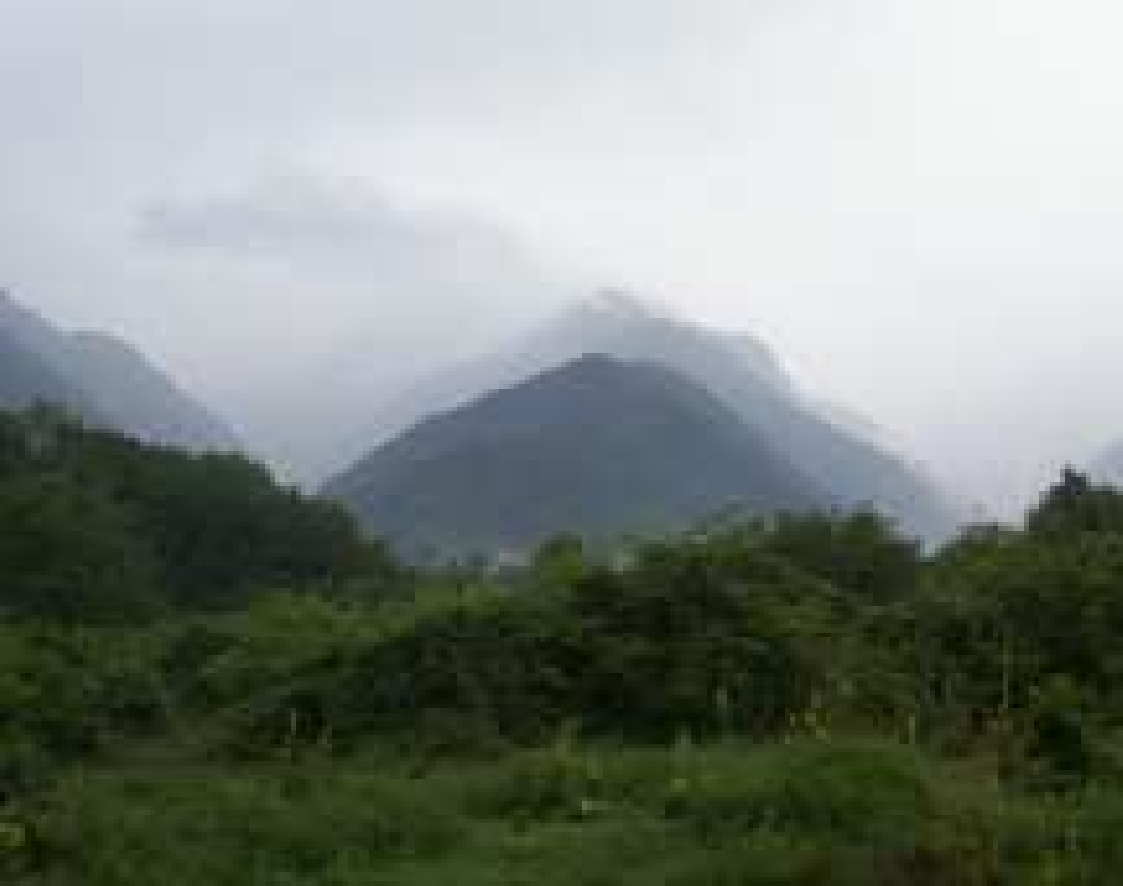}
		\captionsetup{labelformat=empty,,skip=0pt}
		\caption{(a)}
		\label{fig:afog}
	\end{subfigure}
	\begin{subfigure}[b]{0.15\textwidth}
		\includegraphics[width=1\linewidth,height=0.8\linewidth]{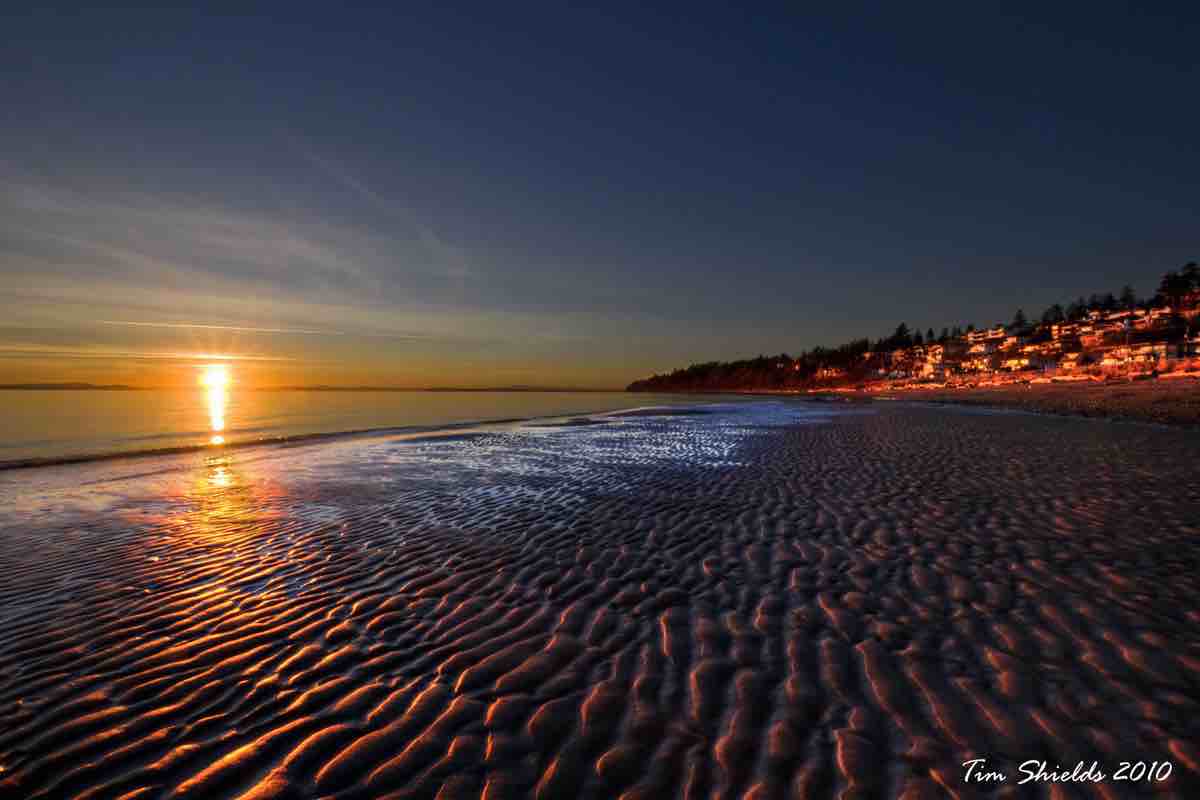}
		\captionsetup{labelformat=empty,skip=0pt}
		\caption{(b)}
		\label{fig:bsun}
	\end{subfigure}
	\begin{subfigure}[b]{0.15\textwidth}
		\includegraphics[width=1\linewidth,height=0.8\linewidth]{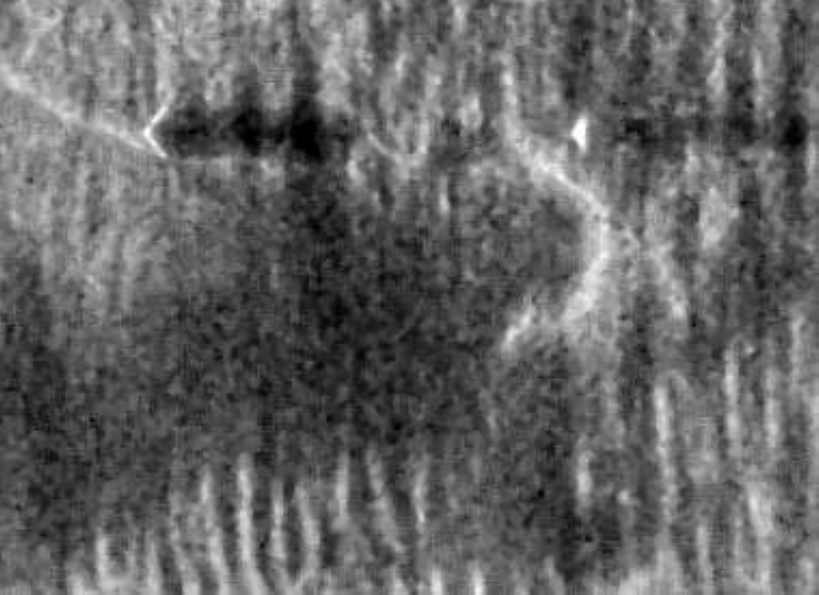}
		\captionsetup{labelformat=empty,skip=0pt}
		\caption{(c)}
		\label{fig:csonar}
	\end{subfigure}
	\caption{Imagery with regions of gradual transition. (a) Image with gradual transition from fog to mountain. 
		(b) Sunset image with gradual transition from sun to sky. 
		(c) SONAR image with gradually vanishing sand ripples. }
	\label{fig:sunset0}
\end{figure}

%
%
%
%
\IEEEPARstart{T}{he} goal of unsupervised semantic image segmentation is to divide an image into semantically distinct coherent regions, \ie, regions corresponding to objects or parts of objects. Inspired by the success of Latent Dirichlet Allocation (LDA) \cite{blei:2003} in discovering semantically meaningful topics from document collections, many have successfully applied LDA or its variants to image segmentation \cite{russell:2006, cao:2007, wang:2008, zhao:2010,andreetto:2012}.  It has been used in a wide range of computer vision applications, such as object recognition and tracking and image retrieval \cite{zhao:2010,shi:2000,comaniciu:2002, felzenszwalb:2004}. Yet, in many images, the widely used crisp image segmentation methods fail to perform.  Specifically, these methods poorly address imagery in which there are smooth gradients and transition regions.    For example, consider the photograph in Fig. \ref{fig:afog} where the gradually thinning fog blurs the boundary between the foggy sky and the mountain, a sharp
 boundary between the ``fog'' and ``mountain'' topics does not exist. Similarly,  notice the lack of a sharp boundary given the gradually fading sunlight in Fig. \ref{fig:bsun} or the gradually vanishing sand ripples shown in the Synthetic Aperture SONAR (SAS) image of the sea floor in Fig. \ref{fig:csonar}.  In this paper, we present Partial Membership Latent Dirichlet Allocation (PM-LDA) to address unsupervised semantic image segmentation in the case of imagery with regions of transition.

Unsupervised semantic image segmentation methods differ from traditional (\ie, non-hierarchical/flat) segmentation methods  (\eg, normalized cuts algorithm \cite{shi:2000}) by estimating and describing additional inter-segment relationships. Namely, these methods over-segment imagery and then group the resulting ``visual words'' into topic clusters such that small segments from the same object class can be combined into a complete object and provide a comprehensive organization of the larger scene.   In other words, unsupervised semantic image segmentation methods cluster imagery hierarchically in which the lower level corresponds to an over-segmentation of the imagery and the higher level groups the over-segmented pieces into topic clusters.  In addition to the direct application of LDA to visual words, methods that consider spatial structure have also been developed.  These include the Spatial Latent Topic model (Spatial-LTM) and Spatial LDA (sLDA) model \cite{cao:2007, wang:2008}.  Zhao, \etal \cite{zhao:2010} developed a Topic Random Field (TRF) model to tackle the problems caused by discarding spatial information and information loss by feature quantization. Andreetto, \etal \cite{andreetto:2012} proposed an Affinity-based Latent Dirichlet Allocation (A-LDA). Yet, under these existing topic models, a visual word is only assigned to one topic (\ie, the word-topic assignment is a binary indicator).  In this paper, we generalize LDA to allow for partial memberships.

\begin{figure*}[htb!]
	\vspace{2mm}
	\begin{subfigure}{0.3\textwidth}
		\centering
		\includegraphics[width=0.8\linewidth]{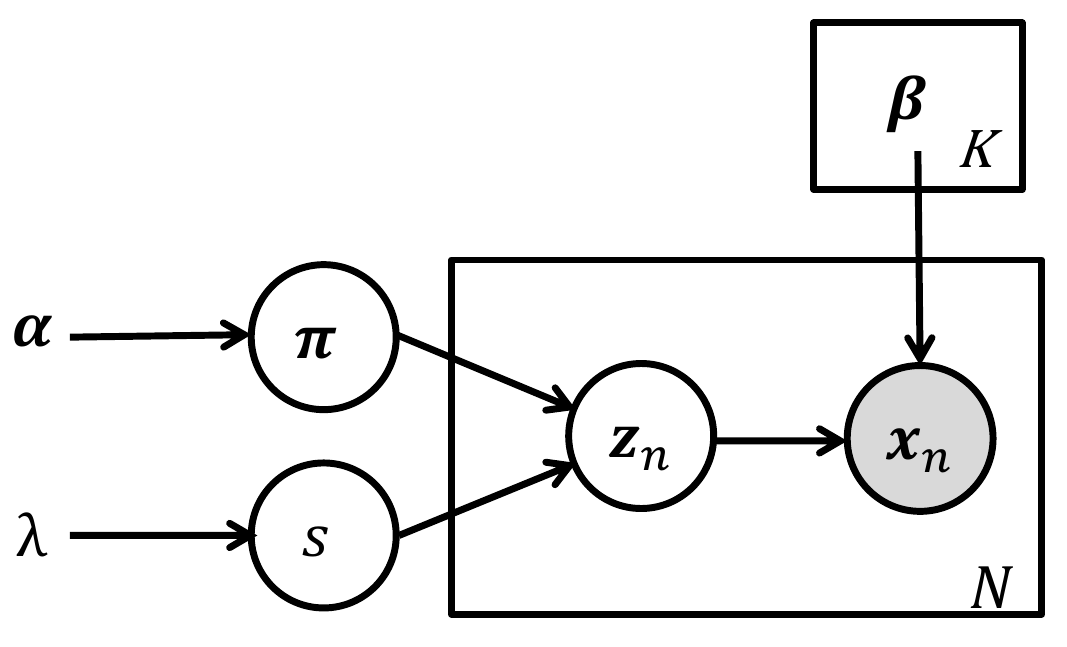}
		\captionsetup{skip=0pt}
		\caption{BPM}
		\label{fig:bmp}
	\end{subfigure}
	\begin{subfigure}{0.3\textwidth}
		\centering
		\includegraphics[width=0.8\linewidth]{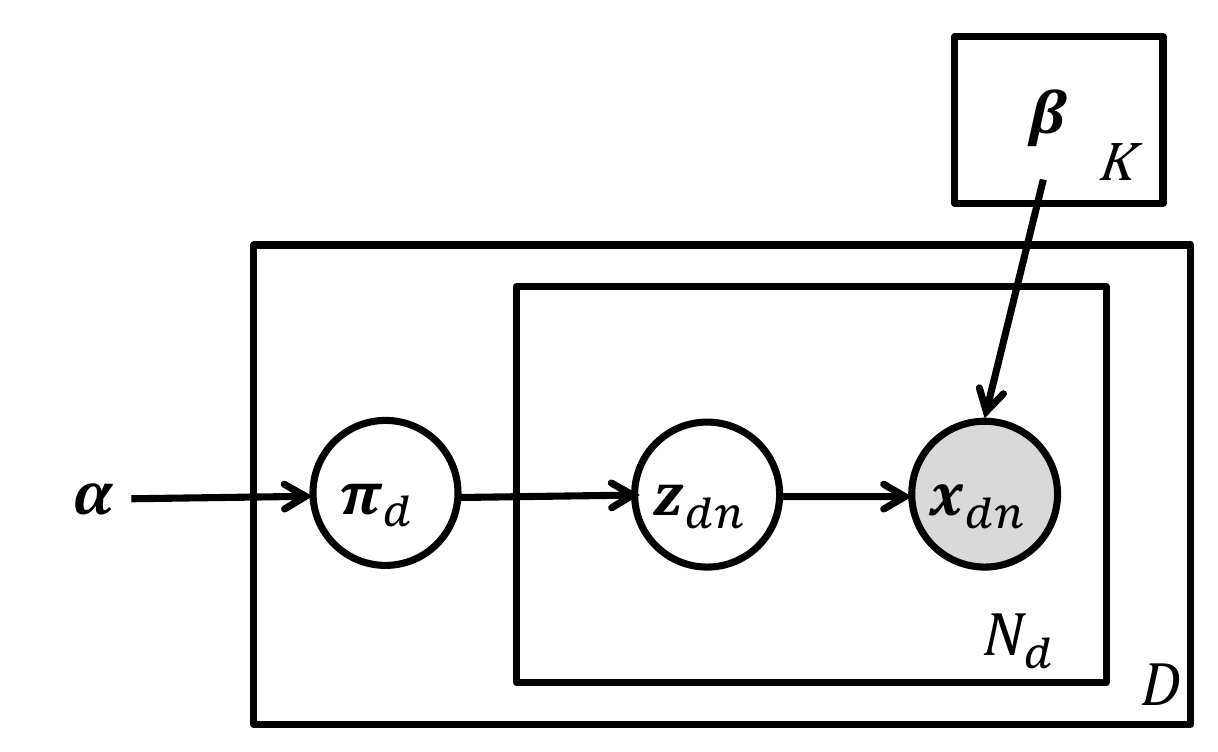}
		\captionsetup{skip=0pt}
		\caption{LDA}
		\label{fig:lda}
	\end{subfigure}
	\begin{subfigure}{0.3\textwidth}
		\centering
		\includegraphics[width=0.8\linewidth]{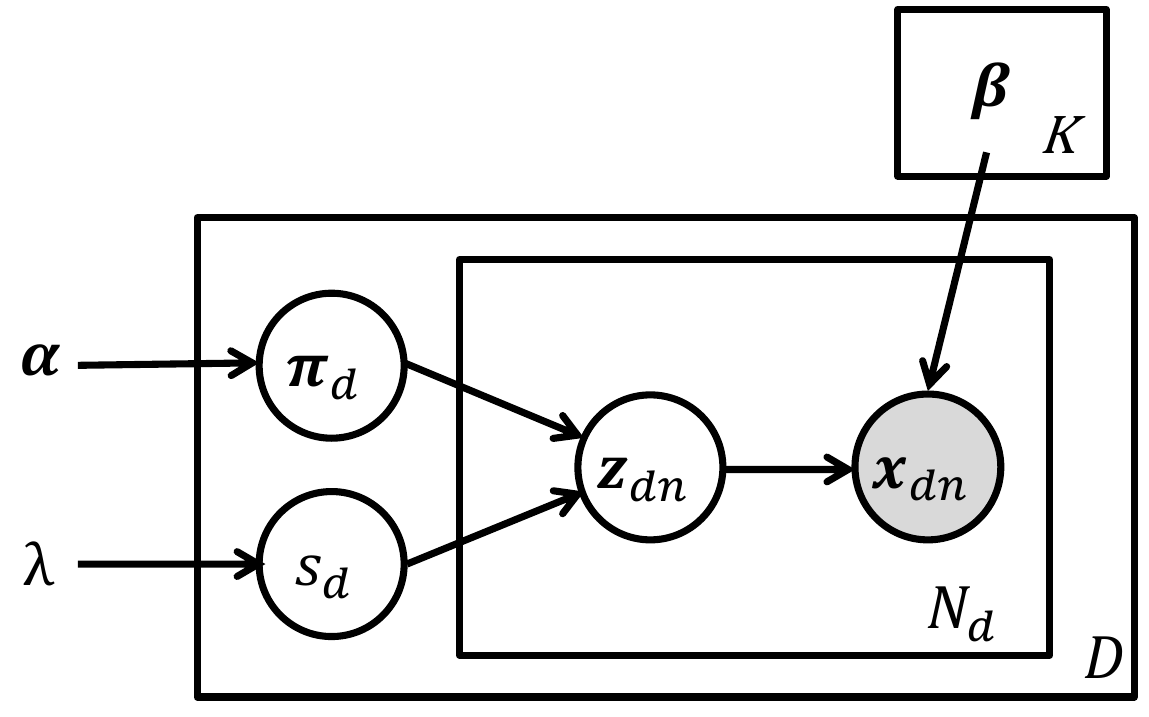}
		\captionsetup{skip=0pt}
		\caption{PM-LDA}
		\label{fig:pmlda}
	\end{subfigure}
	\caption{(a) Graphical model for BPM. (b) Graphical model for LDA. (c) Graphical model for PM-LDA}
\end{figure*}

\section{Partial Membership Models}

Partial membership models and algorithms have been previously developed in the literature.  One prevalent partial membership approach is the Fuzzy C-means algorithm (FCM) \cite{bezdek:1984} and, in particular, FCM extensions for image segmentation \cite{naz:2010,Krinidis:2010,Krinidis:2012}. 
FCM is a centroid-based clustering method which iteratively minimizes the following objective function,
\begin{equation}
J=\displaystyle{\sum_{n=1}^{N}}\displaystyle{\sum_{k=1}^{K}}z_{nk}^{m}d^2(\mathbf{x}_n, \boldsymbol{\mu}_k)
\label{eqn:fcm}
\end{equation}
where $z_{nk}$ represents the (partial) membership of the $n$th data point $\mathbf{x}_n$ to cluster $k$.  The membership values are constrained such that $\sum_{k=1}^{K}z_{nk}=1$ and $z_{nk}\geq0$.  The $m>1$ is the \emph{fuzzifier} parameter which controls the degree to which memberships are mixed or non-binary, $\boldsymbol{\mu}_k$ is the prototype representing cluster $k$, and $d^2(\mathbf{x}_n, \boldsymbol{\mu}_k)$ is generally taken to be the squared Euclidean distance between $\mathbf{x}_n$ and $\boldsymbol{\mu}_k$. 

Although the FCM has been extensively used, until recently, probabilistic interpretations of FCM had not been developed.  These interpretations were contributed by Heller \etal  by introducing the Bayesian Partial Membership model (BPM) \cite{heller:2008} and Glenn \etal  through the introduction of Bayesian Fuzzy Clustering (BFC) \cite{GlennZare:2014,Glenn:2013}.   Both of these Bayesian models for partial membership methods are non-hierarchical.  In contrast, the proposed PM-LDA extends these models to a hierarchical approach that allows for semantic image segmentation like LDA.  BFC and BPM are reviewed in the following sub-sections.  

\subsection{Bayesian Partial Membership Model} \label{sec:heller}

In a finite mixture model, the data likelihood of $\mathbf{x}_n$,  is
\begin{equation}
p(\mathbf{x}_n|\boldsymbol{\beta})=\sum_{k=1}^{K}\pi_kp_k(\mathbf{x}_n|\beta_k),
\end{equation}
where $\left\{ \pi_k \right\}_{k=1}^K$  are the mixture weights and ${\boldsymbol{\beta}}=\{{\beta}_1, {\beta}_2, ..., {\beta}_K\}$ are the mixture component parameters. $p_k(\mathbf{x}_n|\beta_k)$ is the $k^{th}$ mixture component with parameters ${\beta}_k$. In this model, a data point is assumed to come from one (and only one) of the $K$ mixture components. Thus, given its component assignment, $\mathbf{z}_n$, the probability of a data point, $\mathbf{x}_n$,  is defined as $p(\mathbf{x}_n|\mathbf{z}_n,\boldsymbol{\beta})=\prod_{k=1}^{K}p_k(\mathbf{x}_n|\beta_k)^{z_{nk}}$ where $
z_{nk} \in \{0,1\}$, $\sum_{k=1}^{K}z_{nk}=1$, 
and $\mathbf{z}_n=[z_{n1}, z_{n2}, ..., z_{nK}]$ is the binary membership vector.   If $z_{nk}=1$, the data point $\mathbf{x}_n$ is assumed to have been drawn from mixture component $k$. 

In order to obtain a model allowing multiple cluster memberships for a data point, the constraint $z_{nk} \in \{0, 1\}$ is relaxed to $z_{nk} \in [0,1]$.
The modified constraints and the inclusion of prior distributions for several  key parameters results in the Bayesian Partial Membership model, 
\begin{eqnarray}
p(\boldsymbol{\pi},s,\mathbf{z}_n, \mathbf{x}_n|\boldsymbol{\alpha},\lambda,\boldsymbol{\beta})&=&p(\boldsymbol{\pi}|\boldsymbol{\alpha})p(s|\lambda)p(\mathbf{z}_n|\boldsymbol{\pi} s) \nonumber \\
&&\prod_{k=1}^{K}p_k(\mathbf{x}_n|\beta_k)^{z_{nk}}, 
\label{eqn:bpm}
\end{eqnarray}
where $z_{nk}\in[0,1]$, $\sum_{k=1}^{K}z_{nk}=1$, and $\boldsymbol{\pi}$ is the cluster mixing proportion assumed to be distributed according to a Dirichlet distribution with parameter $\boldsymbol{\alpha}$, \emph{i.e.,} $\boldsymbol{\pi} \sim \text{Dir}(\boldsymbol{\alpha})$. 
The scaling factor, $s$, determines the level of cluster mixing and is distributed according to an exponential distribution with mean $1/\lambda$, $s \sim \text{exp}(\lambda)$, and
$\mathbf{z}_n \sim \text{Dir}(\boldsymbol{\pi} s)$ is the membership vector for data point $\mathbf{x}_n$.

As shown in \cite{heller:2008}, if each of the mixture components are exponential family distributions of the same type, then $p(\mathbf{x}_n|\mathbf{z}_n,\boldsymbol{\beta})=\prod_{k=1}^{K}p_k(\mathbf{x}_n|\beta_k)^{z_{nk}}$ with $z_{nk} \in [0,1]$ and $\sum_{k=1}^{K}z_{nk}=1$, can be written as:
\begin{equation}
p(\mathbf{x}_n|\mathbf{z}_n,\boldsymbol{\beta}) = \text{Expon}\left(\sum_k z_{nk}\boldsymbol{\eta}_k\right).
\label{eqn:expfamprod}
\end{equation}
This indicates that the data generating distribution for $\mathbf{x}_n$ is of the same exponential family distribution as the original $K$ clusters, but with new natural parameters $\sum_k z_{nk}\boldsymbol{\eta}_k$.  The new parameters are a convex combination of the natural parameters, $\boldsymbol{\eta}_k$, of the original clusters weighted by $z_{nk}$. This provides the powerful (and convenient) ability to sample directly from the unique mixture distribution for each data point if the natural parameters of the original clusters and the membership vector for the data point are known. A graphical model of BPM is shown in Fig. \ref{fig:bmp}. The generative process of BPM is described as follows: 

\begin{enumerate}
	\item [1.] Draw mixing proportion $\boldsymbol{\pi}$ from a Dirichlet distribution, $\boldsymbol{\pi} \sim \text{Dir}(\boldsymbol{\alpha})$.
	\item [2.] Draw scaling constant $s$ from an exponential distribution $s \sim \text{exp}(\lambda)=\lambda e^{-\lambda s}$.
	\item [3.] For each data point $\mathbf{x}_n$
	\begin{enumerate}
		\item [(a)] Draw the membership vector $\mathbf{z}_n$ from a Dirichlet distribution, $\mathbf{z}_n \sim \text{Dir}(s\boldsymbol{\pi})$.
		\item [(b)] Draw a data point $\mathbf{x}_n \sim \text{Expon}(\sum_k z_{nk}\boldsymbol{\eta}_k)$.
	\end{enumerate}
\end{enumerate}

\subsection{Bayesian Fuzzy Clustering Model}
 Bayesian Fuzzy Clustering (BFC) also models membership vectors as Dirichlet distributed random variables.  Input data points are then modeled as Gaussian random variables.  Since Gaussian distributions are of the exponential family and using the result in \eqref{eqn:expfamprod}, the natural parameters for the Gaussian representing each data point are convex combinations of the natural parameters of the individual clusters weighted by the associated partial membership values. In order to incorporate the fuzzifier parameter used in FCM as shown in \eqref{eqn:fcm}, each membership value is raised to the $m${th} power.  This results in the  data likelihood distribution (named in \cite{GlennZare:2014} as the Fuzzy Data Likelihood (FDL)),
\begin{equation}
p(\mathbf{X}|\mathbf{Z},\mathbf{Y}) =\prod_{n=1}^{N}\frac{1}{Z(\mathbf{z}_n,m,\mathbf{Y})}\prod_{k=1}^{K}\mathcal{N}(\mathbf{x}_n|\boldsymbol{\mu}_k,\mathbf{Q}=z_{nk}^m\mathbf{I}) \nonumber \\
\end{equation}
with
\begin{equation}
\mathbf{Y}=\left\{\boldsymbol{\mu}_1, \cdots, \boldsymbol{\mu}_K \right\},
\end{equation}
and a prior distribution for the cluster membership (called Fuzzy Cluster Prior (FCP)), 
\begin{equation}\label{eqn:bfc_prior}
\tilde{p}(\mathbf{Z}|Y)=\prod_{n=1}^{N}Z(\mathbf{z}_n,m,\mathbf{Y})\left(\prod_{k=1}^K z_{nk}^{-mp/2}\right)\text{Dir}(\mathbf{z}_n|\boldsymbol{\alpha}) 
\end{equation}
and a Gaussian prior distribution on cluster prototypes
\begin{equation}\label{eqn:bfc_prior1}
p(\mathbf{Y})=\prod_{k=1}^{K}\mathcal{N}(\boldsymbol{\mu}_k|\boldsymbol{\mu}_y,\boldsymbol{\Sigma}_y),
\end{equation}
where $Z(\mathbf{z}_n, m, \mathbf{Y})$ is a normalization constant, $\boldsymbol{\mu}_k$ is the mean of the $k$th Gaussian cluster, $p$ is the dimensionality of the data, $\boldsymbol{\mu}_y$ and $\boldsymbol{\Sigma}_y$ are the sample mean and sample covariance, respectively.
The joint likelihood of data and parameters 

\begin{eqnarray}\label{eqn:bfc_dl}
&&p(\mathbf{X},\mathbf{Z},\mathbf{Y})=p(\mathbf{X}|\mathbf{Z},\mathbf{Y})\tilde{p}(\mathbf{Z}|\mathbf{Y})p(\mathbf{Y})\nonumber \\
&&\propto \text{exp}\left\{{-\frac{1}{2}\sum_{n=1}^{N}\sum_{k=1}^{K}z_{nk}^m  \left(\mathbf{x}_n-\boldsymbol{\mu}_k\right)^T\left(\mathbf{x}_n-\boldsymbol{\mu}_k\right)}\right\} \nonumber \\
&&\prod_{n=1}^N\prod_{k=1}^{K}z_{nk}^{\alpha_k-1}\text{exp}\left\{-\frac{1}{2}\sum_{k=1}^K\left(\boldsymbol{\mu}_k-\boldsymbol{\mu}_y\right)^T\boldsymbol{\Sigma}_y^{-1}\left(\boldsymbol{\mu}_k-\boldsymbol{\mu}_y\right)\right\}. \nonumber \\
&&
\end{eqnarray}

Given the cluster prototype $\boldsymbol{\mu}_k$ and membership $z_{nk}$, the distribution of data point $\mathbf{x}_n$ is described as
\begin{eqnarray}\label{eqn:bfc1}
&&p(\mathbf{x}_n|\mathbf{z}_n,\boldsymbol{\mu}_k) \propto \text{exp}\Bigg\{-\frac{1}{2}\Bigg[ \Bigg(\sum_{k=1}^{K}z_{nk}^m \Bigg)\mathbf{x}_n^T\mathbf{x}_n\nonumber\\
&&-2\mathbf{x}_n^T\Bigg(\sum_{k=1}^{K}z_{nk}^m\boldsymbol{\mu}_k \Bigg) 
+\Bigg( \sum_{k=1}^{K}z_{nk}^m\boldsymbol{\mu}_k^T\boldsymbol{\mu}_k \Bigg)\Bigg]\Bigg\}.
\end{eqnarray}

Let $\mathbf{Q}=\sum_{k=1}^{K}z_{nk}^m \mathbf{I}$, and $\boldsymbol{\mu}=\frac{1}{\sum_{k=1}^{K}z_{nk}^m}\left(\sum_{k=1}^{K}z_{nk}^m \boldsymbol{\mu}_k\right)$, Equation (\ref{eqn:bfc1}) can be rewritten as
\begin{eqnarray}
p(\mathbf{x}_n|\mathbf{z}_n,\boldsymbol{\mu}_k) &\propto&\text{exp}\left\{-\frac{1}{2}\left[\mathbf{x}_n^T\mathbf{Q}\mathbf{x}_n-2\mathbf{x}_n^T\mathbf{Q}\boldsymbol{\mu}+\boldsymbol{\mu}^T\mathbf{Q}\boldsymbol{\mu}\right] \right\}	\nonumber \\
&\times& \text{exp}\left\{-\frac{1}{2}\left[\left(\sum_{k=1}^{K}z_{nk}^m\boldsymbol{\mu}_k^T\boldsymbol{\mu}_k\right)-\boldsymbol{\mu}^T\mathbf{Q}\boldsymbol{\mu}\right]\right\},\nonumber \\
&&
\end{eqnarray}
then
\begin{equation}
p(\mathbf{x}_n|\mathbf{z}_n,\boldsymbol{\mu}_k)=\mathscr{N}(\mathbf{x}_n|\boldsymbol{\mu},\mathbf{Q}).
\label{eqn:6}
\end{equation}
For data point $\mathbf{x}_n$, the generative process implicitly defined in the BFC is as follows:
\begin{enumerate}
	\item [(a)] Draw the membership vector $\mathbf{z}_n$ from the FCP.  
	\item [(b)] Draw a data point $\mathbf{x}_n$ from a Gaussian distribution, $\mathbf{x}_n \sim \mathcal{N}(\dot|\boldsymbol{\mu},\mathbf{Q})$, with  
	\begin{equation}
	\mathbf{Q}=\sum_{k=1}^{K}z_{nk}^m \mathbf{I} \quad \quad \text{and} \quad\quad \boldsymbol{\mu}=\frac{1}{\sum_{k=1}^{K}z_{nk}^m}\left(\sum_{k=1}^{K}z_{nk}^m \boldsymbol{\mu}_k\right). \nonumber
	\end{equation}
\end{enumerate}
The FCP is an improper prior which is uninformative about the membership prior belief.  Thus, is it unclear how to sample memberships from the FCP in \cite{GlennZare:2014}. However, it is also noted in \cite{GlennZare:2014} that the FCP can be converted to a proper prior by replacing the second term $\left(\prod_{k=1}^K z_{nk}^{-mp/2}\right)$ with a product of Inverse-Gamma distribution with shape parameter as $mp/2-1$ with a small scale parameter.

\subsection{Unified Partial Membership Model}

Considering both the BPM and BFC, a unified model which combines the two models is developed in this paper.
A comparison of the terms in the BFC and BPM models is shown in Table \ref{tab:comp}. The major difference between the BPM and BFC models is how the two approaches model the degree of mixing between clusters.  The BFC uses both a fixed {fuzzifier} parameter and the membership prior FCP to control the degree of mixing between clusters.  In contrast, in the BPM, the degree of mixing between clusters is controlled only through the scaling hyper-parameter, $s$, found in the prior distribution on partial membership values. The BPM explicitly defines the cluster mixing proportion $\boldsymbol{\pi}$ while the BFC does not. In terms of the data generating distribution, coefficients of the convex combination for BFC are raised to the $m$ power while the BPM does not.

\begin{center}
	\captionof{table}{Table comparison between BFC, BPM and the unifying model.}
	\label{tab:comp} 
	\begin{tabular}{ | p{1.6cm}  | p{1.5cm}  | p{2.2cm}  | p{1.75cm} |}
		\hline
		& \vspace{0.8mm}BFC  & \vspace{0.8mm}BPM & Unifying model \\ \hline
		\emph{fuzzifier} & $m$ & $m=1$ & $m$  \\ \hline
		Scaling \newline Factor & \vspace{0.5mm}N/A &\vspace{0.5mm} $s\sim\text{exp}(\lambda)$ & \vspace{0.5mm} $s\sim\text{exp}(\lambda)$\\ \hline
		Mixing Prop.& \vspace{0.5mm}N/A & \vspace{0.5mm} $\boldsymbol{\pi}\sim\text{Dir}(\boldsymbol{\alpha})$ & \vspace{0.5mm} $\boldsymbol{\pi}\sim\text{Dir}(\boldsymbol{\alpha})$  \\ 	\hline
		\vspace{0.5mm}Membership  & \vspace{0.5mm}$\mathbf{z}\sim$ FCP  & \vspace{0.5mm} $\mathbf{z}\sim\text{Dir}(\boldsymbol{\pi}s)$ & \vspace{0.5mm}$\mathbf{z}\sim$ Modified FCP \\ \hline
		Coeff. of  Data \newline Generating \newline Distribution  &  \vspace{2mm} $\frac{z_{nk}^m }{\sum_k z_{nk}^m}$ & \vspace{2mm} $\frac{z_{nk}^m }{\sum_k z_{nk}^m}, m=1$  & \vspace{2mm} $\frac{z_{nk}^m }{\sum_k z_{nk}^m}$ \\ \hline
	\end{tabular}
\end{center}

In this paper, a unified model combining BFC and BPM is proposed to combine the two approaches and investigate the effect of {fuzzifier} $m$ and scaling factor $s$ on the membership mixing level. The unified model is composed of an exponential prior on scaling factor defined as
\begin{equation}  \label{eqn:um_prior_s}
p(s|\lambda)=\lambda e^{-\lambda s},
\end{equation}
a Dirichlet prior on mixing proportion defined as
\begin{equation} \label{eqn:um_prior_pi}
p(\boldsymbol{\pi}|\boldsymbol{\alpha})=\frac{\Gamma\left(\sum_{k=1}^{K}\alpha_k\right)}{\prod_{k=1}^{K}\Gamma(\alpha_k)}\prod_{k=1}^{K}(\pi_k)^{(\alpha_k-1)},
\end{equation}
a data likelihood in the same form of the FDL, defined as
\begin{equation} \label{eqn:um_dl}
p(\mathbf{X}|\mathbf{Z},\mathbf{Y}) =\prod_{n=1}^{N}\frac{1}{Z(\mathbf{z}_n,m,\mathbf{Y})}\prod_{k=1}^{K}\mathcal{N}(\mathbf{x}_n|\boldsymbol{\mu}_k,\mathbf{Q}=z_{nk}^m\mathbf{I}),  \\
\end{equation}
and a prior distribution for the cluster membership defined as
\begin{equation} \label{eqn:um_mem}
\tilde{p}(\mathbf{Z}|\mathbf{Y})=\prod_{n=1}^{N}Z(\mathbf{z}_n,m,\mathbf{Y})\left(\prod_{k=1}^K z_{nk}^{-mp/2}\right)\text{Dir}(\mathbf{z}_n|{s\boldsymbol{\pi}}),
\end{equation}
which modifies the FCP in the BFC by replacing the Dirichlet parameter with $s\boldsymbol{\pi}$.
The joint likelihood of the unifying model can be computed as
\begin{eqnarray}
&&p(\mathbf{X},\mathbf{Z},\boldsymbol{\pi},s|\mathbf{Y})=p(\mathbf{X}|\mathbf{Z},\mathbf{Y})\tilde{p}(\mathbf{Z}|\mathbf{Y})p(\boldsymbol{\pi})p(s)\nonumber \\
&=&\prod_{n=1}^{N}\prod_{k=1}^{K}\left(2\pi\right)^{-p/2}\exp\left\{-\frac{1}{2}z_{nk}^m(\mathbf{x}_n-\boldsymbol{\mu}_k)^T(\mathbf{x}_n-\boldsymbol{\mu}_k)\right\} \nonumber \\
&&\frac{\Gamma\left(\sum_{k=1}^{K}s{\pi}_k\right)}{\prod_{k=1}^{K}\Gamma(s{\pi}_k)}\prod_{k=1}^{K}(z_{nk})^{(s{\pi}_k-1)} \frac{\Gamma\left(\sum_{k=1}^{K}\alpha_k\right)}{\prod_{k=1}^{K}\Gamma(\alpha_k)}\nonumber \\
&&\prod_{k=1}^{K}(\pi_k)^{(\alpha_k-1)}\lambda e^{-\lambda s}.\nonumber \\
\\ \nonumber
\end{eqnarray} 
Thus, the log of this likelihood is
\begin{eqnarray}
&&\ln p(\mathbf{X},\mathbf{Z},\mathbf{Y},s,\boldsymbol{\pi})=\text{const}+\nonumber \\
&&\sum_{n=1}^N\sum_{k=1}^K\left\{-\frac{1}{2}z_{nk}^m(\mathbf{x}_n-\boldsymbol{\mu}_k)^T(\mathbf{x}_n-\boldsymbol{\mu}_k)\right\}+\sum_{n=1}^{N}\bigg\{\nonumber \\
&&\ln \Gamma \left(\sum_{k=1}^{K}s\pi_{k}\right)-\sum_{k=1}^{K}\ln\Gamma(s\pi_{k})+\sum_{k=1}^{K}(s\pi_{k}-1)\ln z_{nk}\bigg\}+ \nonumber \\
&&\ln \Gamma \left(\sum_{k=1}^{K}\alpha_k\right)-\sum_{k=1}^{K}\ln\Gamma \left(\alpha_k \right) +\sum_{k=1}^{K}(\alpha_k-1)\ln \pi_k+\nonumber \\
&& \ln \lambda - \lambda s.
\end{eqnarray}

\begin{figure*}[!htb]
	\centering
	\includegraphics[width=1.0\textwidth]{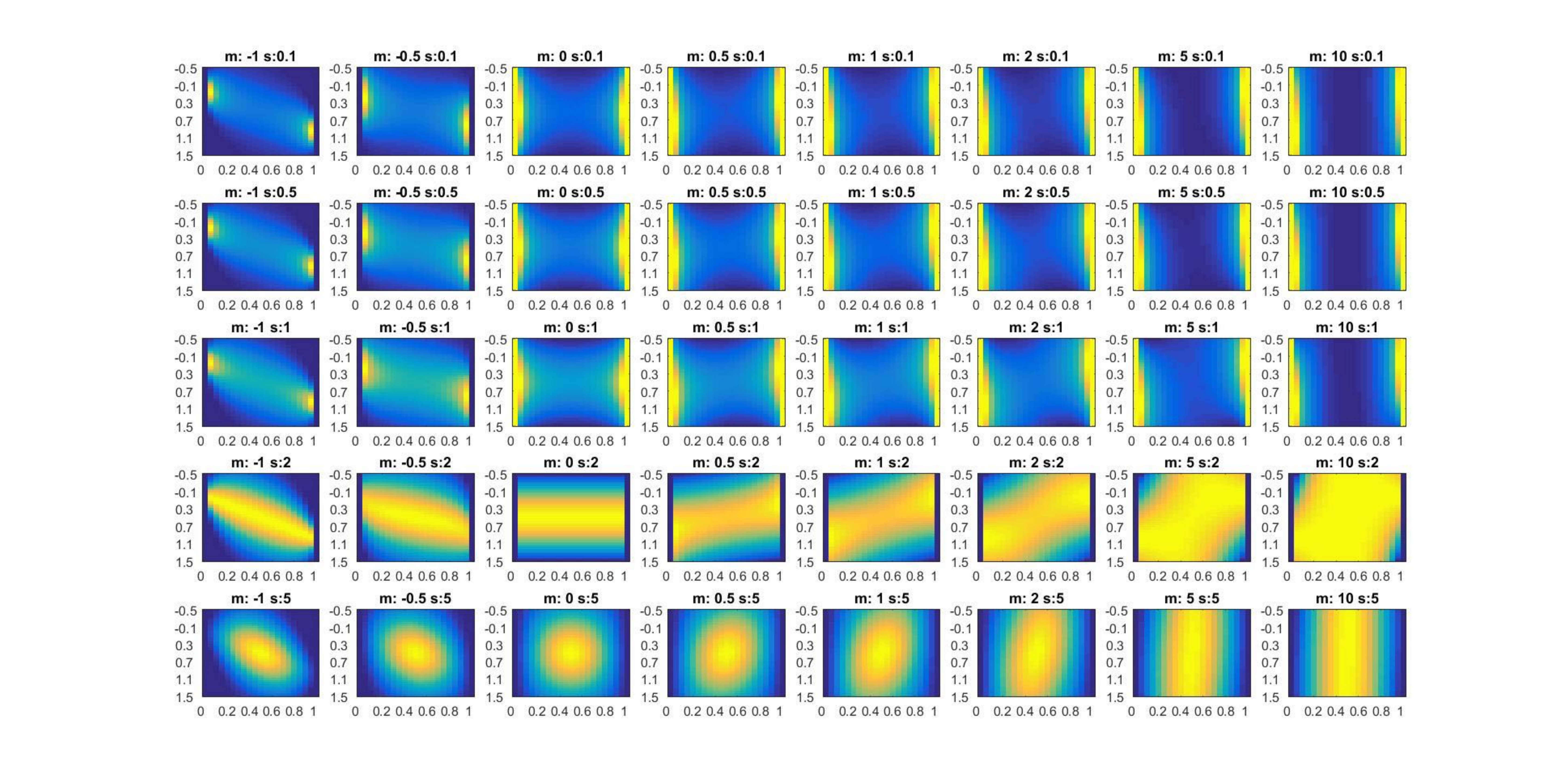}
	\caption{Log likelihood values of the unified model with different $x$, $m$ and $s$ parameter settings. To generate this figure, the unified model was assumed to have two Gaussian clusters, $\mathcal N(0,1)$ and $\mathcal N(1,1)$. $\boldsymbol{\pi}$ = $[0.5,0.5]$,  $\boldsymbol{\alpha}$ = $[1,1]$. The $Y-$axis denotes the value of an input data point, $x$, which varies from $-0.5$ to $1.5$ with increment 0.05. The $X-$axis denotes the membership in the first cluster varied from $0$ to $1$. For each row, the scaling factor $s$ is fixed and the {fuzzifier} $m$ is varied to be different values. For each column, the {fuzzifier} $m$ is fixed and the scaling factor $s$ is varied to be different values. The color indicates the log likelihood value for the fixed data set with a given membership vector, $m$ and $s$ value. Yellow corresponds to a high log likelihood value and blue corrsponds to a low log likelihood value. }
	\label{fig:unifiedModel}
\end{figure*}

Figure \ref{fig:unifiedModel} illustrates the effect of $m$ and $s$ on the membership mixing level in the unified model. To generate this figure, two Gaussian clusters were considered, $\mathcal N(0,1)$ and $\mathcal N(1,1)$.  A sequences of input data points, $\mathbf{X}$, were generated in range $[-0.5, 1.5]$ with increment $0.05$, and a set of membership values for the first cluster, $\mathbf{Z}$, were generated in range $[0, 1]$ with increment $0.05$ (resulting in the membership values in the second cluster being 1-$\mathbf{Z}$). The {fuzzifier} $m$ was varied to be $-1$, $-0.5$, $0$, $0.5$, $1$, $2$, $5$, and $10$. The scaling factor $s$ was varied to be $0.1$, $0.5$, $1$, $2$, and $5$. The parameters $\lambda$, $\boldsymbol{\pi}$, and $\boldsymbol{\alpha}$  were set to be $1$, $[0.5,0.5]$, and $[1,1]$, respectively. Given this generated data set and parameters, the log of joint likelihood in the unifying model was computed and  shown in Figure \ref{fig:unifiedModel}.  Here, the $X-$axis denotes the membership in the first cluster ranging from $0$ to $1$ and the $Y-$axis denotes the input data point value, $x$, ranging from $-0.5$ to $1.5$.  As shown in each column where the {fuzzifier} $m$ is fixed, as $s$ increases, mixing memberships correspond to higher log likelihood values. Memberships with high log likelihood values (bright yellow region) are gradually moved from the left and right margins (corresponding to crisp memberships) to the horizontal center (corresponding to highly mixed memberships). The {fuzzifier} $m$ determines the vertical position of the parameters with high log likelihood values.  As shown in each row, when $m$ increases, data points with high log likelihood values are gradually expanded to the top and bottom margins. This figure illustrates that the scaling factor $s$ and the fuzzifier $m$ impact resulting partial membership values in different ways.  One observation is that the scaling factor $s$ impacts the cluster mixing level by controlling the prior knowledge on memberships and the {fuzzifier} $m$ by controlling the percentage of mixing data points.

 The generative processes for both the BFC and BPM models are similar. The main difference between the BPM and BFC models is that the BFC uses both a fixed {fuzzifier} parameter and a scaling parameter to control the degree of mixing between topics.  In contrast, in the BPM, the degree of mixing between topics is controlled only through a scaling hyper-parameter, $s$, found in the prior distribution on partial membership values. The unified model incorporates both of these parameters to allow a larger degree of freedom in controlling the mixing between topics.

\section{Partial Membership LDA}  \label{sec:PMLDA}

To generalize LDA by allowing for partial memberships, a PM-LDA model is proposed in this paper, where partial memberships are introduced to the standard LDA in a similar fashion to the developed unified model. Both the {fuzzifier} $m$ and the scaling factor $s$ are incorporated in the proposed PM-LDA model. In PM-LDA, the random variable associated with a data point is assumed to be distributed according to multiple topics with a continuous partial membership in each topic. First, we consider the simplest case where the {fuzzifier} $m=1$.  Given, $m=1$, the PM-LDA model is
\begin{eqnarray}
p(\boldsymbol{\pi}_d,s_d, \mathbf{z}_{dn},\mathbf{x}_{dn}|\boldsymbol{\alpha},\lambda,\boldsymbol{\beta})&=&p(\boldsymbol{\pi}_d|\boldsymbol{\alpha})p(s_d|\lambda)p(\mathbf{z}_{dn}|\boldsymbol{\pi}_ds_d)\nonumber \\
&& \prod_{k=1}^{K}p_k(\mathbf{x}_{dn}|\beta_k)^{{z}_{dnk}} 
\label{eqn:pmlda}
\end{eqnarray} 
where $\mathbf{x}_{dn}$ is the $n$th word in document $d$, $\mathbf{z}_{dn}$ is the partial membership vector of $\mathbf{x}_{dn}$, $\boldsymbol{\pi}_d \sim \text{Dir}(\boldsymbol{\alpha})$ and $s_d \sim \text{exp}(\lambda)$ are the topic proportion and  the level of topic mixing in document $d$, respectively. The parameter $\boldsymbol{\alpha}$ corresponds to the topic composition across a document. For example, in  Fig. \ref{fig:csonar}, the image may be composed of $40\%$ ``sand ripple'' topic and $60\%$ ``flat sand'' topic (\ie, $\boldsymbol{\alpha} = [0.4, 0.6]$).  The  parameter $\lambda$ controls how similar the partial membership vector of each word is expected to be to the topic distribution of the document. For example, a small $\lambda$ would correspond to most words in a document to have partial membership vectors very close to $\boldsymbol{\pi}_{d}$. During image segmentation, a small $\lambda$ generally corresponds to large transition regions (\eg, transition from ``flat sand'' to ``sand ripple'' comprises most of the image). For a large $\lambda$, the partial membership vectors for each word can vary significantly from the document mixing proportions.  In general, a large $\lambda$ corresponds to very narrow (tending towards crisp) transition regions during image segmentation (\eg, the SAS image may have $39\%$ of the visual words as pure ``sand ripple'', $59\%$ as pure ``flat sand'', and only $2\%$ mixed).

The vector  $\mathbf{z}_{dn} \sim \text{Dir}(\boldsymbol{\pi}_{d} s_{d})$ represents the partial memberships of data point $\mathbf{x}_{dn}$ in each of the $K$ topics. 
If each topic distribution is assumed to be of the exponential family, $p_k(\cdot|\beta_k) = \text{Expon}(\boldsymbol{\eta}_k)$, then using the result in \eqref{eqn:expfamprod}, $p(\mathbf{x}_{dn}|\mathbf{z}_{dn},\boldsymbol{\beta}) = \text{Expon}(\sum_k z_{dnk}\boldsymbol{\eta}_k)$. The  graphical model  for PM-LDA is shown in Fig. \ref{fig:pmlda} and the generative procedure of PM-LDA is described as follows.
\begin{enumerate}
	\item [1.] For each document $\mathbf{X}_d$, draw topic proportions $\boldsymbol{\pi}_{d} \sim \text{Dir}(\boldsymbol{\alpha})$.	
	\item [2.] Draw scaling factor $s_d$ from an exponential distribution $s_d \sim \text{exp}(\lambda)=\lambda e^{-\lambda s}$.
	\item [3.] For each word $\mathbf{x}_{dn}$
	\begin{enumerate}
		\item [(a)] Draw the membership vector $\mathbf{z}_{dn} \sim \text{Dir}(\boldsymbol{\pi}_d s_d)$.
		\item [(b)] Draw word $\mathbf{x}_{dn}  \sim \text{Expon}(\sum_k z_{dnk}\boldsymbol{\eta}_k)$.
	\end{enumerate}
\end{enumerate}

For any value of the {fuzzifer} $m$, the above step 3(b) is modified to be: Draw word  $\mathbf{x}_{dn}  \sim \text{Expon}(\sum_k \frac{z_{dnk}^m}{\sum_j z_{dnj}^m }\boldsymbol{\eta}_k)$. $z_{dnk}$ in the last term of Equation (\ref{eqn:pmlda}) is correspondingly modified as $\frac{z_{dnk}^m}{\sum_j z_{dnj}^m }$.

\begin{figure}[!tb]
	\vspace{2mm}
	\centering
	\begin{subfigure}[b]{0.22\textwidth}
		\centering
		\includegraphics[width=0.8\linewidth]{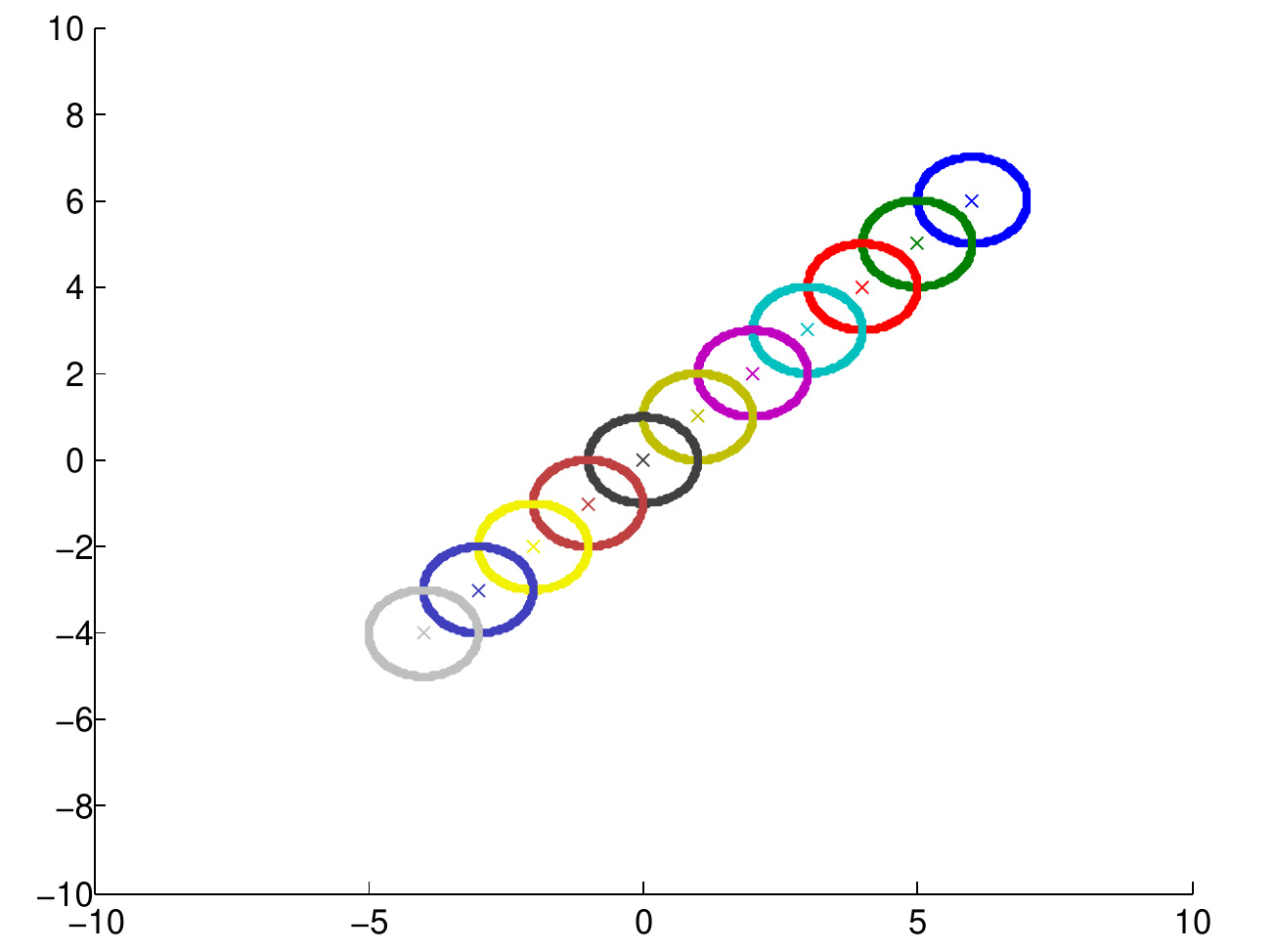}
		\captionsetup{labelformat=empty,skip=0pt}
		\caption{(a)}
	\end{subfigure}
	\begin{subfigure}[b]{0.22\textwidth}
		\centering
		\includegraphics[width=0.8\linewidth]{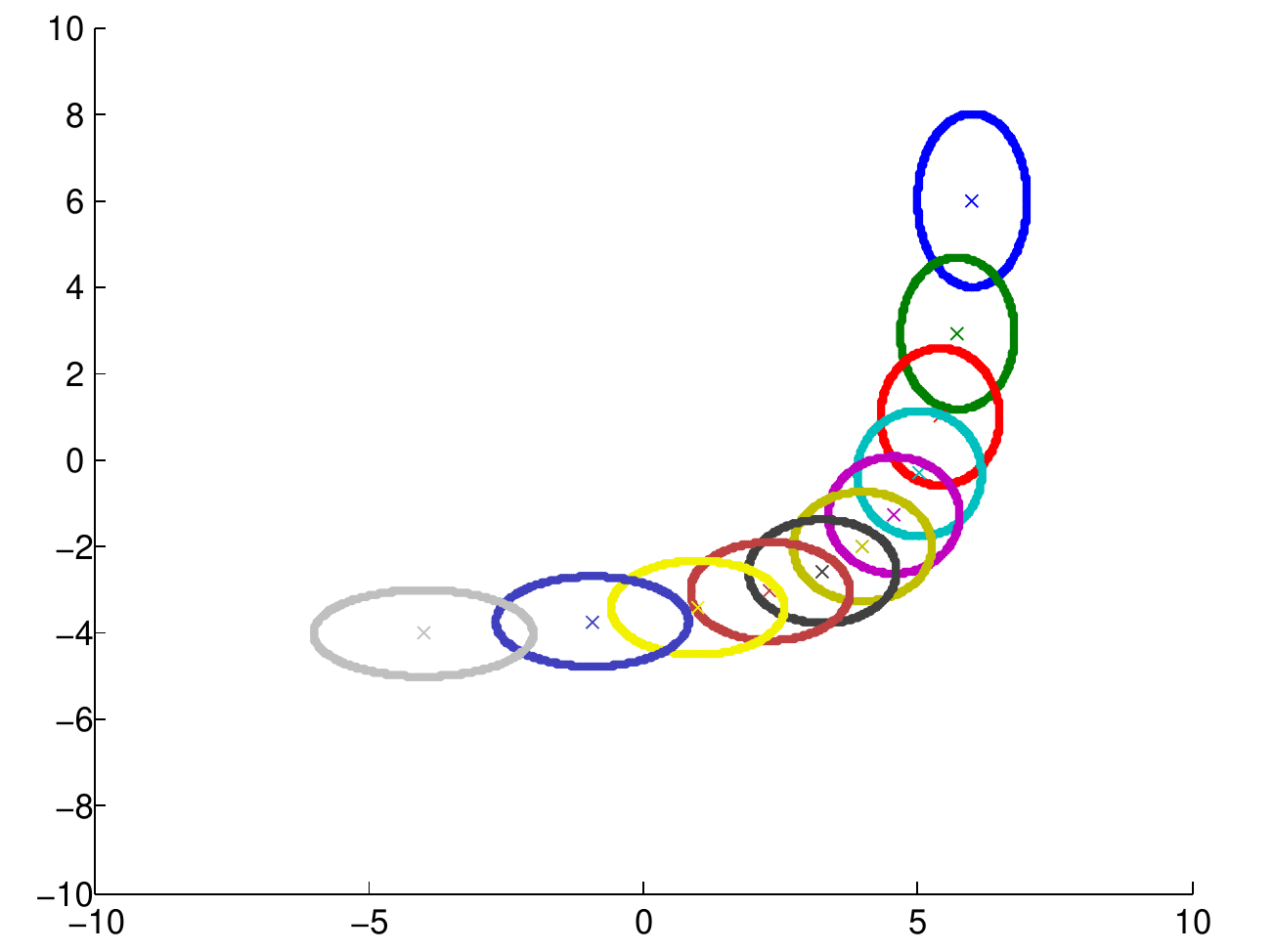}
		\captionsetup{labelformat=empty,skip=0pt}
		\caption{(b)}
	\end{subfigure}
	\caption{Partial membership data generating distributions. In (a), two Gaussian topics with $\mu_1=[-4, -4]; \mu_2=[6, 6]$, $\Sigma_1=\Sigma_2=\mathbf{I}$. In (b), two Gaussian topics with  $\mu_1=[-4,-4]; \mu_2=[6, 6]$, $\Sigma_1=[4, 0; 0, 1],\Sigma_2=[1, 0; 0, 4]$ \cite{heller:2008}.}
	\label{fig:topics}
\end{figure}

In PM-LDA, the membership $\mathbf{z}_{dn}$ is drawn from a Dirichlet distribution which is in contrast to a multinomial distribution as used in LDA. With the infinite number of possible values for $\mathbf{z}_{dn}$, the word generating distributions in PM-LDA are expanded from only $K$ generating distributions (as in LDA) to infinitely many. Fig. \ref{fig:topics}  illustrates this using two Gaussian topic distributions, where the membership value to one topic is varied from $0$ to $1$ with an increment $0.1$. The two original topics are shown as the Gaussian distributions at either end. In LDA,  words are generated from only the two original topic distributions. In PM-LDA, words can be generated from any (of the infinitely many) convex combinations of the topic distributions.  As the scaling factor $s \rightarrow 0$, the PM-LDA model will degrade to the LDA model.

Given the hyperparameters $\boldsymbol{\Psi}=\{\boldsymbol{\alpha},\lambda,\boldsymbol{\beta}\}$, the full PM-LDA model over all words in the $d^{th}$ document is:
\begin{eqnarray}
\label{eqn:joint}
&&p(\boldsymbol{\pi}_d,s_d,\mathbf{Z}_d,\mathbf{X}_d|\boldsymbol{\alpha},\lambda,\boldsymbol{\beta})   \\ \nonumber
&=&p(\boldsymbol{\pi}_d|\boldsymbol{\alpha}) p(s_d|\lambda) \prod_{n=1}^{N_d}p(\mathbf{x}_{dn}|\mathbf{z}_{dn},\boldsymbol{\beta})p(\mathbf{z}_{dn}|\boldsymbol{\pi}_ds_d). \nonumber 
\end{eqnarray}
where $\boldsymbol{\pi}_d$ are the topic proportions, scaling factor $s_d$, partial membership vectors $\mathbf{Z}_d=\{\mathbf{z}_{dn}\}_{n=1}^{N_d}$ and a set of $N_d$  words $\mathbf{X}_d$ for document $d$. 
The log of \eqref{eqn:joint} when considering the specific forms chosen in our model, is shown in \eqref{eqn:onedocument}.
\begin{figure*}[ht!]
	\begin{eqnarray}  \label{eqn:onedocument}
	\pazocal{L}_{d} &=& \ln(p(\boldsymbol{\pi}_d,s_d,\mathbf{Z}_d,\mathbf{X}_d|\boldsymbol{\alpha},\lambda,\boldsymbol{\beta}) ) =  \ln \Gamma \left(\sum_{k=1}^{K}\alpha_k\right)-\sum_{k=1}^{K}\ln\Gamma \left(\alpha_k \right) + \sum_{k=1}^{K}(\alpha_k-1)\ln \pi_{dk} +\ln \lambda - \lambda s_d \nonumber \\
	&+&\sum_{n=1}^{N_d}\ln p(\mathbf{x}_{dn}|\mathbf{z}_{dn},\boldsymbol{\beta}) +\sum_{n=1}^{N_d}\bigg\{\ln \Gamma \left(\sum_{k=1}^{K}s_d\pi_{dk}\right) -\sum_{k=1}^{K}\ln\Gamma(s_d\pi_{dk})+\sum_{k=1}^{K}(s_d\pi_{dk}-1)\ln z_{dnk}\bigg\}
	\end{eqnarray}
\end{figure*}

During parameter estimation, our goal is to maximize $\pazocal{L}=\sum_{d=1}^{D}\pazocal{L}_{d}$ by estimating all the model parameters ${\boldsymbol{\pi}_{d},s_{d}, \mathbf{z}_{dn}, \boldsymbol{\beta}}$. 
In this paper, we employ a Metropolis within Gibbs  \cite{robert:2013, GlennZare:2014} sampling approach.

\section{Parameter Estimation for PM-LDA}
The goal of parameter estimation is to maximize the following  posterior distribution,
\begin{equation}
p(\boldsymbol{\Pi},\mathbf{S},\mathbf{M},\boldsymbol{\beta}|\mathbf{D},\boldsymbol{\alpha},\lambda) \propto p({\boldsymbol{\Pi},\mathbf{S},\mathbf{M},\mathbf{D}|\boldsymbol{\alpha},\lambda,\boldsymbol{\beta})}, 
\label{eqn:post}
\end{equation}
where $\mathbf{D} =\left\{\mathbf{X}_1, \mathbf{X}_2, ..., \mathbf{X}_D \right\}$ includes all training documents and $\boldsymbol{\Pi}, \mathbf{S}, \mathbf{M}$ include all of the topic proportions, scaling factors and membership vectors, respectively.

A Metropolis within Gibbs sampler is employed to perform the MAP inference which can generate samples from the posterior distribution in \eqref{eqn:post}, \cite{robert:2013,GlennZare:2014}. An outline of the sampler is provided in Alg. \ref{alg:clustercenter}. The sampler is a simple and straight-forward implementation composed of a series of draws from candidate distributions for each parameter and then evaluation of the candidate in the appropriate acceptance ratio. Our implementation of the sampler has been posted online.\footnote[1]{Code can be found at: https://github.com/TigerSense/PMLDA}   In our current implementation, we consider the topic distributions to be Gaussian with different means, $\mu_k$, but identical diagonal and isotropic covariance matrices, $\Sigma_k = \sigma^2\mathbf{I}$.

\begin{algorithm}[ht]
	\begin{algorithmic}[1]
		\REQUIRE{A corpus $\mathbf{D}$, the number of topics $K$, and the number of sampling iterations $T$}
		\ENSURE{Collection of all samples: $\boldsymbol{\Pi}^{(t)}, \mathbf{S}^{(t)}, \mathbf{M}^{(t)}$,  $\boldsymbol{\beta}^{(t)}=\left\{\mu^{(t)}_1, \Sigma^{(t)}_1, \mu^{(t)}_2, \Sigma^{(t)}_2..., \mu^{(t)}_K, \Sigma^{(t)}_K\right\}$.
			\FOR{$t=1:T$}
			\FOR{$d=1:D$}
			\STATE \underline{Sample $\boldsymbol{\pi}_d$:} Draw candidate: $\boldsymbol{\pi}^\dagger \sim \text{Dir}(\boldsymbol{\alpha})$ \\
			Accept candidate with probability:\\ $a_{\boldsymbol{\pi}}=\min \left \{ 1, \frac{p(\boldsymbol{\pi}^{\dagger}, s^{(t-1)}, \mathbf{Z}^{(t-1)}, \mathbf{X}|\boldsymbol{\Psi}) p(\boldsymbol{\pi}^{(t-1)}|\boldsymbol{\alpha})}{ p(\boldsymbol{\pi}^{(t-1)}, s^{(t-1)}, \mathbf{Z}^{(t-1)}, \mathbf{X}|\boldsymbol{\Psi}) p(\boldsymbol{\pi}^\dagger|\boldsymbol{\alpha})}\right \}$

			\STATE \underline{Sample $s_d$:} Draw candidate: $s^\dagger \sim \text{exp}(\lambda)$\\
			Accept candidate with probability:\\
			$a_s=\min \left \{ 1, \frac{p(\boldsymbol{\pi}^{(t)}, s^{\dagger}, \mathbf{Z}^{(t-1)}, \mathbf{X}|\boldsymbol{\Psi})p(s^{(t-1)}|\lambda)}{p(\boldsymbol{\pi}^{(t)}, s^{(t-1)}, \mathbf{Z}^{(t-1)}, \mathbf{X}|\boldsymbol{\Psi})p(s^{\dagger}|\lambda)}\right \}$

			\FOR{$n=1:N_d$}
			\STATE \underline{Sample $\mathbf{z}_{dn}$:} Draw candidate: $\mathbf{z}_n^\dagger \sim \text{Dir}(\mathbf{1}_K)$\\
			Accept candidate with probability:\\
			$a_{\mathbf{z}}=\min \left\{1, \frac{p(\boldsymbol{\pi}^{(t)}, s^{(t)}, \mathbf{z}_n^\dagger, \mathbf{x}_n|\boldsymbol{\Psi})}{p(\boldsymbol{\pi}^{(t)}, s^{(t)}, \mathbf{z}_n^{(t-1)}, \mathbf{x}_n|\boldsymbol{\Psi})}\right\} $

			\ENDFOR
			\ENDFOR
			
			\FOR{$k=1:K$}
			\STATE \underline{Sample $\mu_k$:} Draw proposal: ${\mu}_k^{\dagger}\sim\mathcal{N}(\cdot|{\mu}_{\mathbf{D}},f{\Sigma}_{\mathbf{D}})$\\ 
			${\mu}_{\mathbf{D}}$ and ${\Sigma}_{\mathbf{D}}$ are  mean and covariance of the data\\
			Accept candidate with probability:\\
			$a_k=\small{\min\left\{1, \frac{p\left(\boldsymbol{\Pi}^{(t)}, \mathbf{S}^{(t)}, \mathbf{M}^{(t)}, \mathbf{D}|{\mu}_k^{\dagger}\right)\mathcal{N}(\mu_k^{(t-1)}|\mu_\mathbf{D}, \Sigma_\mathbf{D})}{p\left(\boldsymbol{\Pi}^{(t)}, \mathbf{S}^{(t)}, \mathbf{M}^{(t)}, \mathbf{D}|\boldsymbol{\mu}_k^{(t-1)}\right)\mathcal{N}(\mu_k^{\dagger}|\mu_\mathbf{D}, \Sigma_\mathbf{D})} \right\}}$

			\ENDFOR
			\STATE \underline{Sample covariance matrices $\boldsymbol{\Sigma}= \sigma^2\mathbf{I}$:}\\
			Draw candidate from:\\ $\sigma^2 = \frac{1}{2}\left\{\max_{\mathbf{x}_n}d^2(\mathbf{x}_n-{\mu}_{\mathbf{D}})-\min_{\mathbf{x}_n}d^2(\mathbf{x}_n-{\mu}_{\mathbf{D}}) \right\}$\\
			Accept candidate with probability: \\
			$a_{\boldsymbol{\Sigma}}= \min\left\{1, \frac{p\left(\boldsymbol{\Pi}^{(t)}, \mathbf{S}^{(t)}, \mathbf{M}^{(t)}, \mathbf{D}|\boldsymbol{\Sigma}^\dagger\right)}{p\left(\boldsymbol{\Pi}^{(t)}, \mathbf{S}^{(t)}, \mathbf{M}^{(t)}, \mathbf{D}|\boldsymbol{\Sigma}^{(t-1)}\right)} \right\}.$

			\ENDFOR}
	\end{algorithmic}
	\caption{Metropolis-within-Gibbs Sampling Method for Parameter Estimation}
	\label{alg:clustercenter}
\end{algorithm}

The proposed Metropolis within Gibbs scheme will return the full distribution of parameter values given the desired posterior. We use the MAP sample (i.e., the sample with the largest log posterior value) as the final estimate, $\left\{\boldsymbol{\Pi}^{*}, \mathbf{S}^{*},\mathbf{M}^{*},\boldsymbol{\mu}^{*}, \boldsymbol{\Sigma}^{*}\right\}$.

\section{Data \& Experimental Results}
In this section, we show  results of image segmentation on two datasets: (i) Synthetic Aperture SONAR (SAS) imagery and (ii) Sunset imagery,. In the following experiments, unless otherwise specified, the {fuzzifier} $m$ value is set to $1$. 

\subsection{Synthetic Aperture Sonar (SAS) Imagery Dataset} 
The first set of experiments considers segmentation of SAS imagery. 

 \subsubsection{SAS sub-image segmentation and comparison to LDA and FCM} 
\begin{figure*}
	\vspace{2mm}
	\centering
	\begin{subfigure}[t]{0.11\textwidth}
		\centering
		\includegraphics[width=1\linewidth,,height=0.8\linewidth]{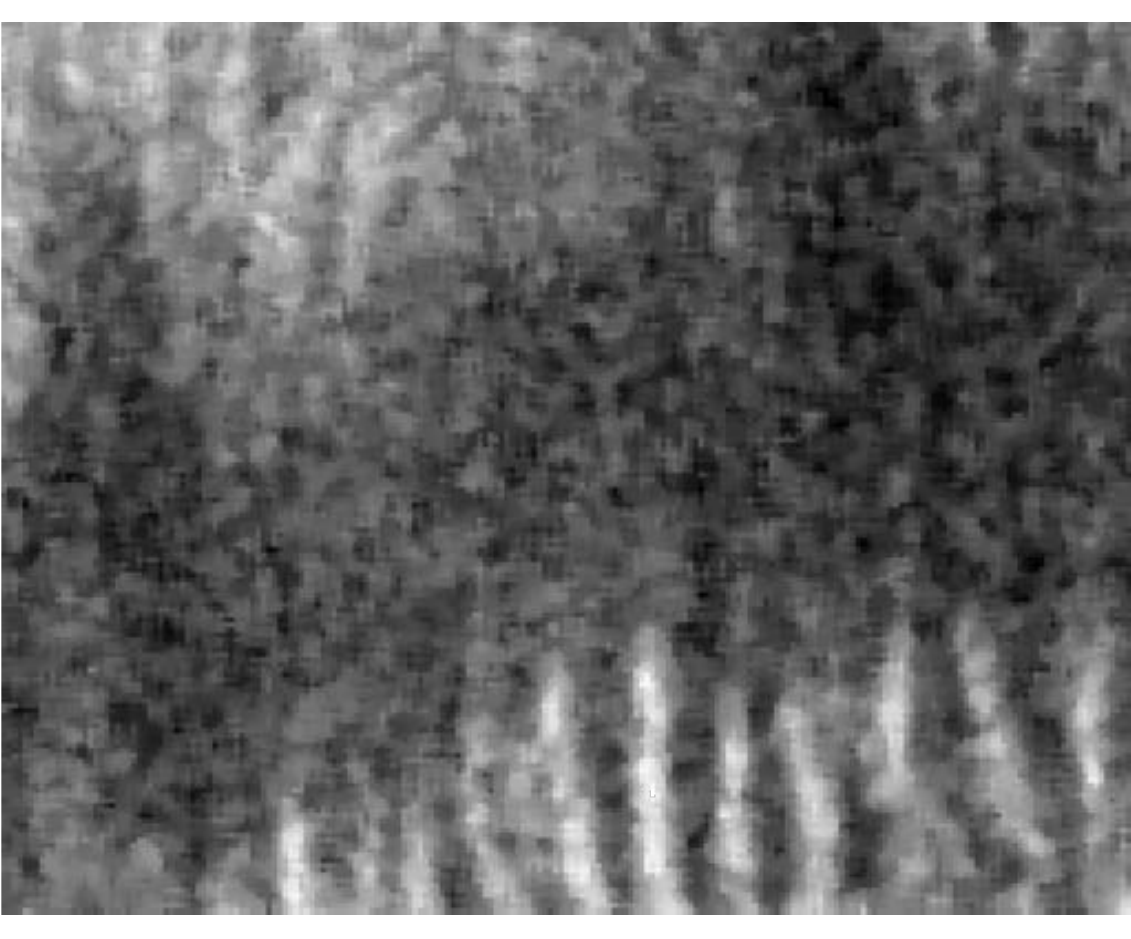}
		\captionsetup{labelformat=empty,width=1\linewidth,justification=centering,skip=0pt} 
		\caption{{(a) Image 1}} \label{fig:c1}
	\end{subfigure}
	\begin{subfigure}[t]{0.11\textwidth}
		\centering
		\includegraphics[width=1\linewidth,,height=0.8\linewidth]{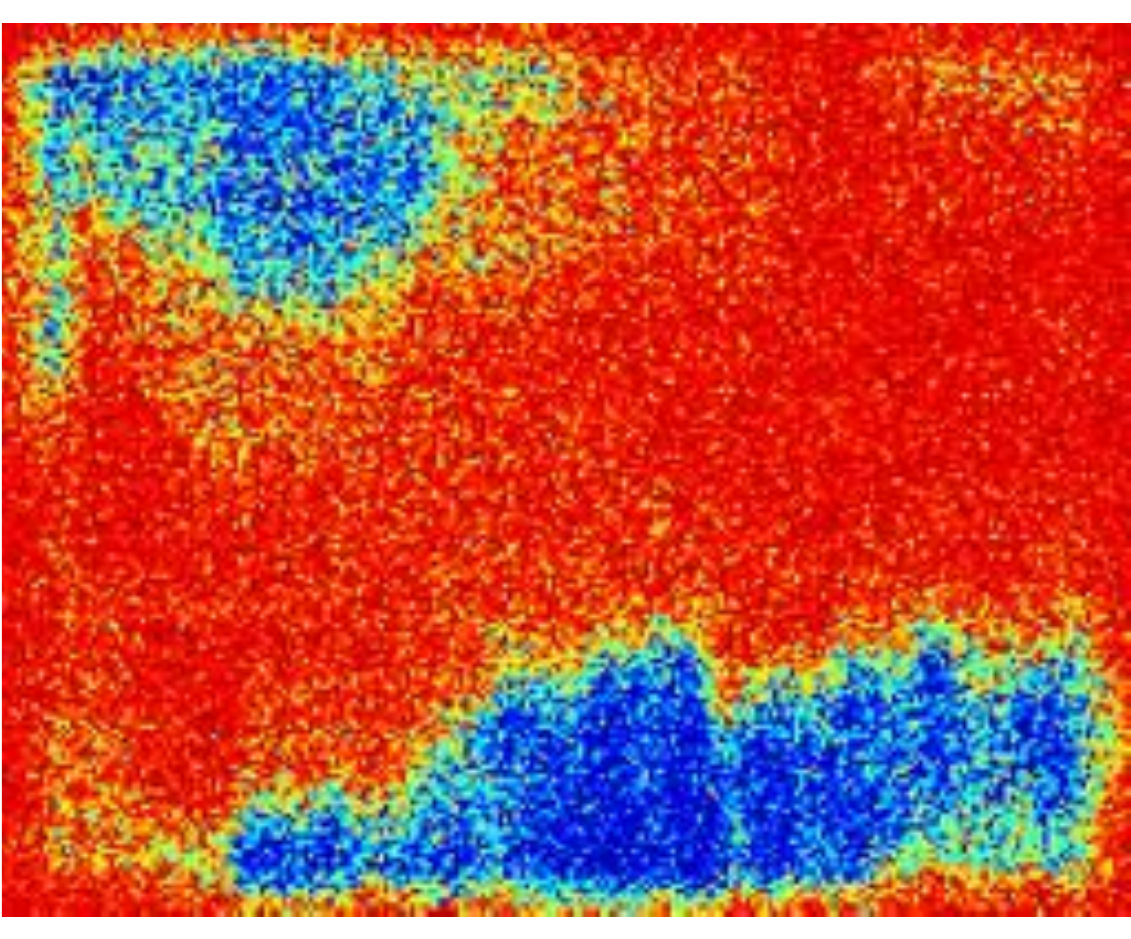}
		\captionsetup{labelformat=empty,width=1\linewidth,justification=centering,skip=0pt} 
		\caption{{(b) PM-LDA:1}}
	\end{subfigure}
	\begin{subfigure}[t]{0.11\textwidth}
		\centering
		\includegraphics[width=1\linewidth,,height=0.8\linewidth]{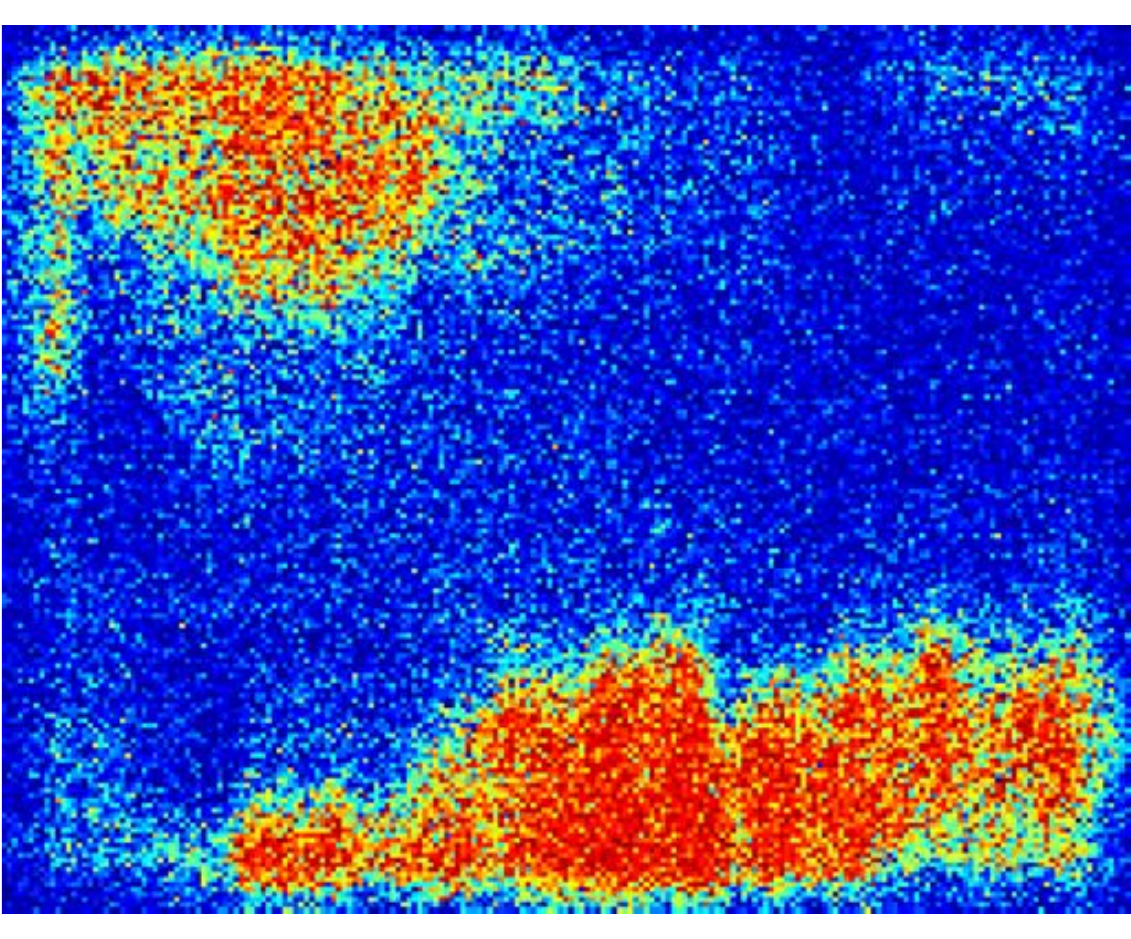}
		\captionsetup{labelformat=empty,width=1\linewidth,justification=centering,skip=0pt} 
		\caption{{(c) PM-LDA:2}} \label{fig:1c}
	\end{subfigure}
	\begin{subfigure}[t]{0.11\textwidth}
		\centering
		\includegraphics[width=1\linewidth,,height=0.8\linewidth]{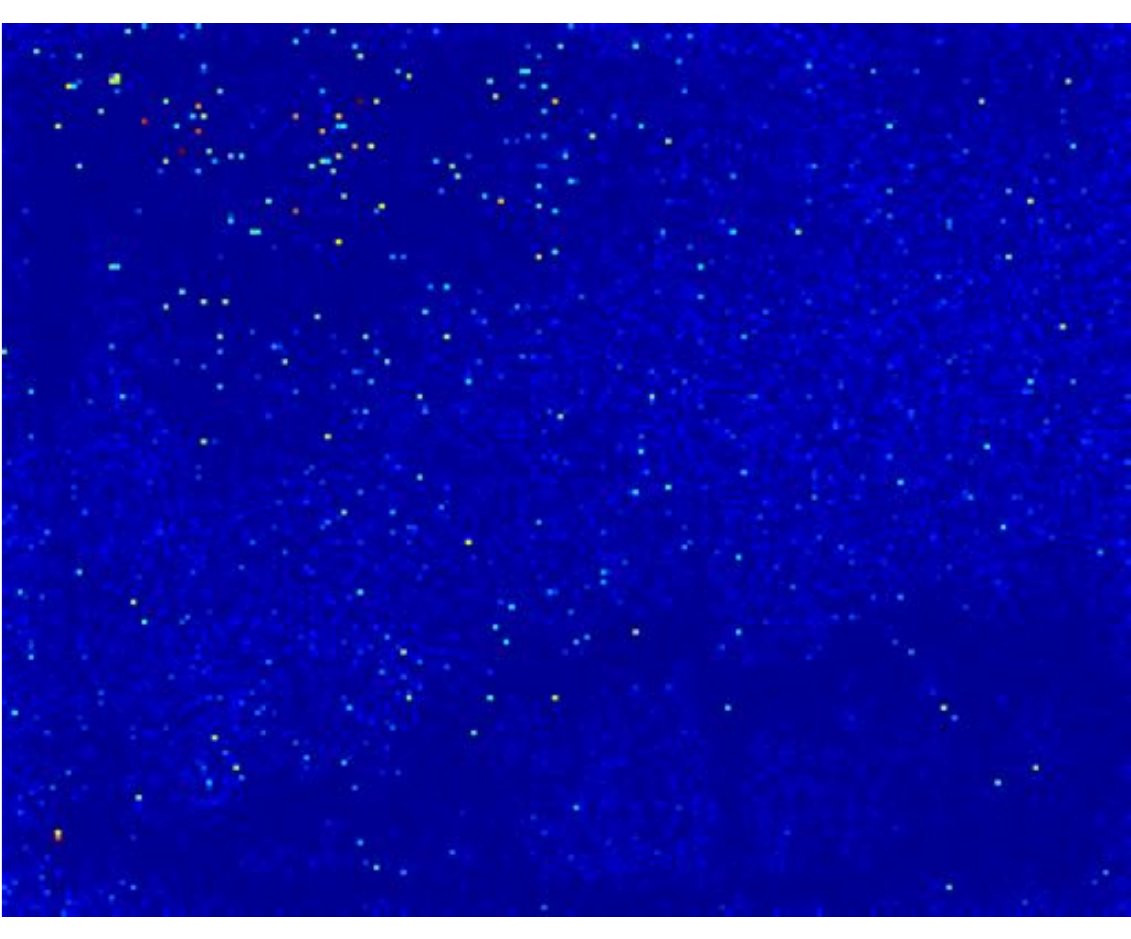}
		\captionsetup{labelformat=empty,width=1.1\linewidth,justification=centering,skip=-5pt} 
		\caption{{(d) PM-LDA:3}}
	\end{subfigure}  
	\begin{subfigure}[t]{0.015\textwidth}
		\centering
		\includegraphics[width=0.91\linewidth]{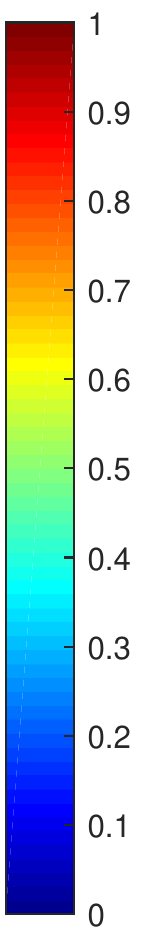}
		\captionsetup{labelformat=empty,width=1\linewidth,justification=centering,skip=0pt} 
		\caption{}
	\end{subfigure}
	\setcounter{subfigure}{4}
	\begin{subfigure}[t]{0.11\textwidth}
		\centering
		\includegraphics[width=1\linewidth,,height=0.8\linewidth]{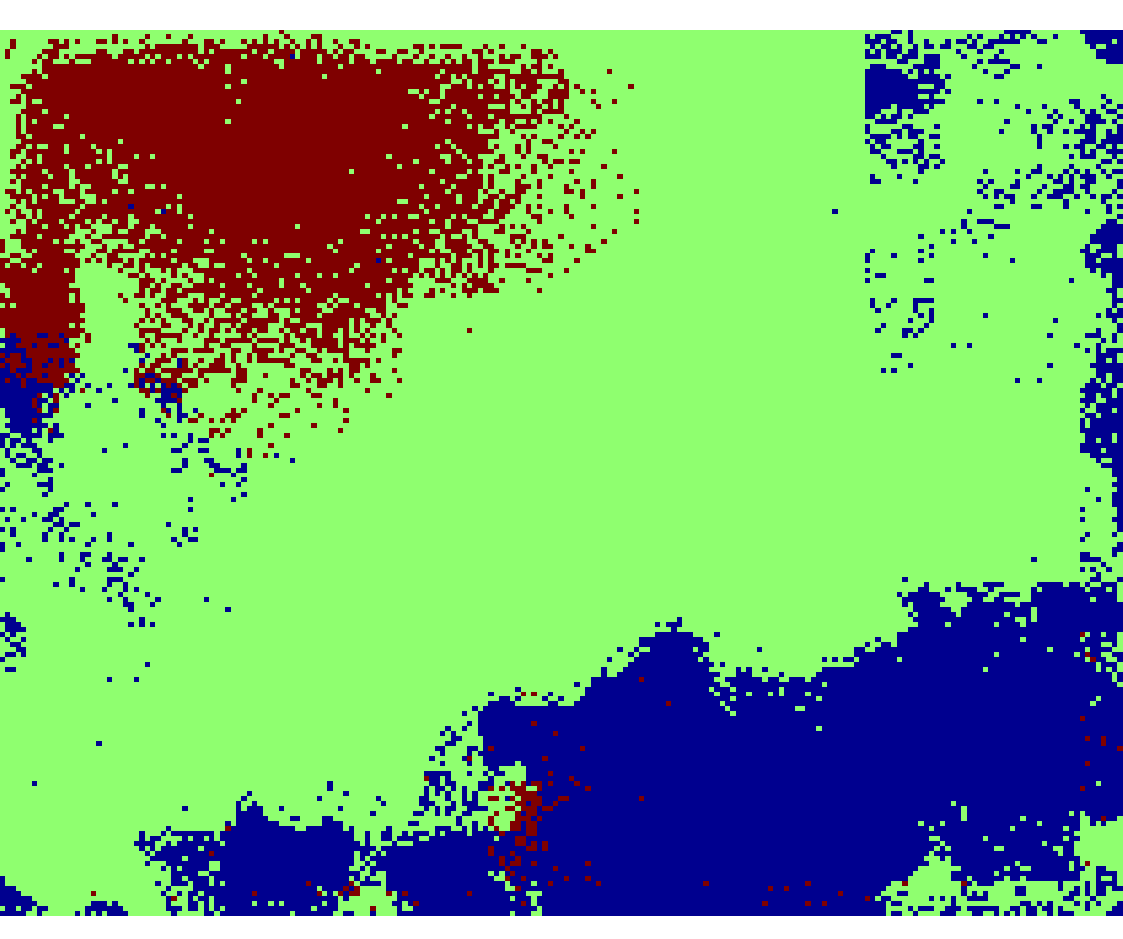}
		\captionsetup{labelformat=empty,width=1\linewidth,justification=centering,skip=0pt} 
		\caption{{(e) LDA}} \label{fig:1e}
	\end{subfigure}   
	\begin{subfigure}[t]{0.11\textwidth}
		\centering
		\includegraphics[width=1\linewidth,,height=0.8\linewidth]{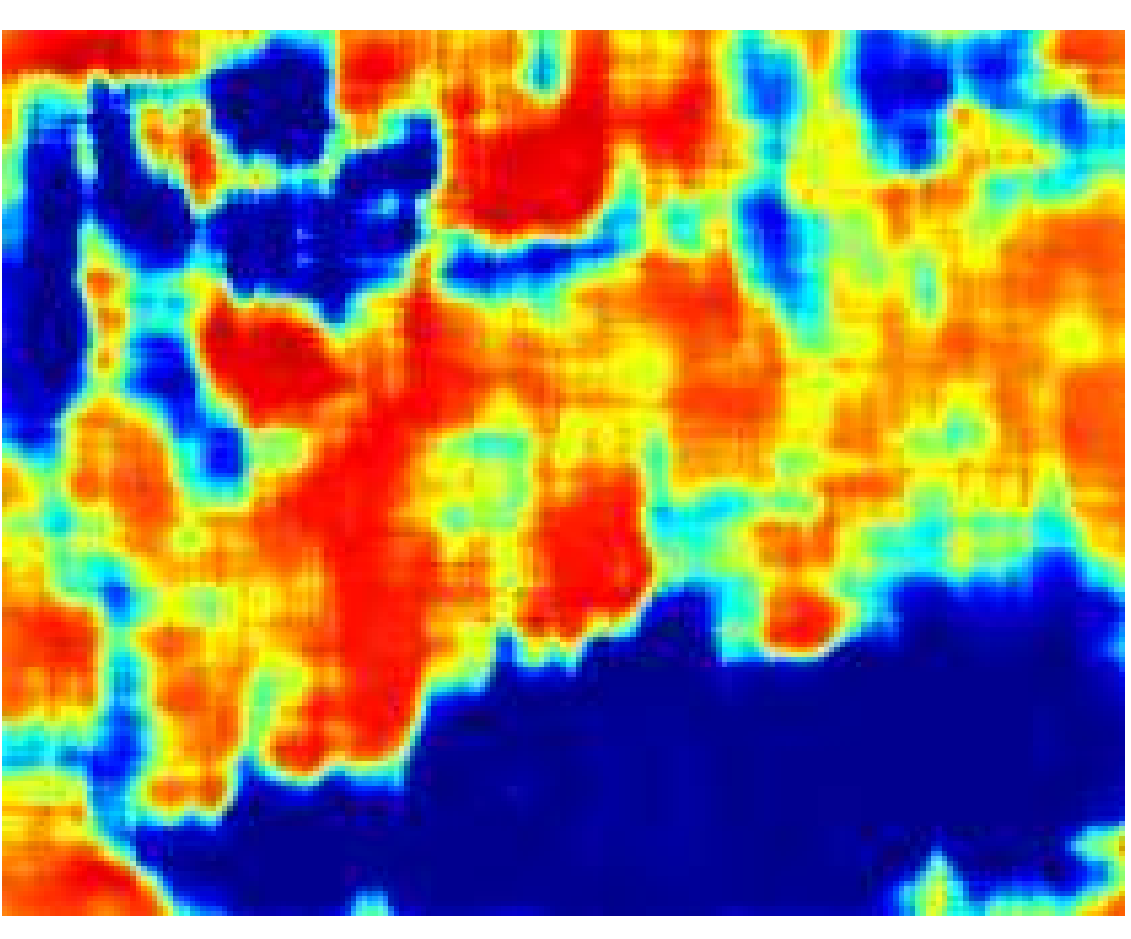}
		\captionsetup{labelformat=empty,width=1\linewidth,justification=centering,skip=0pt} 
		\caption{{(f) FCM:1}}
	\end{subfigure}
	\begin{subfigure}[t]{0.11\textwidth}
		\centering
		\includegraphics[width=1\linewidth,,height=0.8\linewidth]{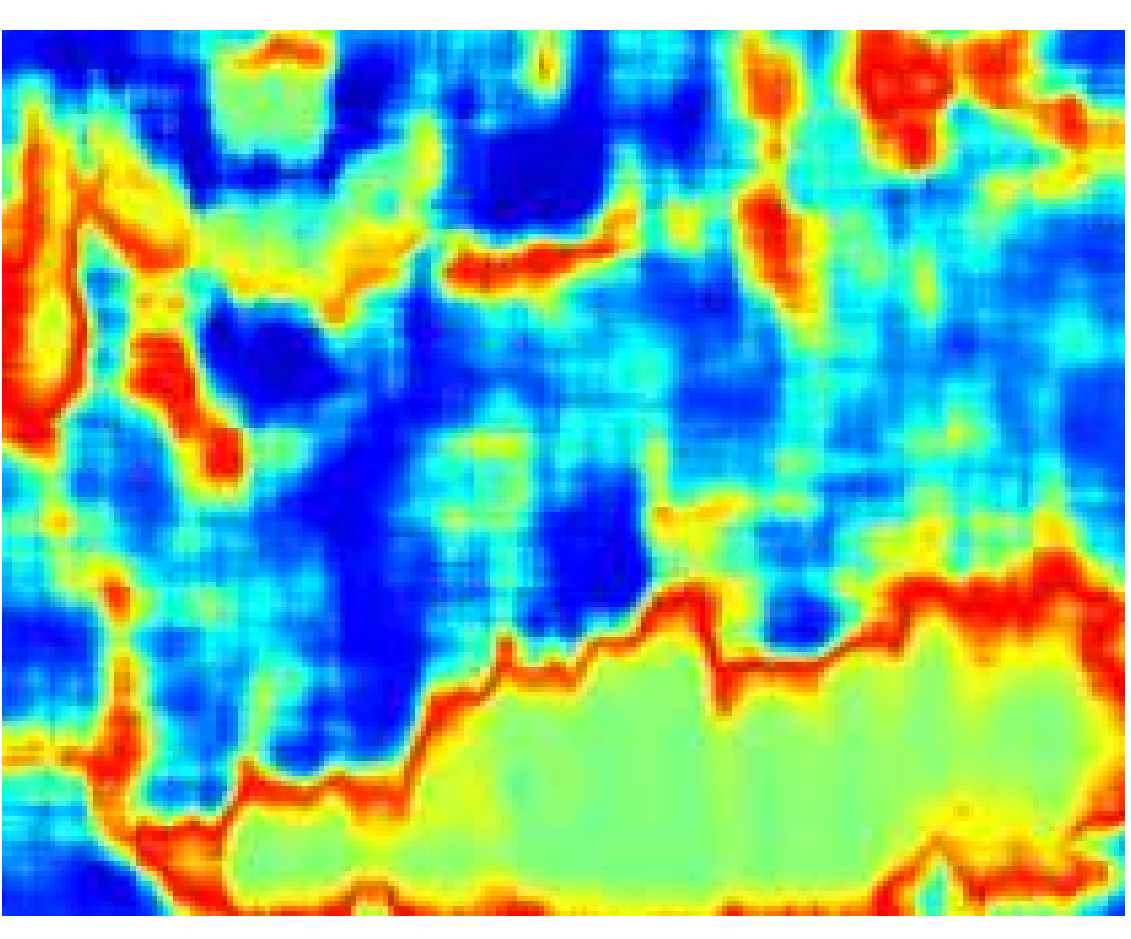}
		\captionsetup{labelformat=empty,width=1\linewidth,justification=centering,skip=0pt} 
		\caption{{(g) FCM:2}}
	\end{subfigure}
	\begin{subfigure}[t]{0.11\textwidth}
		\centering
		\includegraphics[width=1\linewidth,height=0.8\linewidth]{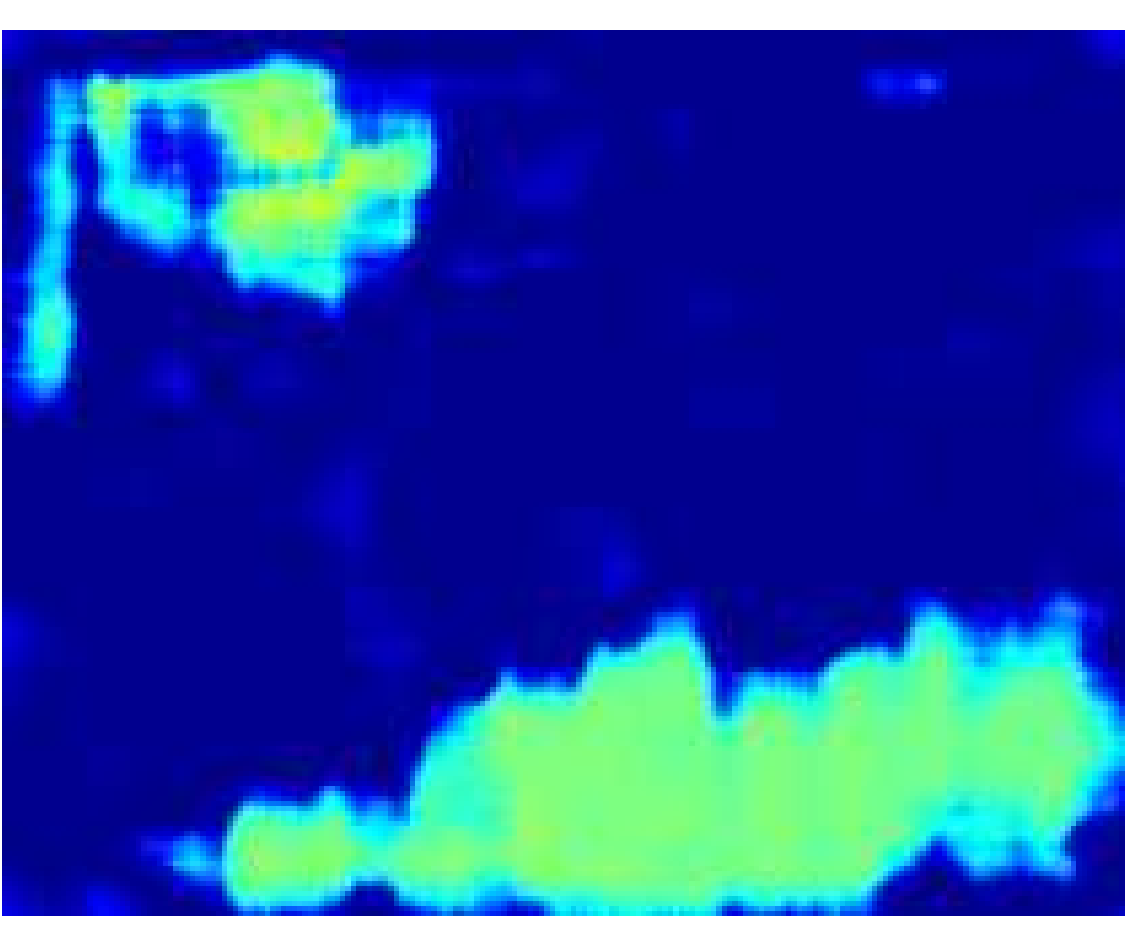}
		\captionsetup{labelformat=empty,width=1\linewidth,justification=centering,skip=0pt} 
		\caption{{(h) FCM:3}}
	\end{subfigure}
	\begin{subfigure}[t]{0.015\textwidth}
		\centering
		\includegraphics[width=0.91\linewidth]{jetcolorbar.pdf}
		\captionsetup{labelformat=empty,width=1\linewidth,justification=centering,skip=0pt} 
	\end{subfigure}
	\setcounter{subfigure}{8}
	\begin{subfigure}[t]{0.11\textwidth}
		\centering
		\includegraphics[width=1\linewidth,height=0.8\linewidth]{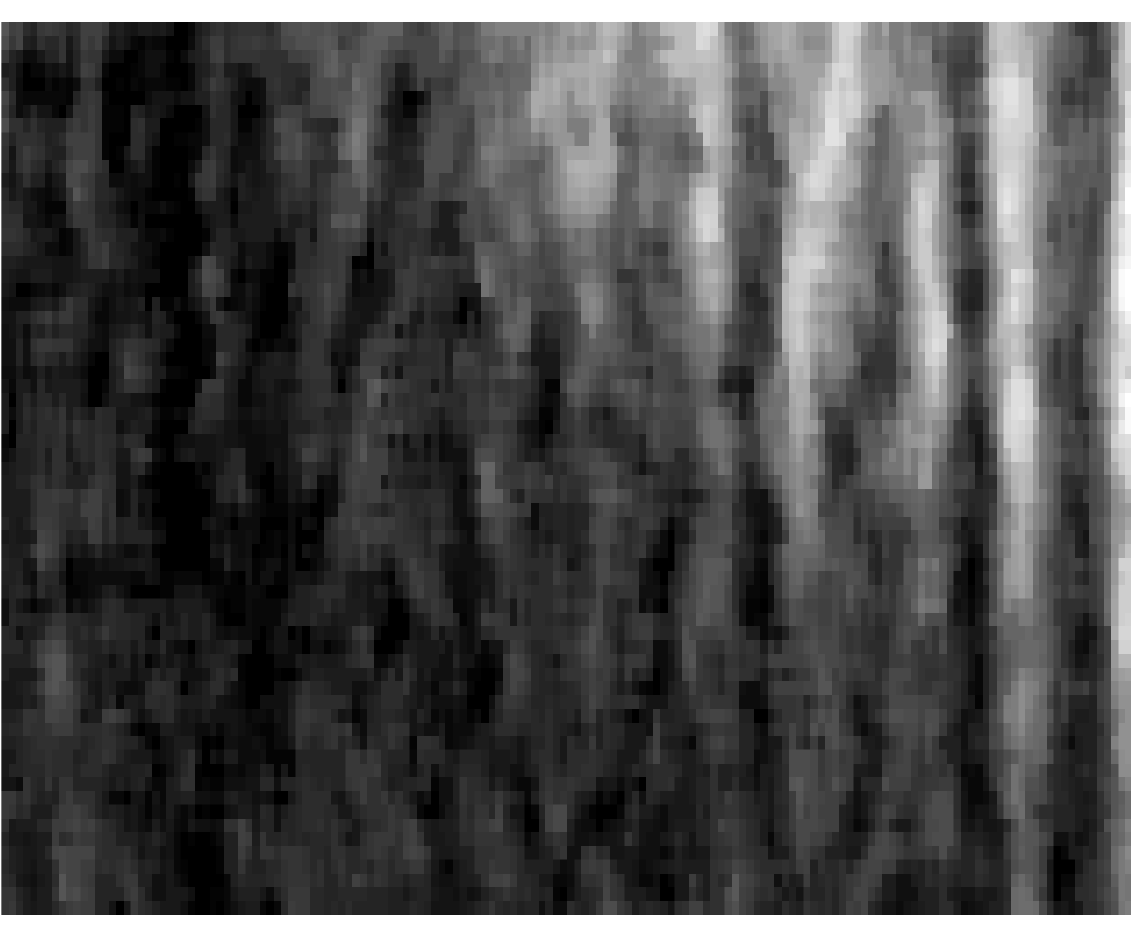}
		\captionsetup{labelformat=empty,skip=0pt}
		\caption{(i) Image 2} \label{fig:c2}
	\end{subfigure}
	\begin{subfigure}[t]{0.11\textwidth}
		\centering
		\includegraphics[width=1\linewidth,,height=0.8\linewidth]{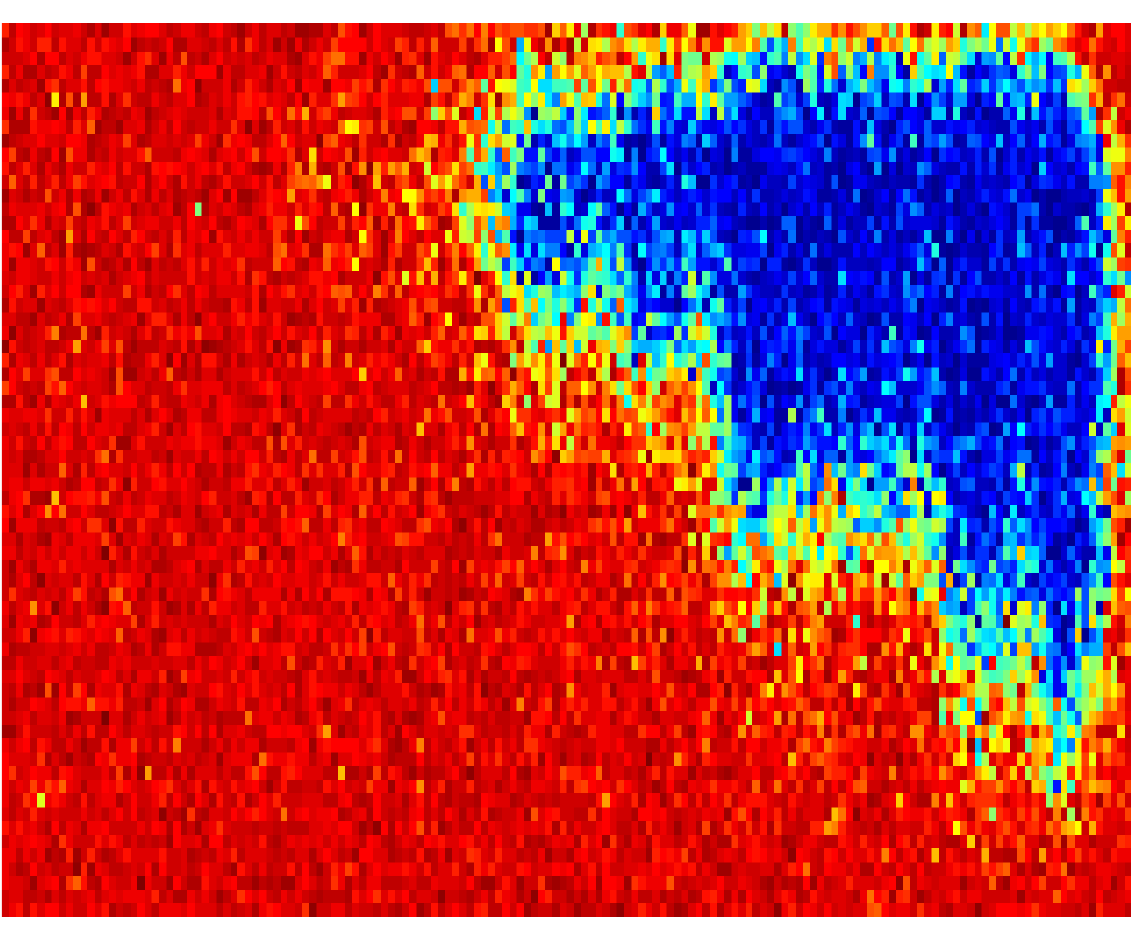} 
		\captionsetup{labelformat=empty,skip=0pt}
		\caption{(j) PM-LDA:1}
	\end{subfigure}
	\begin{subfigure}[t]{0.11\textwidth}
		\centering
		\includegraphics[width=1\linewidth,,height=0.8\linewidth]{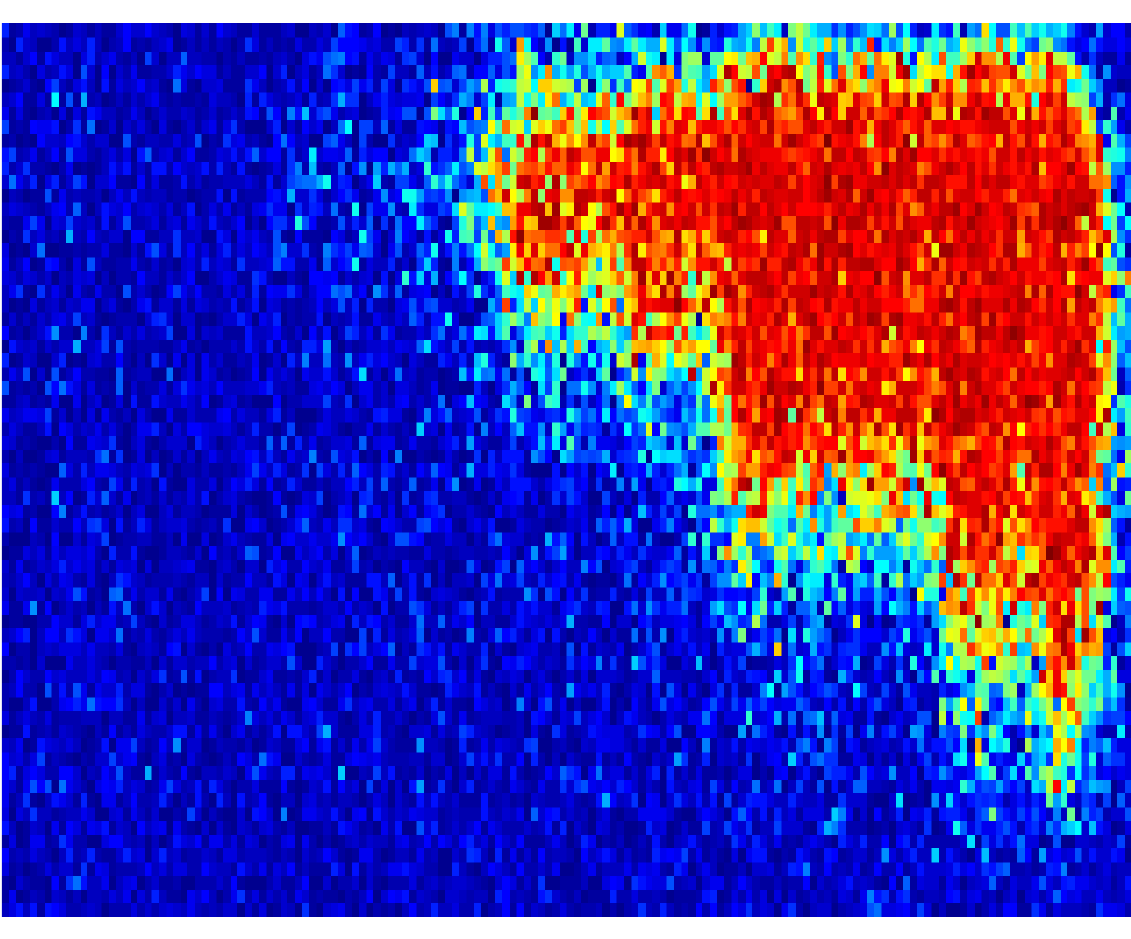}
		\captionsetup{labelformat=empty,skip=0pt}
		\caption{(k) PM-LDA:2} \label{fig:2c}
	\end{subfigure}
	\begin{subfigure}[t]{0.11\textwidth}
		\centering
		\includegraphics[width=1\linewidth,,height=0.8\linewidth]{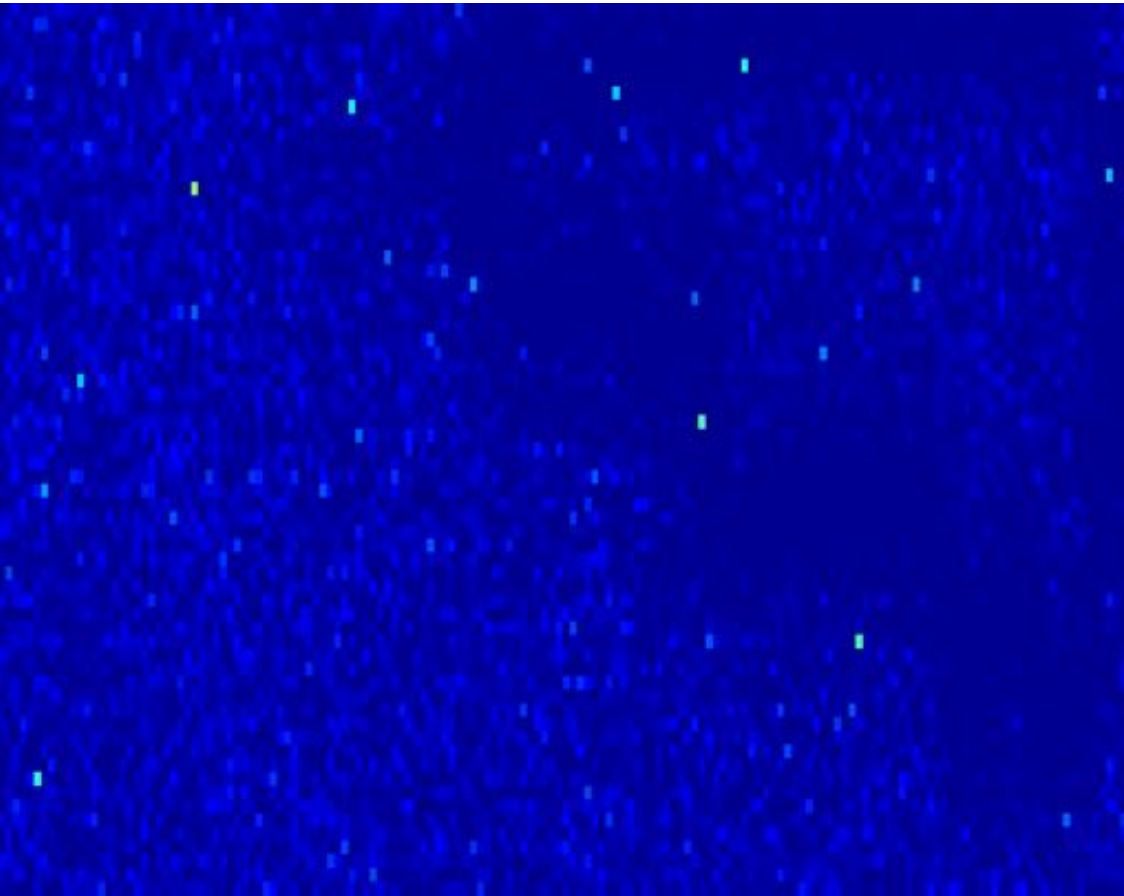}
		\captionsetup{labelformat=empty,skip=0pt}
		\caption{(l) PM-LDA:3}
	\end{subfigure}
	\begin{subfigure}[t]{0.015\textwidth}
		\centering
		\includegraphics[width=0.91\linewidth]{jetcolorbar.pdf}
		\captionsetup{labelformat=empty,skip=0pt}
		\caption{}
	\end{subfigure}
	\setcounter{subfigure}{12}
	\begin{subfigure}[t]{0.11\textwidth}
		\centering
		\includegraphics[width=1\linewidth,,height=0.8\linewidth]{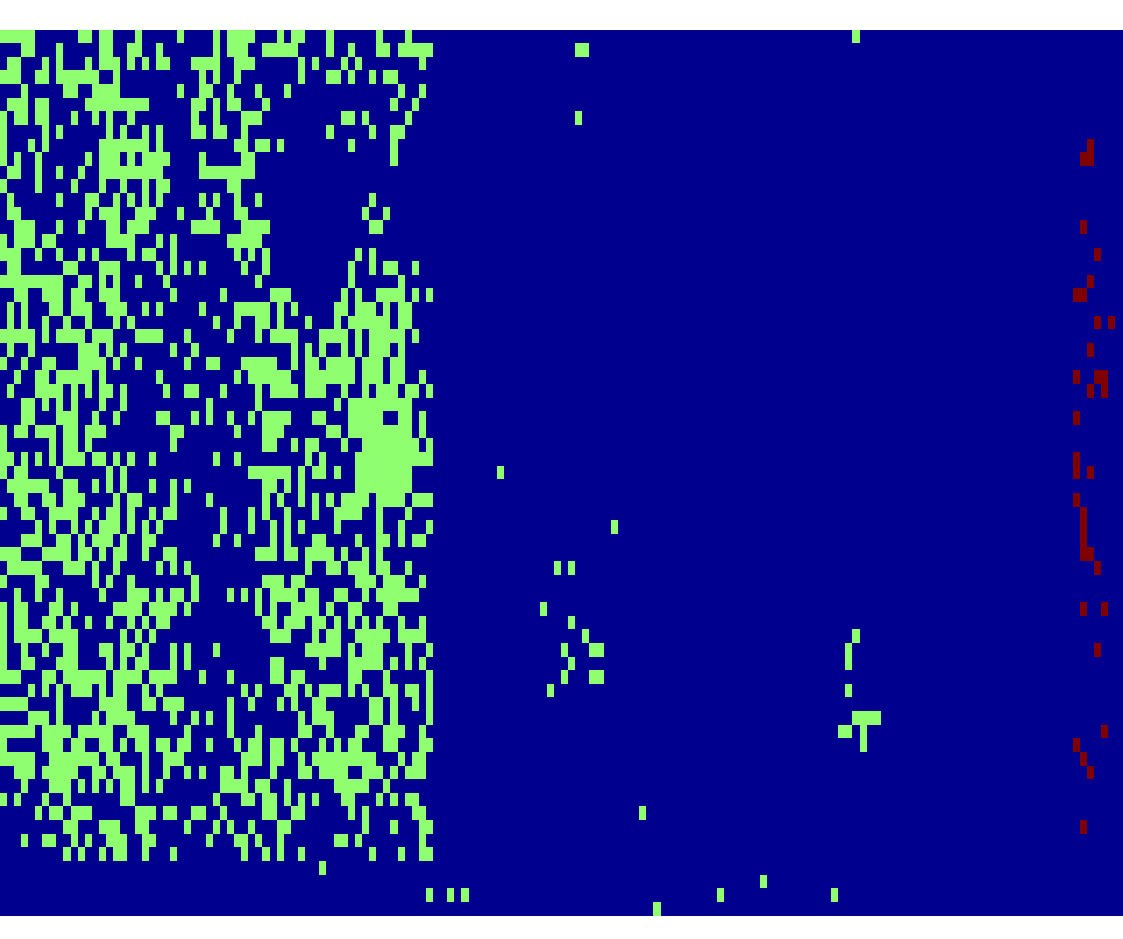}
		\captionsetup{labelformat=empty,skip=0pt}
		\caption{(m) LDA} \label{fig:2e}
	\end{subfigure} 
	\begin{subfigure}[t]{0.11\textwidth}
		\centering
		\includegraphics[width=1\linewidth,,height=0.8\linewidth]{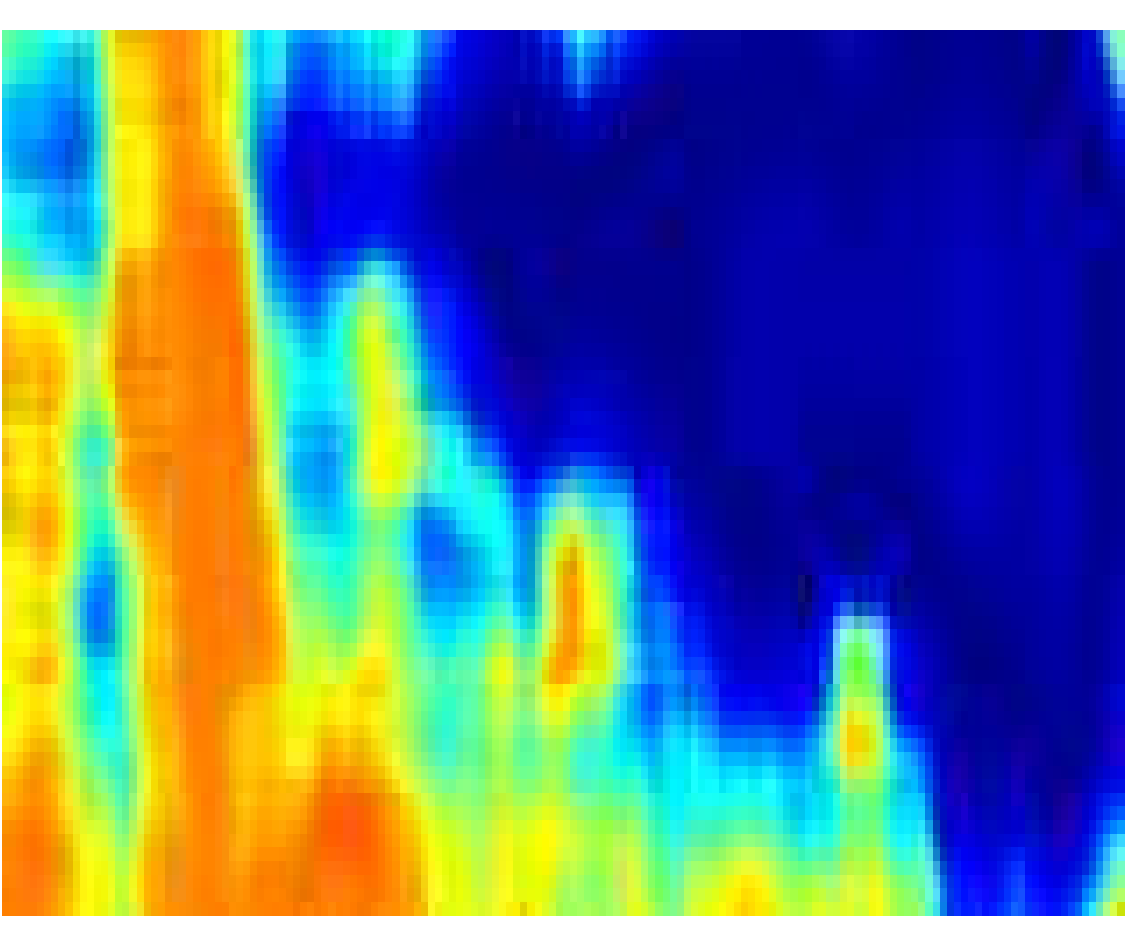}
		\captionsetup{labelformat=empty,skip=0pt}
		\caption{(n) FCM:1} 
	\end{subfigure}
	\begin{subfigure}[t]{0.11\textwidth}
		\centering
		\includegraphics[width=1\linewidth,,height=0.8\linewidth]{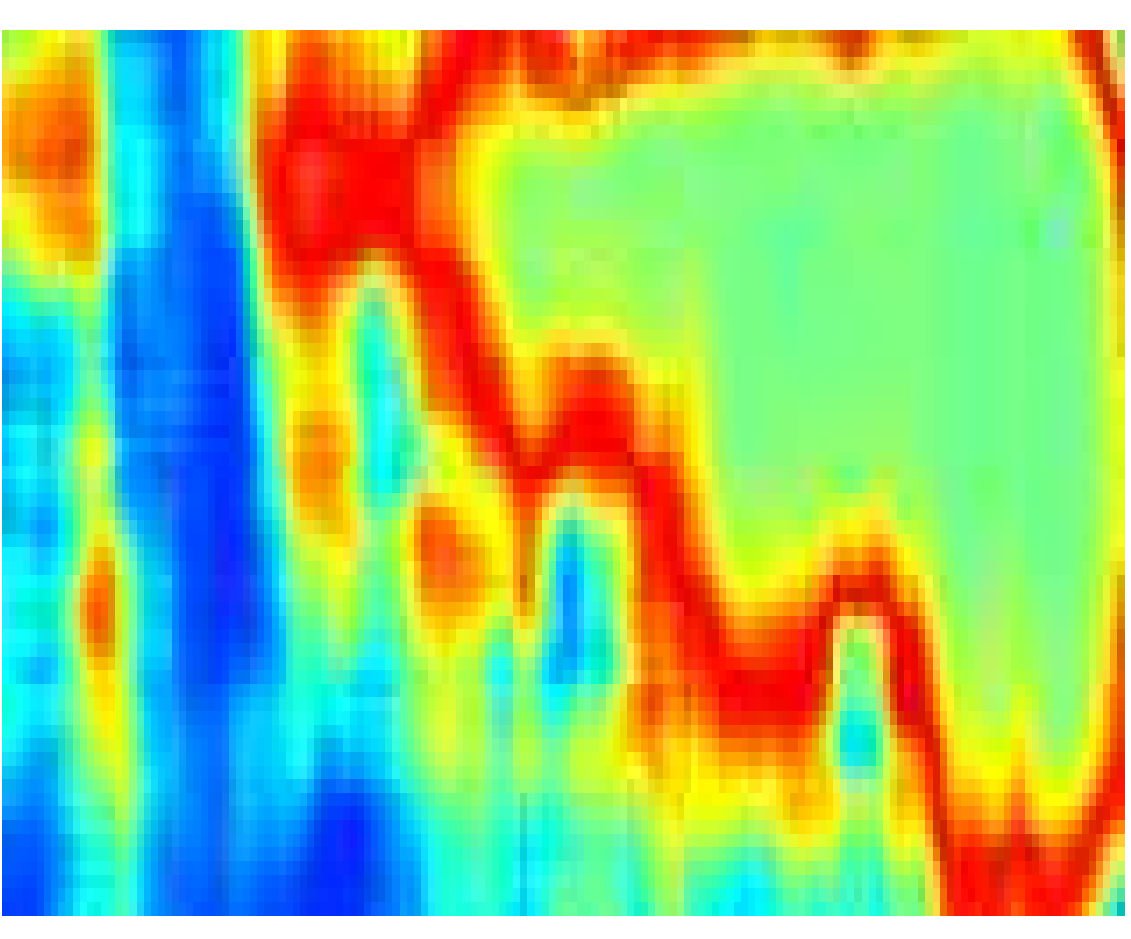}
		\captionsetup{labelformat=empty,skip=0pt}
		\caption{(o) FCM:2} \label{fig:2g}
	\end{subfigure}
	\begin{subfigure}[t]{0.11\textwidth}
		\centering
		\includegraphics[width=1\linewidth,,height=0.8\linewidth]{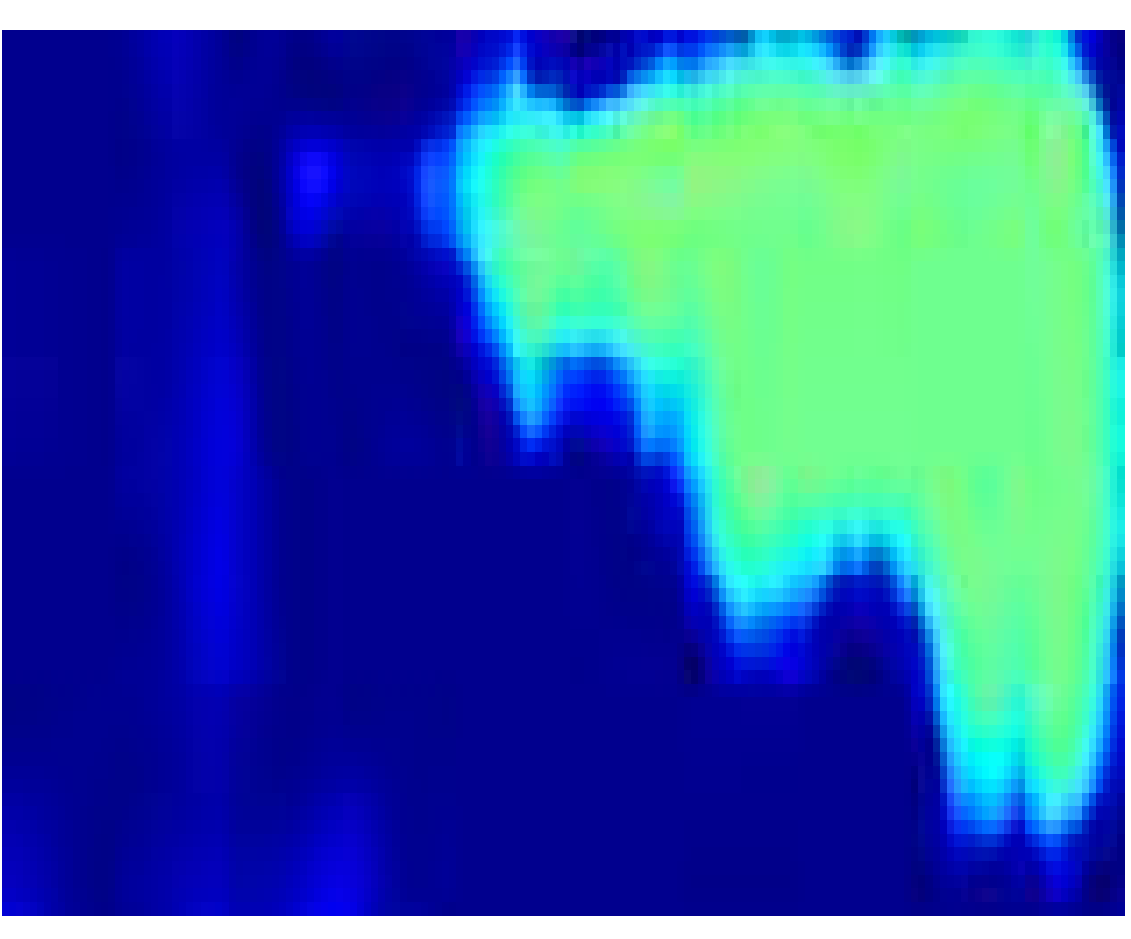}
		\captionsetup{labelformat=empty,skip=0pt}
		\caption{(p) FCM:3} \label{fig:2h}
	\end{subfigure}	
	\begin{subfigure}[t]{0.015\textwidth}
		\centering
		\includegraphics[width=0.91\linewidth]{jetcolorbar.pdf}
		\captionsetup{labelformat=empty,skip=0pt}
	\end{subfigure}
	\setcounter{subfigure}{16}
	\begin{subfigure}[t]{0.11\textwidth}
		\centering
		\includegraphics[width=1\linewidth,height=0.8\linewidth]{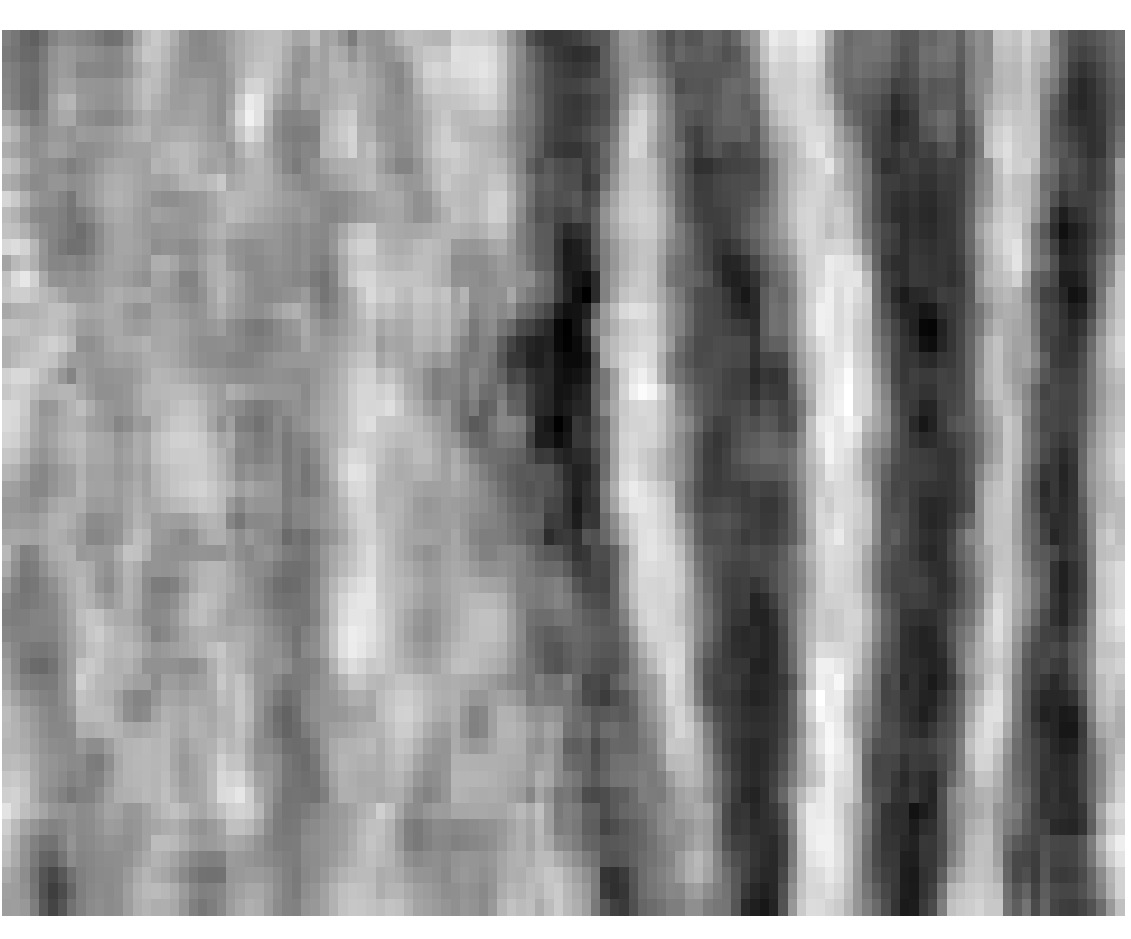}
		\captionsetup{labelformat=empty,skip=0pt}
		\caption{(q) Image 3} \label{fig:c3}
	\end{subfigure}
	\begin{subfigure}[t]{0.11\textwidth}
		\centering
		\includegraphics[width=1\linewidth,height=0.8\linewidth]{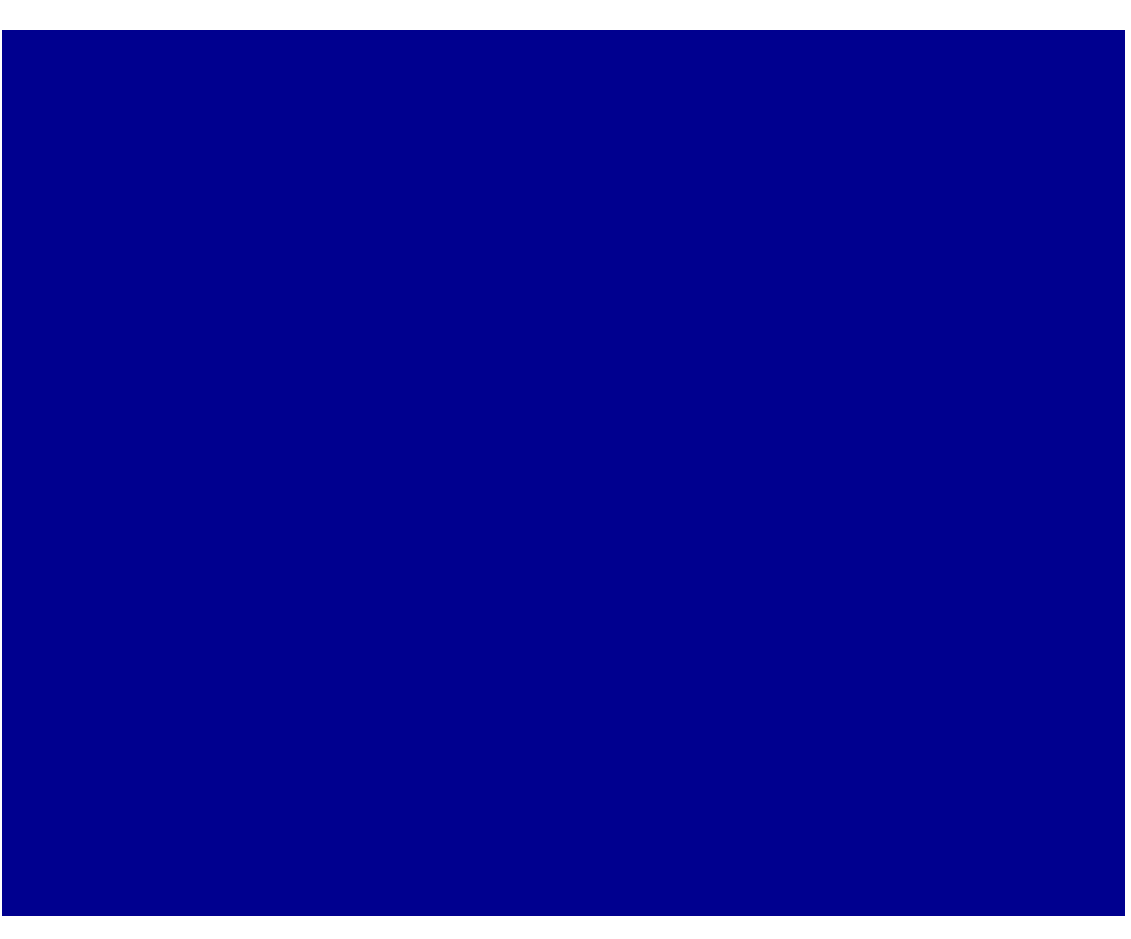}
		\captionsetup{labelformat=empty,skip=0pt}
		\caption{(r) PM-LDA:1}
	\end{subfigure}
	\begin{subfigure}[t]{0.11\textwidth}
		\centering
		\includegraphics[width=1\linewidth,height=0.8\linewidth]{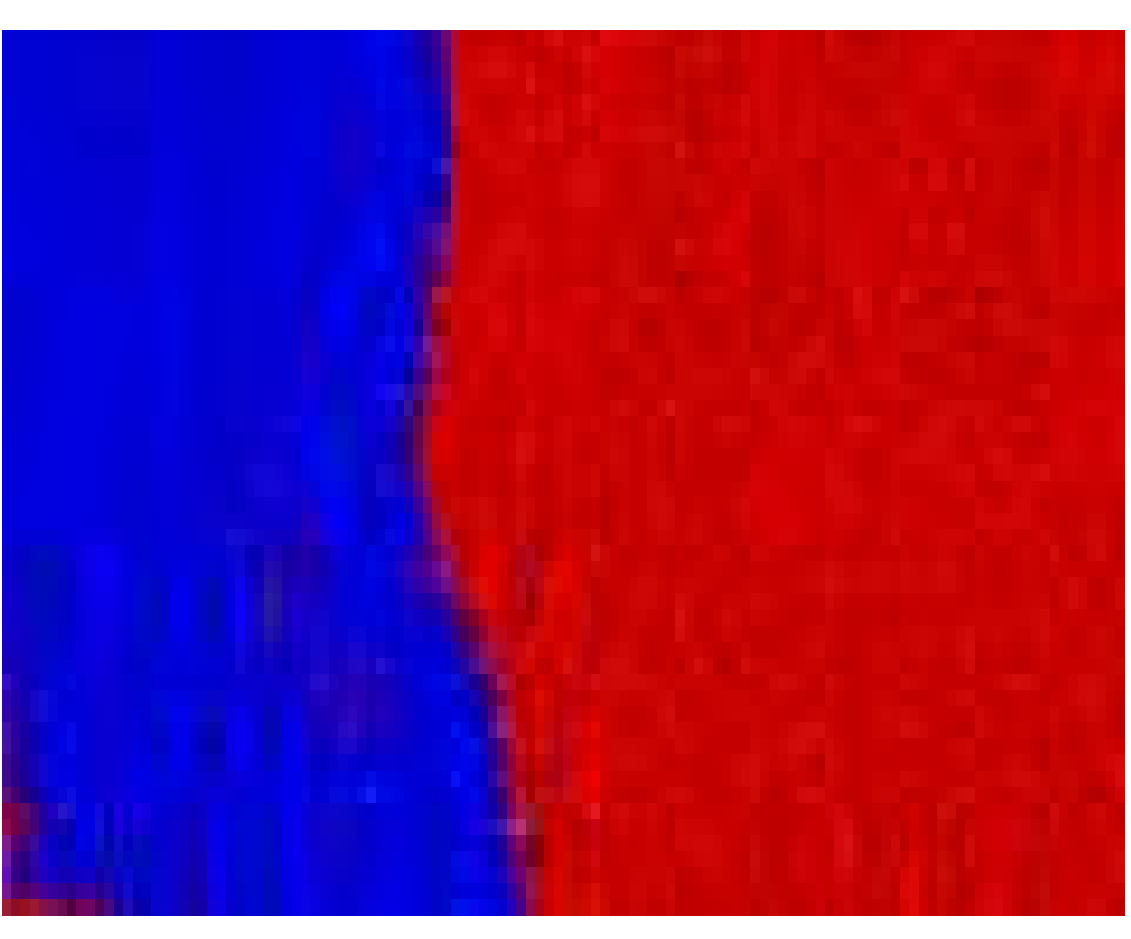}
		\captionsetup{labelformat=empty,skip=0pt}
		\caption{(s) PM-LDA:2}
	\end{subfigure}
	\begin{subfigure}[t]{0.11\textwidth}
		\includegraphics[width=1\linewidth,height=0.8\linewidth]{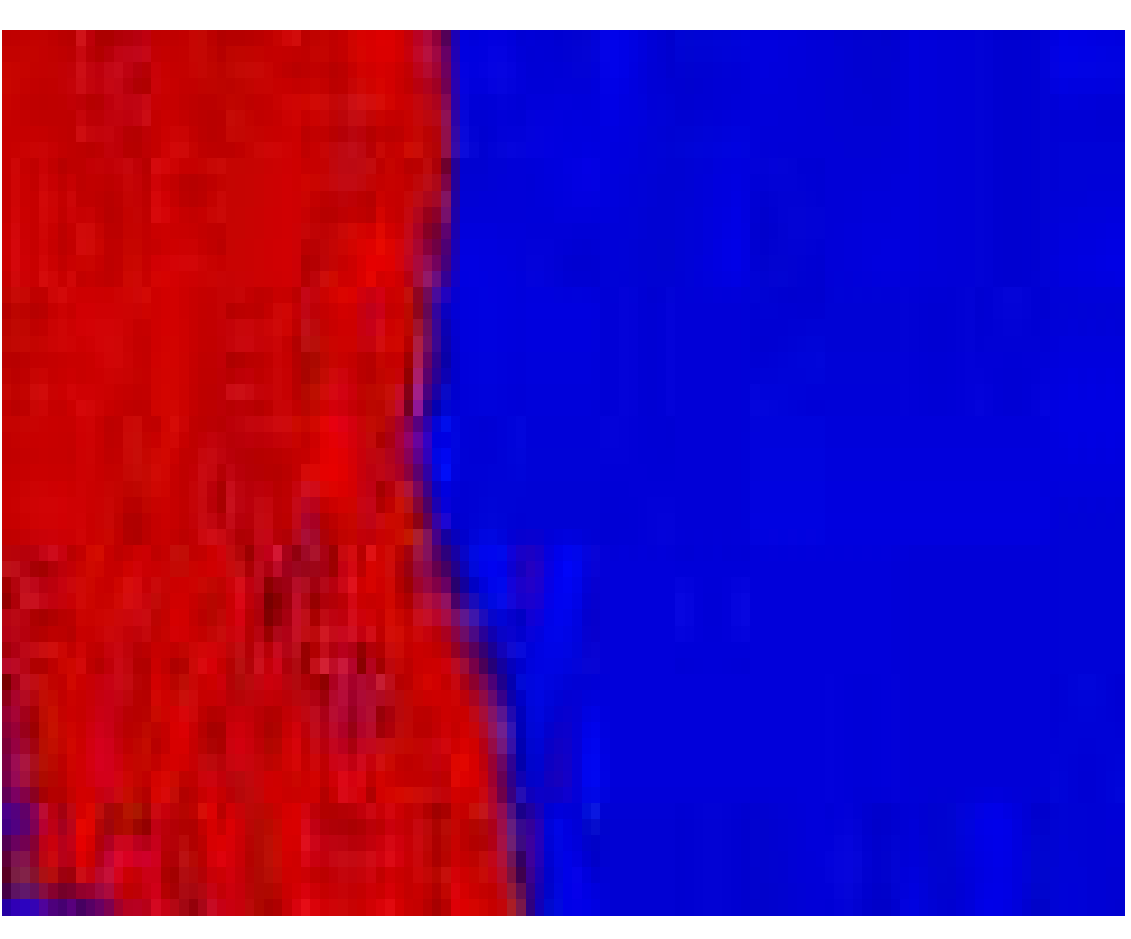}
		\centering
		\captionsetup{labelformat=empty,skip=0pt}
		\caption{(t) PM-LDA:3}\label{fig:3d}
	\end{subfigure}
	\begin{subfigure}[t]{0.015\textwidth}
		\centering
		\includegraphics[width=0.91\linewidth]{jetcolorbar.pdf}
		\captionsetup{labelformat=empty,skip=0pt}
	\end{subfigure}
	\setcounter{subfigure}{20}
	\begin{subfigure}[t]{0.11\textwidth}
		\centering
		\includegraphics[width=1\linewidth,height=0.8\linewidth]{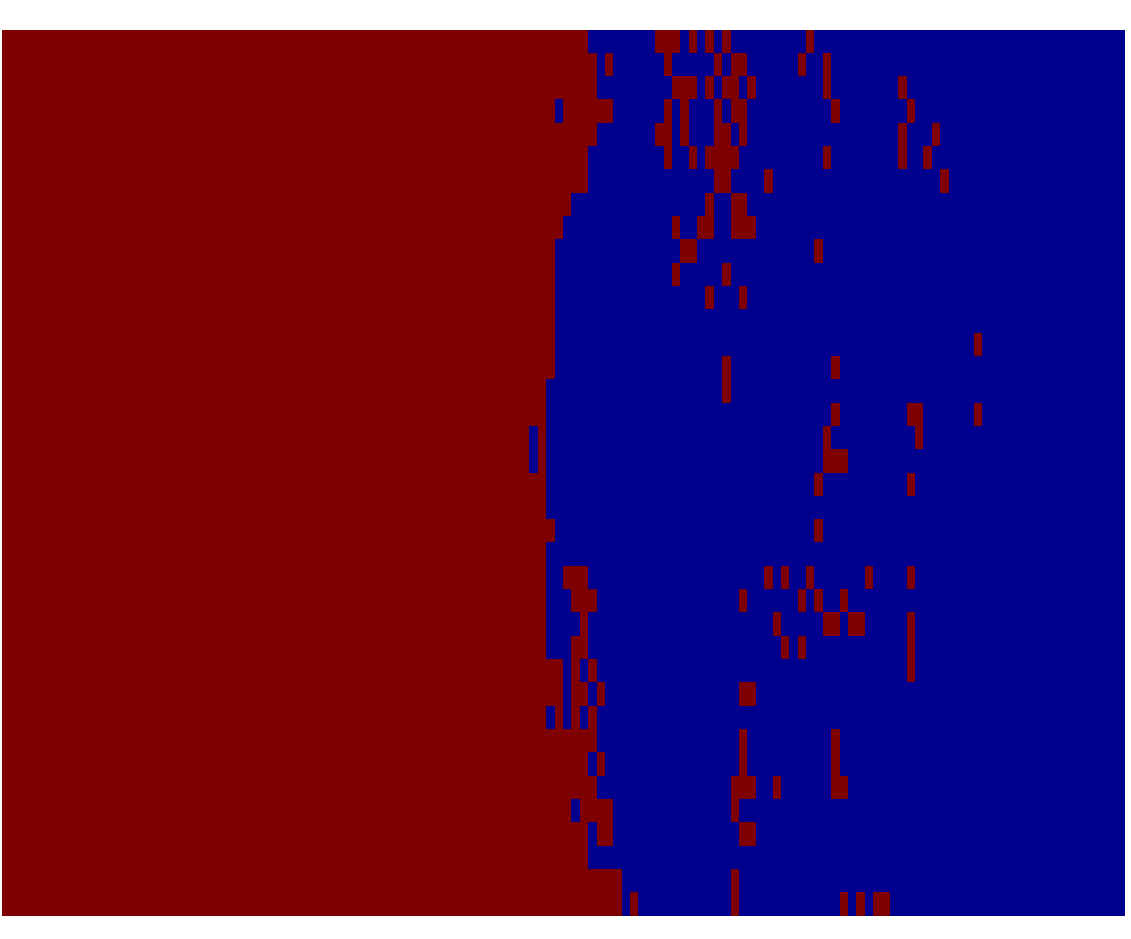}
		\captionsetup{labelformat=empty,skip=0pt}
		\caption{(u) LDA} \label{fig:3e}
	\end{subfigure} 
	\begin{subfigure}[t]{0.11\textwidth}
		\centering
		\includegraphics[width=1\linewidth,height=0.8\linewidth]{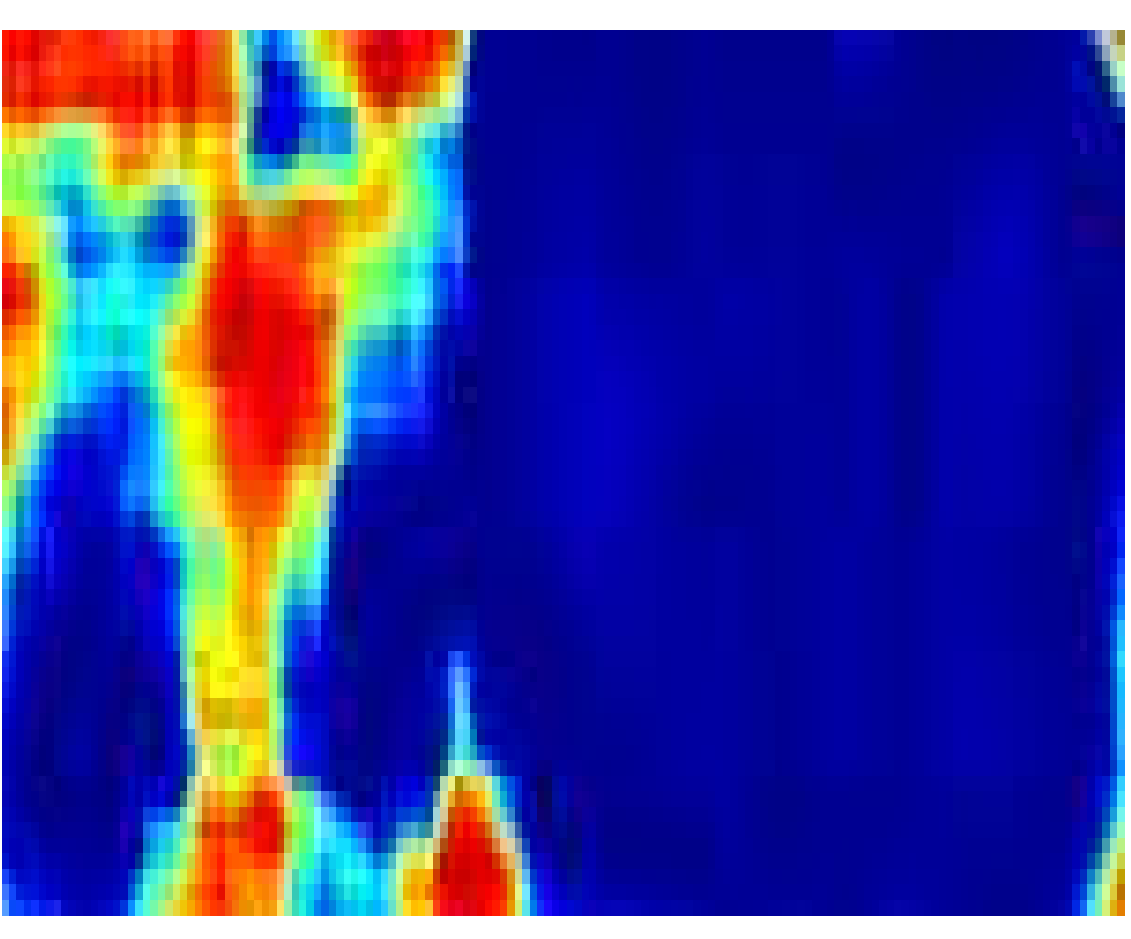}
		\captionsetup{labelformat=empty,skip=0pt}
		\caption{(v) FCM:1}
	\end{subfigure}
	\begin{subfigure}[t]{0.11\textwidth}
		\centering
		\includegraphics[width=1\linewidth,height=0.8\linewidth]{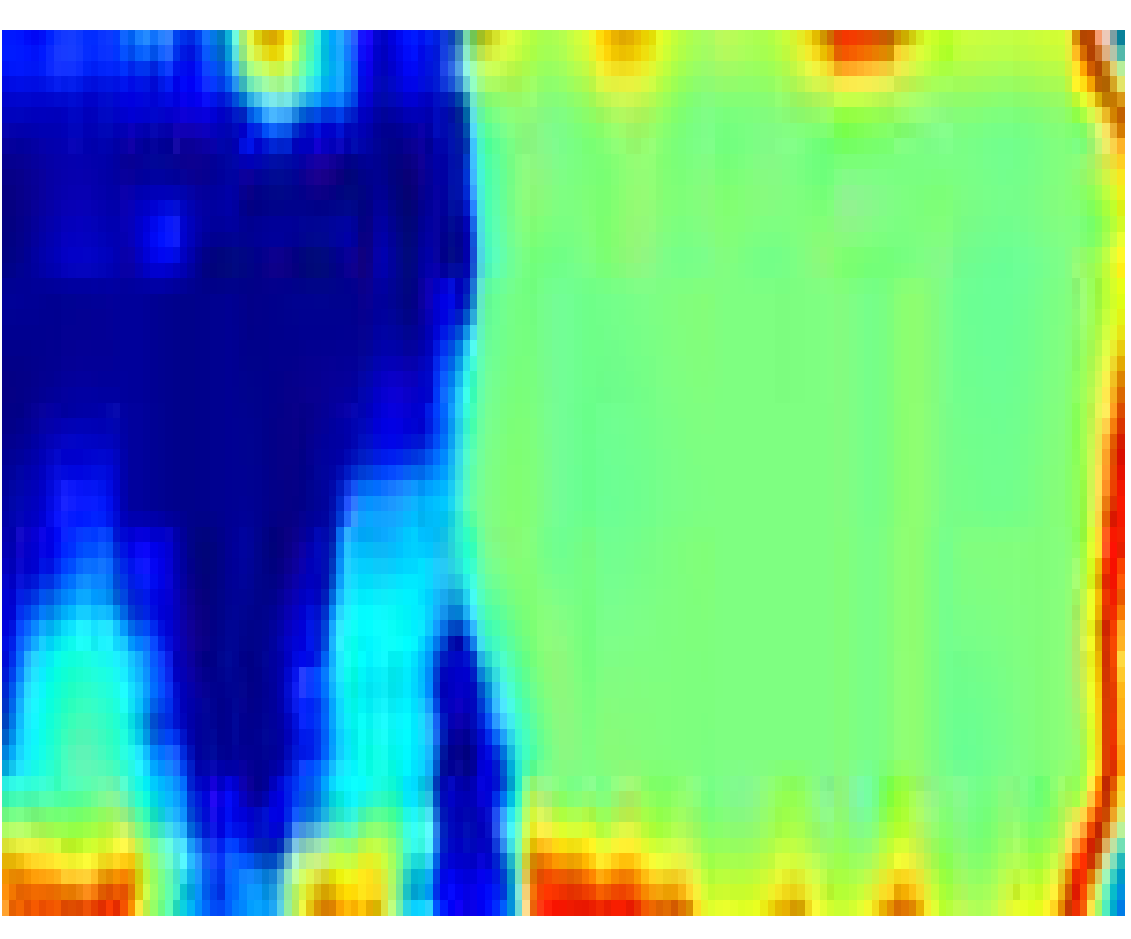}
		\captionsetup{labelformat=empty,skip=0pt}
		\caption{(w) FCM:2} \label{fig:3g}
	\end{subfigure}
	\begin{subfigure}[t]{0.11\textwidth}
		\centering
		\includegraphics[width=1\linewidth,height=0.81\linewidth]{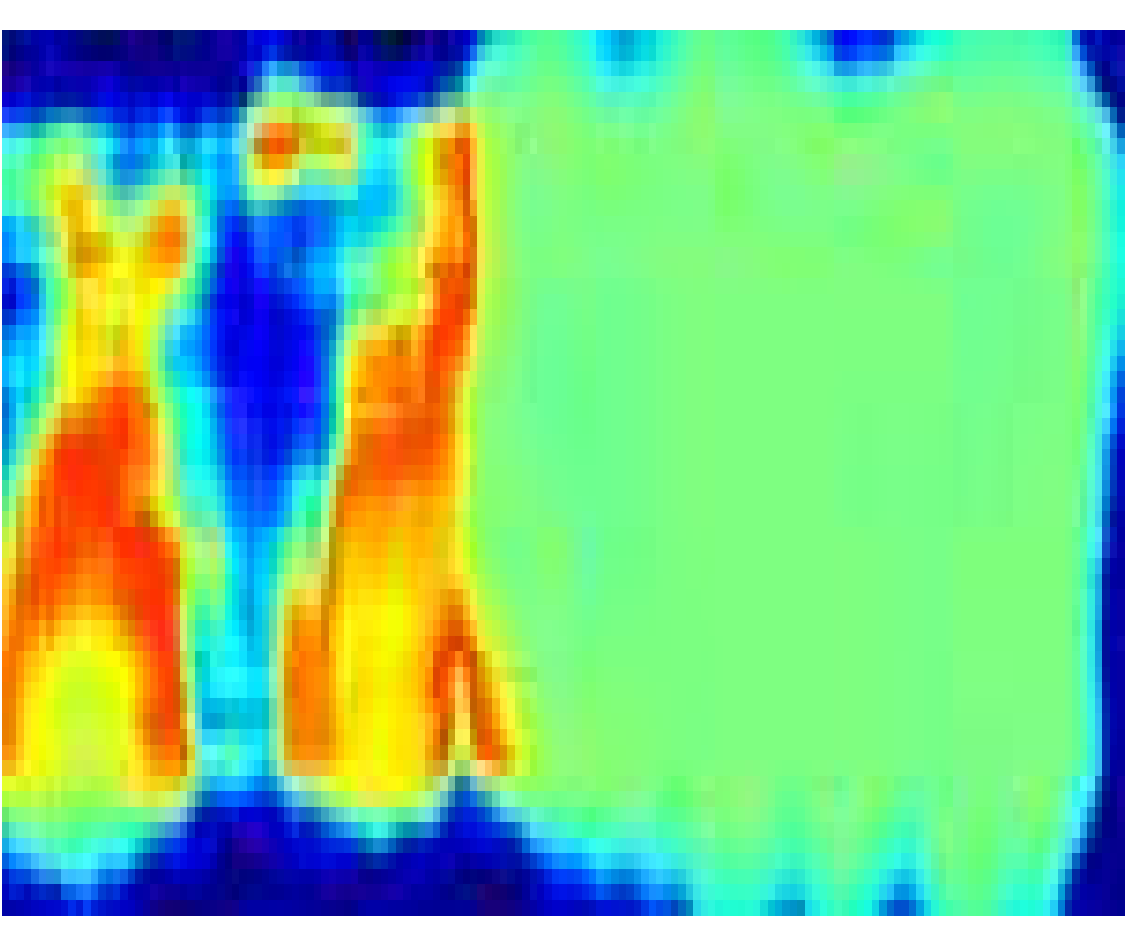}
		\captionsetup{labelformat=empty,skip=0pt}
		\caption{(x) FCM:3}	\label{fig:3h}
	\end{subfigure}	
	\begin{subfigure}[t]{0.015\textwidth}
		\centering
		\includegraphics[width=0.91\linewidth]{jetcolorbar.pdf}
		\captionsetup{labelformat=empty,skip=0pt}
	\end{subfigure}
	\setcounter{subfigure}{24}
	
	\begin{subfigure}[t]{0.11\textwidth}
		\centering
		\includegraphics[width=1\linewidth,height=0.8\linewidth]{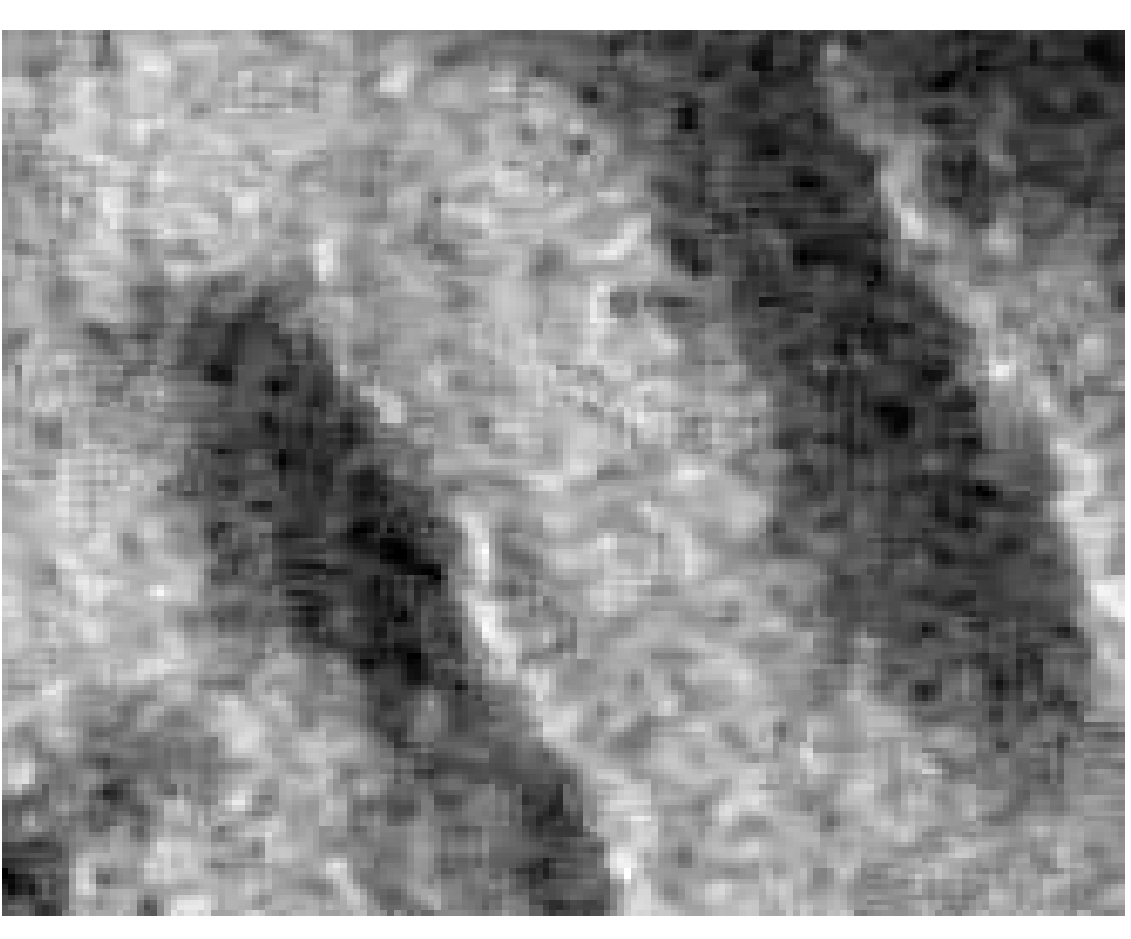}
		\captionsetup{labelformat=empty,skip=0pt}
		\caption{(y) Image 4} \label{fig:c4}
	\end{subfigure}
	\begin{subfigure}[t]{0.11\textwidth}
		\centering
		\includegraphics[width=1\linewidth,height=0.8\linewidth]{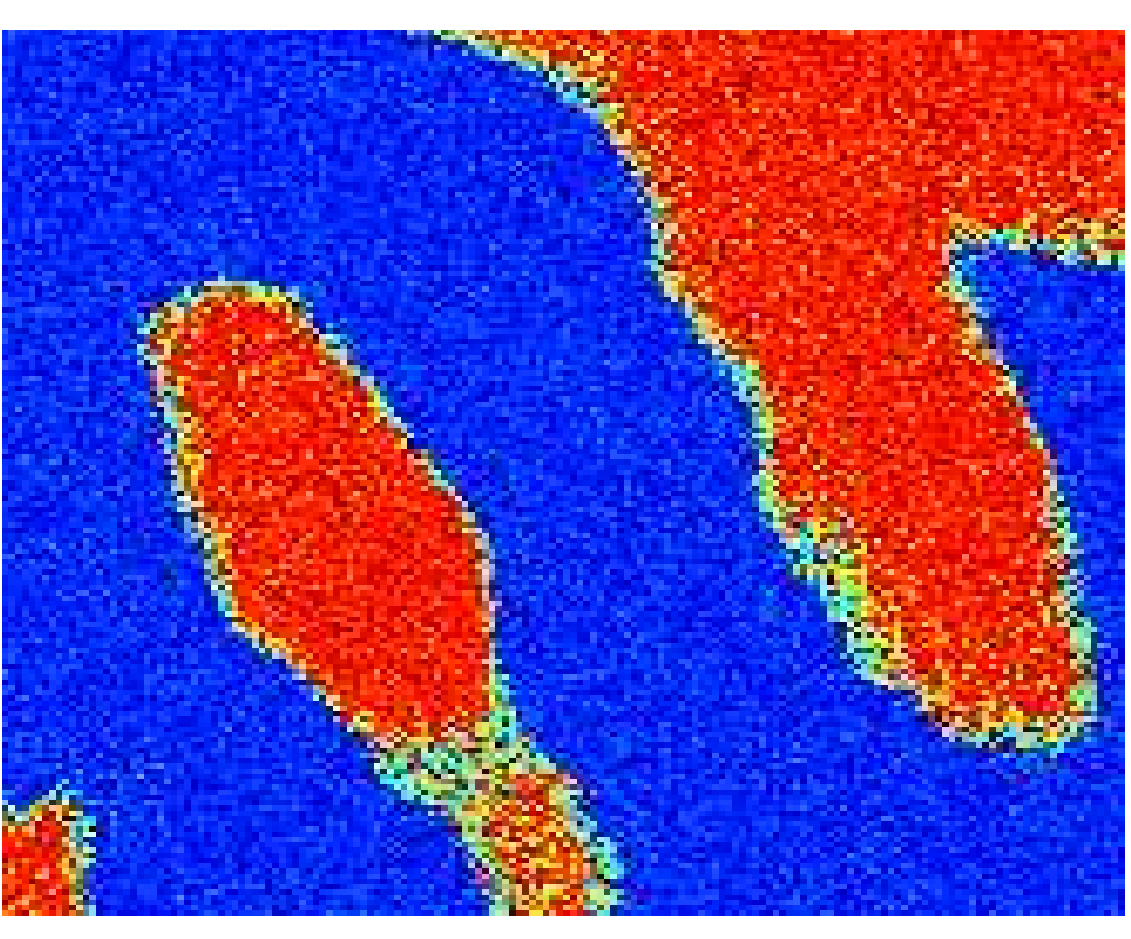}
		\captionsetup{labelformat=empty,skip=0pt}
		\caption{(z) PM-LDA:1}
	\end{subfigure}
	\begin{subfigure}[t]{0.11\textwidth}
		\centering
		\includegraphics[width=1\linewidth,height=0.8\linewidth]{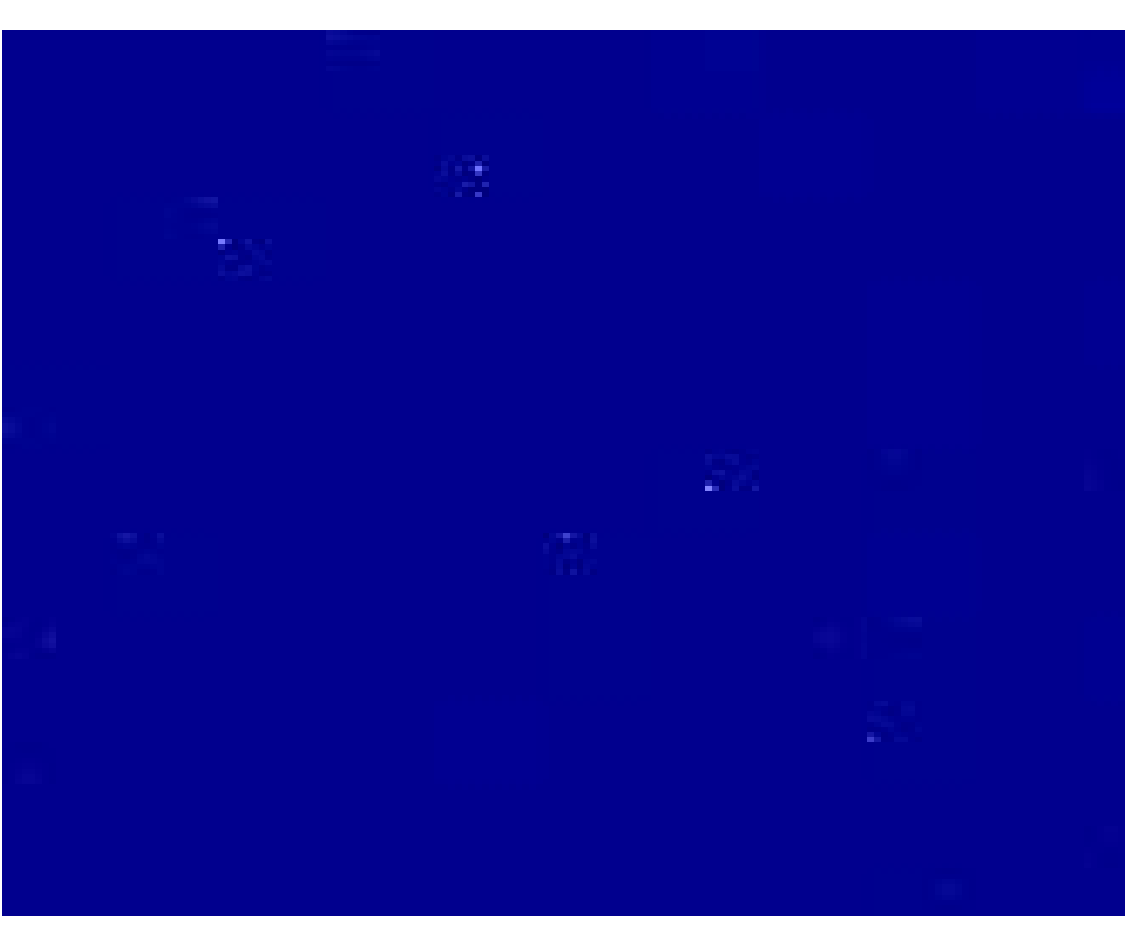}
		\captionsetup{labelformat=empty,skip=0pt}
		\caption{(aa) PM-LDA:2}
	\end{subfigure}
	\begin{subfigure}[t]{0.11\textwidth}
		\centering
		\includegraphics[width=1\linewidth,height=0.8\linewidth]{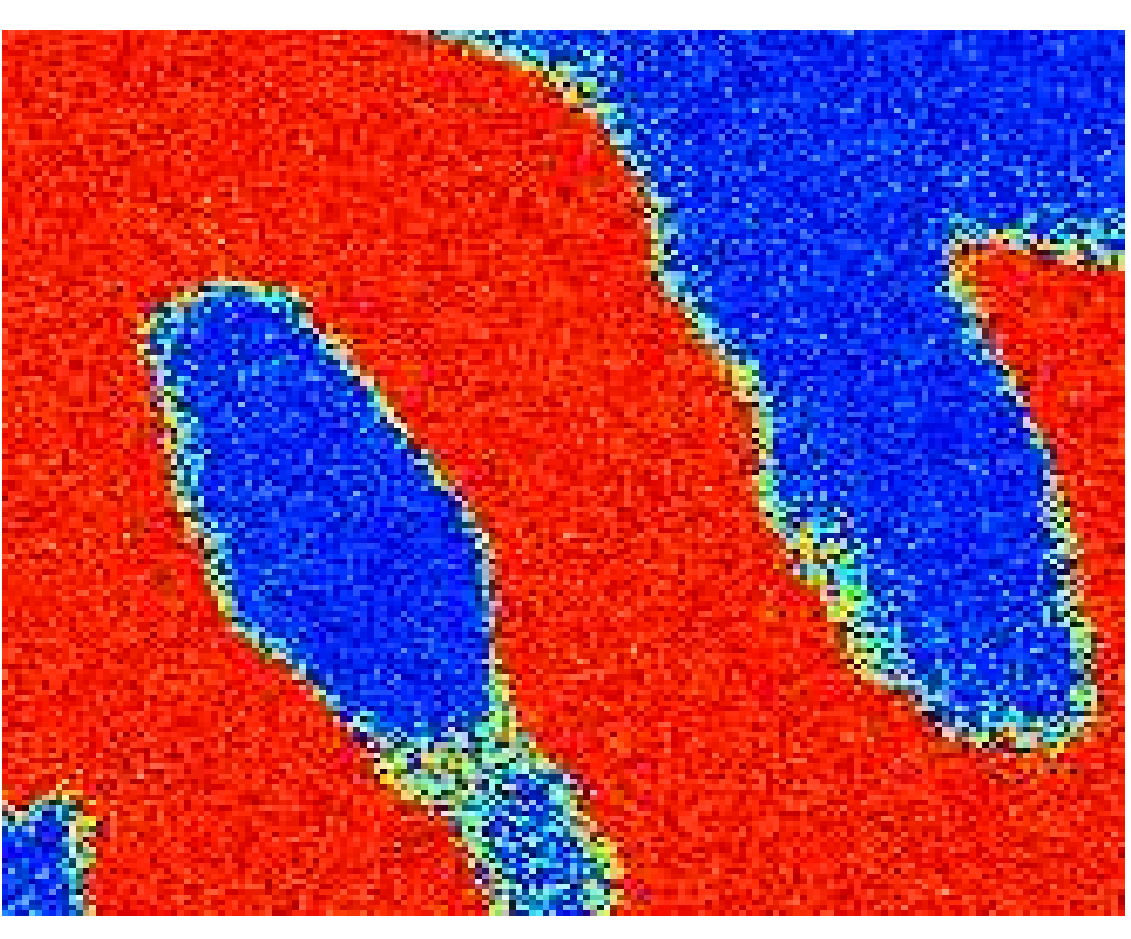}
		\captionsetup{labelformat=empty,skip=0pt}
		\caption{(ab) PM-LDA:3} \label{fig:4d}
	\end{subfigure}
	\begin{subfigure}[t]{0.015\textwidth}
		\centering
		\includegraphics[width=0.91\linewidth]{jetcolorbar.pdf}
		\captionsetup{labelformat=empty,skip=0pt}
	\end{subfigure}
	\setcounter{subfigure}{28}
	\begin{subfigure}[t]{0.11\textwidth}
		\centering
		\includegraphics[width=1\linewidth,height=0.8\linewidth]{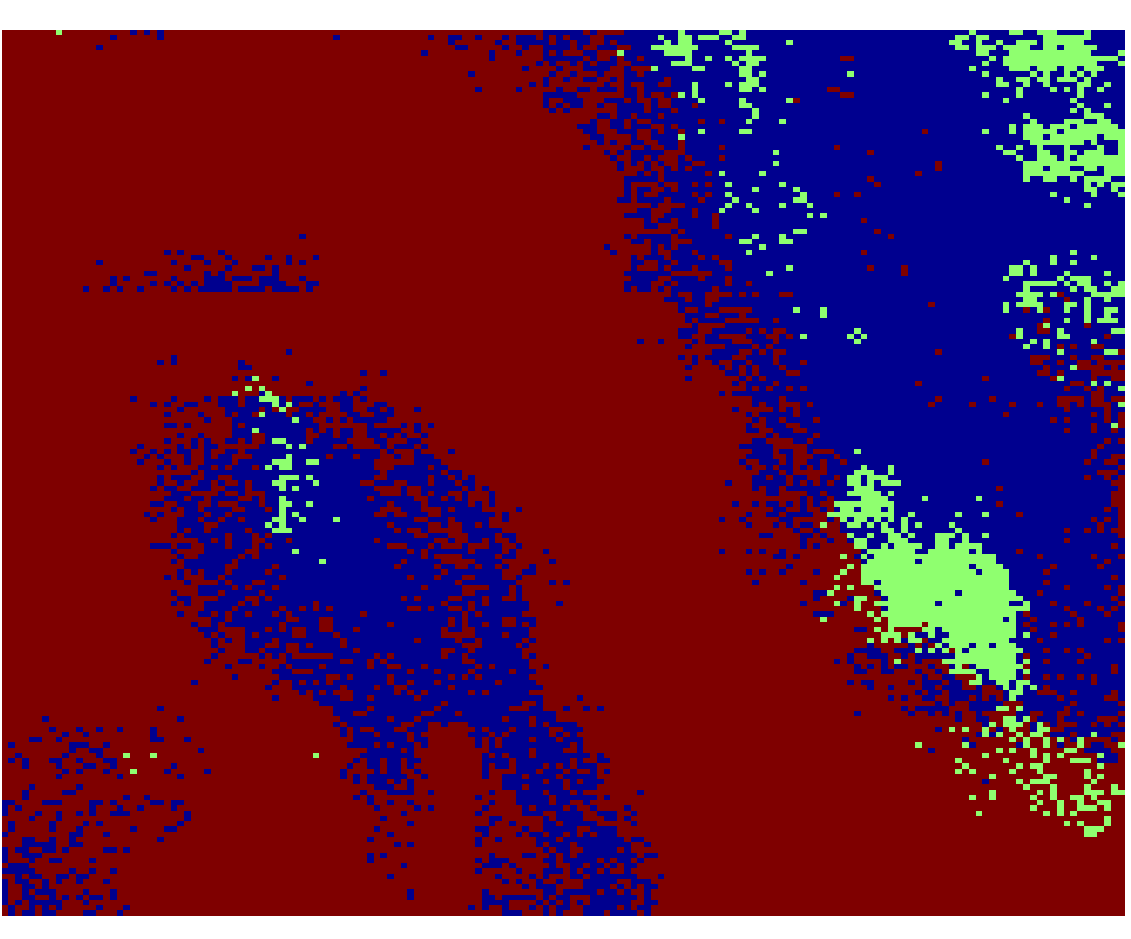}
		\captionsetup{labelformat=empty,skip=0pt}
		\caption{(ac) LDA}\label{fig:4e}
	\end{subfigure} 
	\begin{subfigure}[t]{0.11\textwidth}
		\centering
		\includegraphics[width=1\linewidth,height=0.8\linewidth]{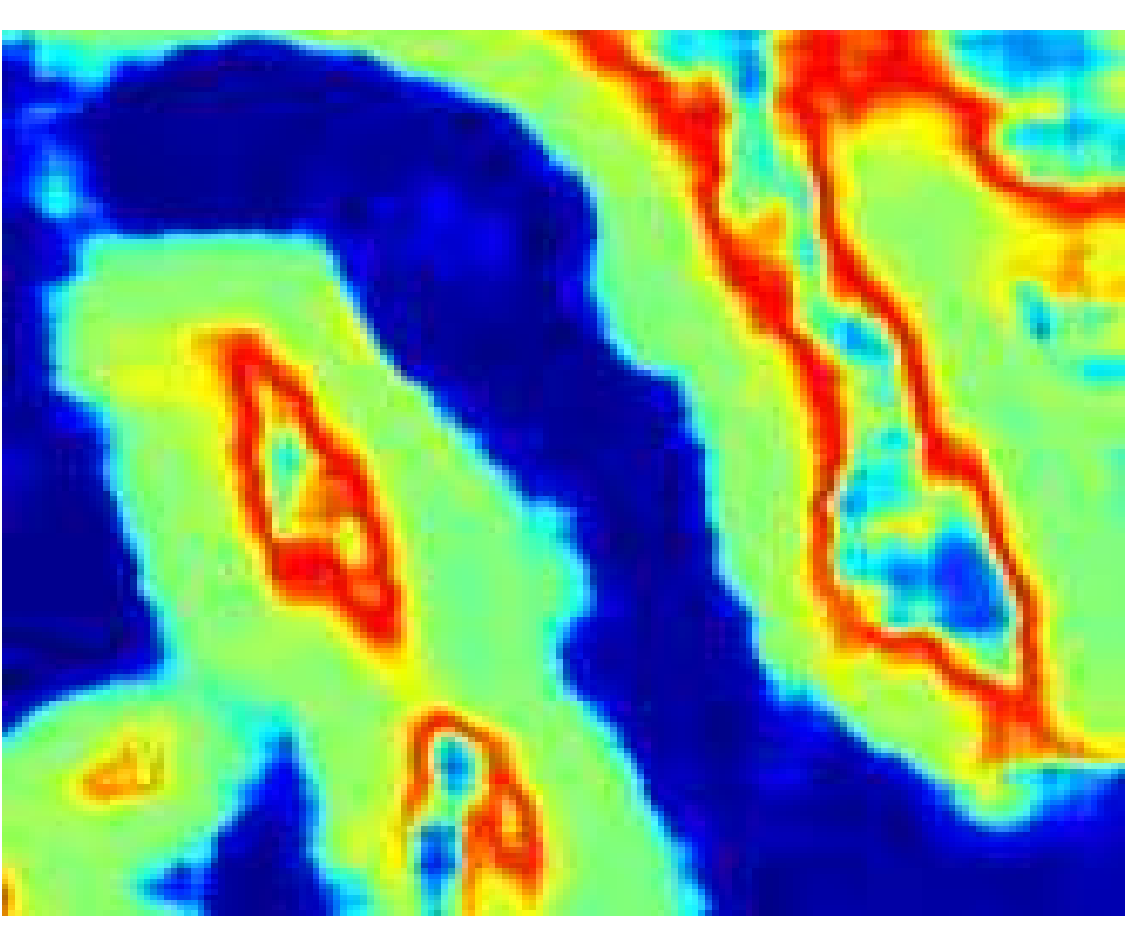}
		\captionsetup{labelformat=empty,skip=0pt}
		\caption{(ad) FCM:1}
	\end{subfigure}
	\begin{subfigure}[t]{0.11\textwidth}
		\centering
		\includegraphics[width=1\linewidth,height=0.8\linewidth]{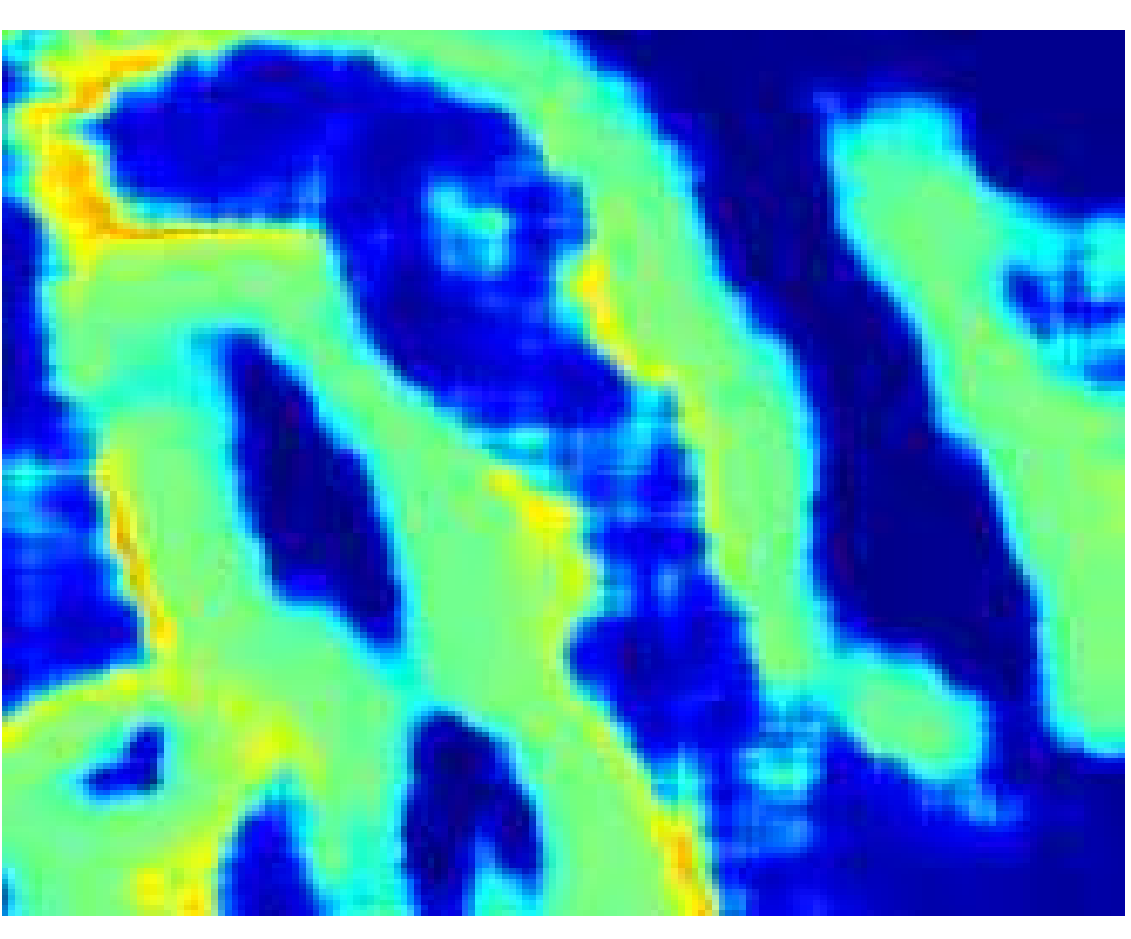}
		\captionsetup{labelformat=empty,skip=0pt}
		\caption{(ae) FCM:2}
	\end{subfigure}
	\begin{subfigure}[t]{0.11\textwidth}
		\centering
		\includegraphics[width=1\linewidth,height=0.8\linewidth]{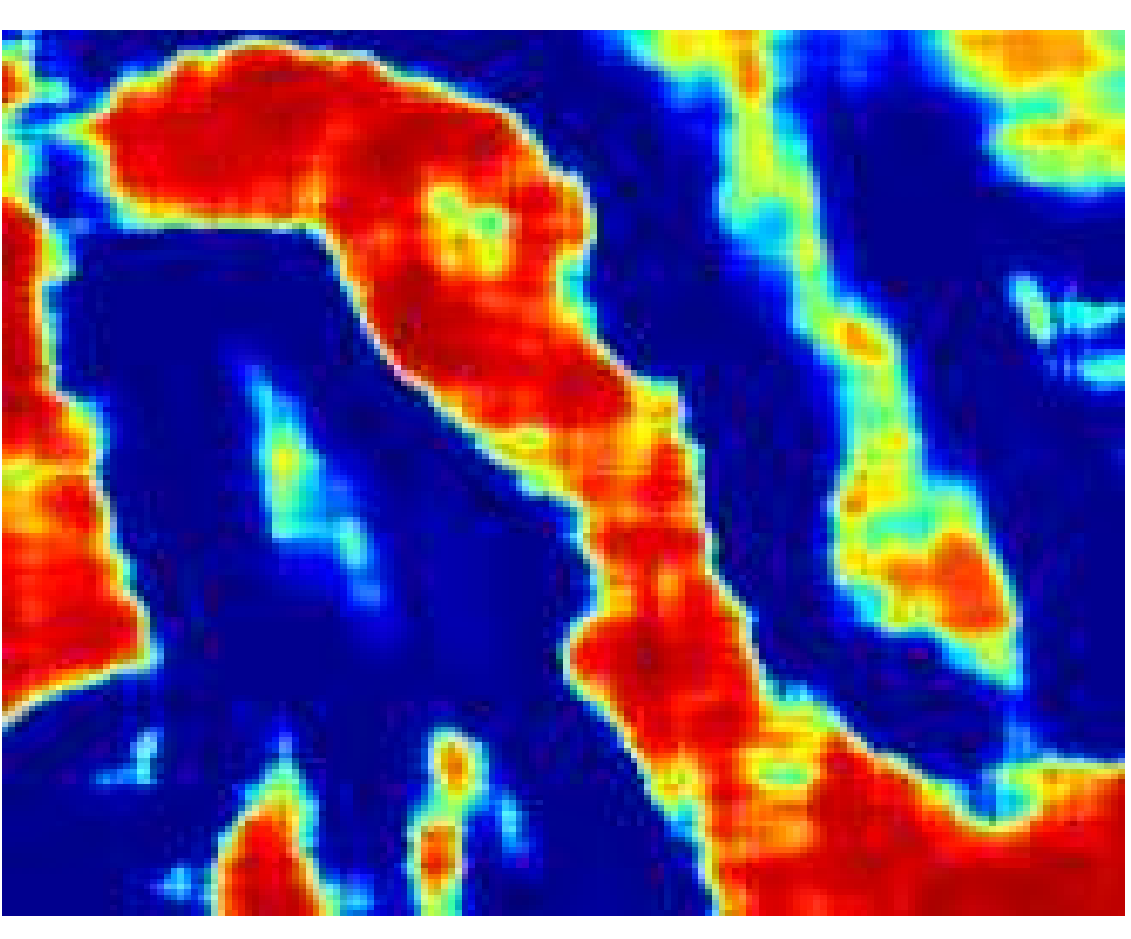}
		\captionsetup{labelformat=empty,skip=0pt}
		\caption{(af) FCM:3}
	\end{subfigure}		
	\begin{subfigure}[t]{0.015\textwidth}
		\centering
		\includegraphics[width=0.91\linewidth]{jetcolorbar.pdf}
		\captionsetup{labelformat=empty,skip=0pt}
	\end{subfigure}
	
	\caption{Segmentation results of Image 1 - 4 using PM-LDA, FCM and LDA. (a): SAS Image. (b)-(d): PM-LDA partial membership map in the ``dark flat sand,'' ``sand ripple,'' ``bright flat sand'' topics, respectively. (e): LDA result where color indicates topic label. (f)-(h): FCM  partial membership map in the first, second, and third cluster, respectively. Subfigure captions in Row 2 - 4 follow those in Row 1. In PM-LDA and FCM results,  color indicates the degree of membership of a visual word in a topic or cluster. }
	\label{fig:all}
\end{figure*}

Our first experiment considers segmentation of a subset of four sub-images from our SAS image database (shown in the first column in Fig. \ref{fig:all}).  For each image, we simply compute the average intensity value and entropy within a 21 $\times$ 21 window as feature values. The average intensity value is scaled ($\times 10$) to roughly the same magnitude of the average entropy value. Each image is divided into multiple documents using a sliding window approach. A document consists of all of the feature vectors associated with each pixel (\ie, visual words) in the window. The number of topics in this dataset is set to $3$. For LDA, a dictionary of size $100$ is built by clustering all the computed feature values using the K-means.  FCM results with $m=1.5$. Parameters for LDA and FCM were selected manually to provide the best results. Due to the lack of ground-truth, qualitative segmentation results in Fig. \ref{fig:all} is provided. In the first row, Subfigures (b), (c), and (d) show the partial membership maps in the ``dark flat sand" , ``sand ripple" , and  ``bright flat sand'' topics using PM-LDA, respectively.  Subfigures (f), (g), and (h) show the partial membership maps in each of the three clusters using FCM, respectively.  In (b) - (d) and (f) - (h), the color indicates the degree of membership of a visual word in a topic where red corresponds to a full membership of $1$ and dark blue color corresponds to a membership value of $0$. The LDA result is shown in (e) where color indicates topic assignment. Subfigures in Row 2-4 follow the same subfigure captions in Row 1.

From the experimental results,  we can see that PM-LDA achieves much better results than FCM and LDA. As shown in Fig. \ref{fig:1c} and \ref{fig:2c}, the segmentation results of PM-LDA show a gradual change from ``sand ripple" to ``dark flat sand".   FCM captures the gradual transition to some extent but is not able to clearly differentiate between clusters. For example, as shown in Fig. \ref{fig:2g} - \ref{fig:2h} and Fig. \ref{fig:3g} - \ref{fig:3h},  using FCM, the rippled region in Images 2 and 3 is assigned to 2 clusters with nearly equal partial memberships.  As LDA cannot generate partial memberships, in Fig. \ref{fig:1e} and \ref{fig:2e}, Image 1 and 2 are simply partitioned into different topics using LDA. Yet, by comparing Fig. \ref{fig:3e} with \ref{fig:3d} and Fig. \ref{fig:4e} with \ref{fig:4d}, we can see that on Image 3 and 4 that do not contain transition regions, LDA achieves similar segmentation result to PM-LDA. 

 \subsubsection{Complete SAS image segmentation and comparison to LDA and sLDA} Experiments are then extended to the complete SONAR images. Five complete high-frequency SAS (indicated as HF-00, HF-01,..., HF-04, are segmented into 200 superpixels \cite{cobb2014boundary} and each superpixel is considered as a document. In addition to the mean and entropy features used in the previous experiment, a ``sand ripple''filter response is also used as the third feature in this experiment. The filter is built based on the sand ripple characterization algorithm proposed in \cite{chen:2015}. It is hypothesized that each superpixel is a rippled region with certain ripple frequency, $f_\text{ripple}$ (number of ripples per meter). The sand ripple characterization algorithm is applied to each superpixel to estimate its ripple frequency. As the range resolution of the sonar imagery is $0.025m$, a complete ripple length $L$ can be computed as $40/f_\text{ripple}$. To capture the ripple repeating pattern, a filter is built as $[-\mathbf{1},\mathbf{0}, \mathbf{1},-\mathbf{1},\mathbf{0}, \mathbf{1}]$, where $\mathbf{1}$ and $\mathbf{0}$ are matrices with height of $11$ and width of $\left\lceil L/3\right\rceil$. The filter is applied to the corresponding superpixel and the filter response is used as the third feature for that superpixel.  Non-rippled superpixels have low filter responses and ripple superpixels have high filter response. 

\begin{figure*}
	\centering
		\begin{subfigure}[b]{0.15\textwidth}
			\includegraphics[width=1\linewidth,height=0.8\linewidth]{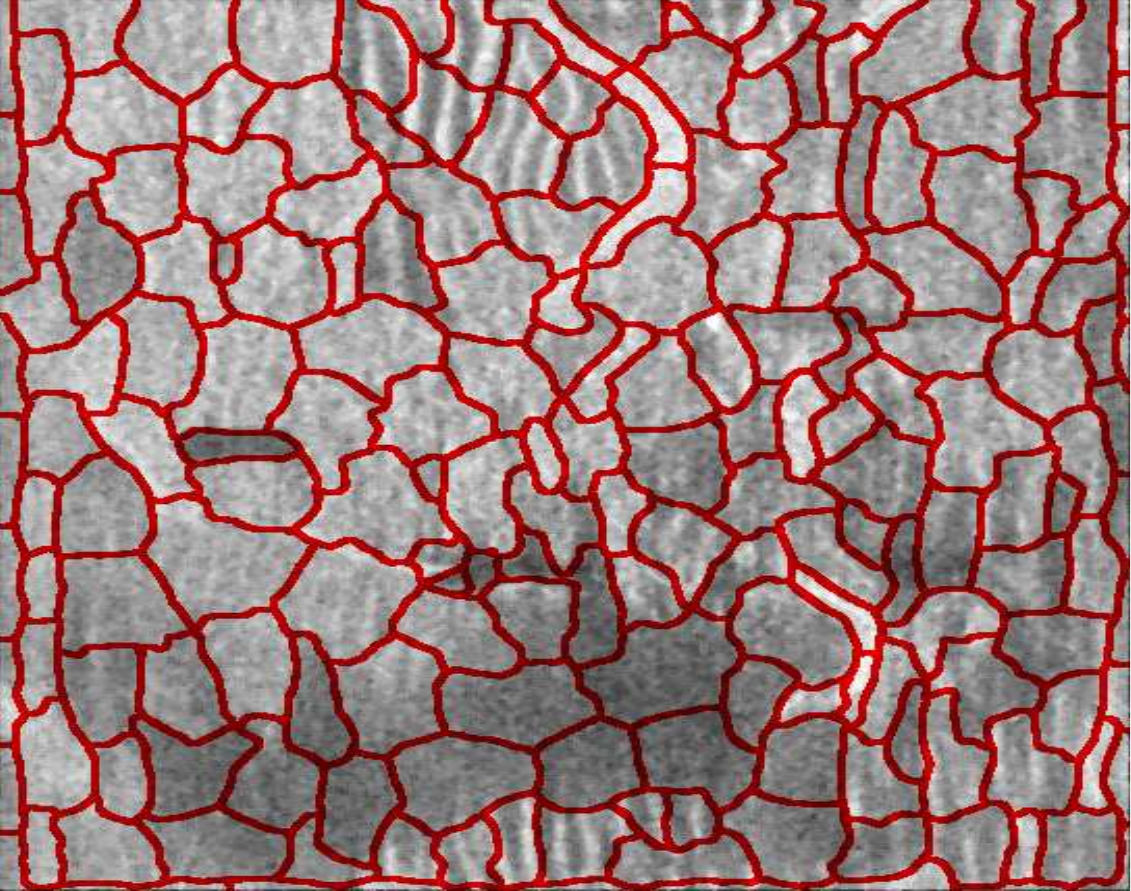}
			\captionsetup{justification=centering,labelformat=empty,skip=0pt}
			\caption{}
		\end{subfigure}
		\begin{subfigure}[b]{0.15\textwidth}
			\includegraphics[width=1\linewidth,height=0.8\linewidth]{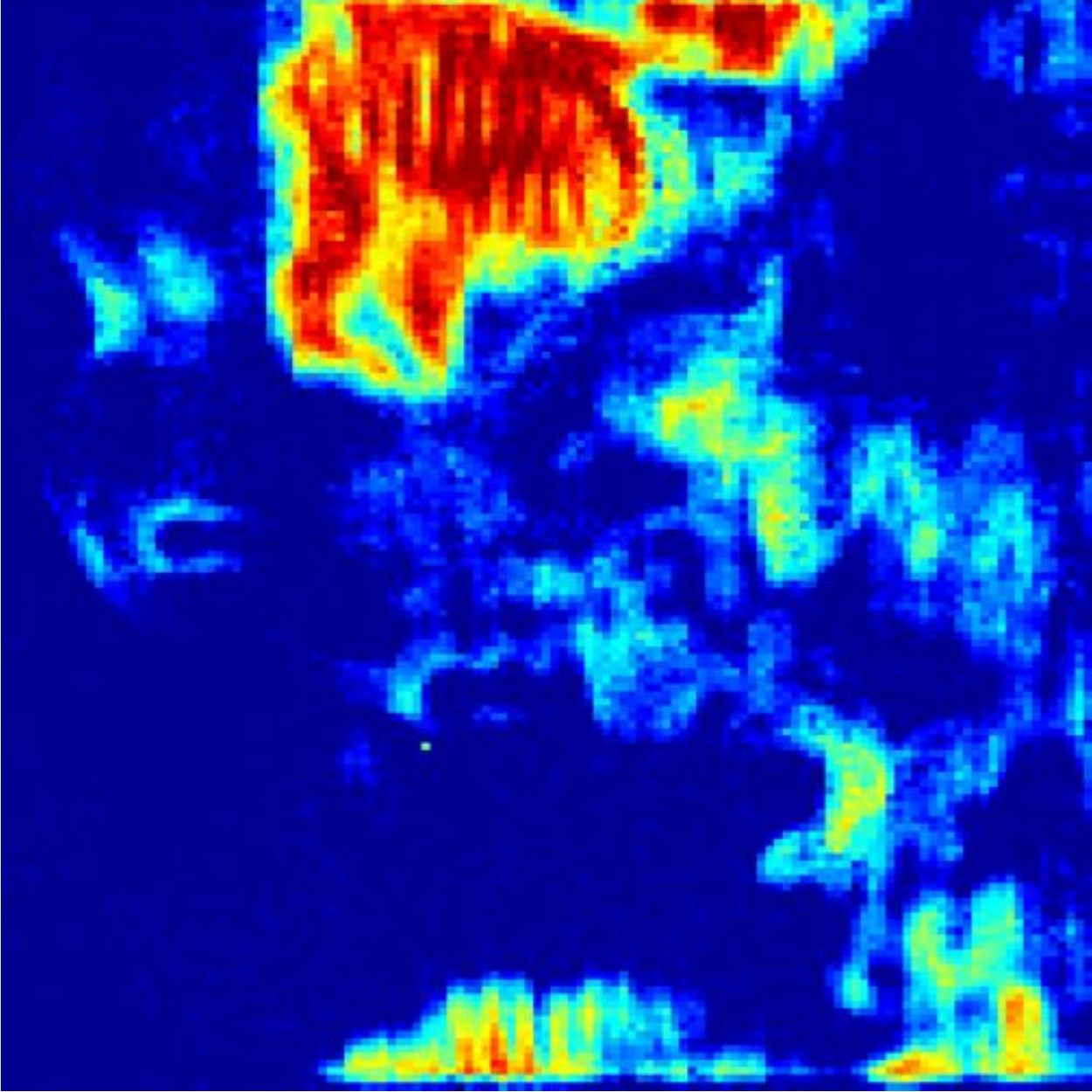}
			\captionsetup{justification=centering,labelformat=empty,skip=0pt}
			\caption{}
		\end{subfigure}
		\begin{subfigure}[b]{0.15\textwidth}
			\includegraphics[width=1\linewidth,height=0.8\linewidth]{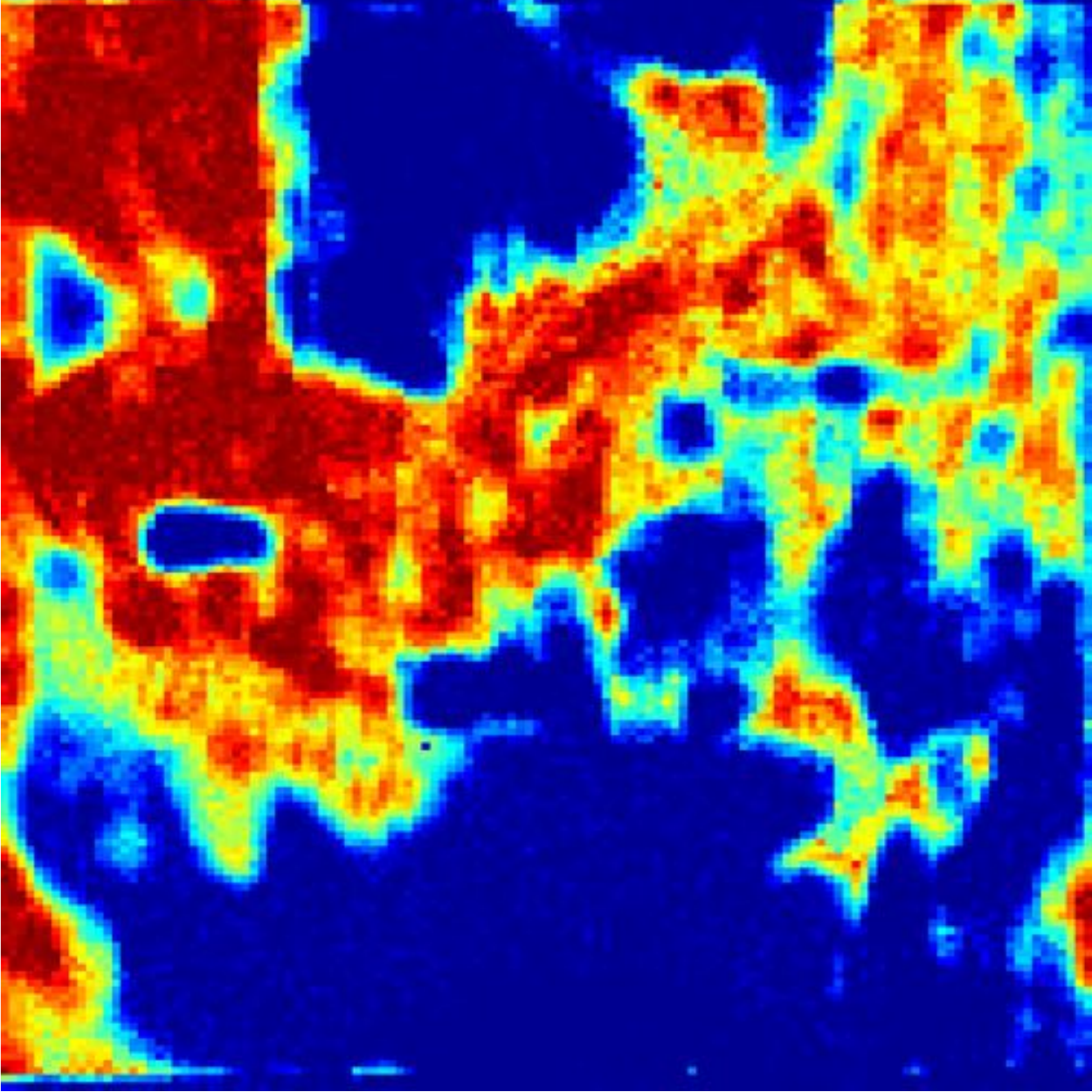}
			\captionsetup{justification=centering,labelformat=empty,skip=0pt}
			\caption{}
		\end{subfigure}
		\begin{subfigure}[b]{0.15\textwidth}
			\includegraphics[width=1\linewidth,height=0.8\linewidth]{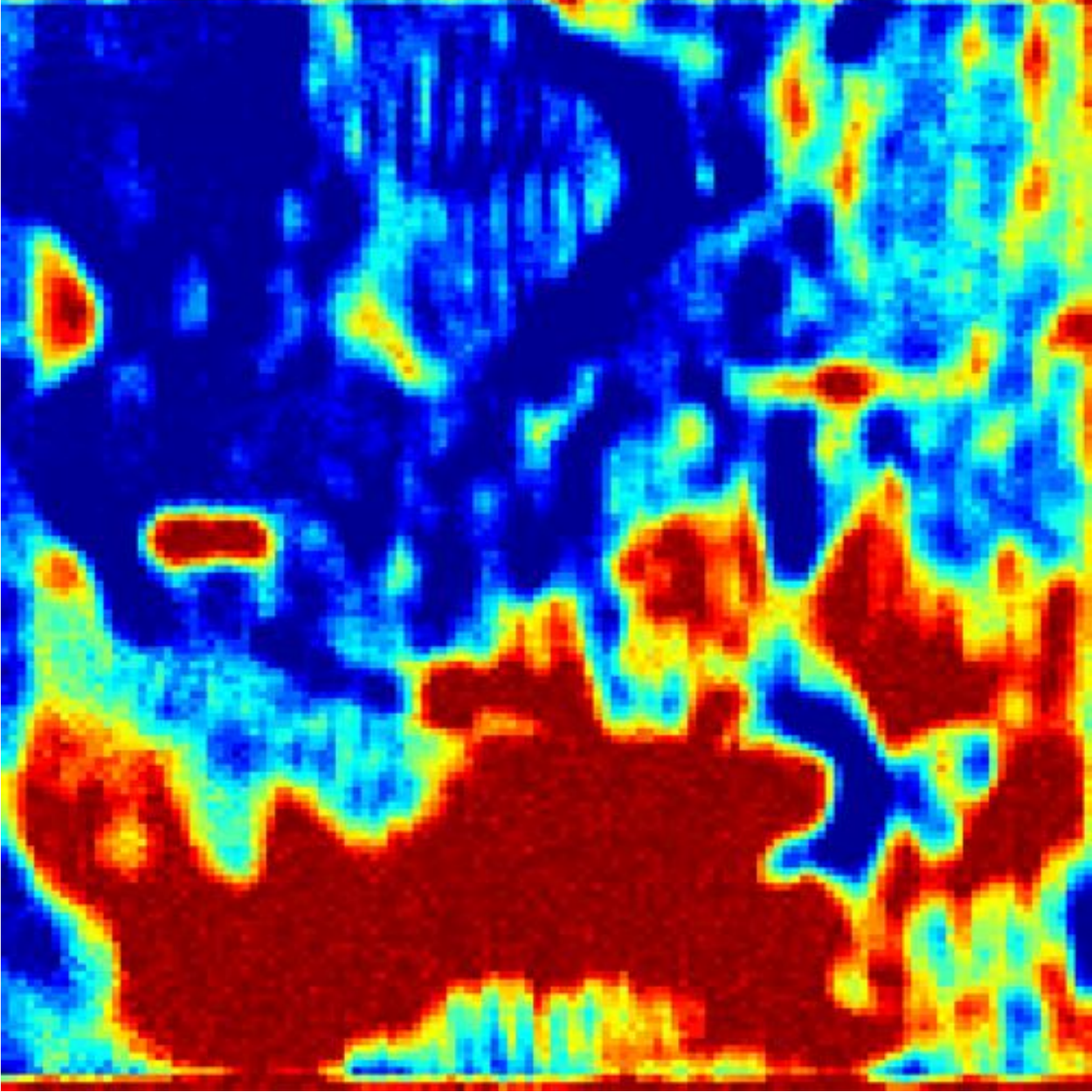}
			\captionsetup{justification=centering,labelformat=empty,skip=0pt}
			\caption{}
		\end{subfigure}
		\begin{subfigure}[b]{0.15\textwidth}
			\includegraphics[width=1\linewidth,height=0.8\linewidth]{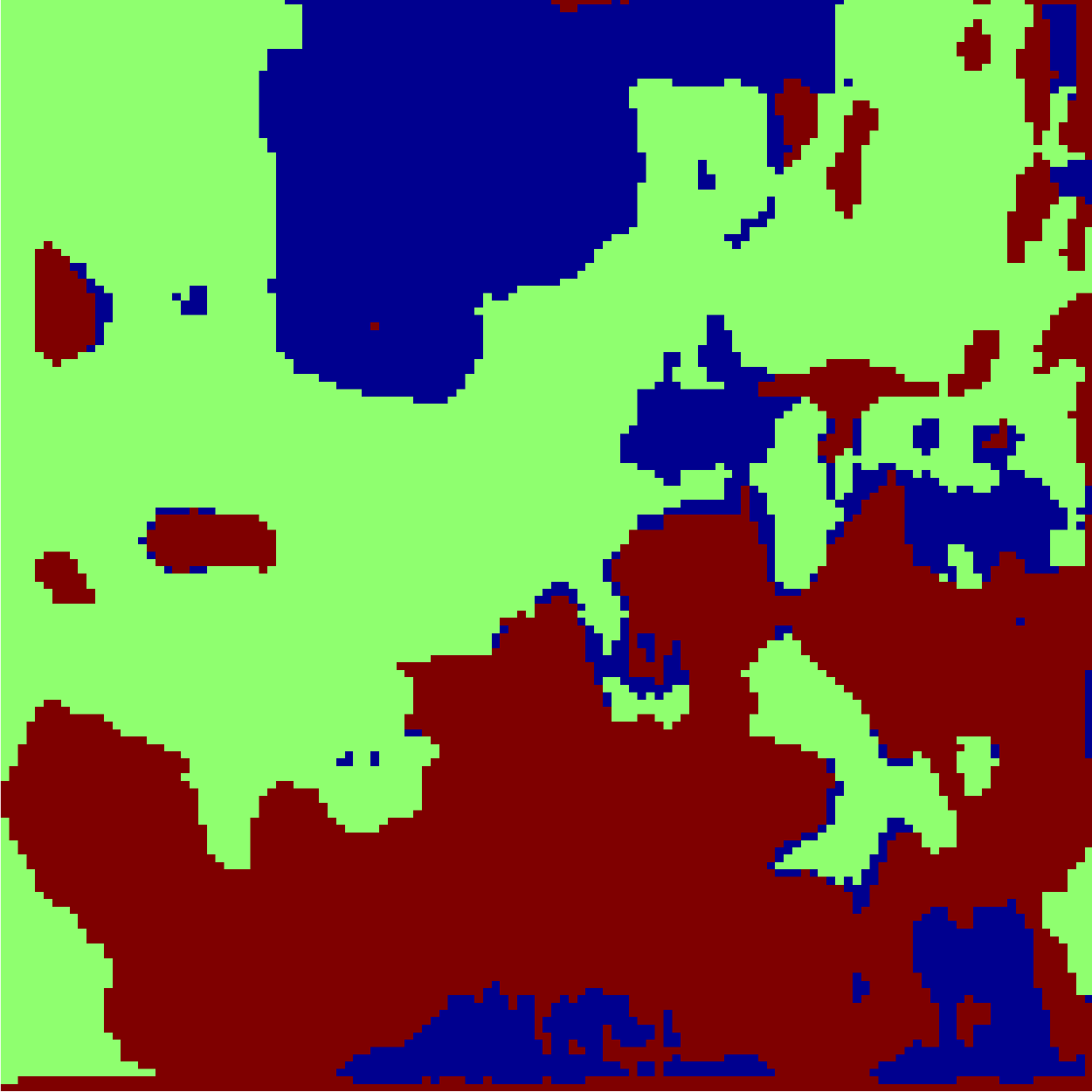}
			\captionsetup{justification=centering,labelformat=empty,skip=0pt}
			\caption{}
		\end{subfigure}
		\begin{subfigure}[b]{0.15\textwidth}
			\includegraphics[width=1\linewidth,height=0.8\linewidth]{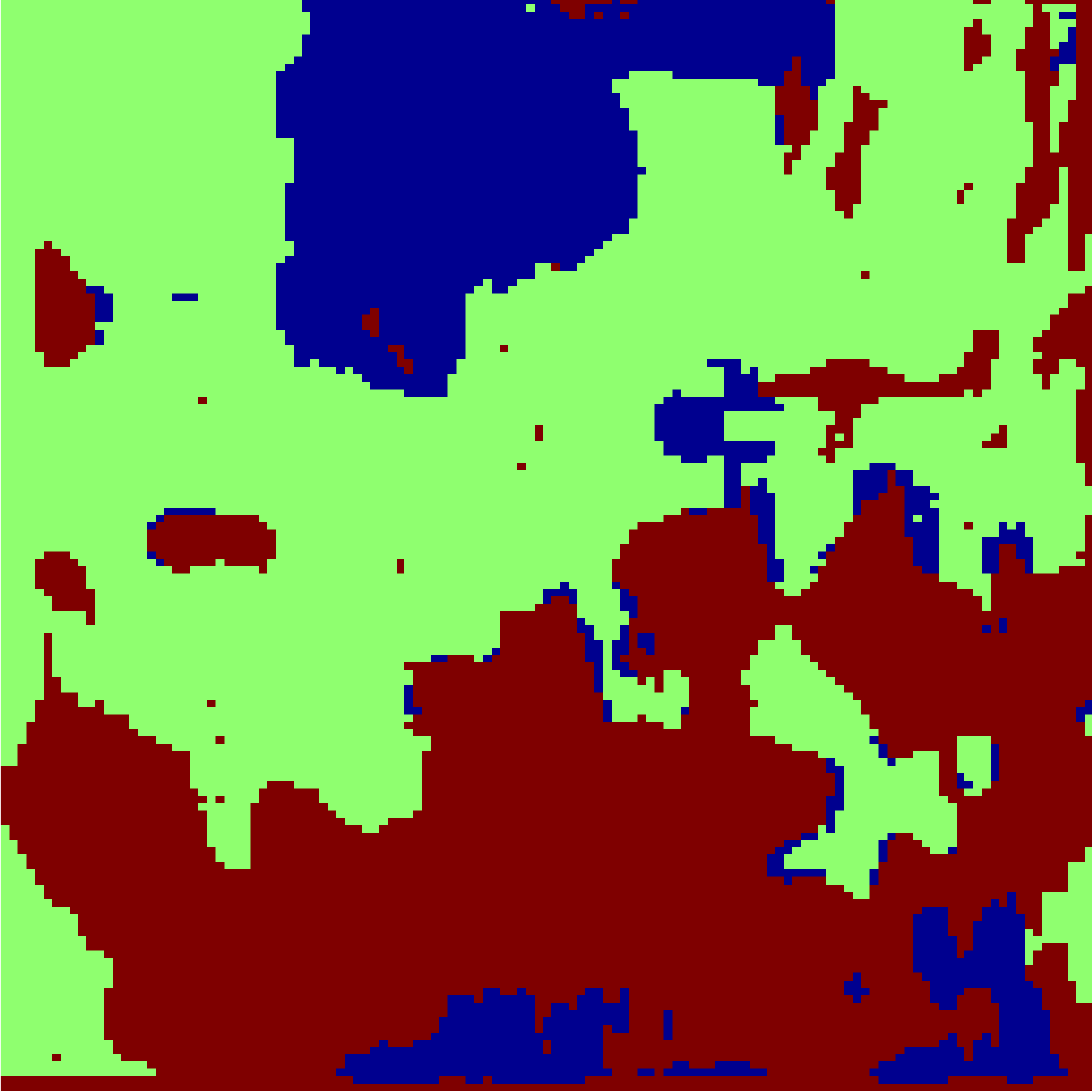}
			\captionsetup{justification=centering,labelformat=empty,skip=0pt}
			\caption{}
			\end{subfigure}		
	
		\begin{subfigure}[b]{0.15\textwidth}
			\includegraphics[width=1\linewidth,height=0.8\linewidth]{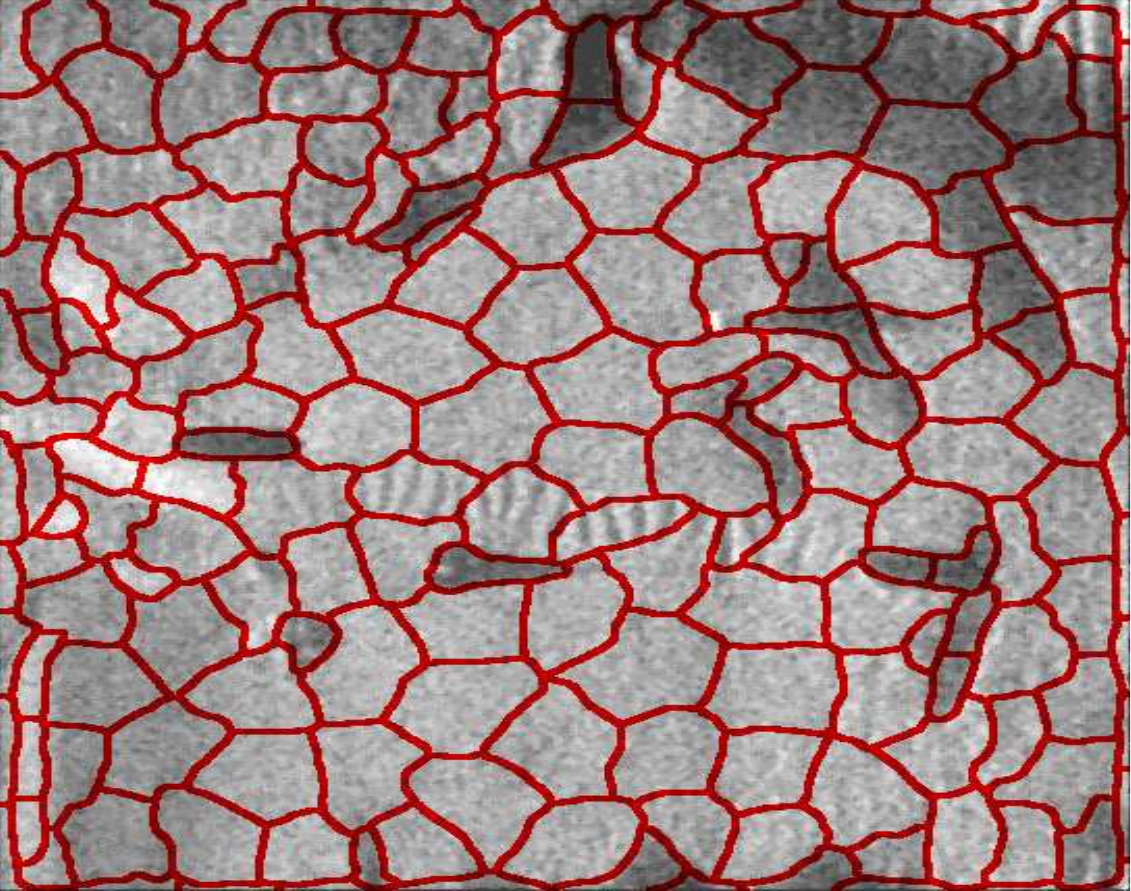}
			\captionsetup{justification=centering,labelformat=empty,,skip=0pt}
			\caption{}
		\end{subfigure}
		\begin{subfigure}[b]{0.15\textwidth}
			\includegraphics[width=1\linewidth,height=0.8\linewidth]{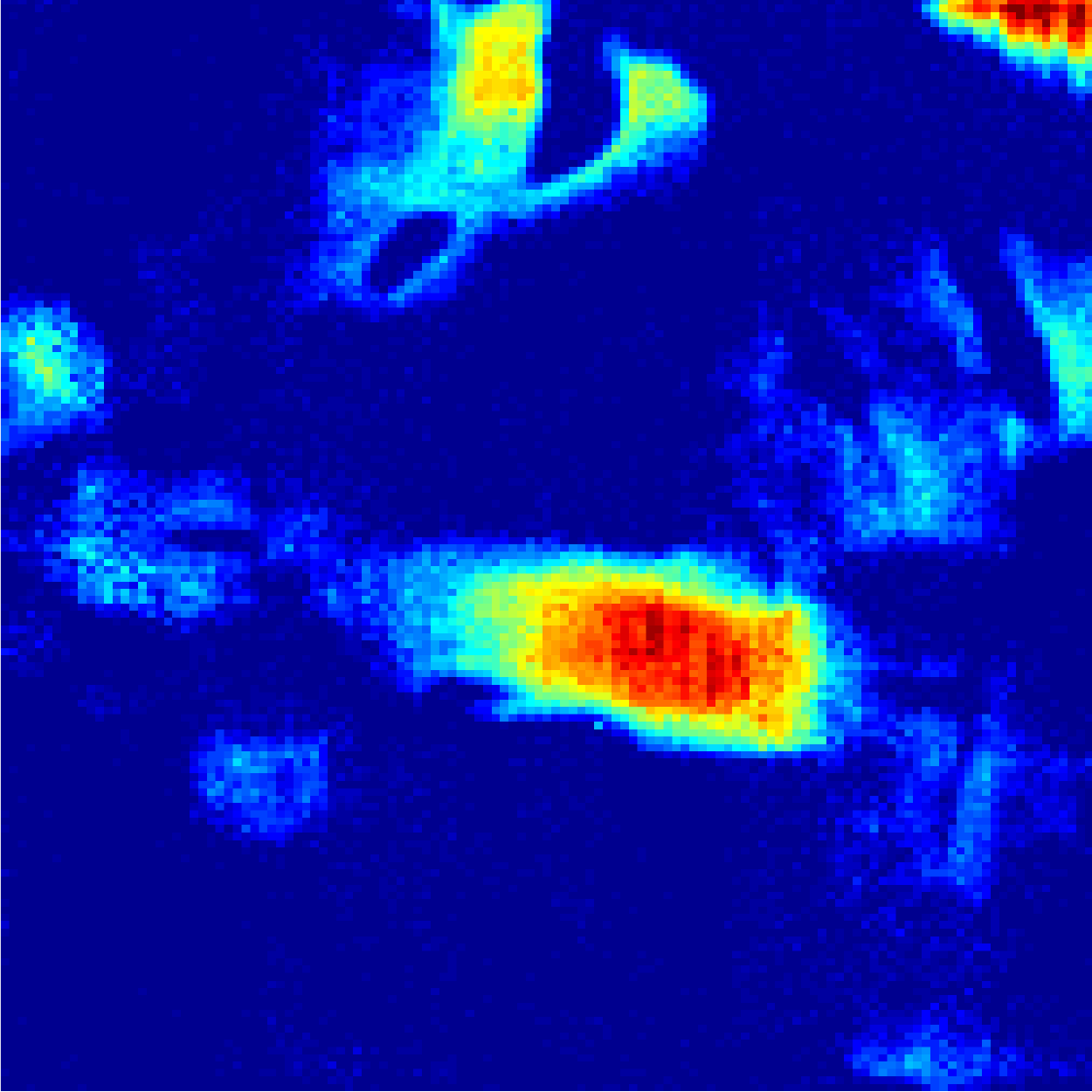}
			\captionsetup{justification=centering,labelformat=empty,skip=0pt}
			\caption{}
		\end{subfigure}
		\begin{subfigure}[b]{0.15\textwidth}
			\includegraphics[width=1\linewidth,height=0.8\linewidth]{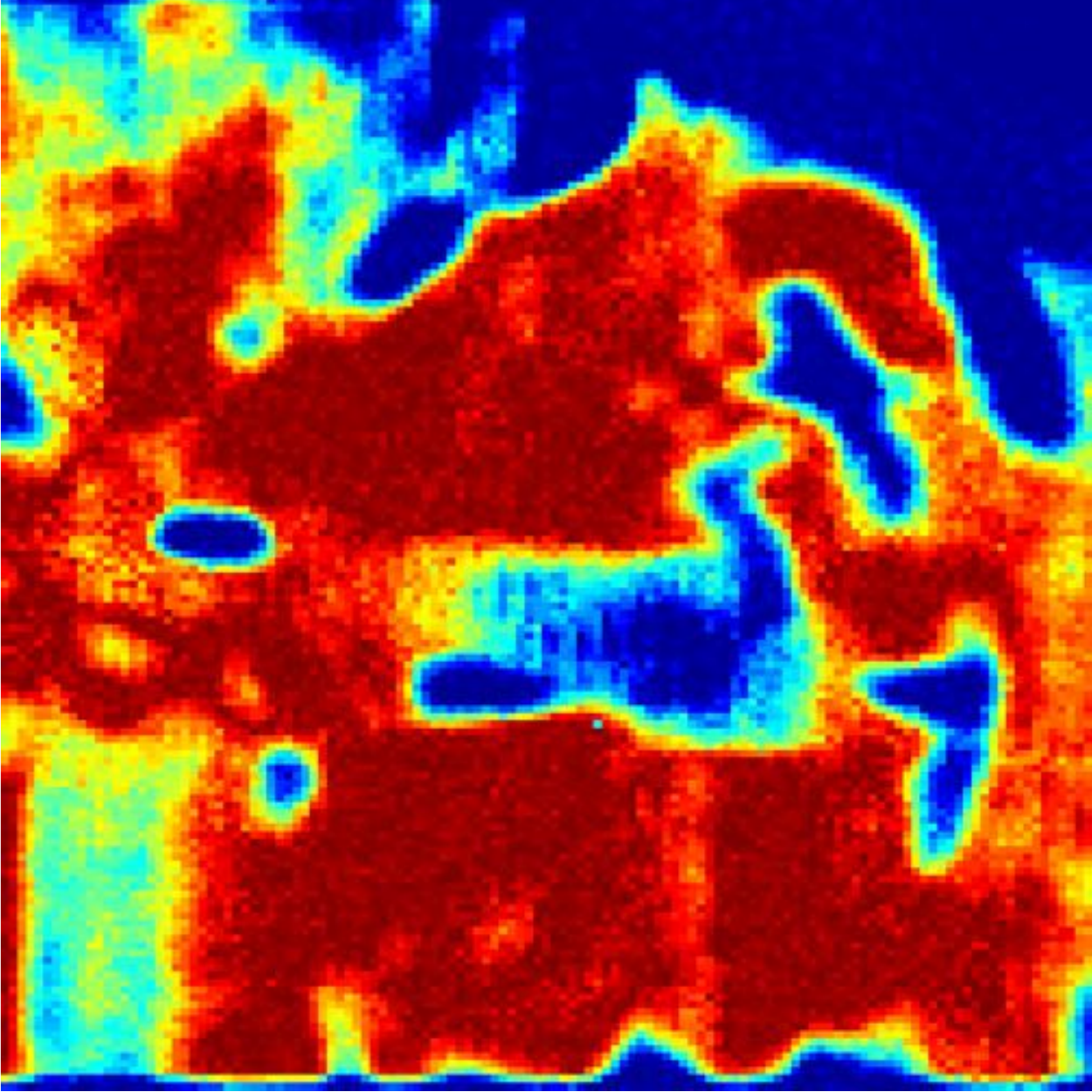}
			\captionsetup{justification=centering,labelformat=empty,skip=0pt}
			\caption{}
		\end{subfigure}
		\begin{subfigure}[b]{0.15\textwidth}
			\includegraphics[width=1\linewidth,height=0.8\linewidth]{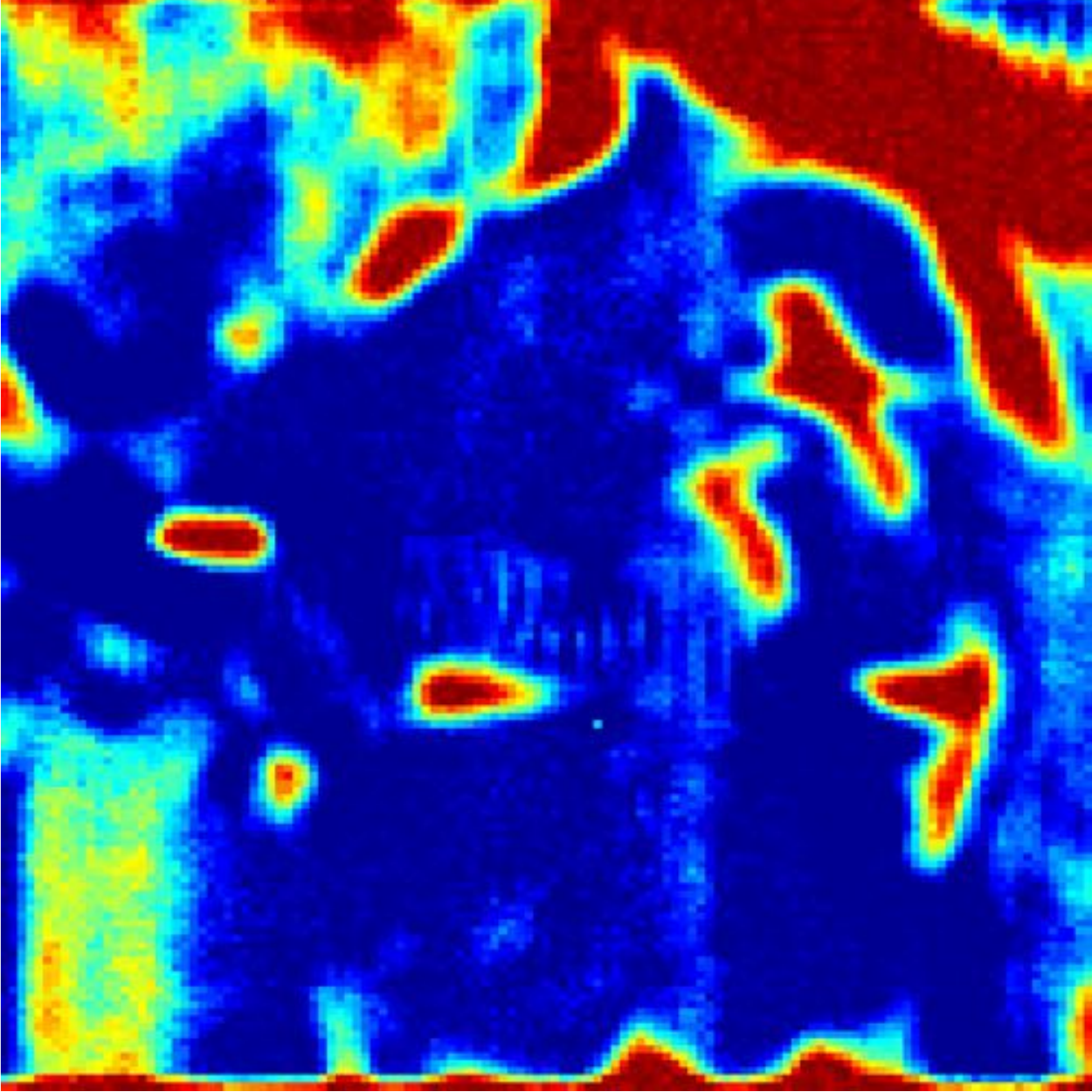}
			\captionsetup{justification=centering,labelformat=empty,skip=0pt}
			\caption{}
		\end{subfigure}	
		\begin{subfigure}[b]{0.15\textwidth}
			\includegraphics[width=1\linewidth,height=0.8\linewidth]{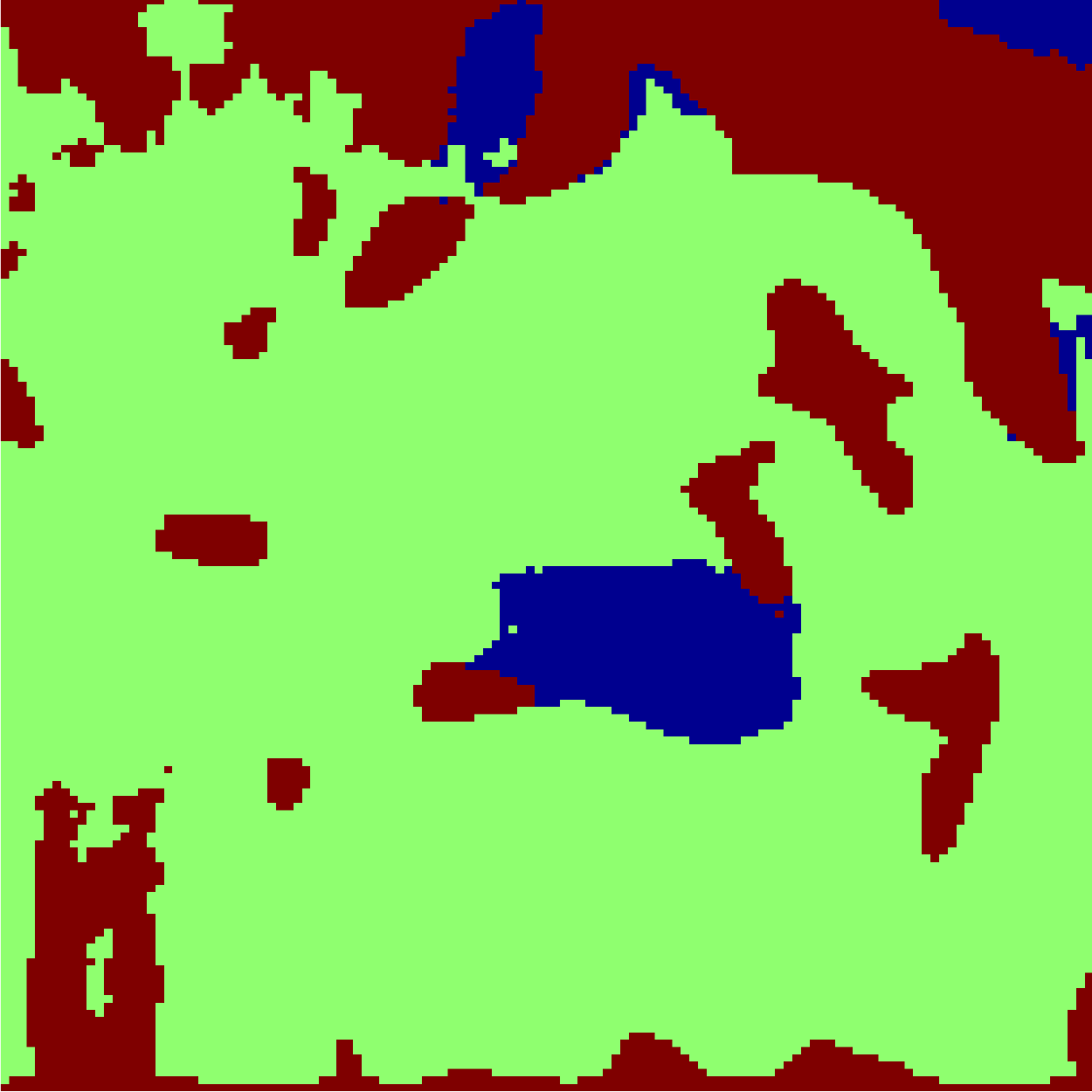}
			\captionsetup{justification=centering,labelformat=empty,skip=0pt}
			\caption{}
		\end{subfigure}
		\begin{subfigure}[b]{0.15\textwidth}
			\includegraphics[width=1\linewidth,height=0.8\linewidth]{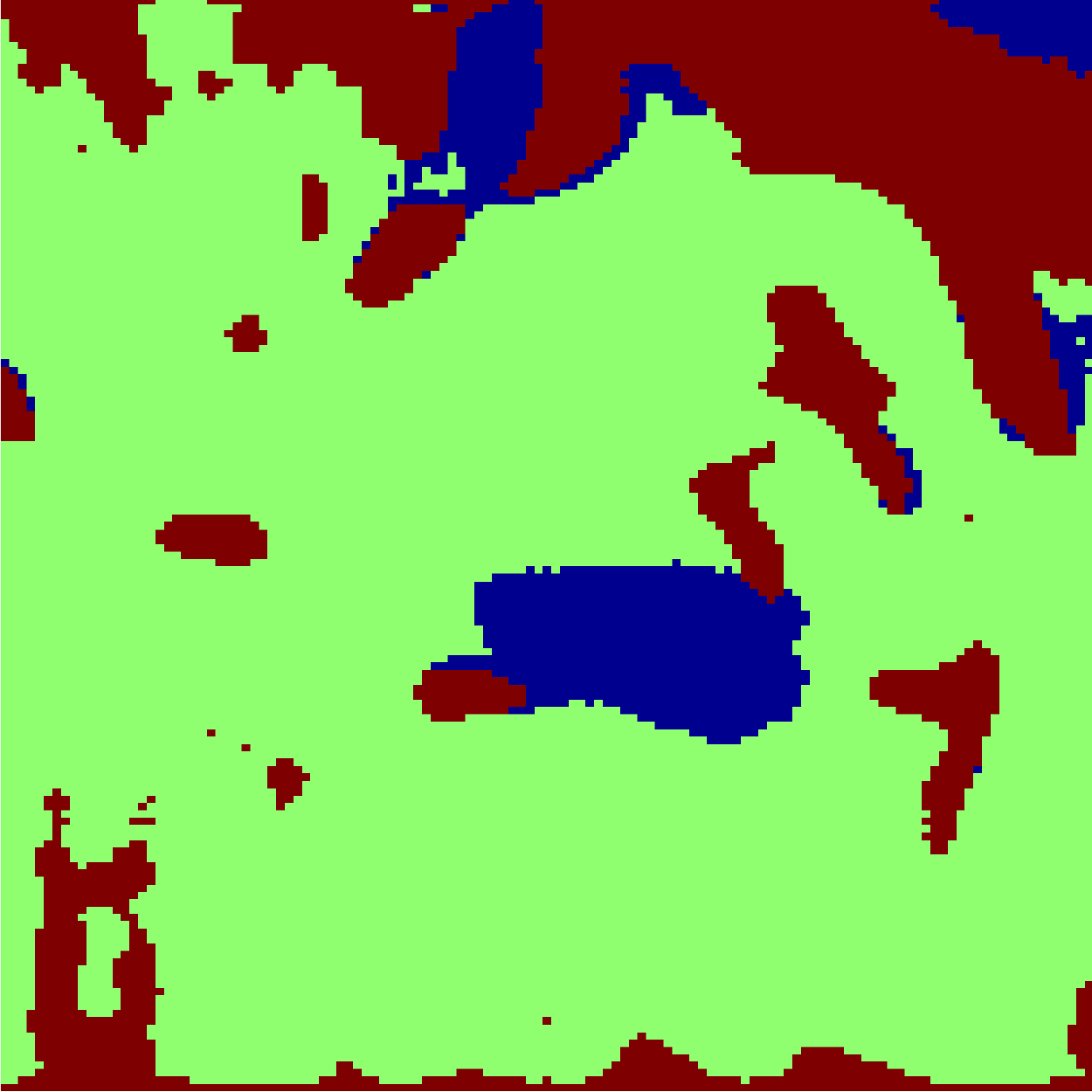}
			\captionsetup{justification=centering,labelformat=empty,skip=0pt}
			\caption{}
		\end{subfigure}		
		
		\begin{subfigure}[b]{0.15\textwidth}
			\includegraphics[width=1\linewidth,height=0.8\linewidth]{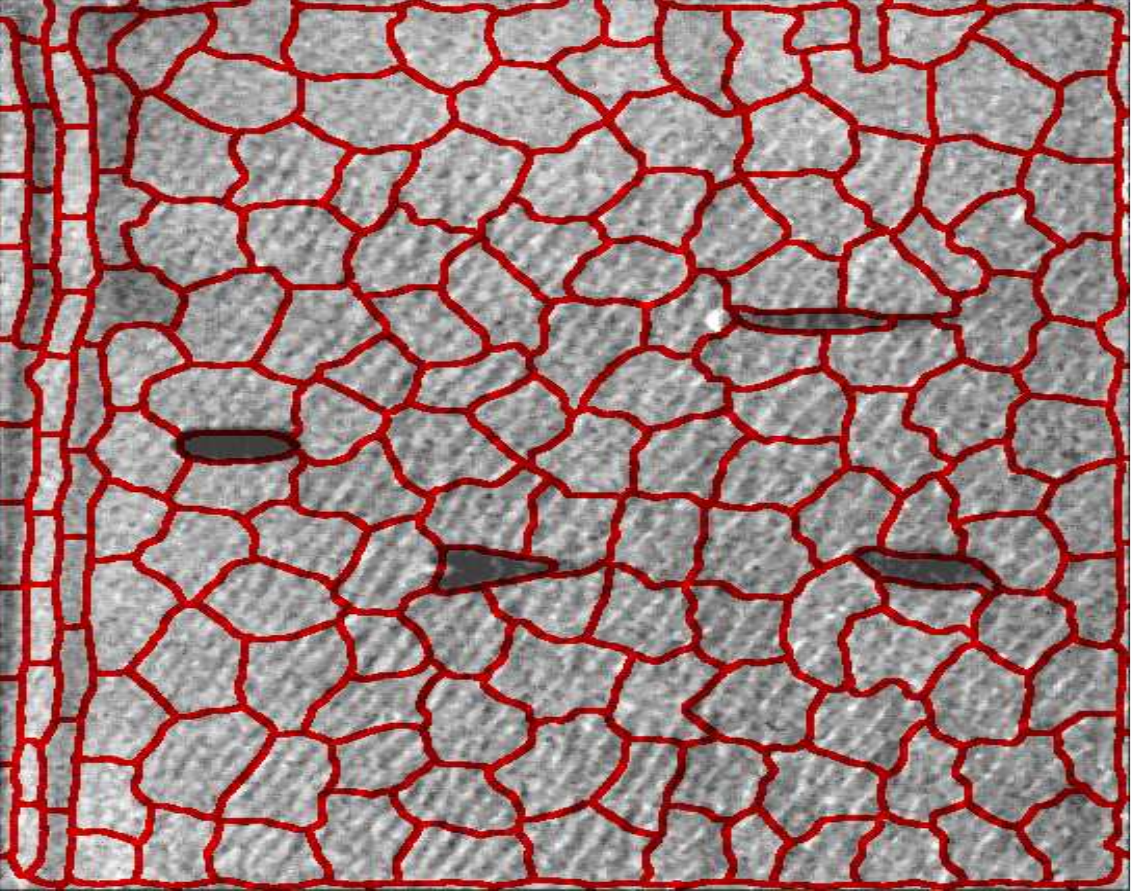}
			\captionsetup{justification=centering,labelformat=empty,,skip=0pt}
			\caption{}
		\end{subfigure}
		\begin{subfigure}[b]{0.15\textwidth}
			\includegraphics[width=1\linewidth,height=0.8\linewidth]{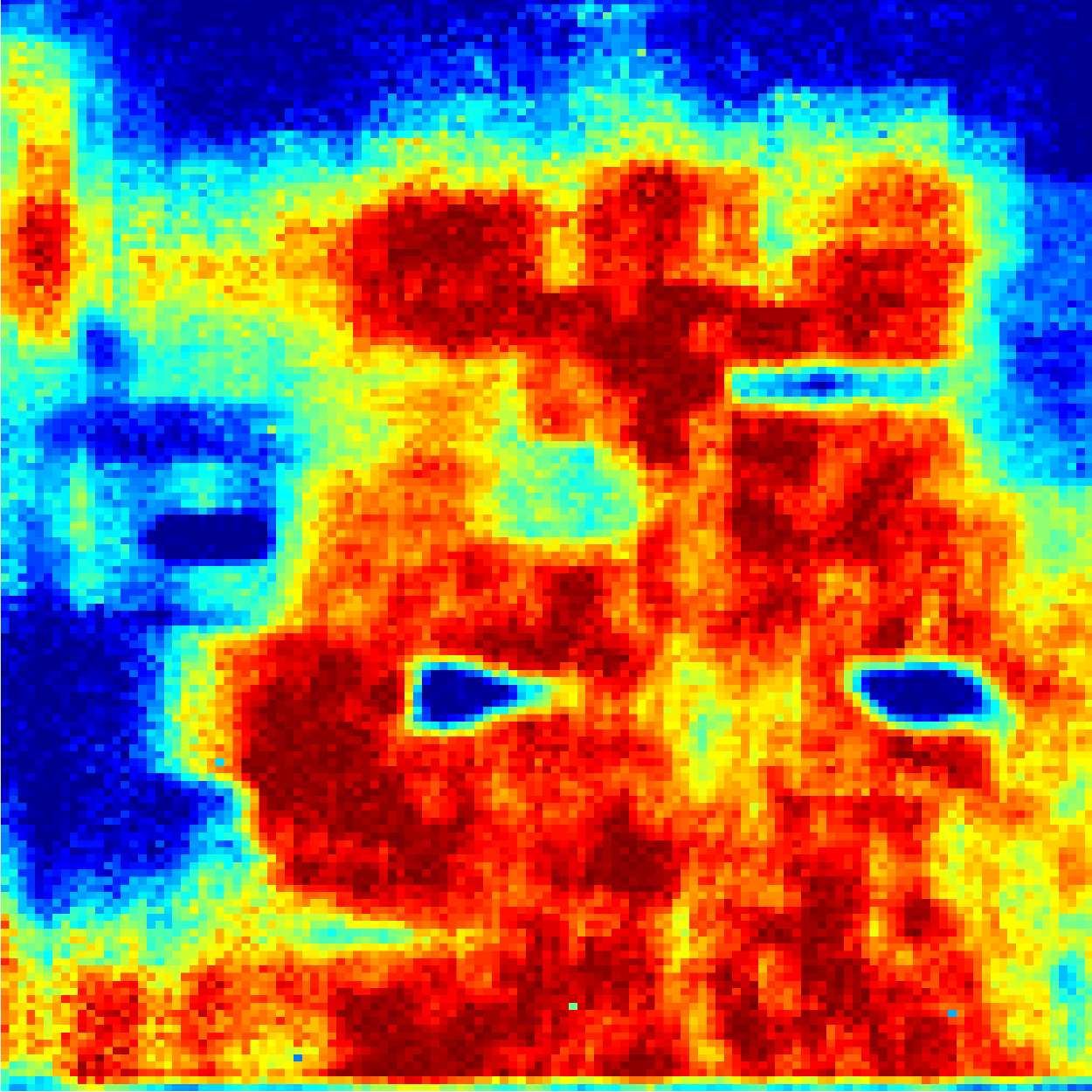}
			\captionsetup{justification=centering,labelformat=empty,skip=0pt}
			\caption{}
		\end{subfigure}
		\begin{subfigure}[b]{0.15\textwidth}
			\includegraphics[width=1\linewidth,height=0.8\linewidth]{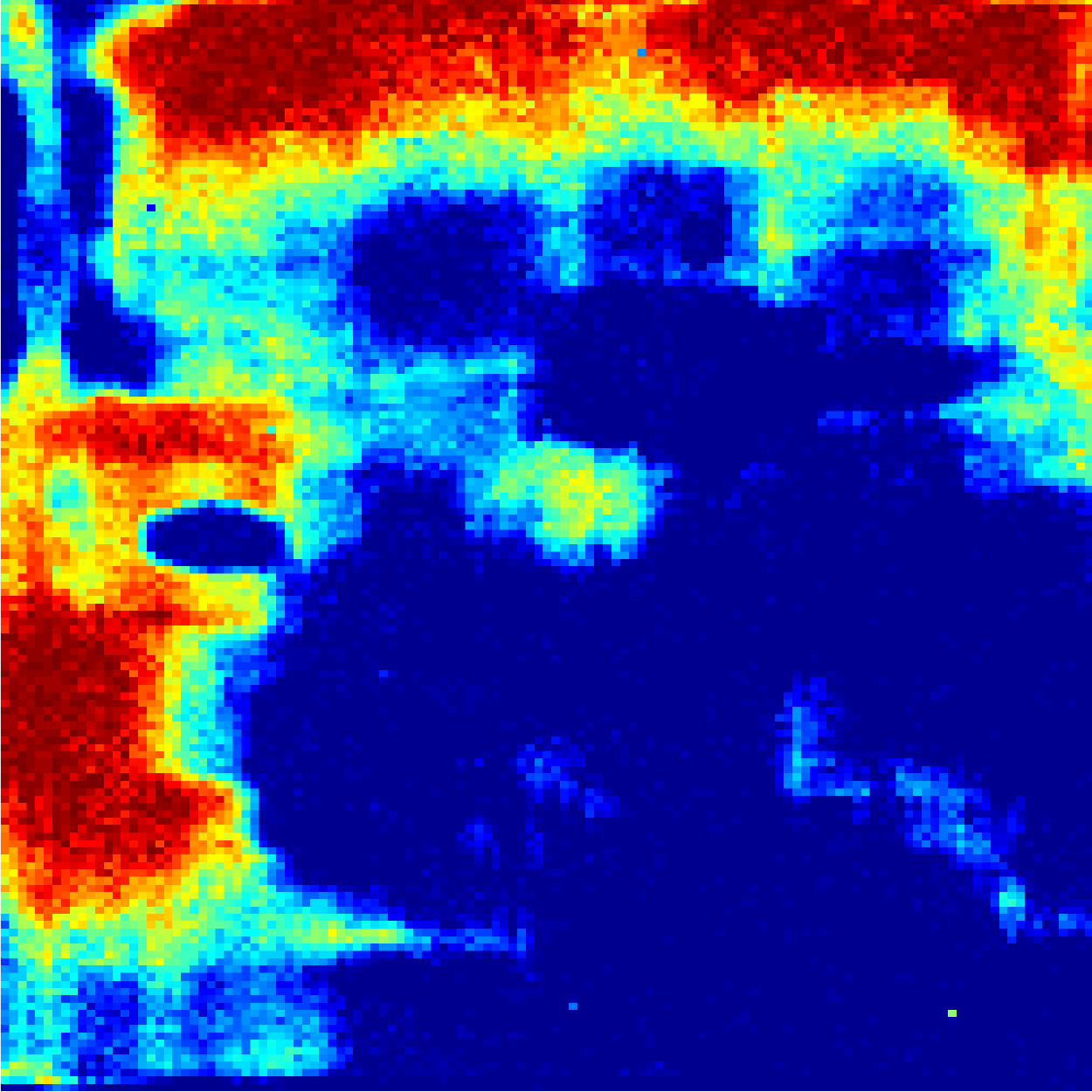}
			\captionsetup{justification=centering,labelformat=empty,skip=0pt}
			\caption{}
		\end{subfigure}
		\begin{subfigure}[b]{0.15\textwidth}
			\includegraphics[width=1\linewidth,height=0.8\linewidth]{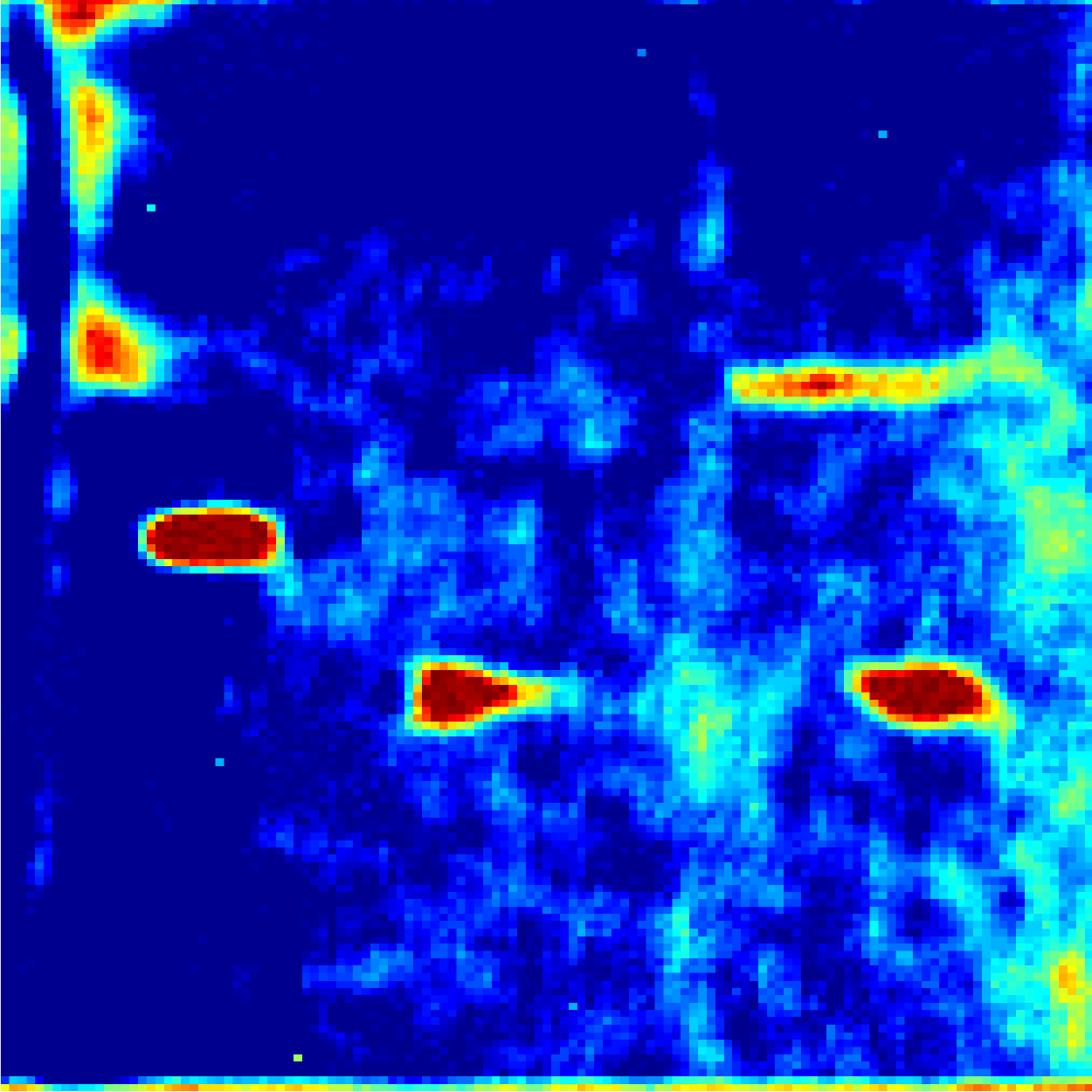}
			\captionsetup{justification=centering,labelformat=empty,skip=0pt}
			\caption{}
		\end{subfigure}	
		\begin{subfigure}[b]{0.15\textwidth}
			\includegraphics[width=1\linewidth,height=0.8\linewidth]{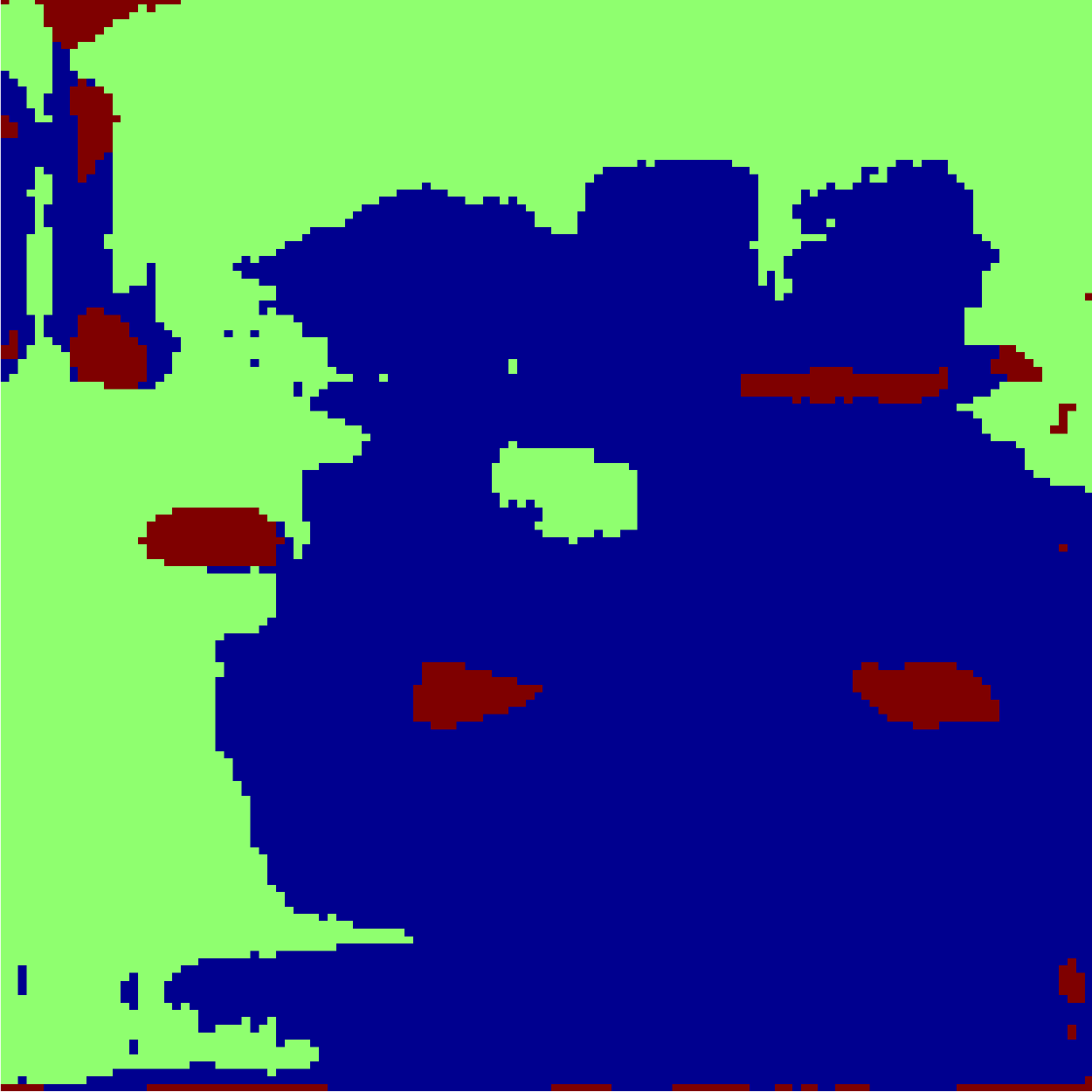}
			\captionsetup{justification=centering,labelformat=empty,skip=0pt}
			\caption{}
		\end{subfigure}
		\begin{subfigure}[b]{0.15\textwidth}
			\includegraphics[width=1\linewidth,height=0.8\linewidth]{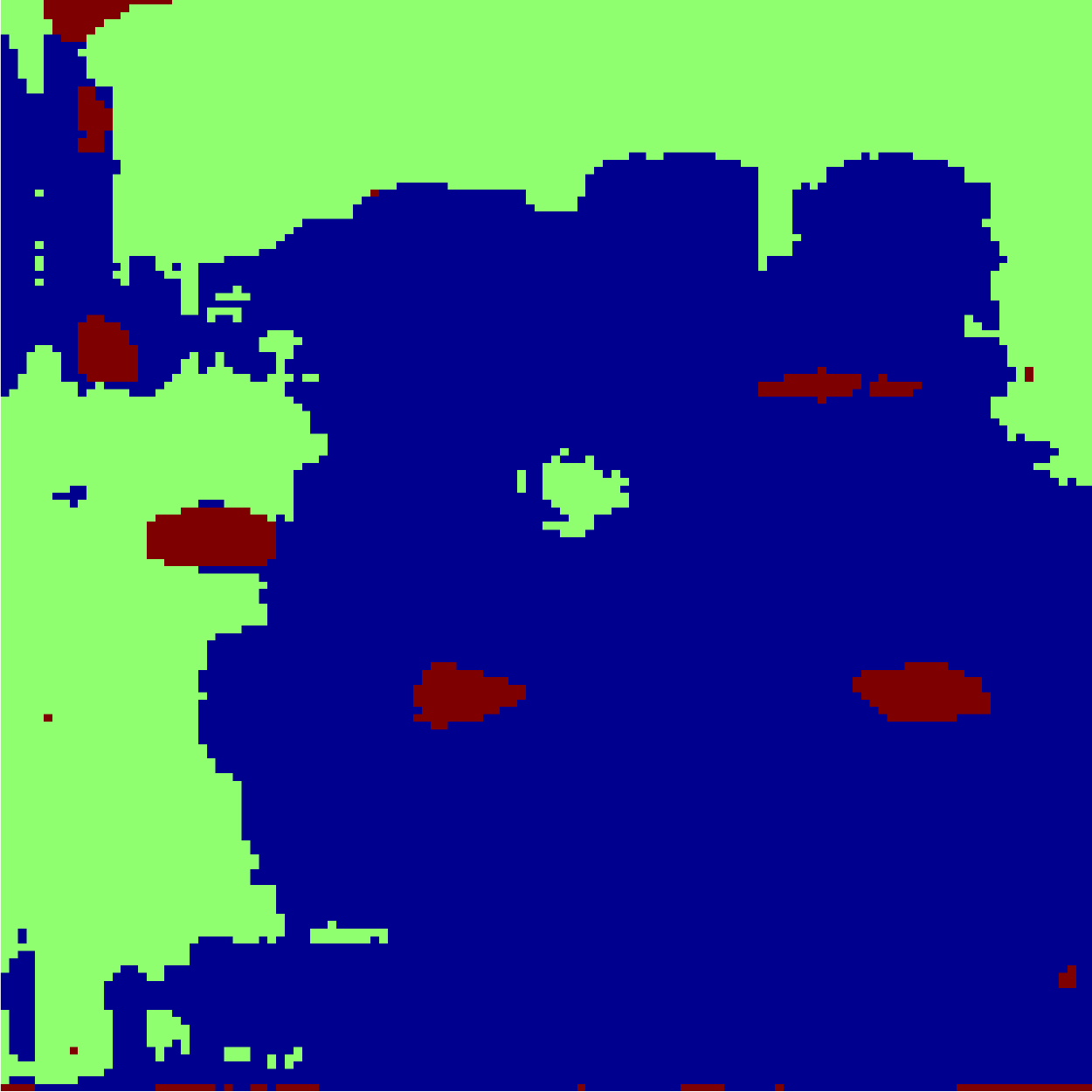}
			\captionsetup{justification=centering,labelformat=empty,skip=0pt}
			\caption{}
		\end{subfigure}		
		
		\begin{subfigure}[b]{0.15\textwidth}
			\includegraphics[width=1\linewidth,height=0.8\linewidth]{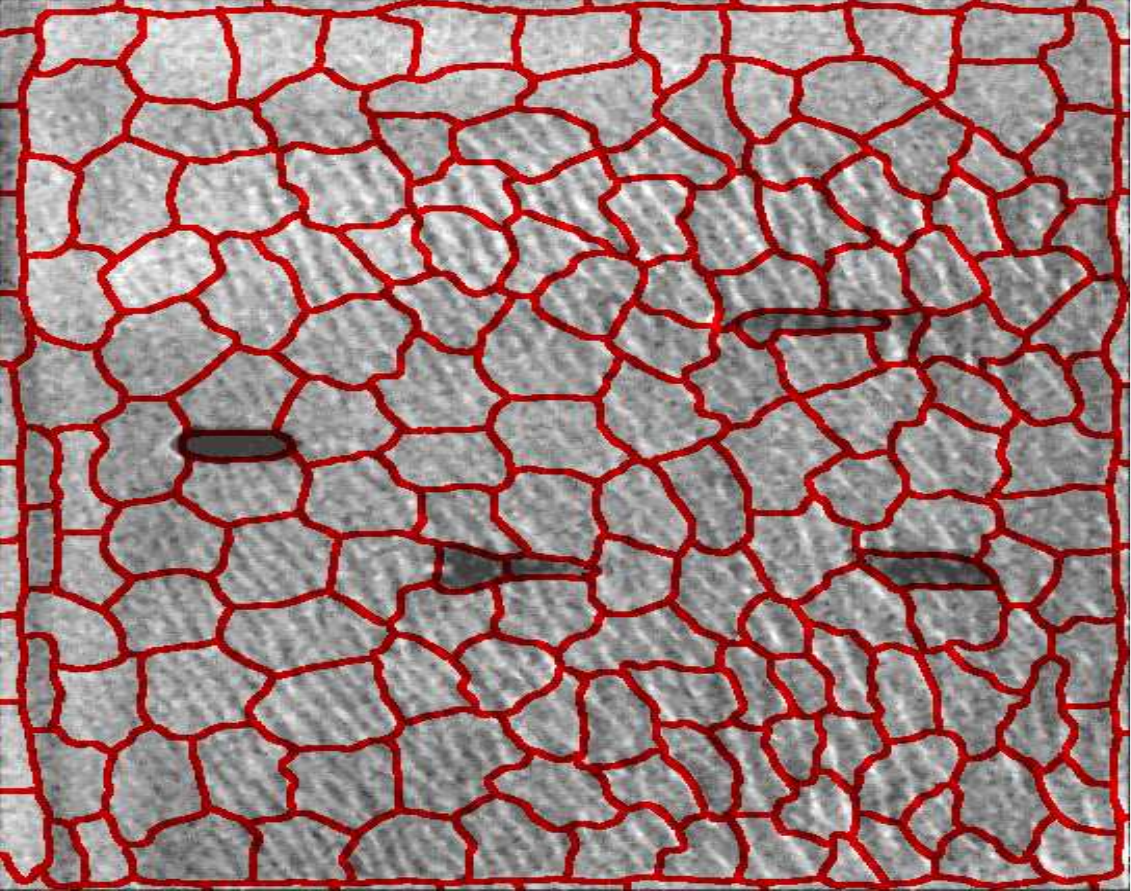}
			\captionsetup{justification=centering,labelformat=empty,,skip=0pt}
			\caption{}
	   \end{subfigure}
	   \begin{subfigure}[b]{0.15\textwidth}
			\includegraphics[width=1\linewidth,height=0.8\linewidth]{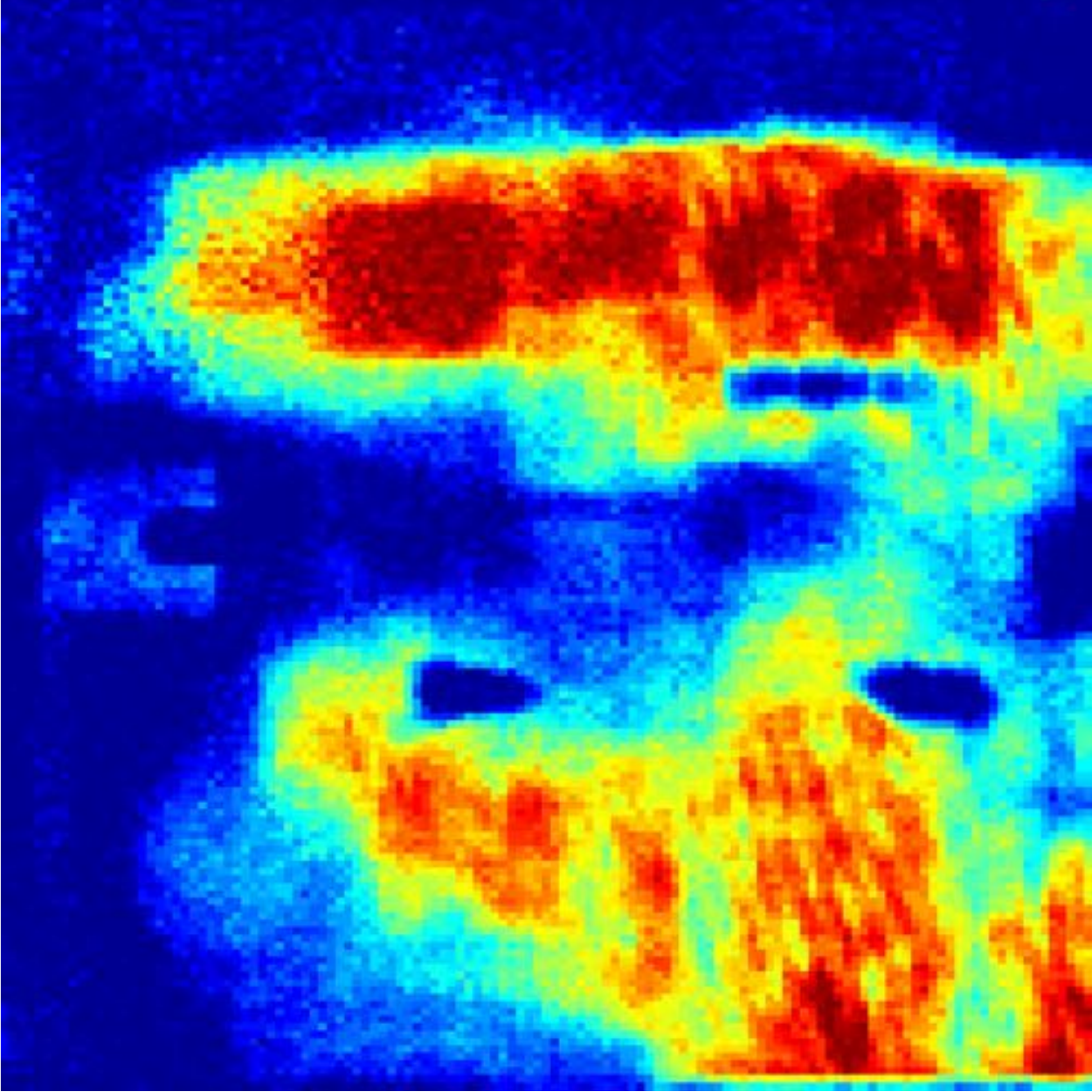}
			\captionsetup{justification=centering,labelformat=empty,skip=0pt}
	 		\caption{}
	  \end{subfigure}
	  \begin{subfigure}[b]{0.15\textwidth}
		  \includegraphics[width=1\linewidth,height=0.8\linewidth]{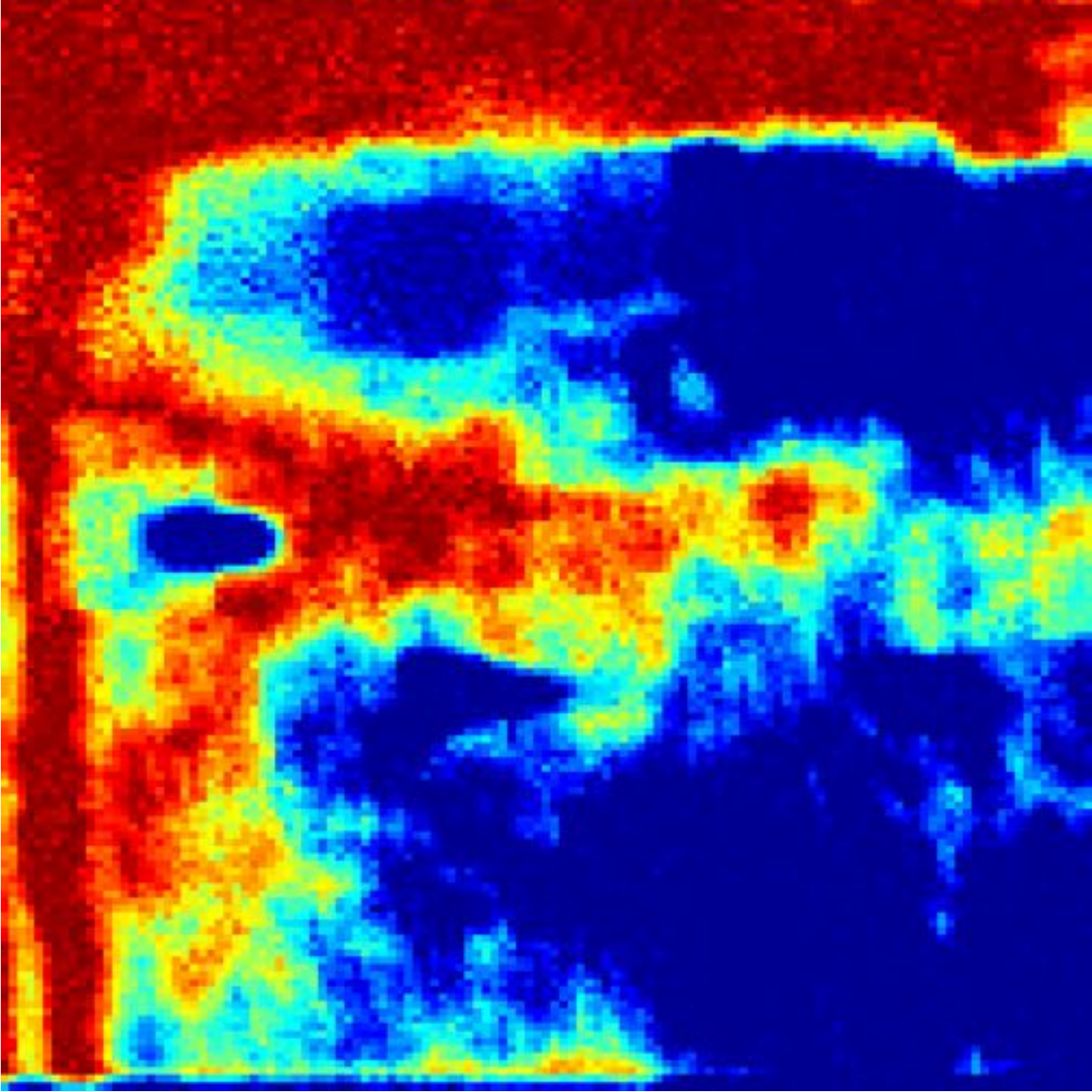}
		  \captionsetup{justification=centering,labelformat=empty,skip=0pt}
		 \caption{}
  	  \end{subfigure}
	  \begin{subfigure}[b]{0.15\textwidth}
		 \includegraphics[width=1\linewidth,height=0.8\linewidth]{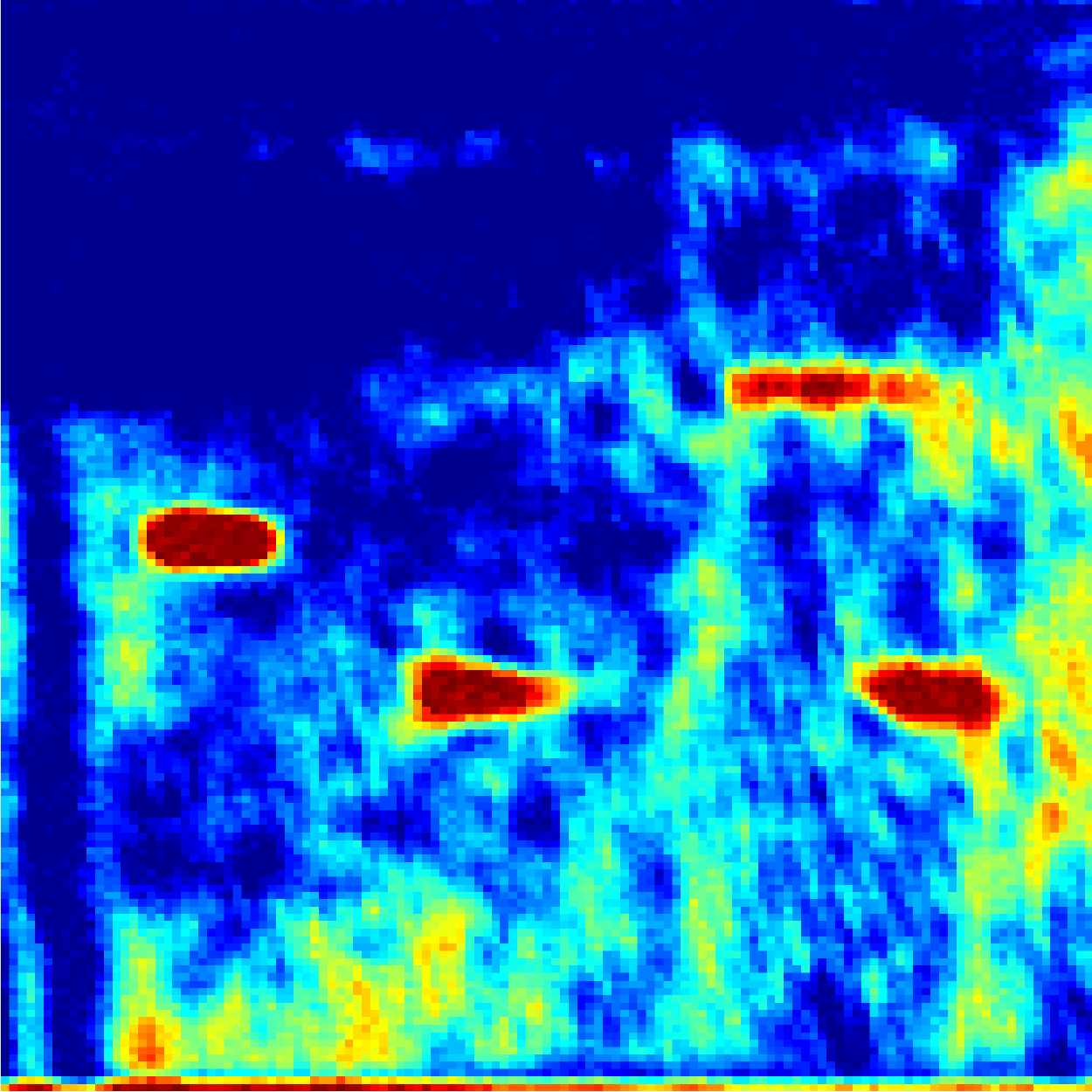}
		 \captionsetup{justification=centering,labelformat=empty,skip=0pt}
	 	 \caption{}
	  \end{subfigure}	
	  \begin{subfigure}[b]{0.15\textwidth}
	  	\includegraphics[width=1\linewidth,height=0.8\linewidth]{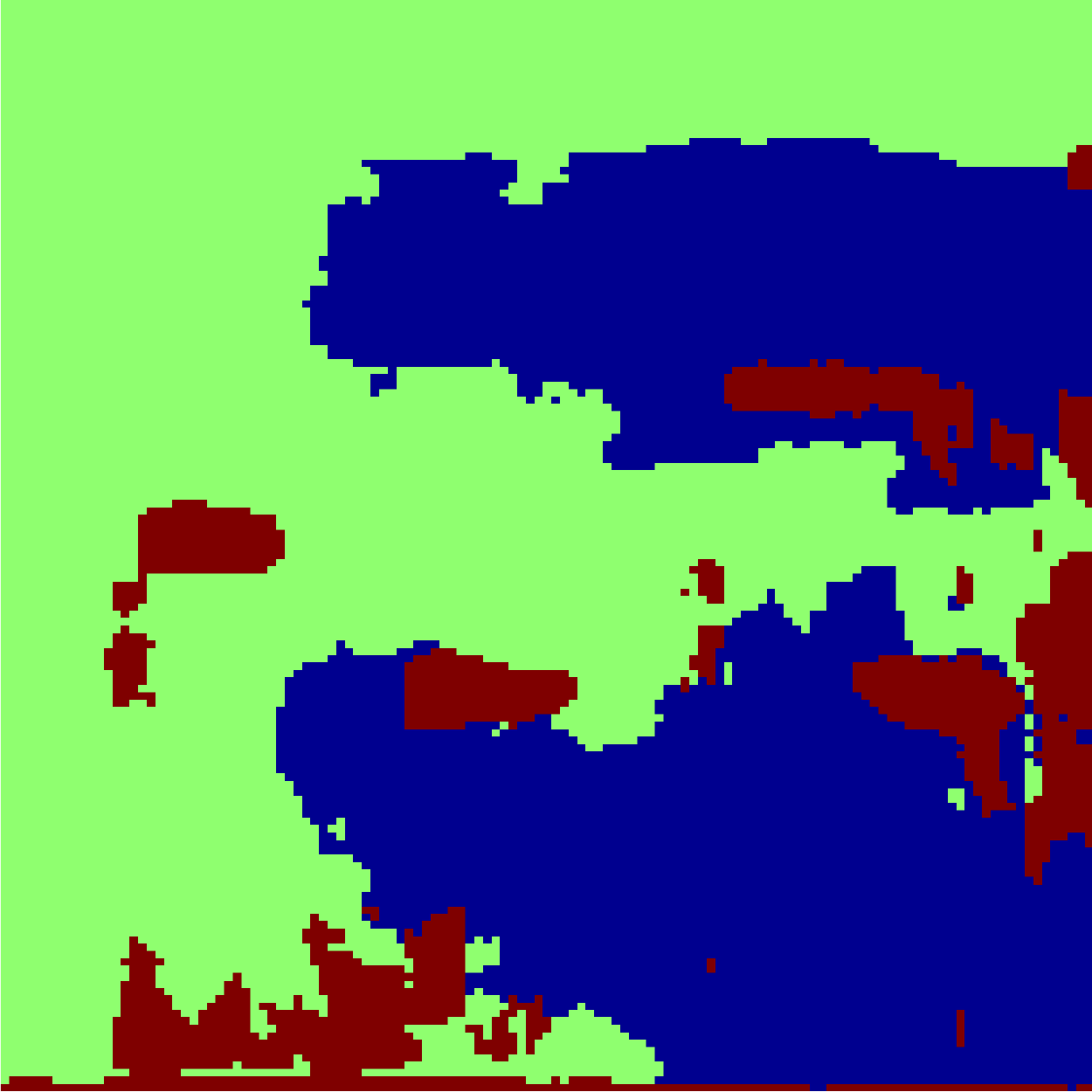}
	  	\captionsetup{justification=centering,labelformat=empty,skip=0pt}
	  	\caption{}
	  \end{subfigure}
	  \begin{subfigure}[b]{0.15\textwidth}
	  	\includegraphics[width=1\linewidth,height=0.8\linewidth]{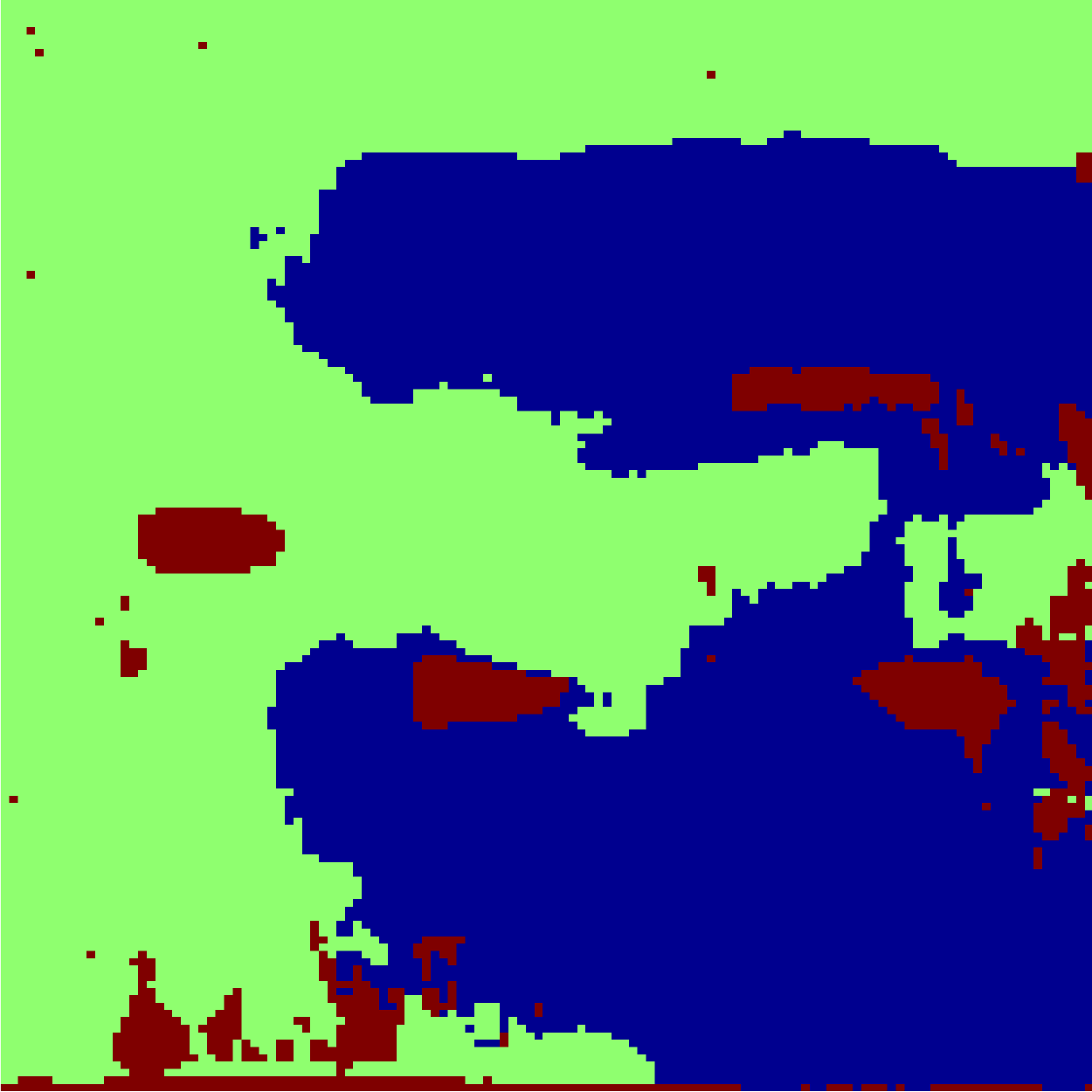}
	  	\captionsetup{justification=centering,labelformat=empty,skip=0pt}
	  	\caption{}
	  \end{subfigure}		
	
			\begin{subfigure}[t]{0.15\textwidth}
				\includegraphics[width=1\linewidth,height=0.8\linewidth]{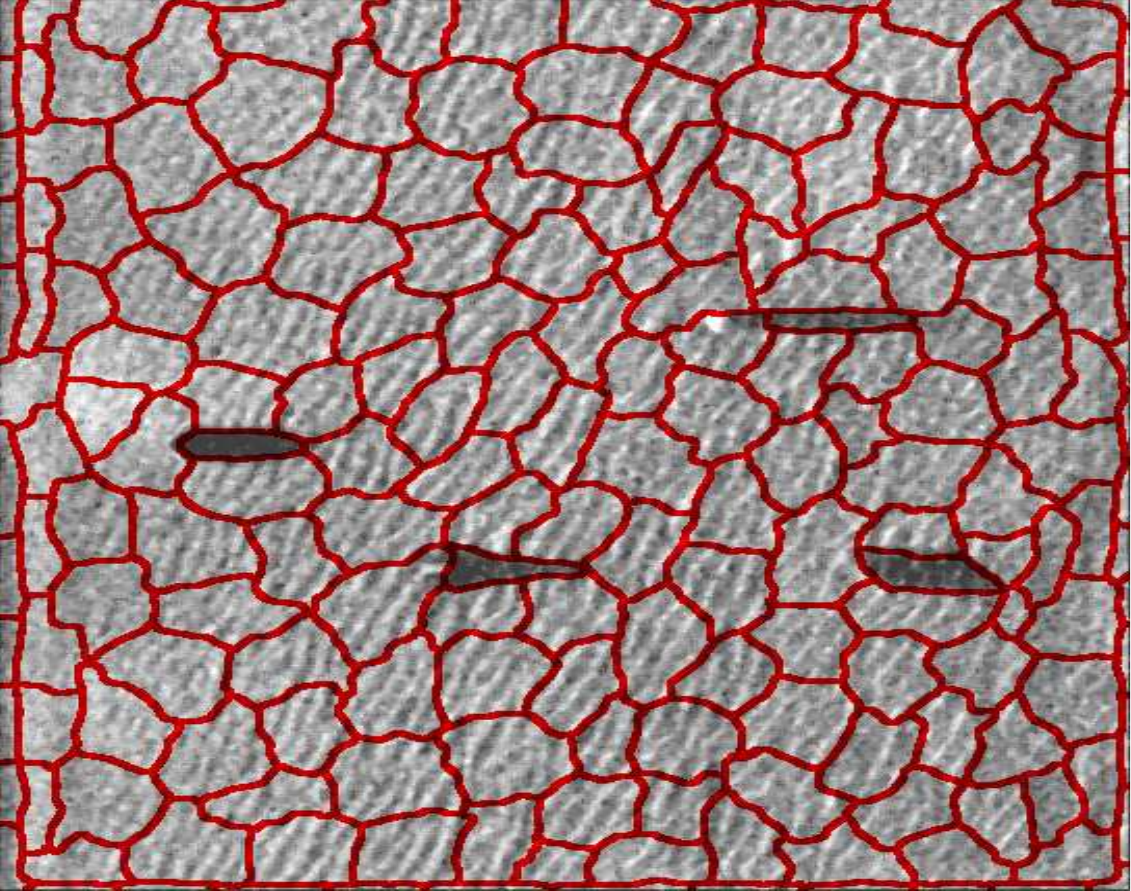}
				\captionsetup{justification=centering,labelformat=empty,skip=2pt}
				\caption{(a) SONAR imagery with super-pixel boundary}
			\end{subfigure}
			\begin{subfigure}[t]{0.15\textwidth}
				\includegraphics[width=1\linewidth,height=0.8\linewidth]{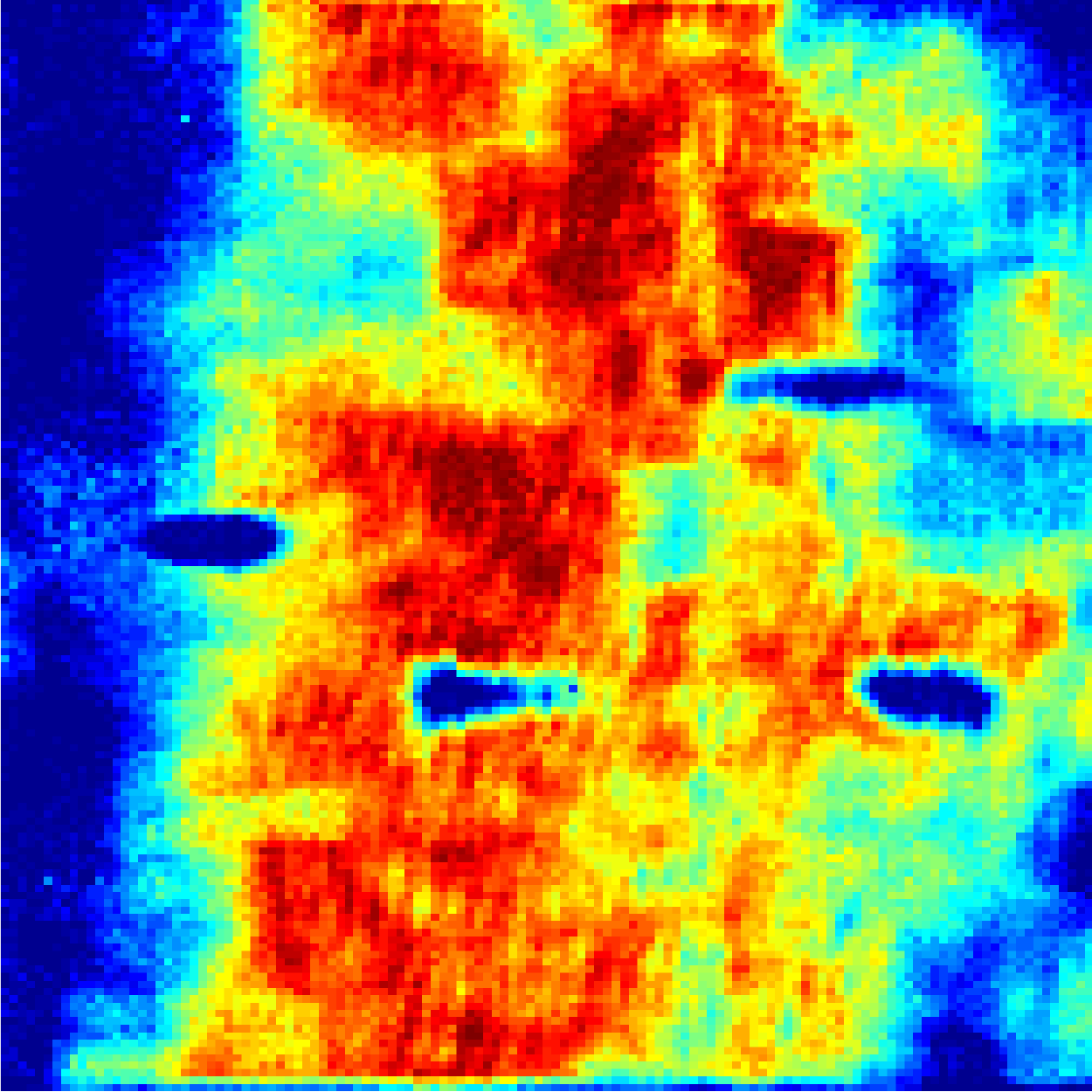}
				\captionsetup{justification=centering,labelformat=empty,skip=2pt}
				\caption{(b) PM-LDA: \\sand ripple}
			\end{subfigure}
			\begin{subfigure}[t]{0.15\textwidth}
				\includegraphics[width=1\linewidth,height=0.8\linewidth]{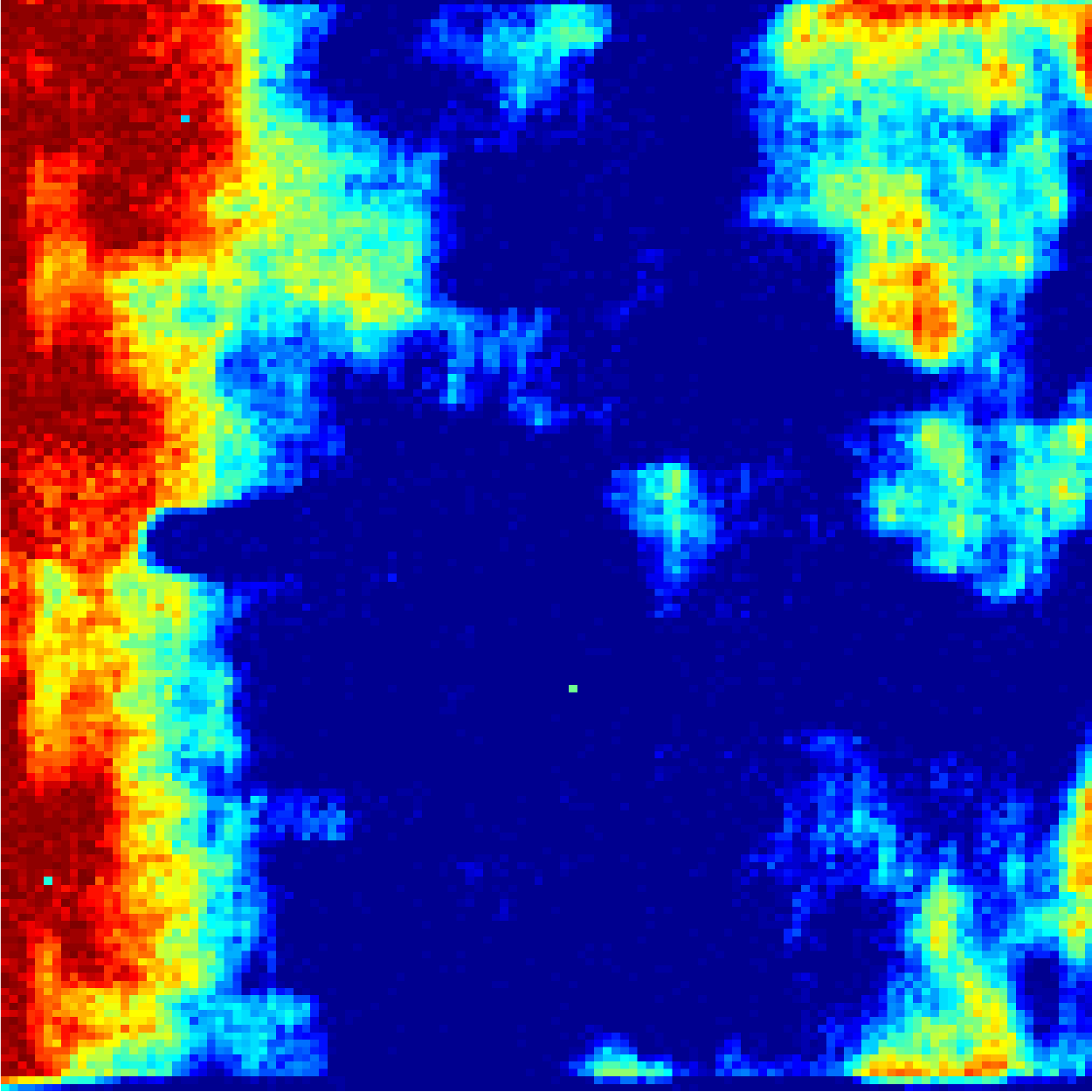}
				\captionsetup{justification=centering,labelformat=empty,skip=2pt}
				\caption{(c) PM-LDA: \\sea grass}
			\end{subfigure}
			\begin{subfigure}[t]{0.15\textwidth}
				\includegraphics[width=1\linewidth,height=0.8\linewidth]{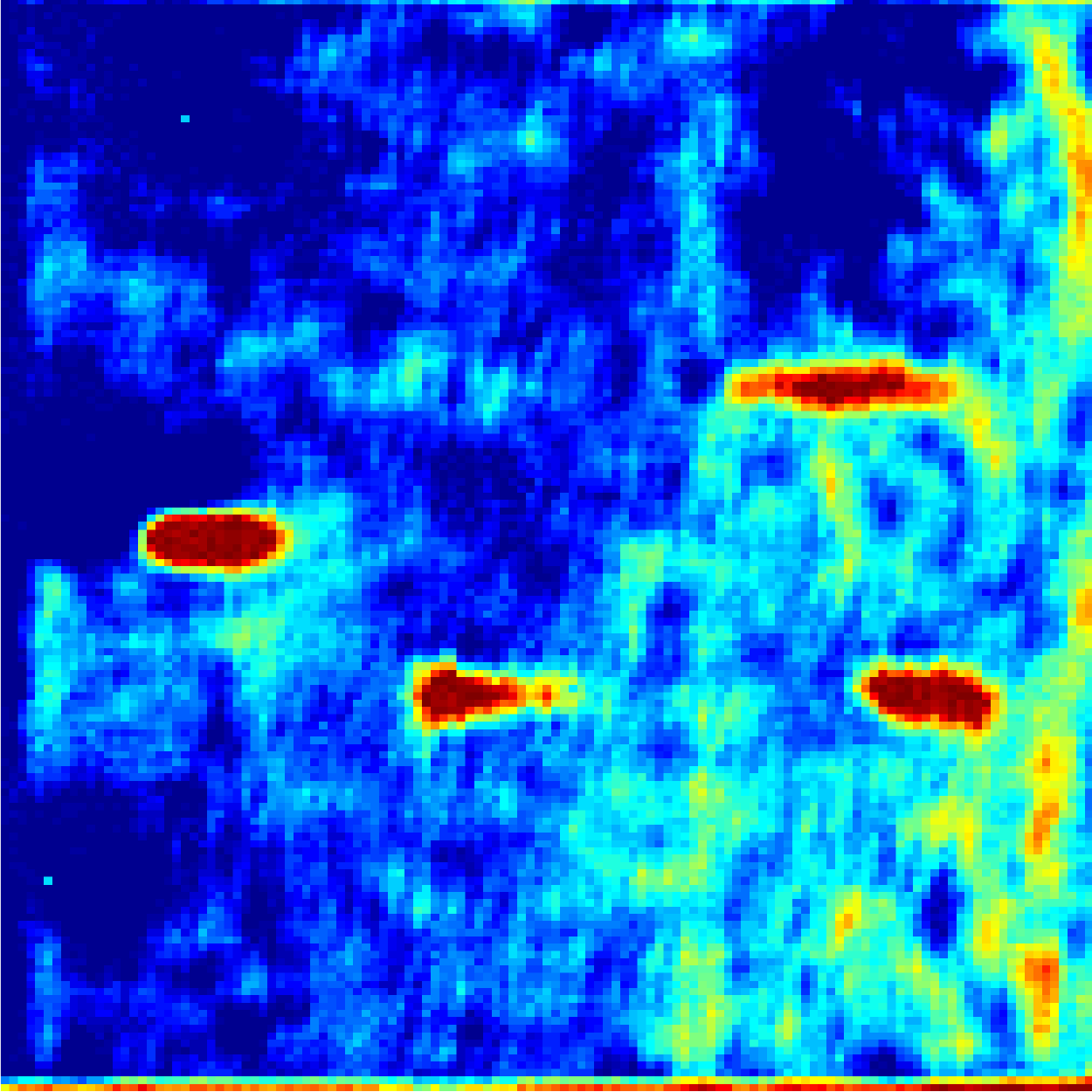}
				\captionsetup{justification=centering,labelformat=empty,skip=2pt}
				\caption{(d) PM-LDA:\\dark flat sand}
			\end{subfigure}	
			\begin{subfigure}[t]{0.15\textwidth}
				\includegraphics[width=1\linewidth,height=0.8\linewidth]{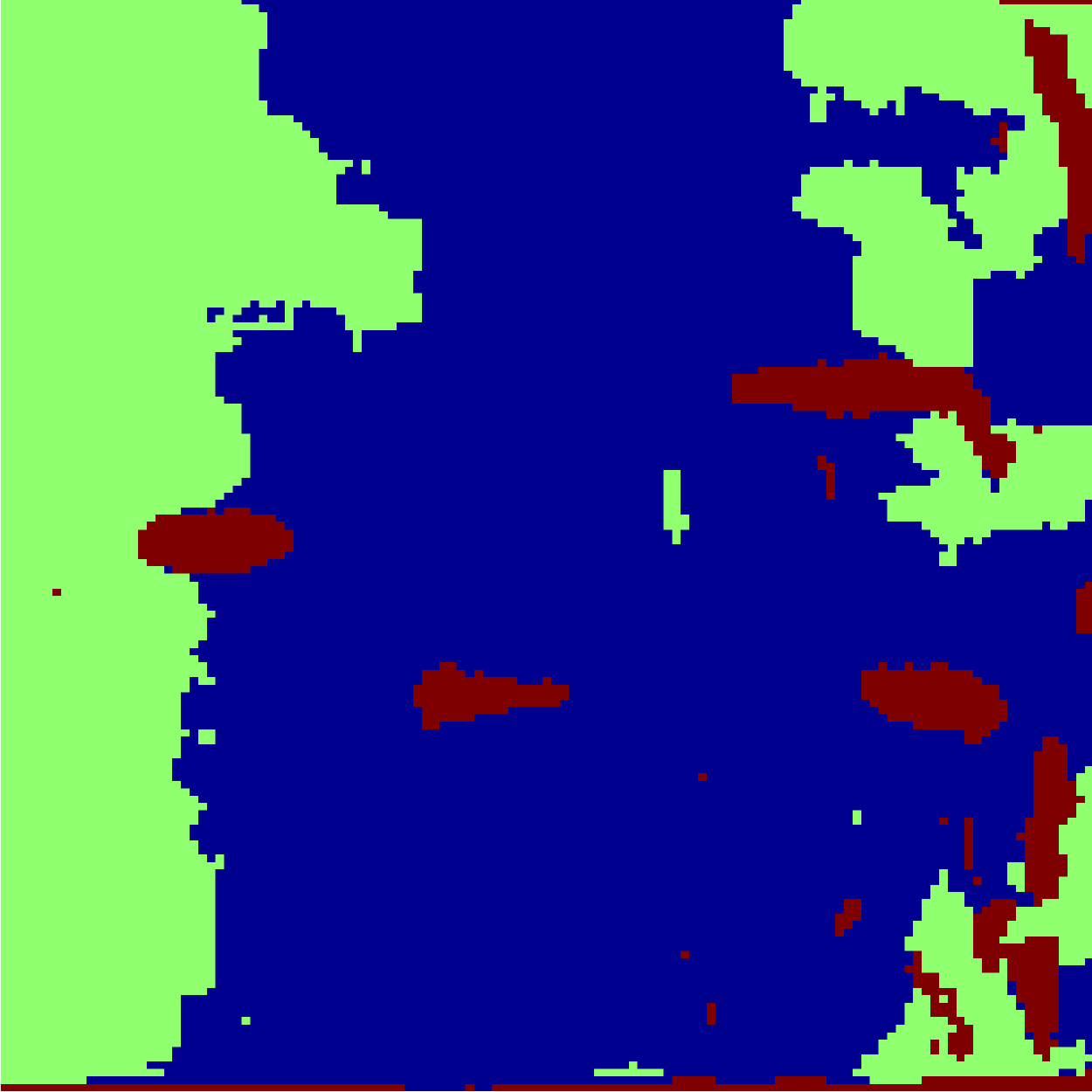}
				\captionsetup{justification=centering,labelformat=empty,skip=2pt}
				\caption{(e) LDA }
				\end{subfigure}
			\begin{subfigure}[t]{0.15\textwidth}
				\includegraphics[width=1\linewidth,height=0.8\linewidth]{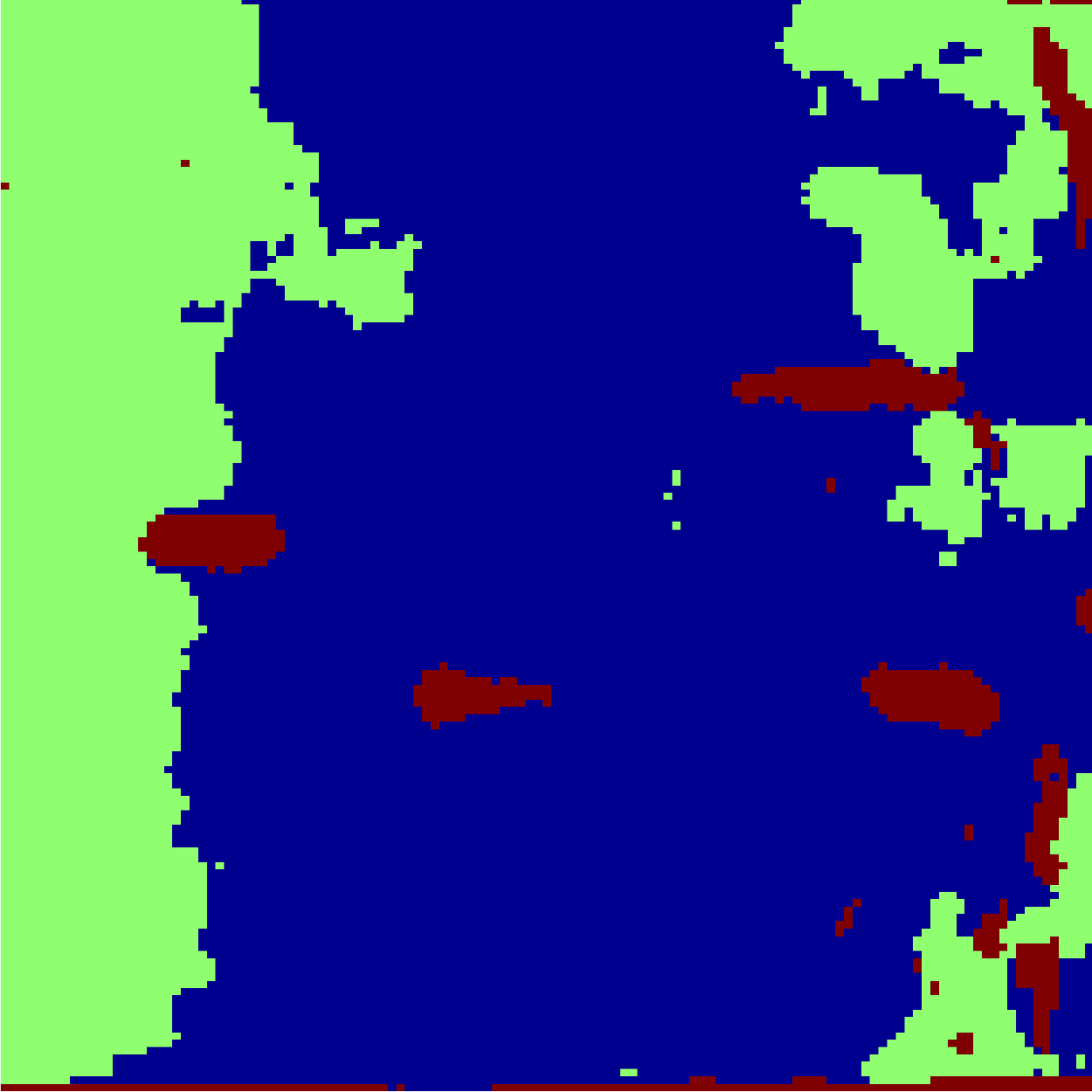}
				\captionsetup{justification=centering,labelformat=empty,skip=2pt}
				\caption{(f) sLDA }
			\end{subfigure}						
		\caption{Segmentation results of PM-LDA, LDA, and sLDA. Column (a) represents the original images HF-00 to HF-04 with super-pixel boundaries. Column (b)-(d) are the PM-LDA results, which represent the partial membership maps in ``sand ripple'', ``sea grass''  and ``dark flat sand'' topic, respectively. Column (e) and (f) are the LDA and sLDA results, respectively. The color indicates the topic number.}
		\label{fig:pmlda-lda-slda}
\end{figure*}

In this experiment, we compare PM-LDA with LDA and sLDA by running the parameter estimation on five SONAR images simultaneously. For PM-LDA, the hyper-parameter $\lambda$ and $\boldsymbol{\alpha}$ is set to be $0.5$ and $\mathbf{1}_K$. For sLDA, $\sigma$ is set to be $0.1$ and each document is repeated for $4$ times. The topic number is set to $K=3$. The segmentation results of PM-LDA, LDA, and sLDA are shown in Fig. \ref{fig:pmlda-lda-slda}.  Column (a) are the five complete SONAR images HF-00 to HF-04 with super-pixel boundaries. Column (b)-(d) are the PM-LDA results, which represent the partial membership maps in ``sand ripple'', ``sea grass''  and ``dark flat sand'' topic, respectively. Column (e) and (f) are the LDA and sLDA results, respectively. The color indicates the topic number. As shown in Fig. \ref{fig:pmlda-lda-slda}, PM-LDA achieves similar segmentation results to LDA and sLDA on SONAR imagery. All of them are able to learn the ``sand ripple'',  ``sea grass'', and  ``dark flat sand'' topics, and for each topic, the corresponding regions localized by PM-LDA, LDA, and sLDA are similar to each other. For example, ``sand ripple'' region learned by PM-LDA  (the red regions in Column (b)  are almost the same as ``sand ripple'' region learned by LDA and sLDA (the dark blue regions in Column (e) and (f)).  However, PM-LDA has an exclusive capability of localizing the gradual transition regions between different topics while both LDA and sLDA only provide crispy boundaries.  These partial membership values mostly occur at the boundary between two topics. Thus, PM-LDA is able to identify when the feature vector contains information from multiple topics (as the feature vector is being computed over a window that contains more than one topic).  This is a powerful result showing the effectiveness of PM-LDA in providing semantic image understanding.

\subsubsection{Varying scaling factor $s$}

\begin{figure}[htb]
	\centering
	\includegraphics[height=0.3\linewidth]{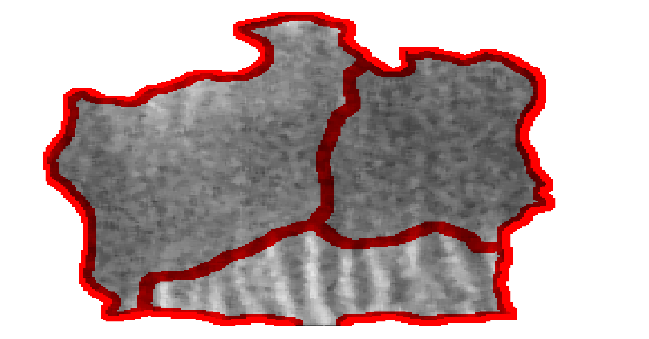}
	\caption{A subregion of three superpixels}
	\label{fig:3sup}
	\vspace{-2mm}
\end{figure}

\begin{figure}[!htb]
	\centering
	\begin{subfigure}[t]{0.15\textwidth}
		\centering
		\includegraphics[height=0.5\linewidth]{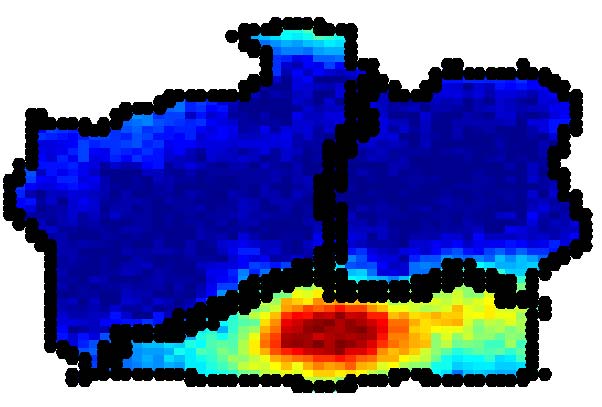}
		\captionsetup{labelformat=empty,skip=0pt}
		\caption{(a) $s=3$}
	\end{subfigure} 
	\begin{subfigure}[t]{0.15\textwidth}
		\centering
		\includegraphics[height=0.5\linewidth]{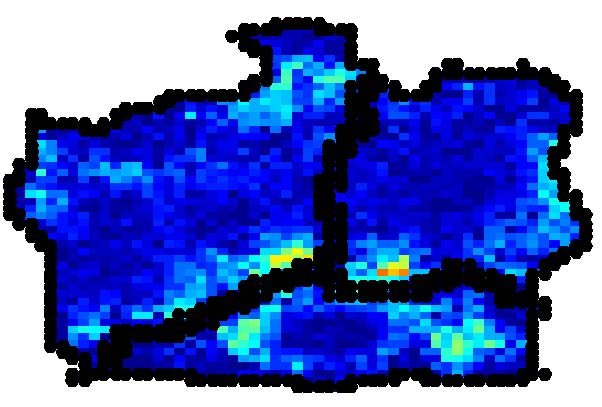}
		\captionsetup{labelformat=empty,skip=0pt}
		\caption{(b) }
	\end{subfigure} 
	\begin{subfigure}[t]{0.15\textwidth}
		\centering
		\includegraphics[height=0.5\linewidth]{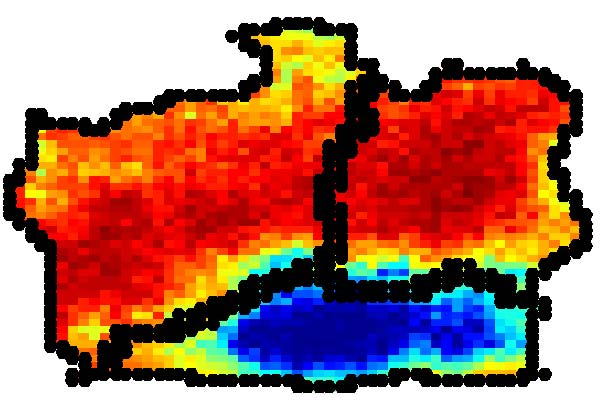}
		\captionsetup{labelformat=empty,skip=0pt}
		\caption{(c) }
	\end{subfigure}
	
	\begin{subfigure}[t]{0.15\textwidth}
		\centering
		\includegraphics[height=0.5\linewidth]{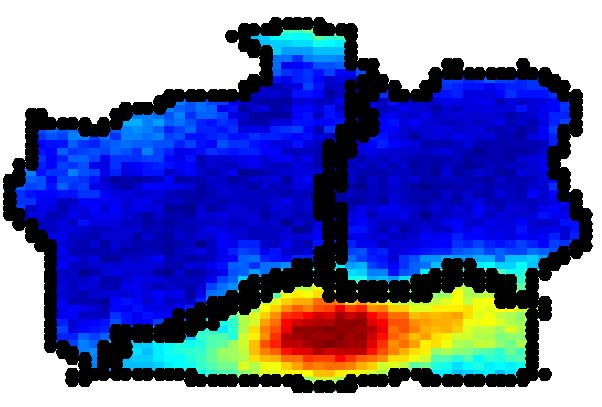}
		\captionsetup{labelformat=empty,skip=0pt}
		\caption{(a) $s=10$}
	\end{subfigure}
	\begin{subfigure}[t]{0.15\textwidth}
		\centering
		\includegraphics[height=0.5\linewidth]{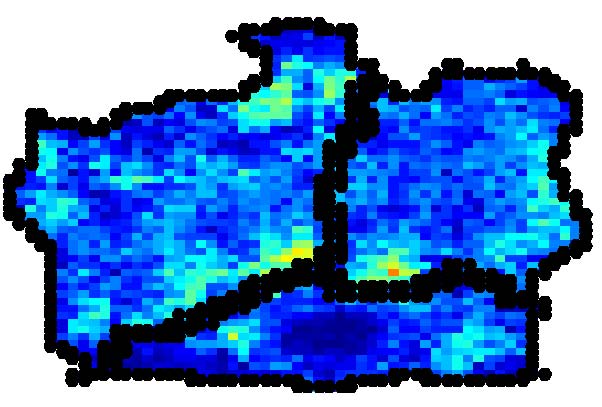}
		\captionsetup{labelformat=empty,skip=0pt}
		\caption{(b)}
	\end{subfigure}
	\begin{subfigure}[t]{0.15\textwidth}
		\centering
		\includegraphics[height=0.5\linewidth]{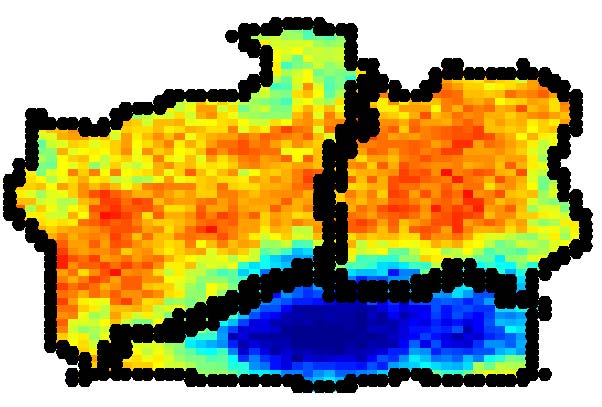}
		\captionsetup{labelformat=empty,skip=0pt}
		\caption{(c)}
	\end{subfigure}
	
	\begin{subfigure}[t]{0.15\textwidth}
		\centering
		\includegraphics[height=0.5\linewidth]{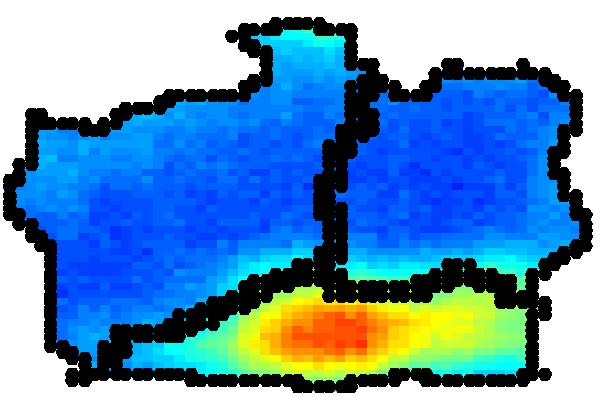}
		\captionsetup{labelformat=empty,skip=0pt}
		\caption{(a) $s=300$}
	\end{subfigure}
	\begin{subfigure}[t]{0.15\textwidth}
		\centering
		\includegraphics[height=0.5\linewidth]{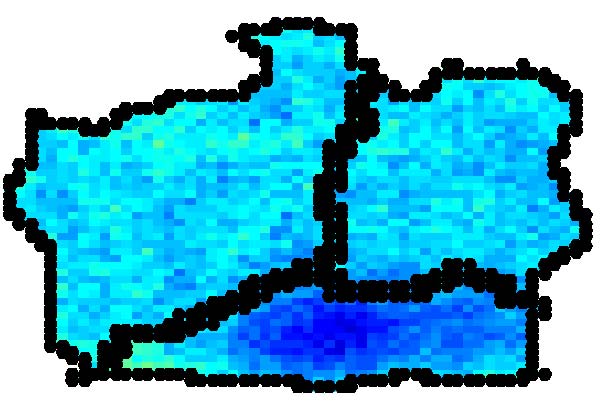}
		\captionsetup{labelformat=empty,skip=0pt}
		\caption{(b) }
	\end{subfigure}
	\begin{subfigure}[t]{0.15\textwidth}
		\centering
		\includegraphics[height=0.5\linewidth]{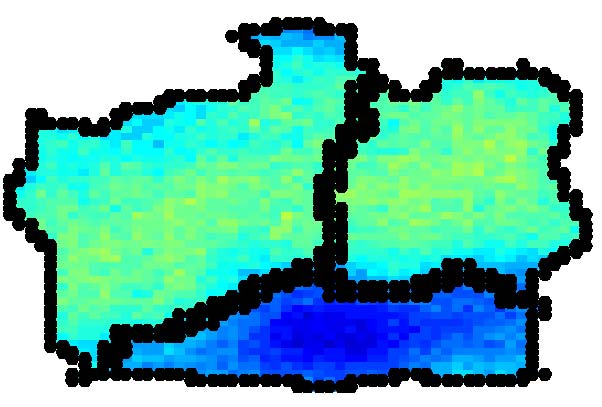}
		\captionsetup{labelformat=empty,skip=0pt}
		\caption{(c) }
	\end{subfigure}
	
	\begin{subfigure}[t]{0.15\textwidth}
		\centering
		\includegraphics[height=0.5\linewidth]{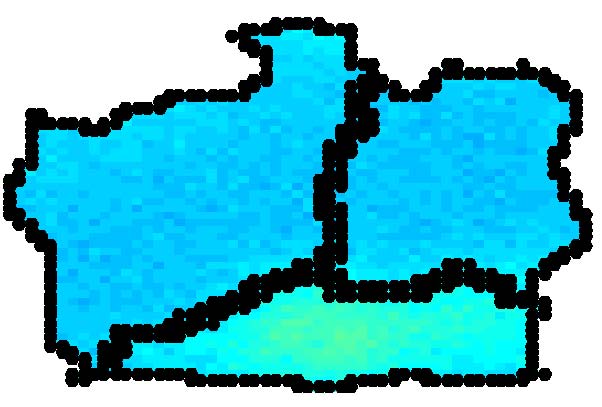}
		\captionsetup{labelformat=empty,skip=0pt}
		\caption{(a) $s=30000$}
	\end{subfigure}
	\begin{subfigure}[t]{0.15\textwidth}
		\centering
		\includegraphics[height=0.5\linewidth]{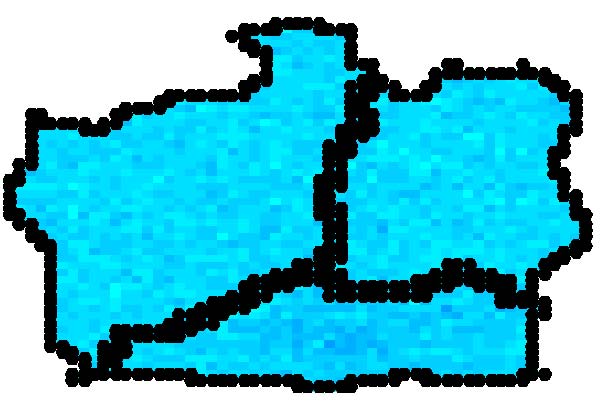}
		\captionsetup{labelformat=empty,skip=0pt}
		\caption{(b)}
	\end{subfigure}
	\begin{subfigure}[t]{0.15\textwidth}
		\centering
		\includegraphics[height=0.5\linewidth]{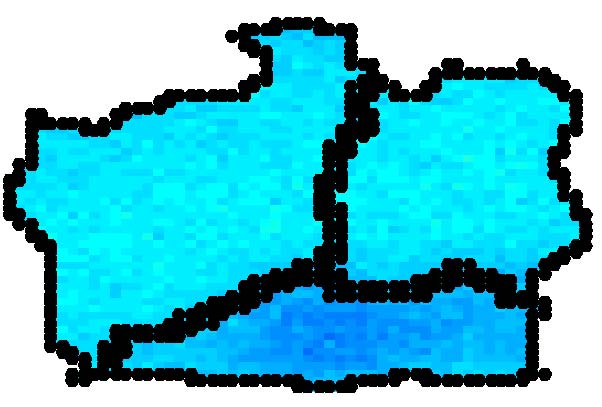}
		\captionsetup{labelformat=empty,skip=0pt}
		\caption{(c) }
	\end{subfigure}	
	
	\caption[Partial membership maps  with varying $s$]{Partial membership maps  with varying $s$. Each row shows the estimated membership maps of the three estimated topics. The black contour indicates the superpixel boundary. The superpixels are results published in \cite{cobb:2013multi}.}
	\label{fig:varyings}
\end{figure}

As discussed in Section \ref{sec:PMLDA}, the scaling factor $s_d$ determines the similarity of the partial membership vector of each word, $\mathbf{z}_{dn}$, to the topic proportion $\boldsymbol{\pi}_d$. In this experiment, we investigated the effect of $s$ by estimating the memberships and topics with fixed topic proportion. A subregion consisting of three superpixels \cite{cobb:2013multi} is used in this experiment and shown in Fig. \ref{fig:3sup}. Each superpixel is treated as a document. The topic proportion $\boldsymbol{\pi}_d$ is set to be $[1,1,1]/3$ and the scaling factor $s$ is varied to be $3, 10, 300, 30000$. The membership estimation results are shown in Figure \ref{fig:varyings}. Column (a)-(c) represent the membership maps in ``sand ripple'', ``sea-grass'', and ``dark flat sand'' topic, respectively.  In this set of superpixels, there are no regions of ``sea-grass'' which PM-LDA correctly determines.  As can be seen, within each superpixel, as the scaling factor $s$ increases, the partial memberships gradually approach the topic proportion $[1,1,1]/3$ and the membership map becomes more smooth. This can be considered as a way of incorporating spatial information. Visual words in a superpixel are spatial neighbors. They are assumed to have similar attributes and should have similar membership vectors. By using superpixels as documents and increasing the value of scaling factor $s$, spatial information is implicitly incorporated to PM-LDA.  

\subsubsection{Varying the number of superpixels} 
\begin{figure*}[htb!]
	\centering
		\begin{subfigure}[t]{0.15\textwidth}
			\centering
			\includegraphics[height=0.9\linewidth]{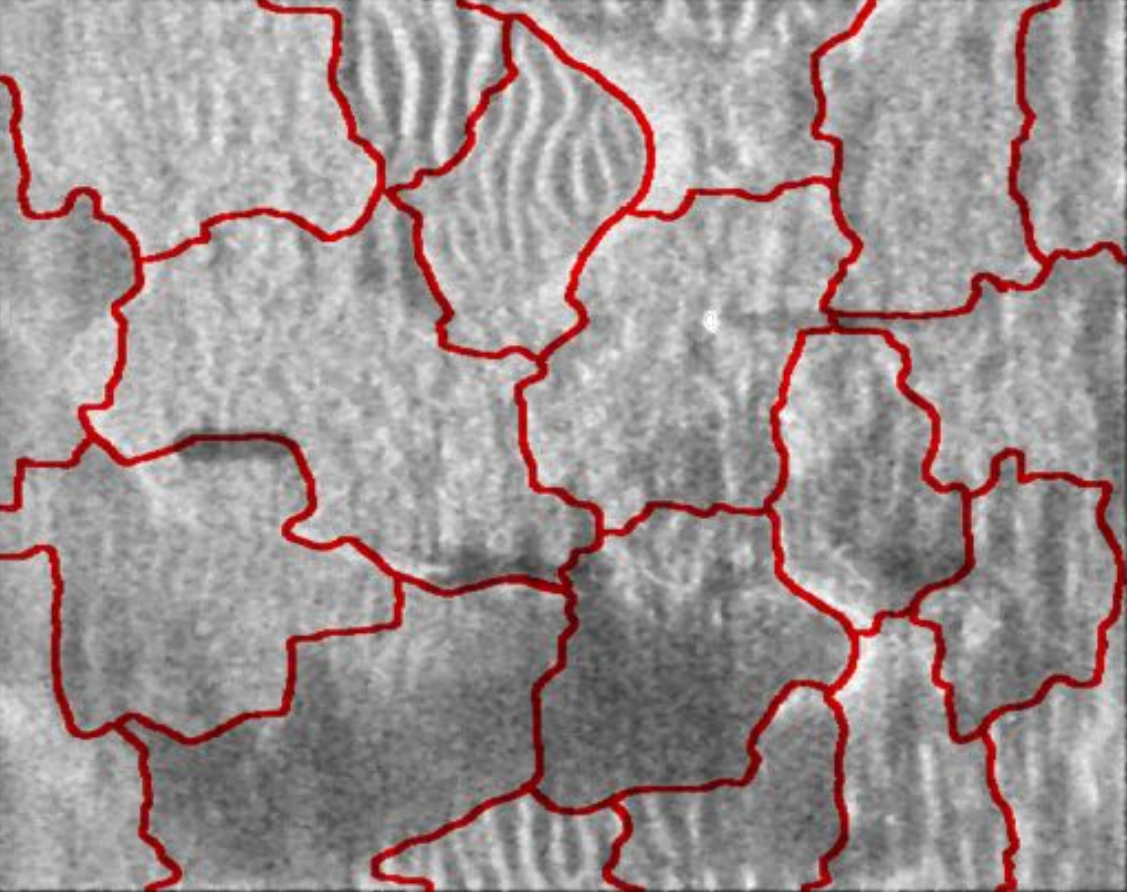}
			\captionsetup{labelformat=empty,skip=0pt}
			\caption{ 20 superpixels}
		\end{subfigure}	
	\begin{subfigure}[t]{0.15\textwidth}
		\centering
		\includegraphics[height=0.9\linewidth]{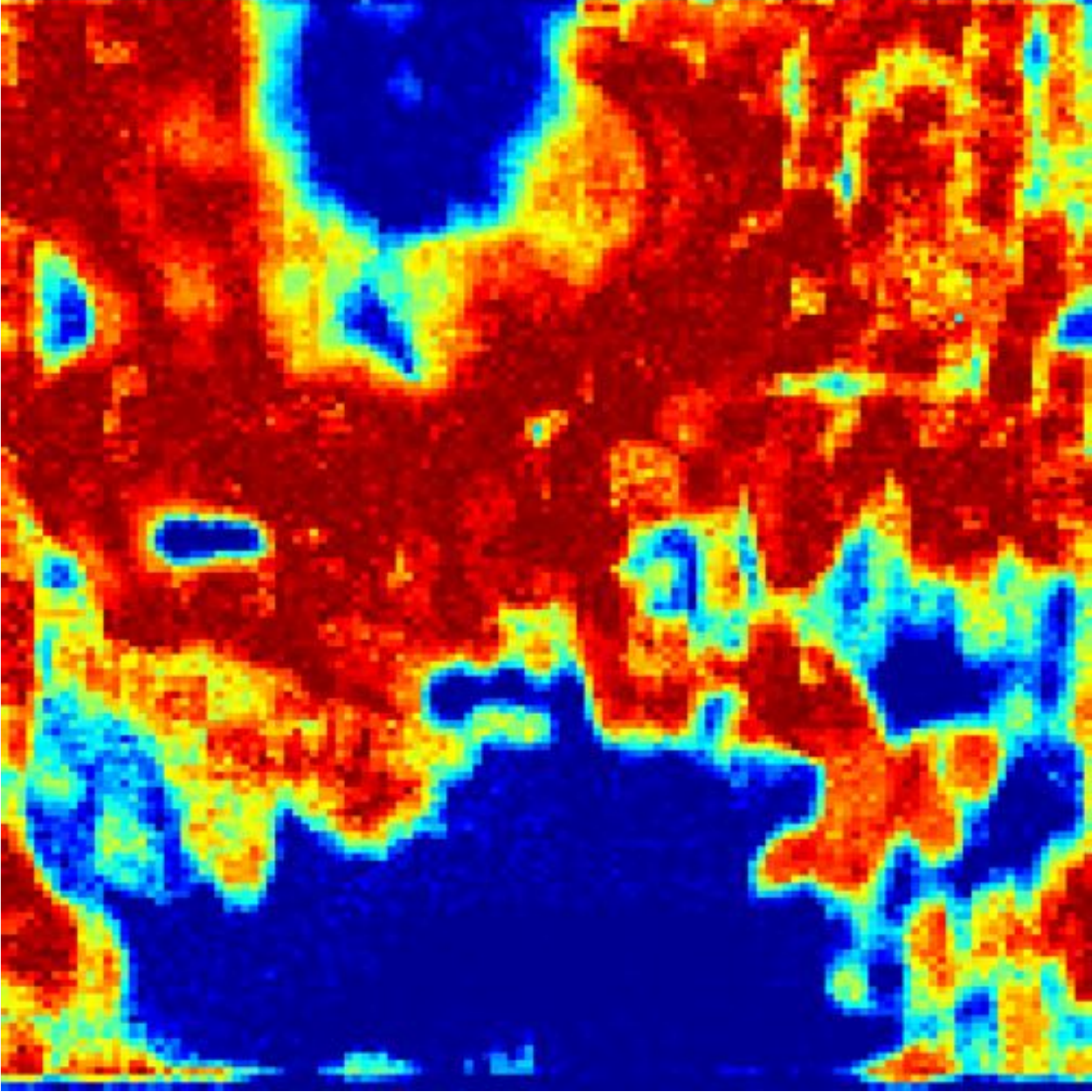}
		\captionsetup{labelformat=empty,skip=0pt}
		\caption{}
	\end{subfigure}	
	\begin{subfigure}[t]{0.15\textwidth}
		\centering
		\includegraphics[height=0.9\linewidth]{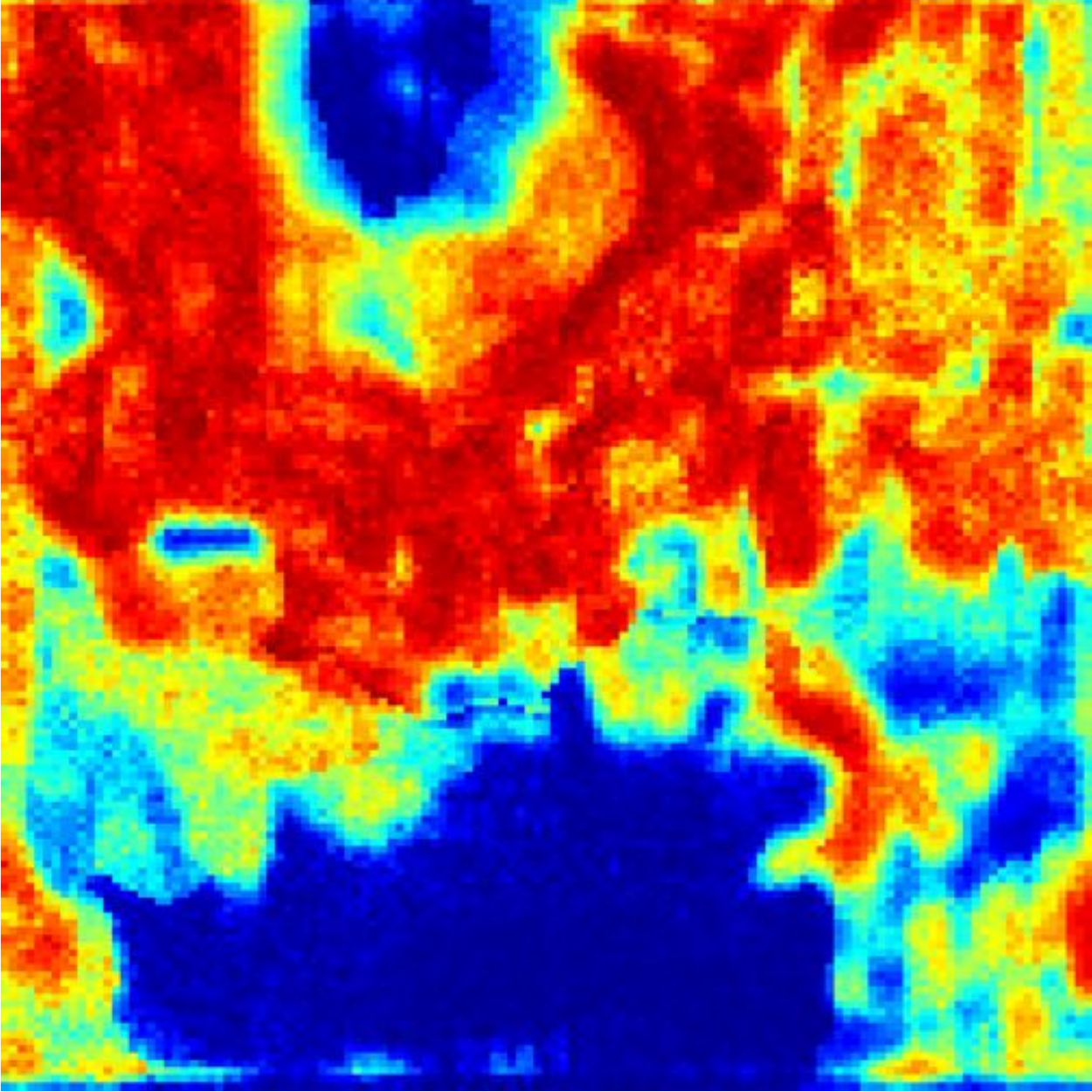}
		\captionsetup{labelformat=empty,skip=0pt}
		\caption{}
	\end{subfigure}	
	\begin{subfigure}[t]{0.15\textwidth}
		\centering
		\includegraphics[height=0.9\linewidth]{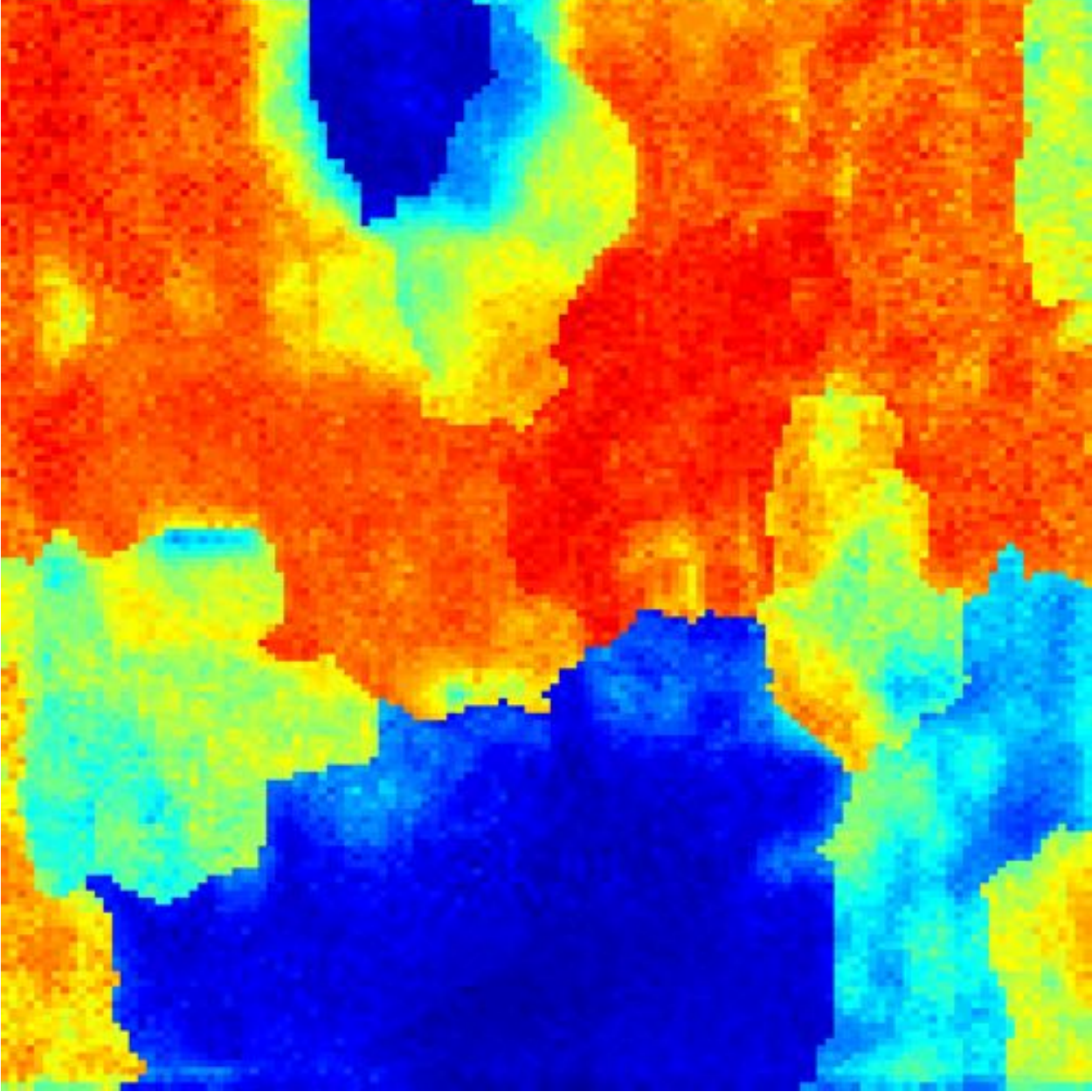}
		\captionsetup{labelformat=empty,skip=0pt}
		\caption{}
	\end{subfigure}	
	\begin{subfigure}[t]{0.15\textwidth}
		\centering
		\includegraphics[height=0.9\linewidth]{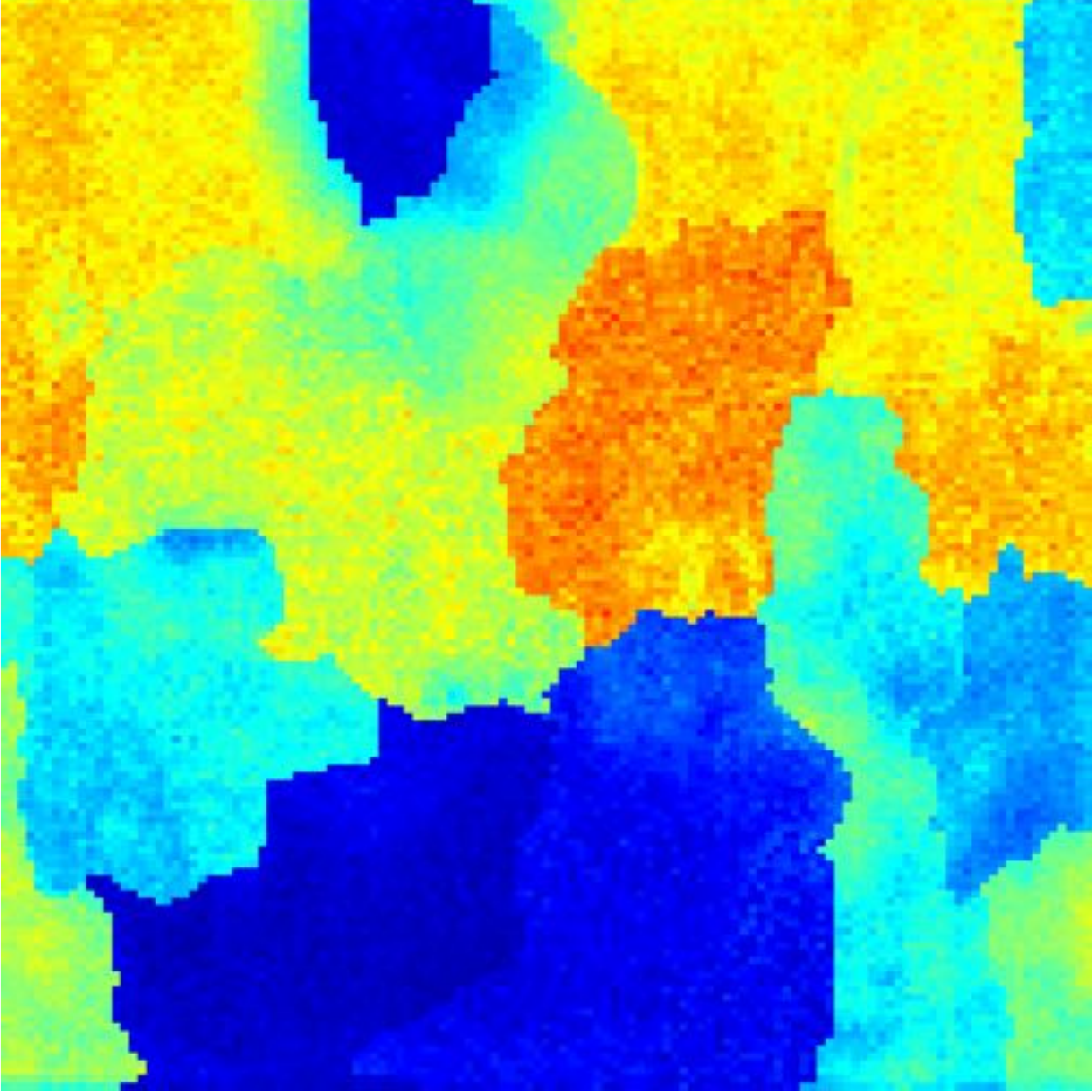}
		\captionsetup{labelformat=empty,skip=0pt}
		\caption{}
	\end{subfigure}
	
		\begin{subfigure}[t]{0.15\textwidth}
			\centering
			\includegraphics[height=0.9\linewidth]{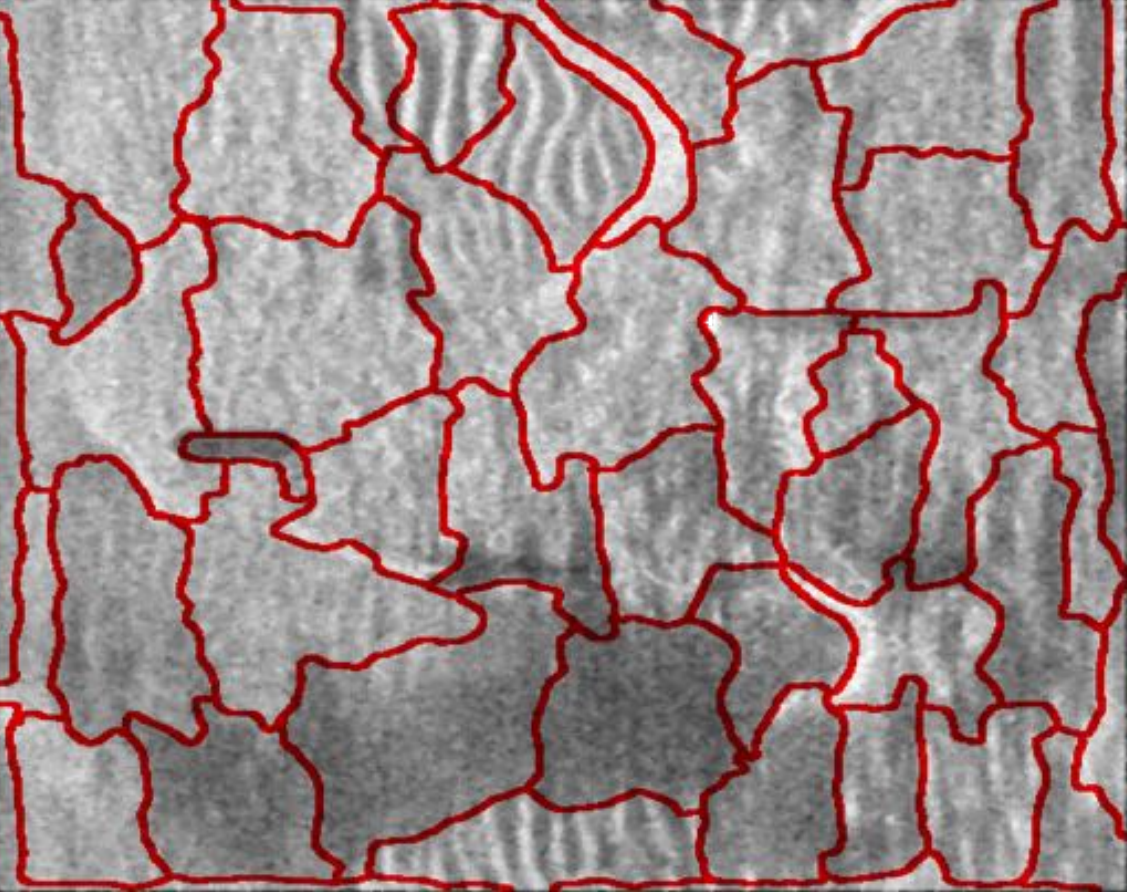}
			\captionsetup{labelformat=empty,skip=0pt}
			\caption{ 48 superpixels}
		\end{subfigure}	
	\begin{subfigure}[t]{0.15\textwidth}
		\centering
		\includegraphics[height=0.9\linewidth]{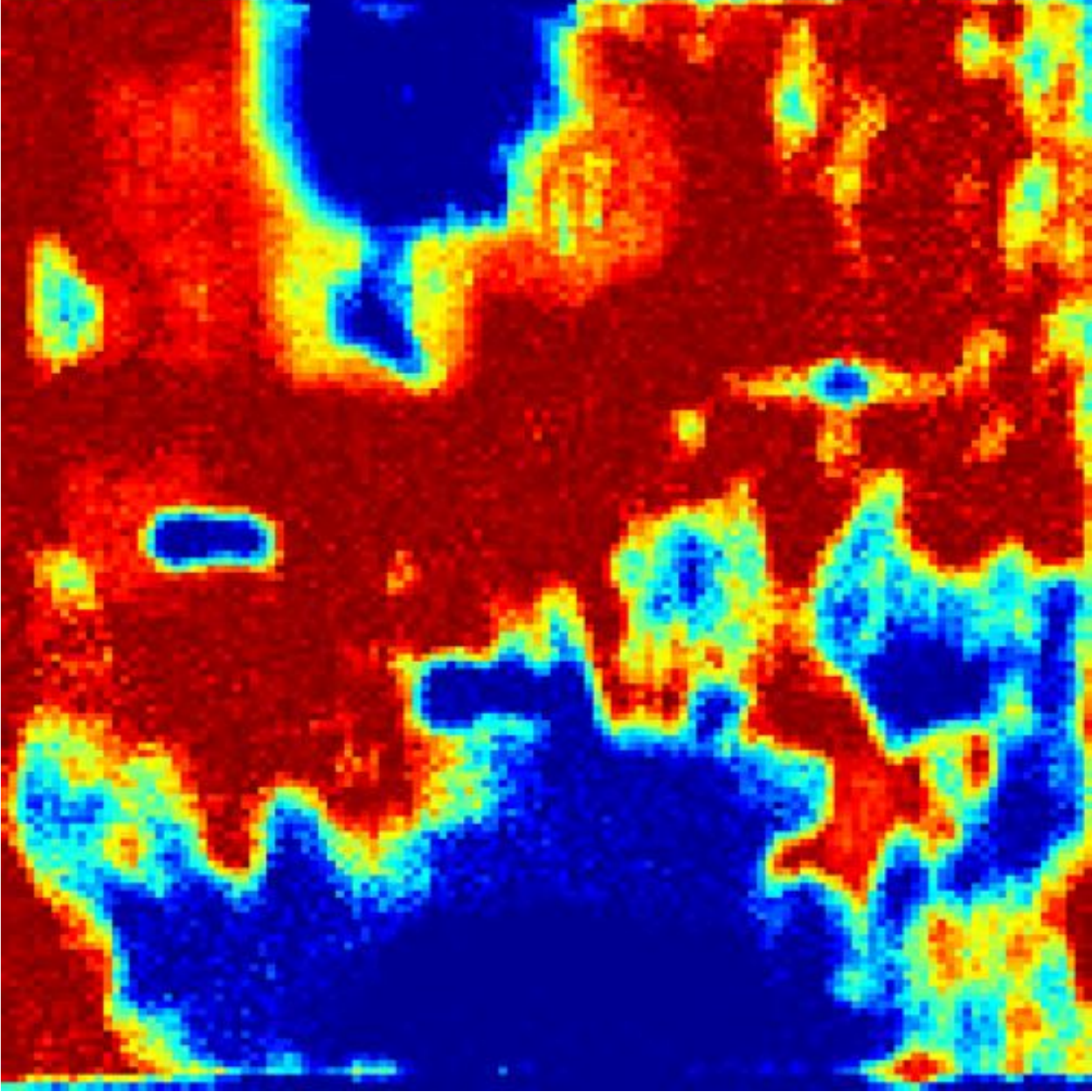}
		\captionsetup{labelformat=empty,skip=0pt}
		\caption{}
	\end{subfigure}	
	\begin{subfigure}[t]{0.15\textwidth}
		\centering
		\includegraphics[height=0.9\linewidth]{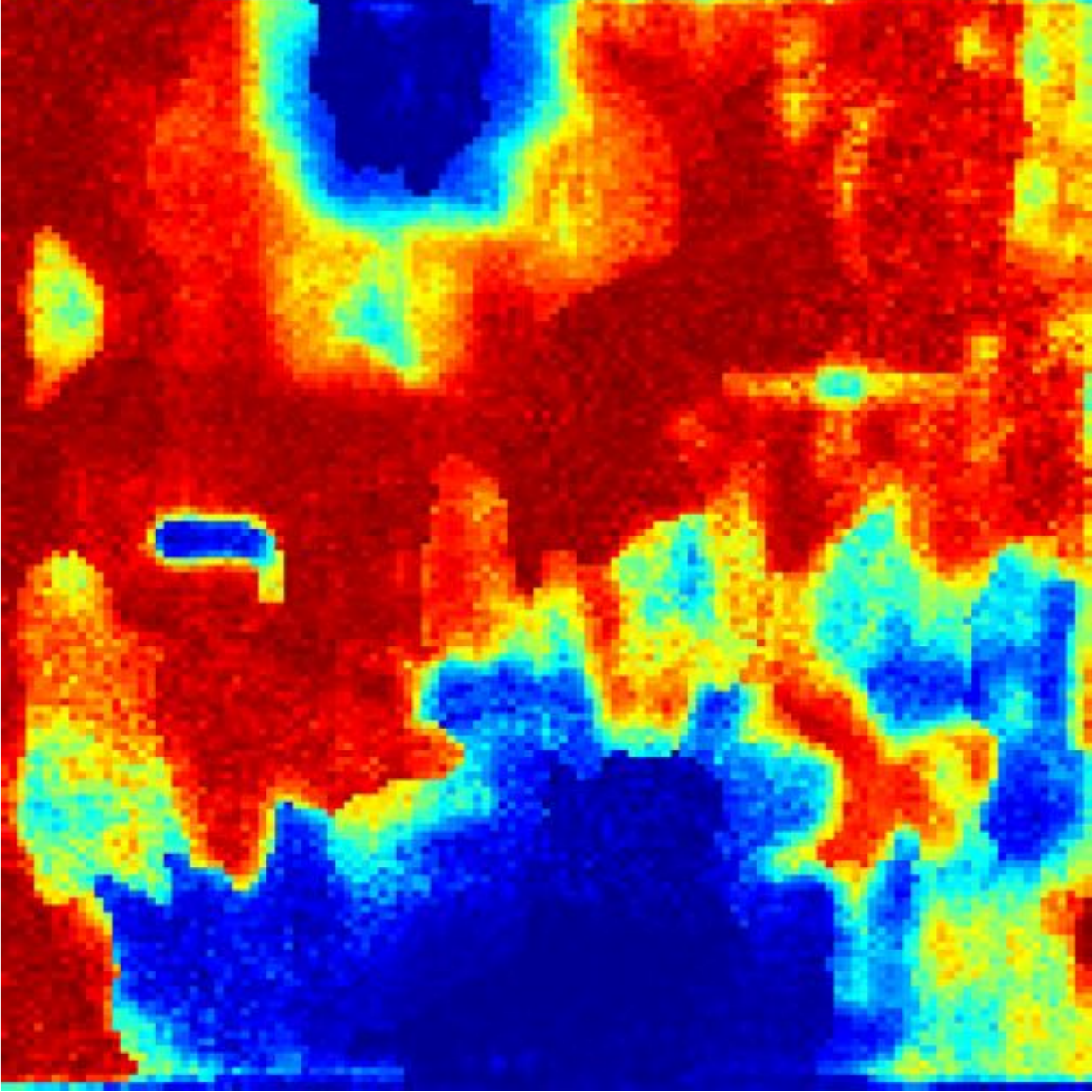}
		\captionsetup{labelformat=empty,skip=0pt}
		\caption{}
	\end{subfigure}	
	\begin{subfigure}[t]{0.15\textwidth}
		\centering
		\includegraphics[height=0.9\linewidth]{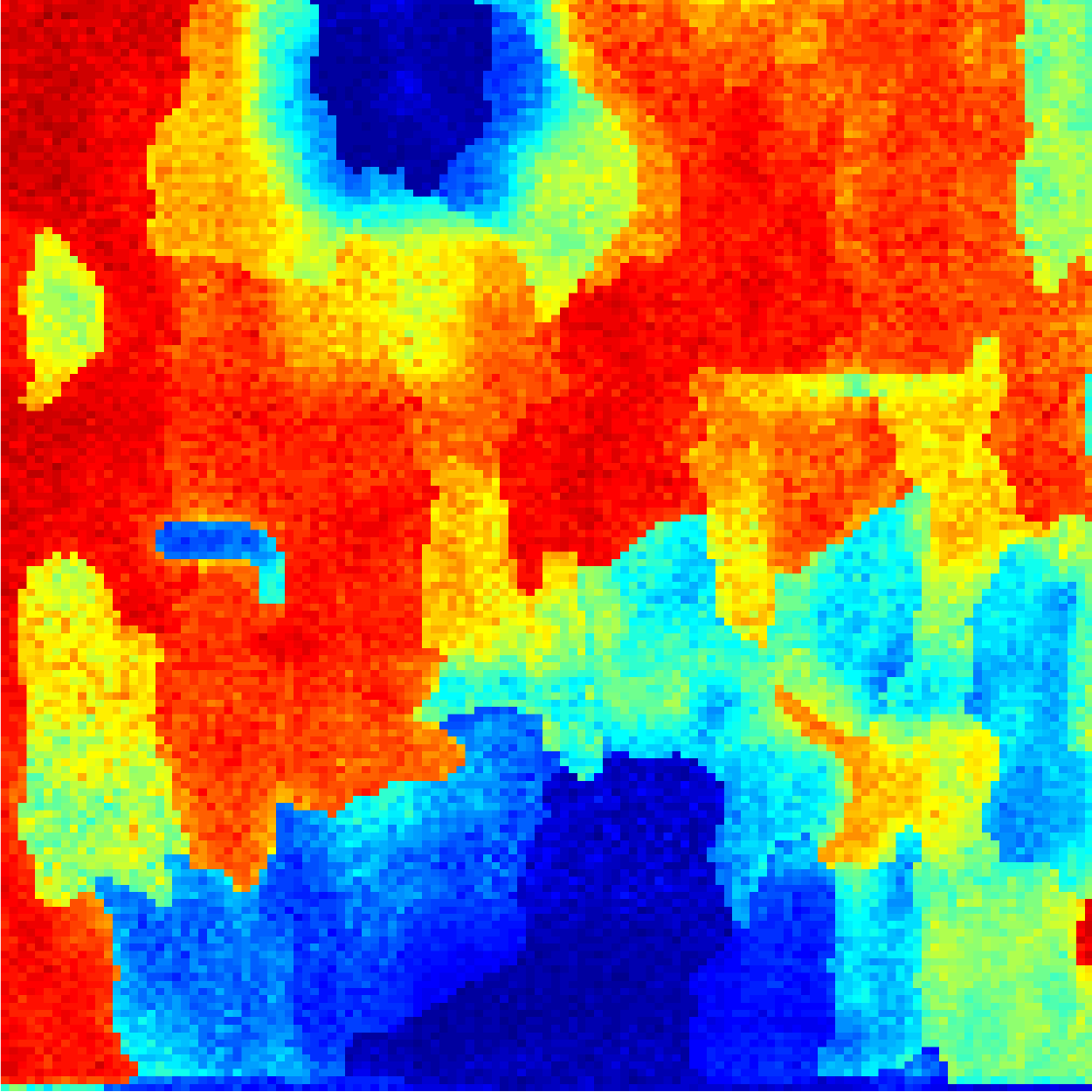}
		\captionsetup{labelformat=empty,skip=0pt}
		\caption{}
	\end{subfigure}	
	\begin{subfigure}[t]{0.15\textwidth}
		\centering
		\includegraphics[height=0.9\linewidth]{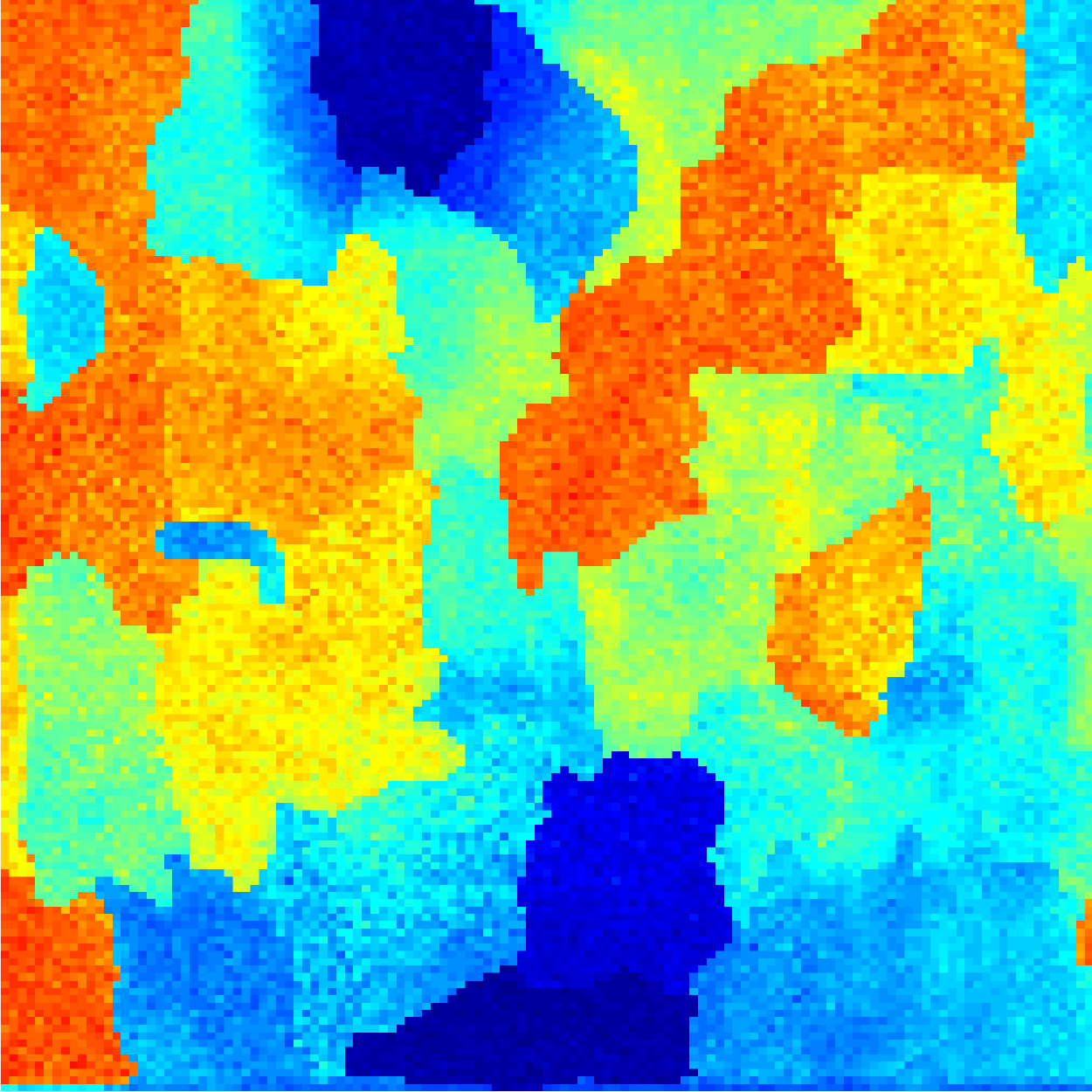}
		\captionsetup{labelformat=empty,skip=0pt}
		\caption{}
	\end{subfigure}
	
		\begin{subfigure}[t]{0.15\textwidth}
			\centering
			\includegraphics[height=0.9\linewidth]{HF_00_Seg200.pdf}
			\captionsetup{labelformat=empty,skip=0pt}
			\caption{ 200 superpixels}
		\end{subfigure}	
	\begin{subfigure}[t]{0.15\textwidth}
		\centering
		\includegraphics[height=0.9\linewidth]{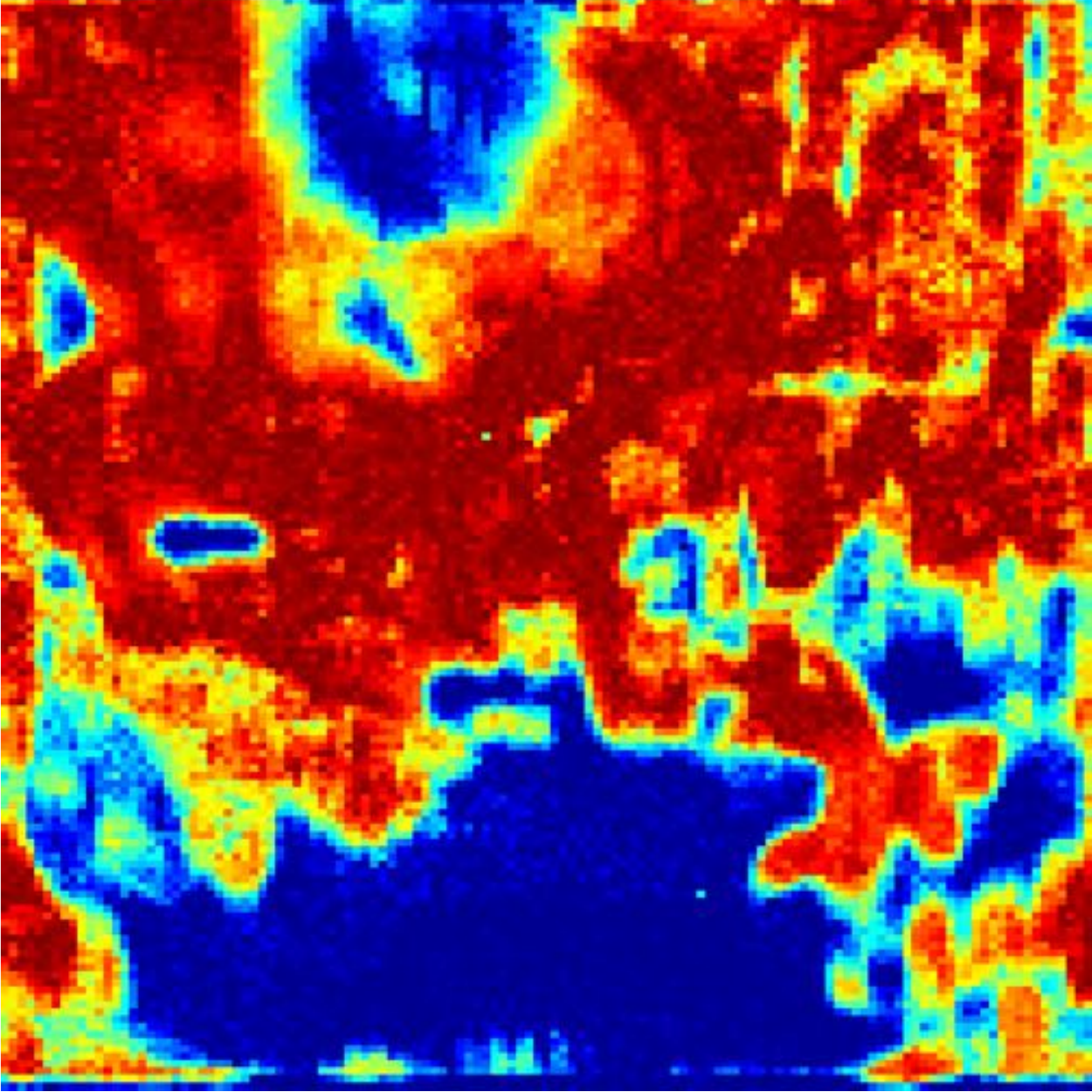}
		\captionsetup{labelformat=empty,skip=0pt}
		\caption{s=1}
	\end{subfigure}	
	\begin{subfigure}[t]{0.15\textwidth}
		\centering
		\includegraphics[height=0.9\linewidth]{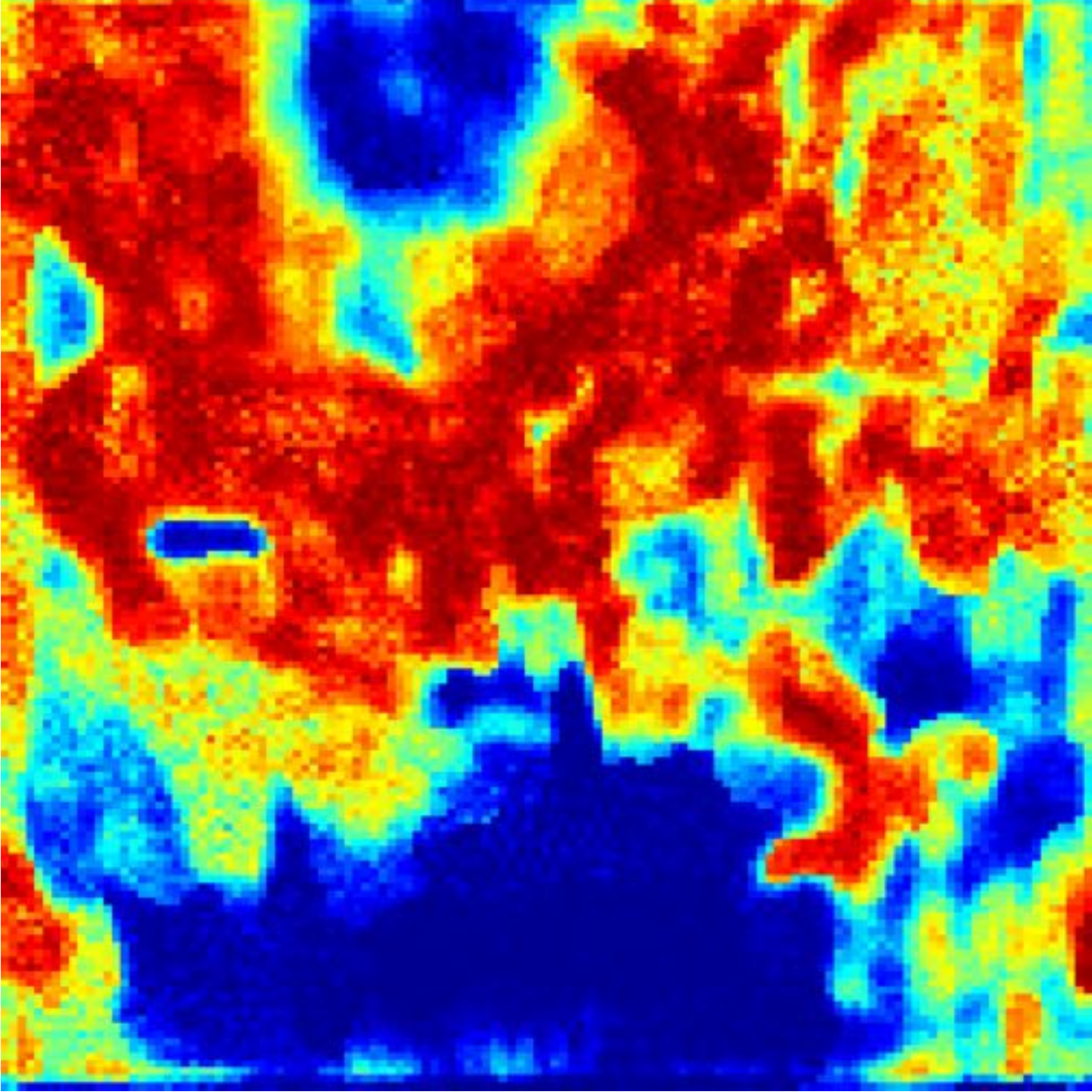}
		\captionsetup{labelformat=empty,skip=0pt}
		\caption{s=100}
	\end{subfigure}	
	\begin{subfigure}[t]{0.15\textwidth}
		\centering
		\includegraphics[height=0.9\linewidth]{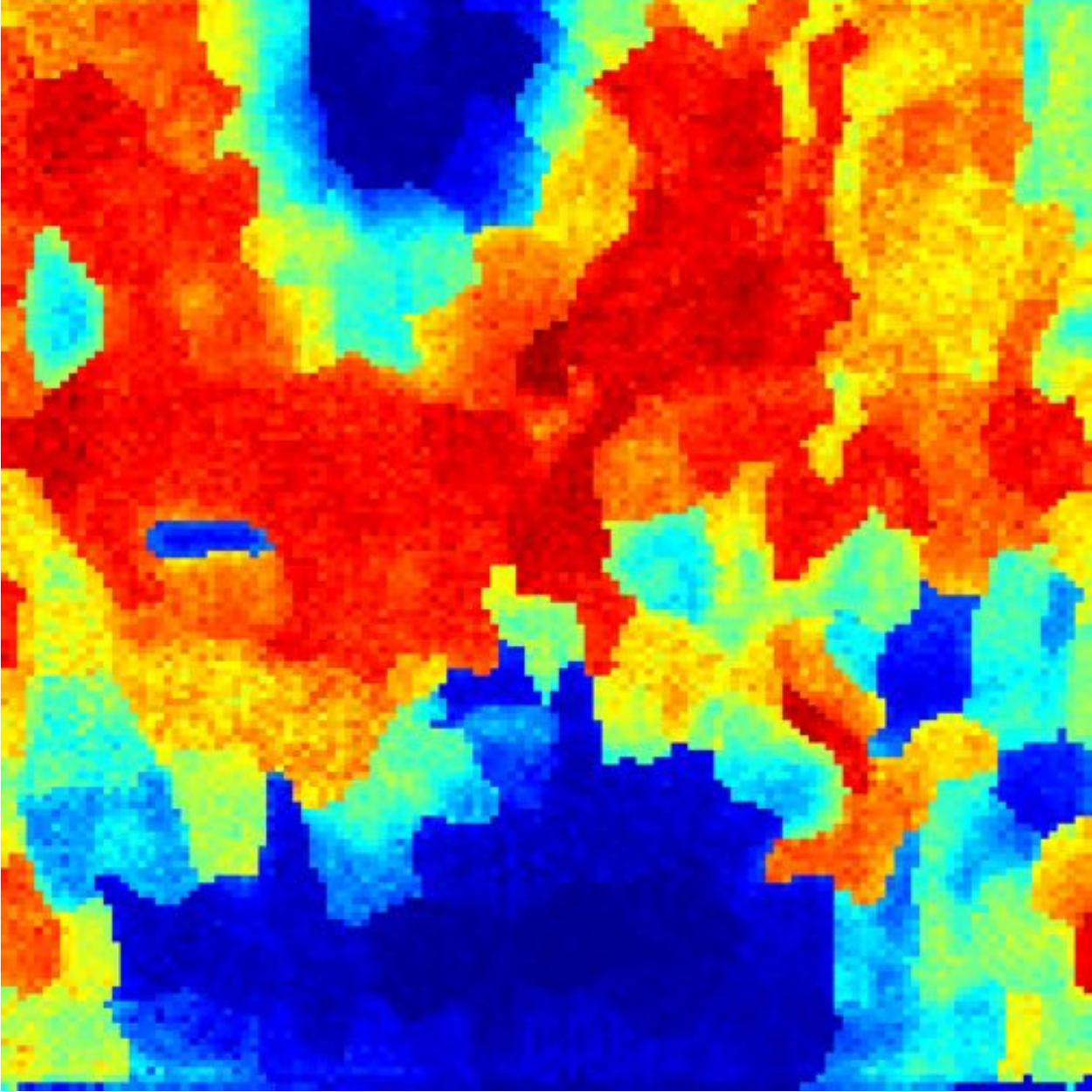}
		\captionsetup{labelformat=empty,skip=0pt}
		\caption{s=500}
	\end{subfigure}	
	\begin{subfigure}[t]{0.15\textwidth}
		\centering
		\includegraphics[height=0.9\linewidth]{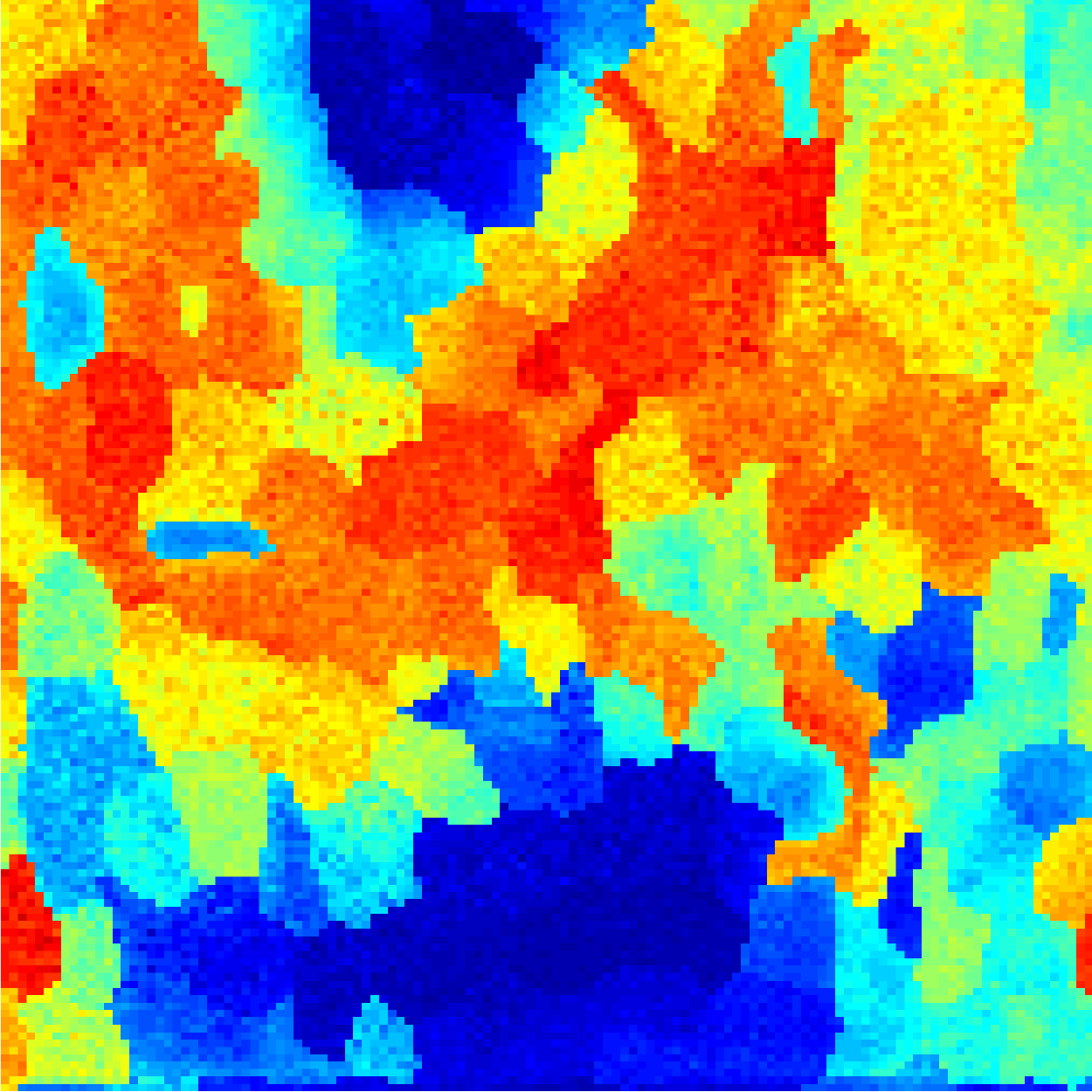}
		\captionsetup{labelformat=empty,skip=0pt}
		\caption{s=1000}
	\end{subfigure}
	\caption{Partial membership maps for HF-00 with varying number of superpixels and varying scaling factor $s$ for topic 1 - sea grass. Column 2 to 5 correspond to scaling factor $s=1, 100, 500, 1000$, respectively.}
	\label{fig:hf00_1}
\end{figure*}

\begin{figure*}[htb!]
	\centering	
	\begin{subfigure}[t]{0.15\textwidth}
		\centering
		\includegraphics[height=0.9\linewidth]{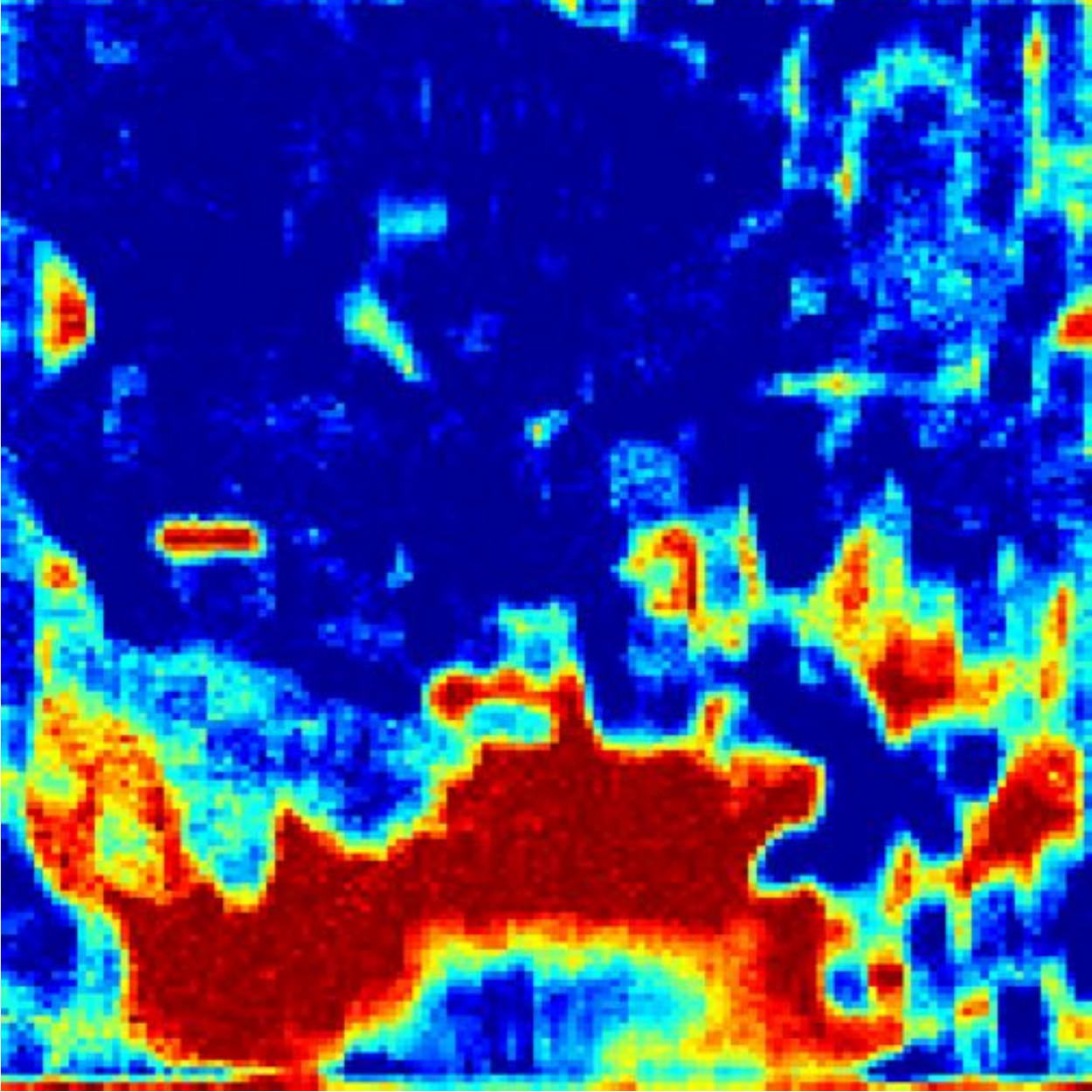}
		\captionsetup{labelformat=empty,skip=0pt}
		\caption{20 superpixels}
	\end{subfigure}	
	\begin{subfigure}[t]{0.15\textwidth}
		\centering
		\includegraphics[height=0.9\linewidth]{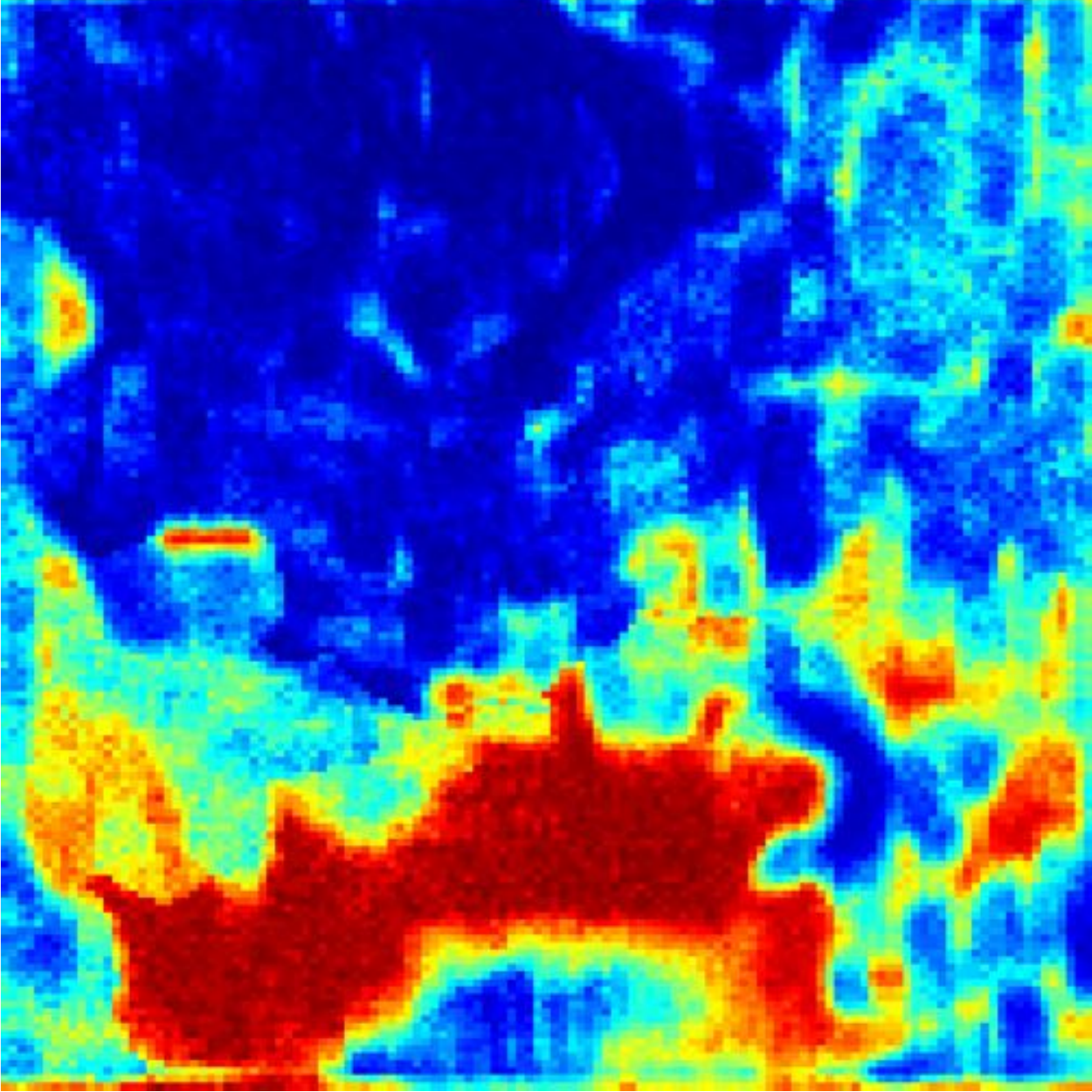}
		\captionsetup{labelformat=empty,skip=0pt}
		\caption{}
	\end{subfigure}	
	\begin{subfigure}[t]{0.15\textwidth}
		\centering
		\includegraphics[height=0.9\linewidth]{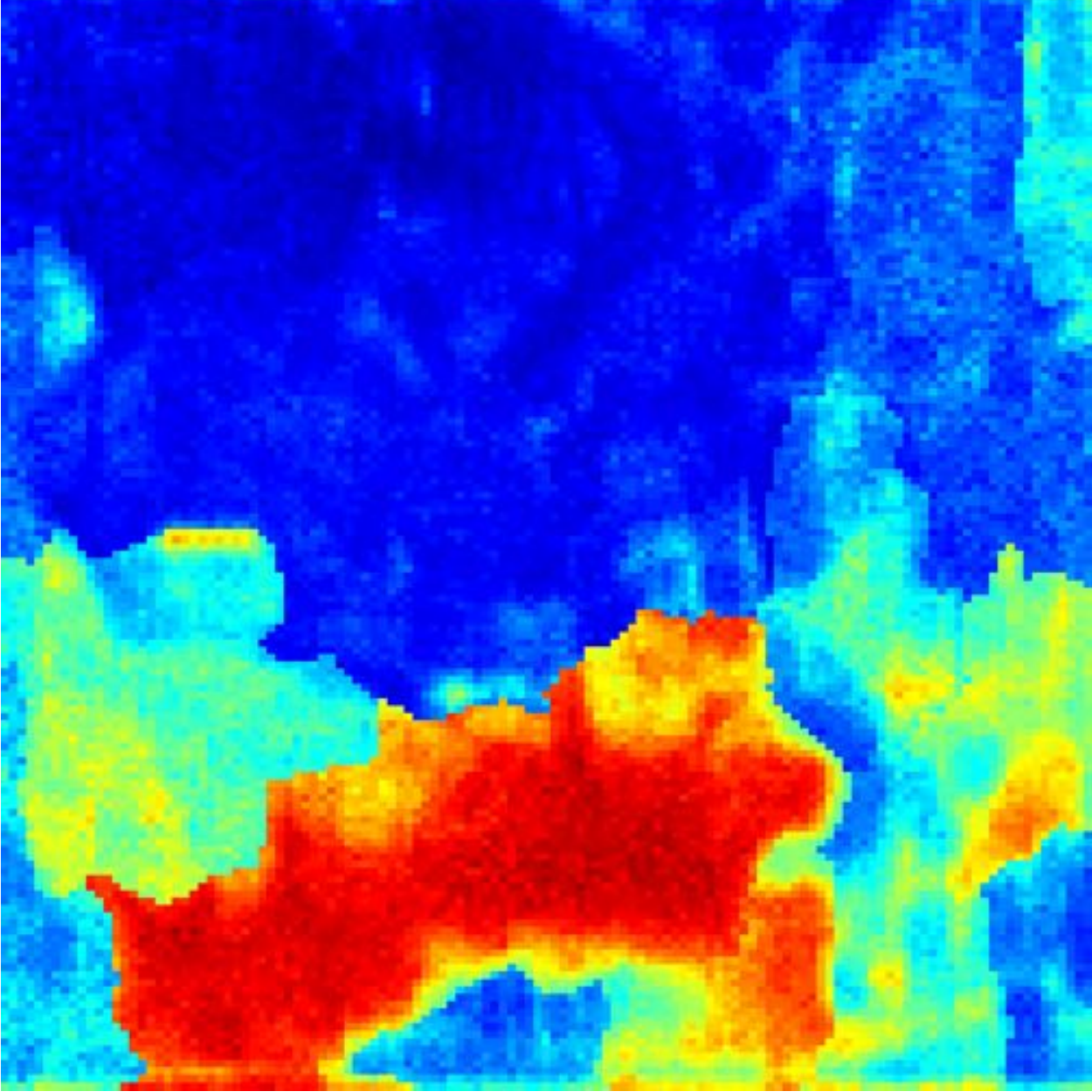}
		\captionsetup{labelformat=empty,skip=0pt}
		\caption{}
	\end{subfigure}	
	\begin{subfigure}[t]{0.15\textwidth}
		\centering
		\includegraphics[height=0.9\linewidth]{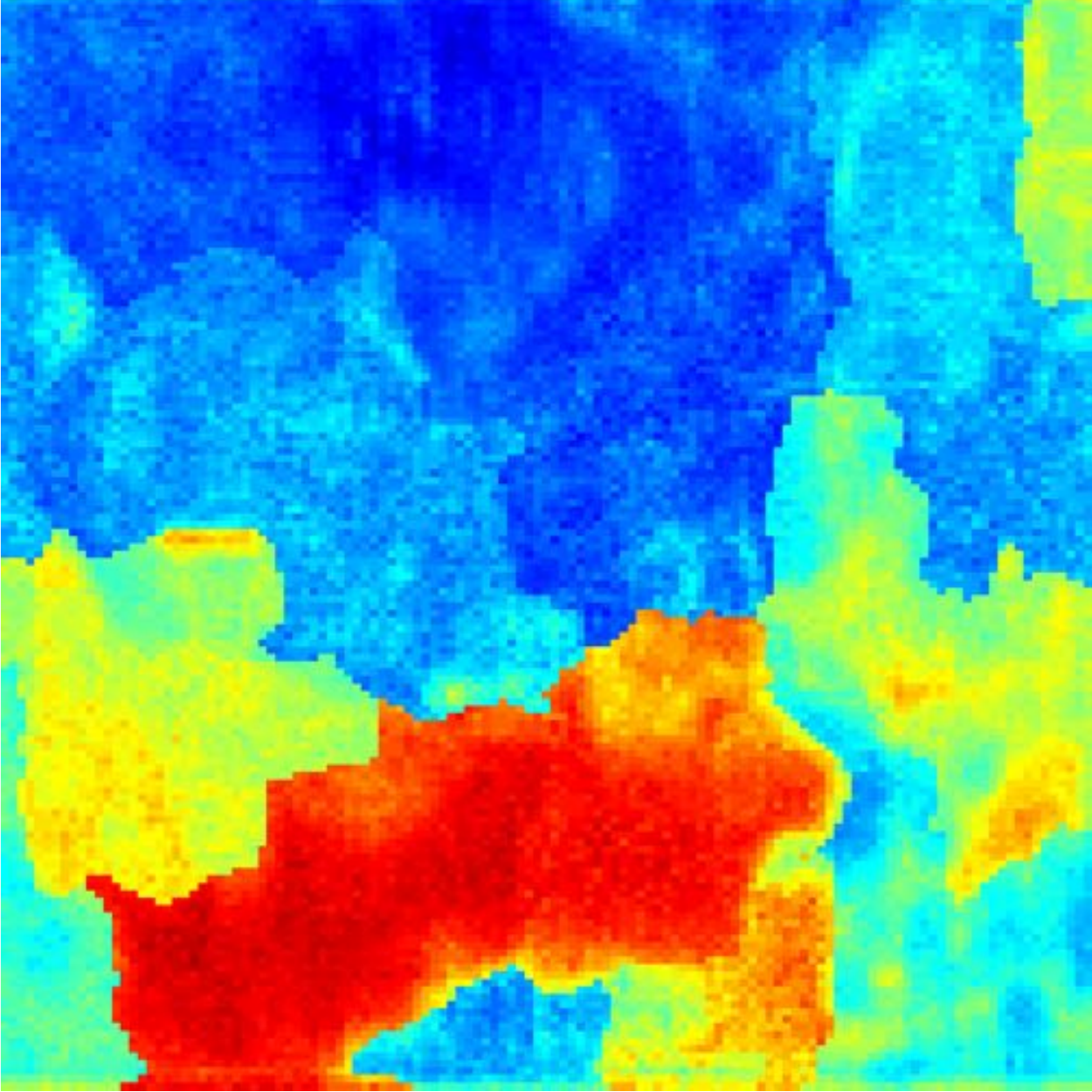}
		\captionsetup{labelformat=empty,skip=0pt}
		\caption{}
	\end{subfigure}
	
	\begin{subfigure}[t]{0.15\textwidth}
		\centering
		\includegraphics[height=0.9\linewidth]{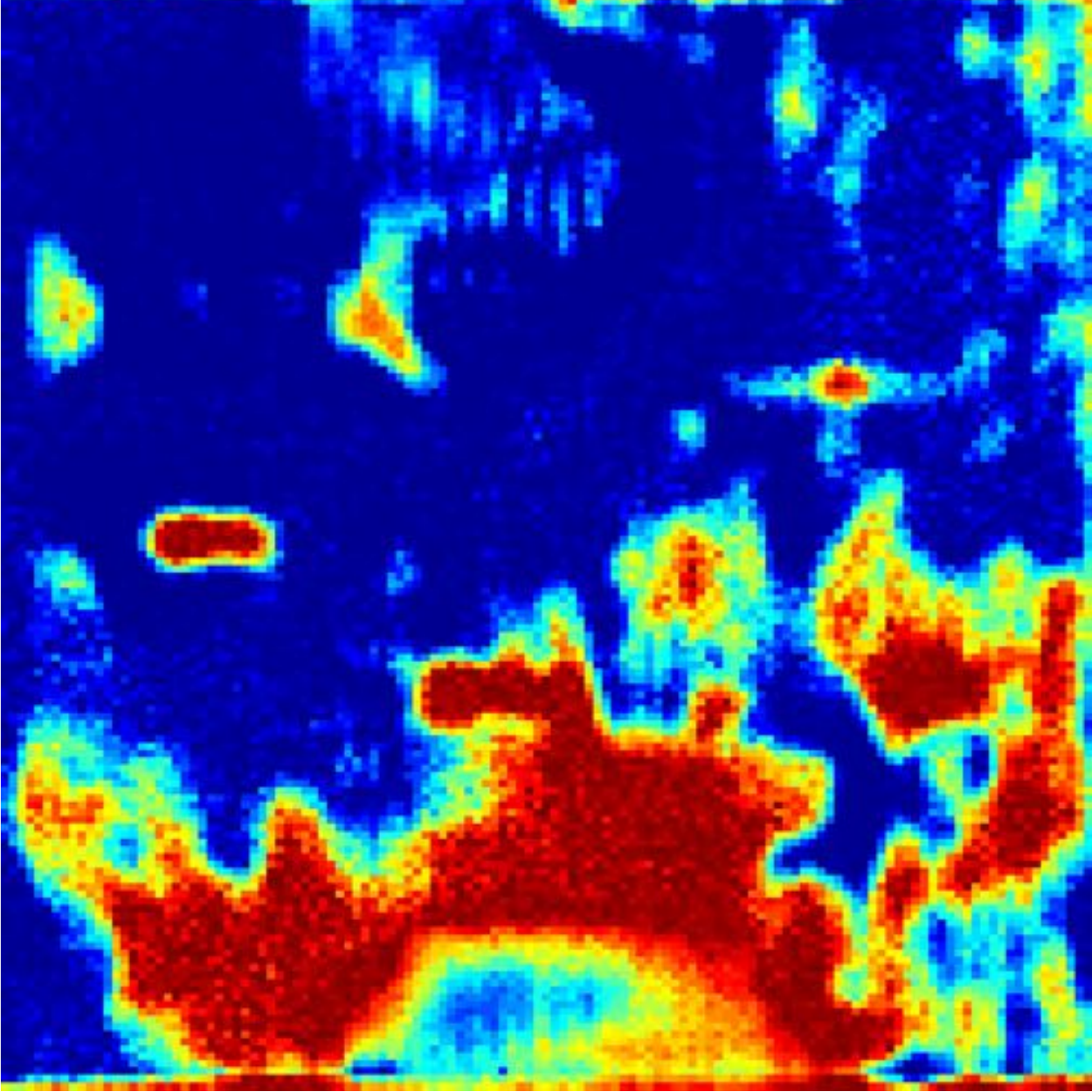}
		\captionsetup{labelformat=empty,skip=0pt}
		\caption{48 superpixels}
	\end{subfigure}	
	\begin{subfigure}[t]{0.15\textwidth}
		\centering
		\includegraphics[height=0.9\linewidth]{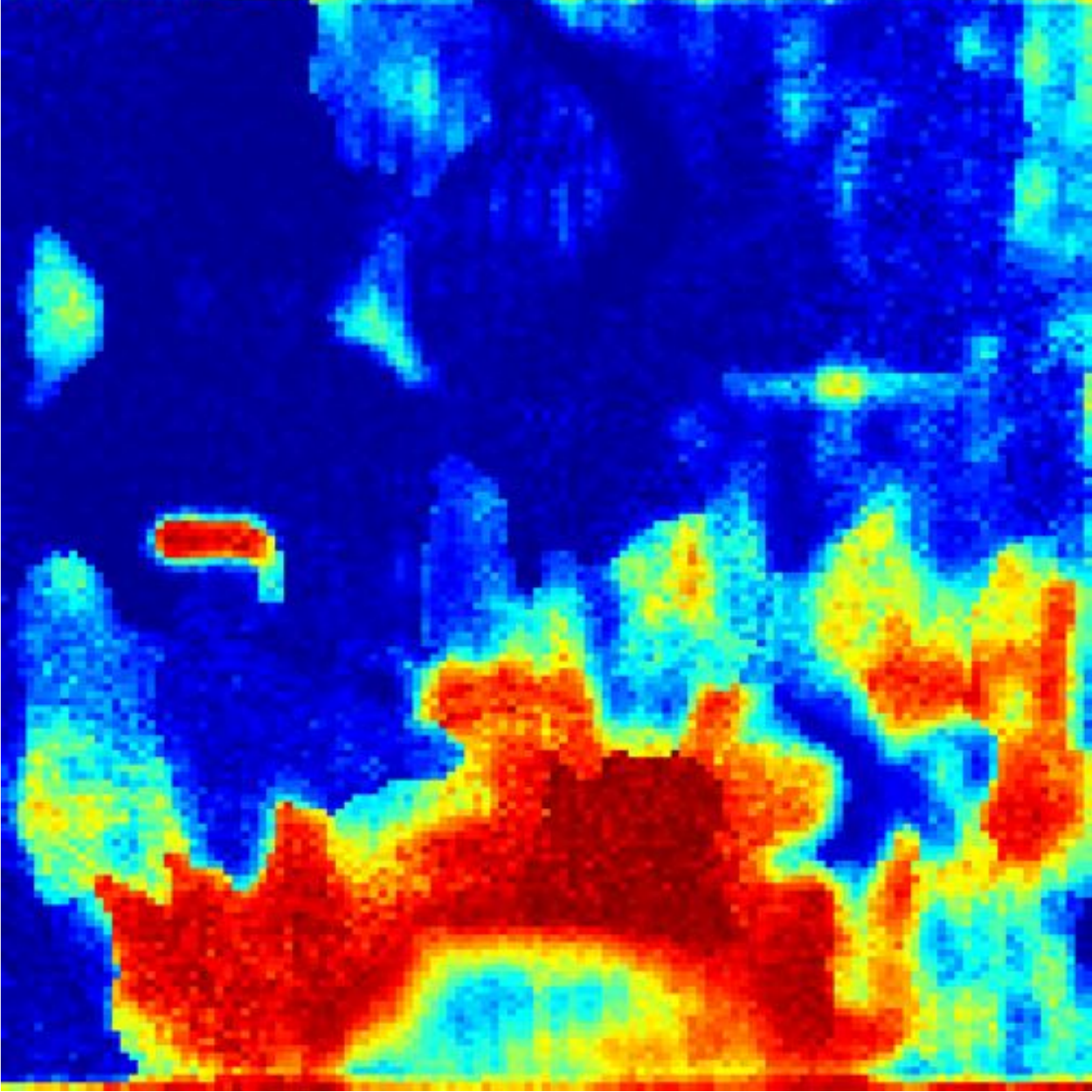}
		\captionsetup{labelformat=empty,skip=0pt}
		\caption{}
	\end{subfigure}	
	\begin{subfigure}[t]{0.15\textwidth}
		\centering
		\includegraphics[height=0.9\linewidth]{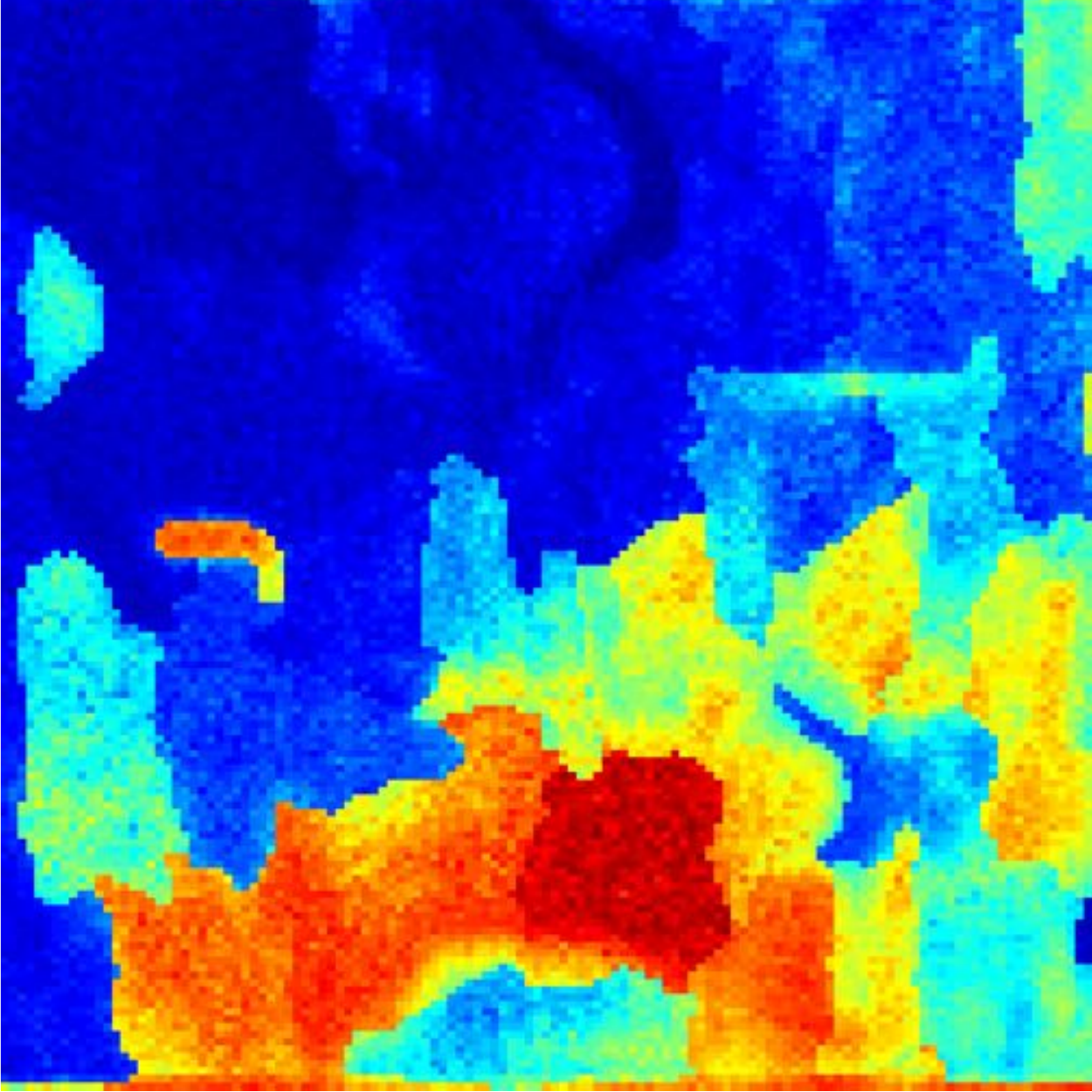}
		\captionsetup{labelformat=empty,skip=0pt}
		\caption{}
	\end{subfigure}	
	\begin{subfigure}[t]{0.15\textwidth}
		\centering
		\includegraphics[height=0.9\linewidth]{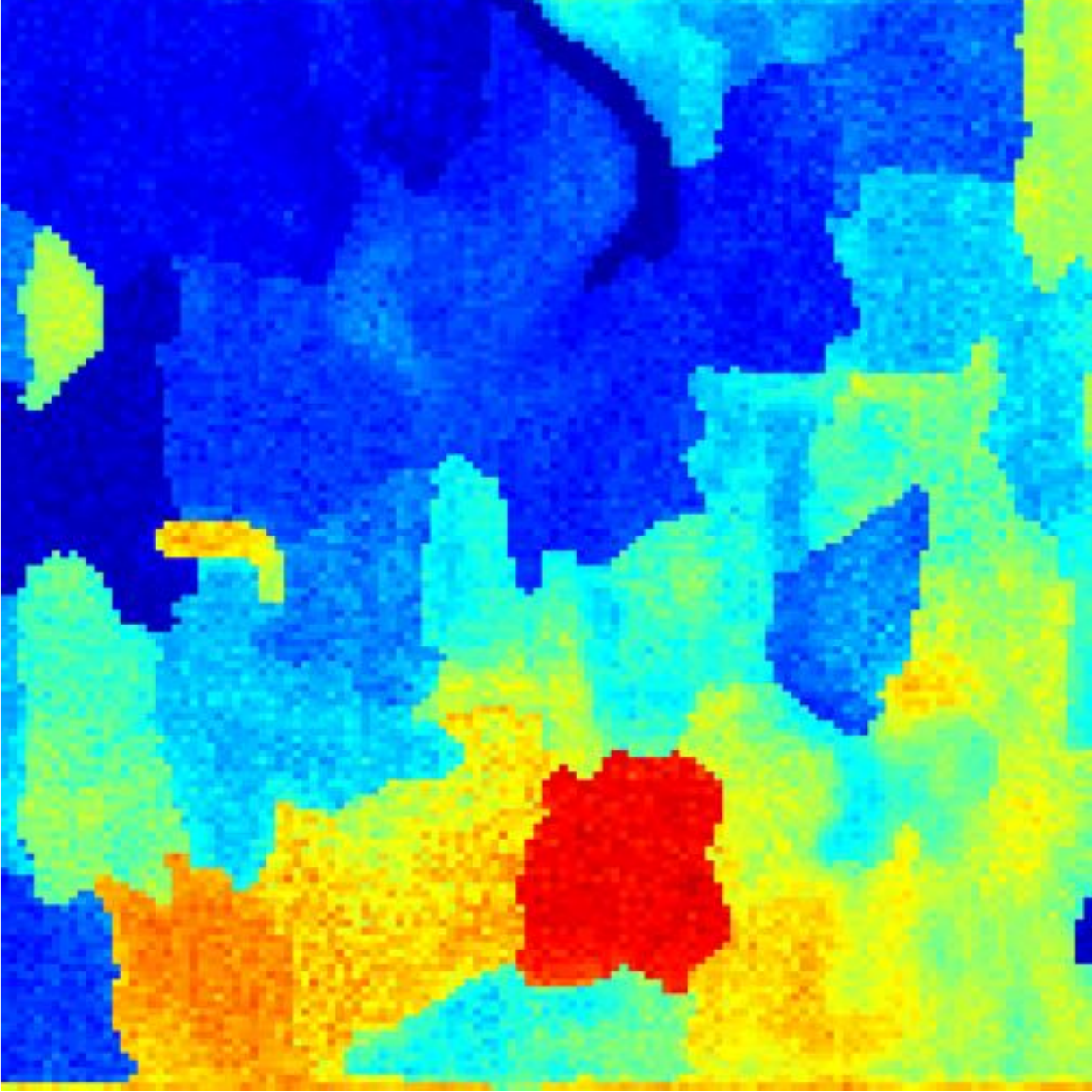}
		\captionsetup{labelformat=empty,skip=0pt}
		\caption{}
	\end{subfigure}
	
	\begin{subfigure}[t]{0.15\textwidth}
		\centering
		\includegraphics[height=0.9\linewidth]{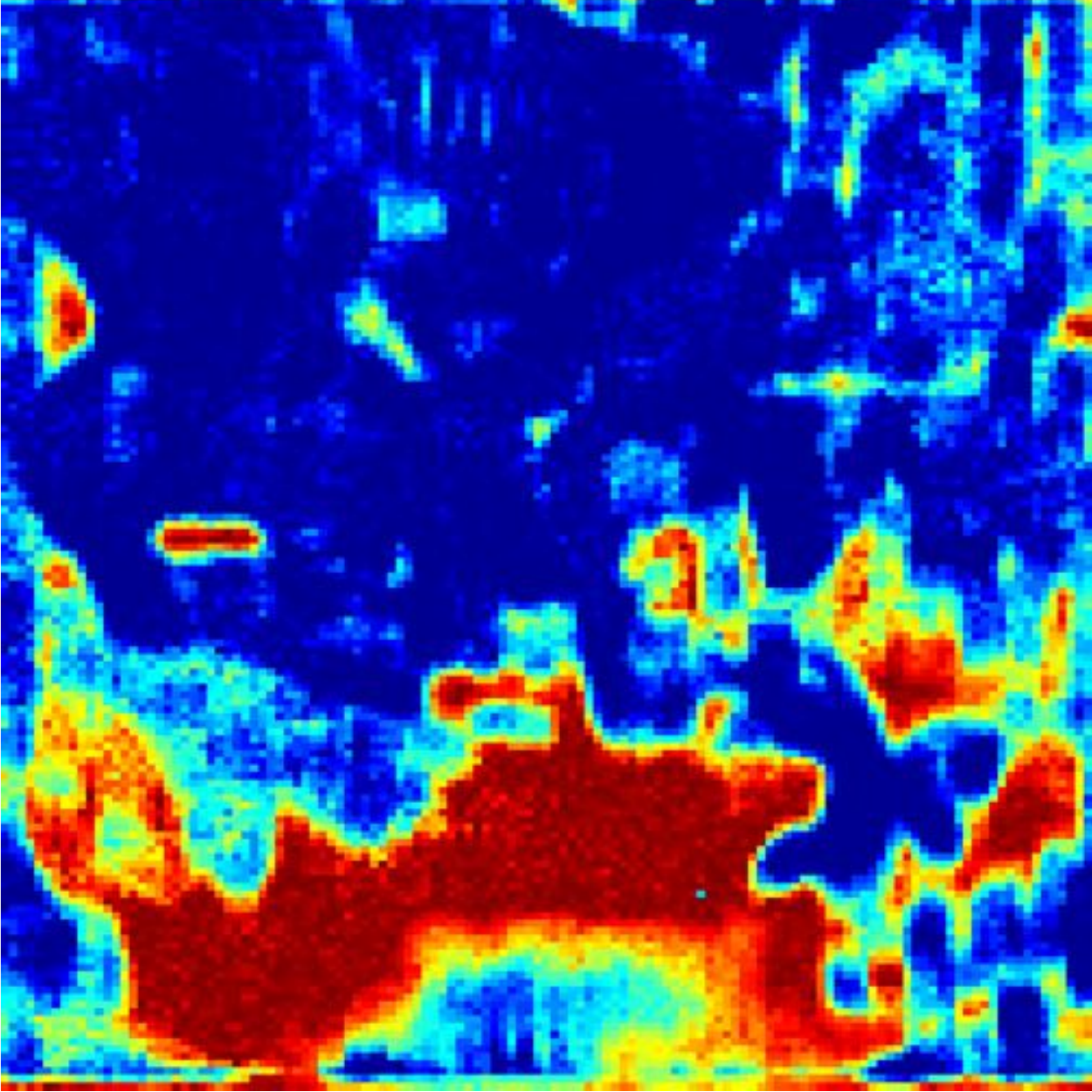}
		\captionsetup{labelformat=empty,skip=0pt}
		\caption{200 superpixels \\ s=1}
	\end{subfigure}	
	\begin{subfigure}[t]{0.15\textwidth}
		\centering
		\includegraphics[height=0.9\linewidth]{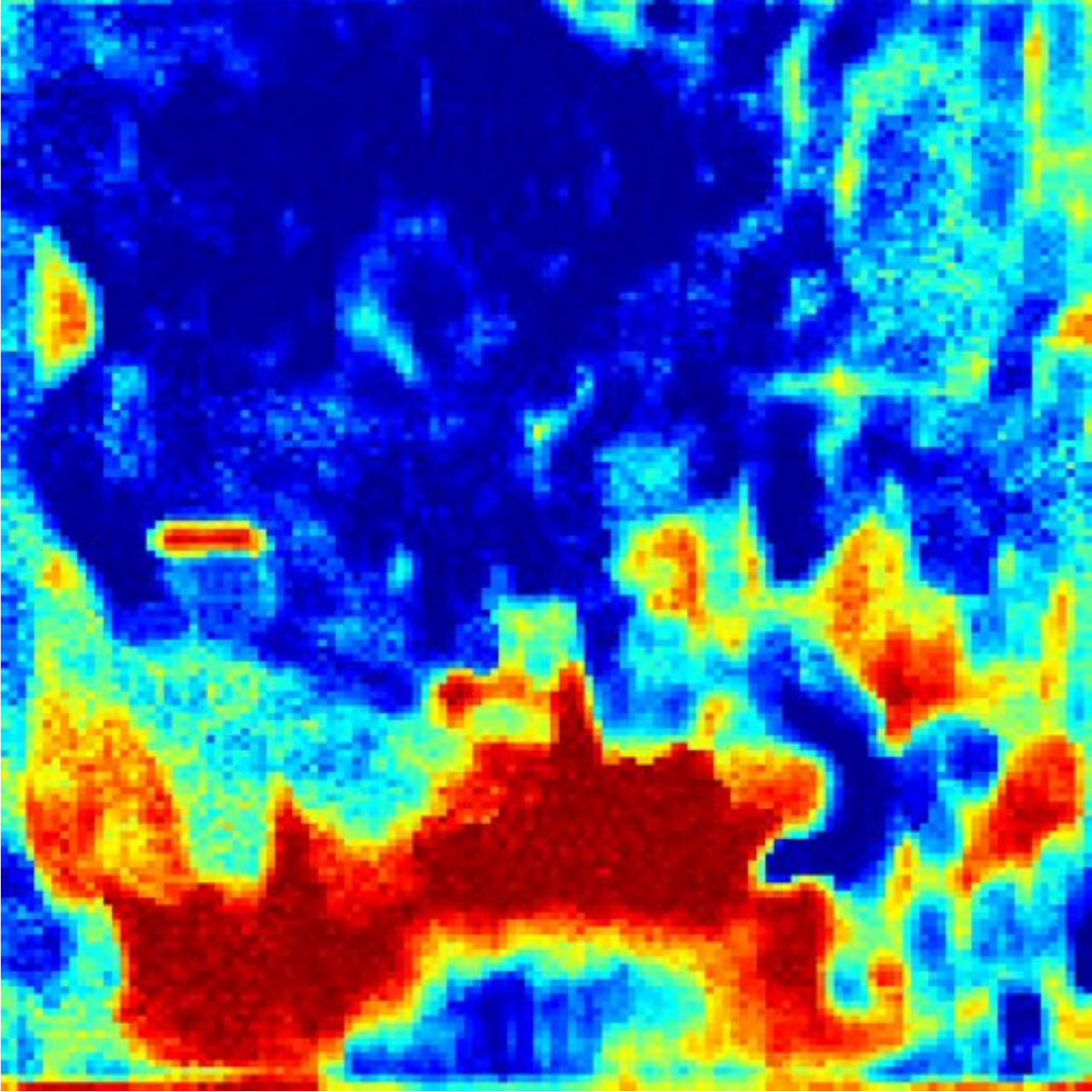}
		\captionsetup{labelformat=empty,skip=0pt}
		\caption{ \\s=100}
	\end{subfigure}	
	\begin{subfigure}[t]{0.15\textwidth}
		\centering
		\includegraphics[height=0.9\linewidth]{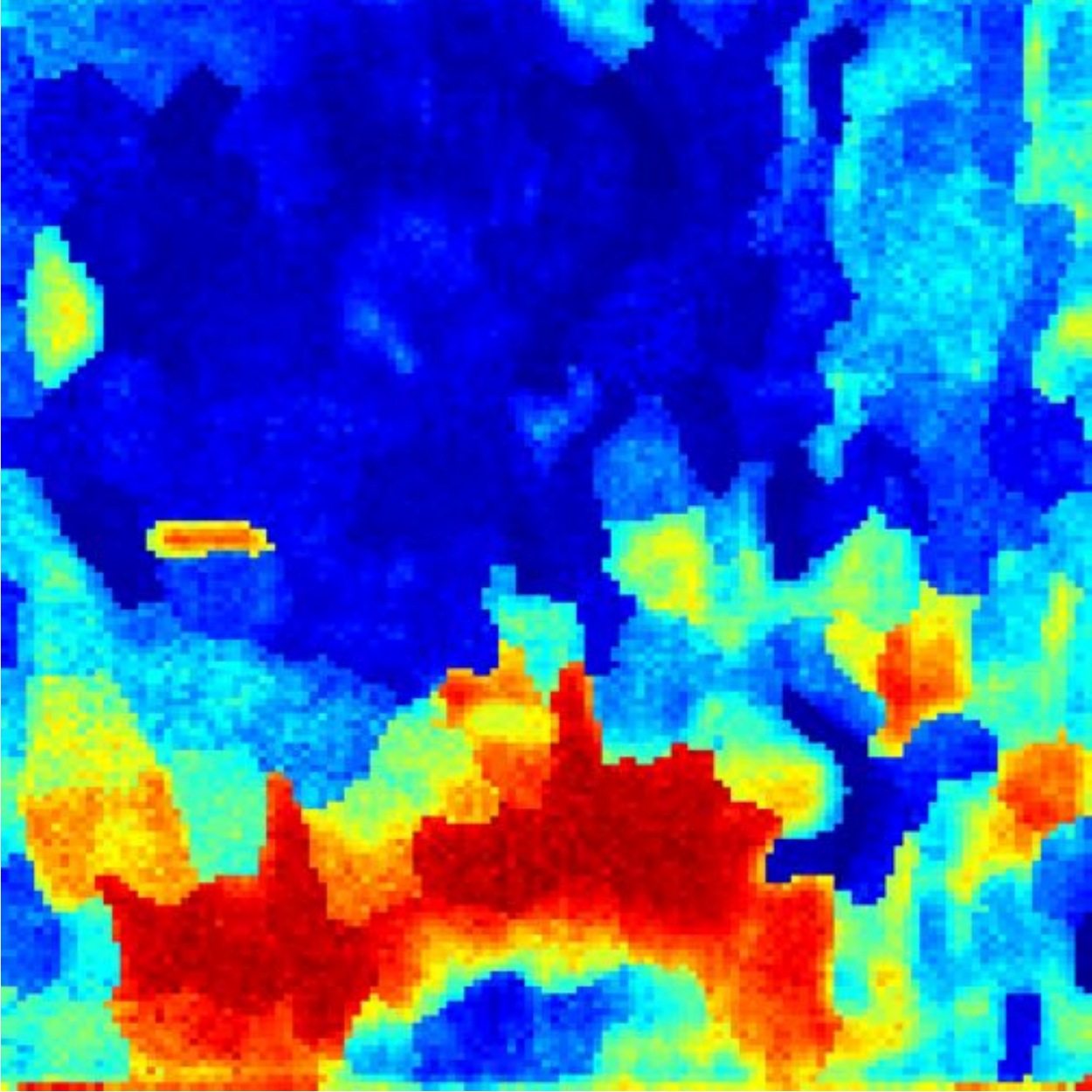}
		\captionsetup{labelformat=empty,skip=0pt}
		\caption{ \\s=500}
	\end{subfigure}	
	\begin{subfigure}[t]{0.15\textwidth}
		\centering
		\includegraphics[height=0.9\linewidth]{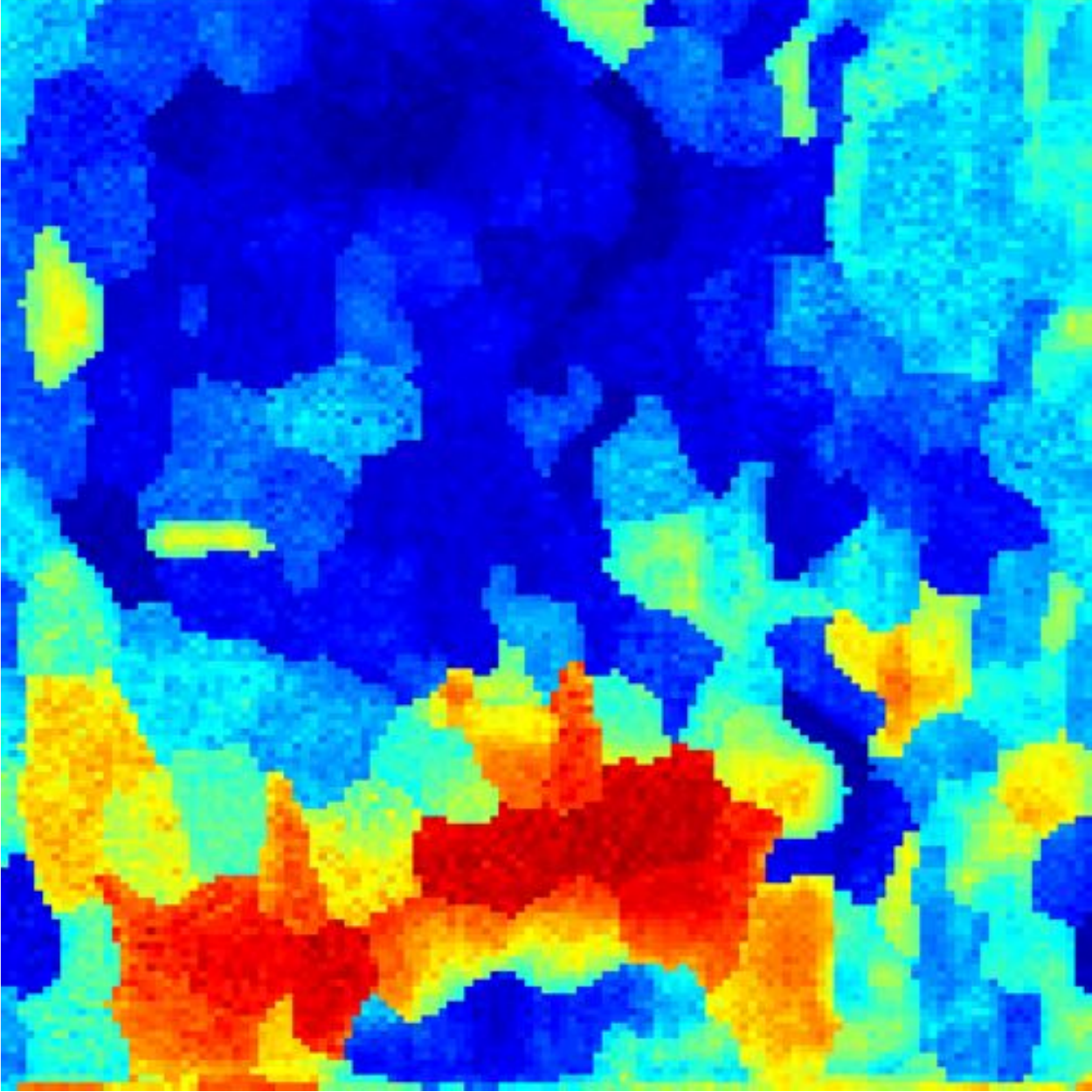}
		\captionsetup{labelformat=empty,skip=0pt}
		\caption{ \\s=1000}
	\end{subfigure}
	\caption{Partial membership maps of HF-00 with varying number of superpixels and varying scaling factor $s$ for topic 2 - dark flat sand. Column 1 to 5 correspond to scaling factor $s=1, 100, 500, 1000$, respectively.}
	\label{fig:hf00_2}
\end{figure*}

\begin{figure*}[htb!]
	\centering
	\begin{subfigure}[t]{0.15\textwidth}
		\centering
		\includegraphics[height=0.9\linewidth]{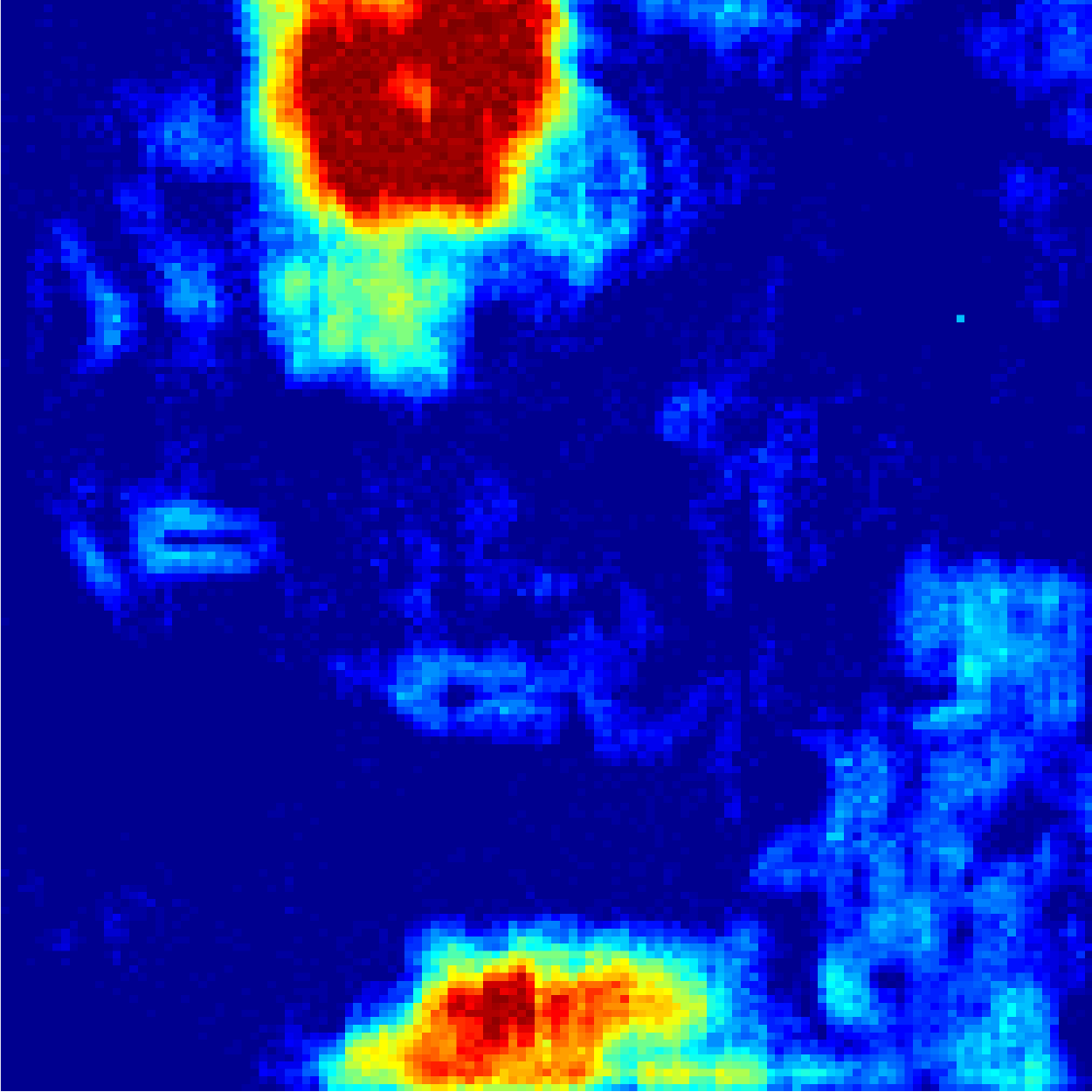}
		\captionsetup{labelformat=empty}
		\caption{20 superpixels}
	\end{subfigure}	
	\begin{subfigure}[t]{0.15\textwidth}
		\centering
		\includegraphics[height=0.9\linewidth]{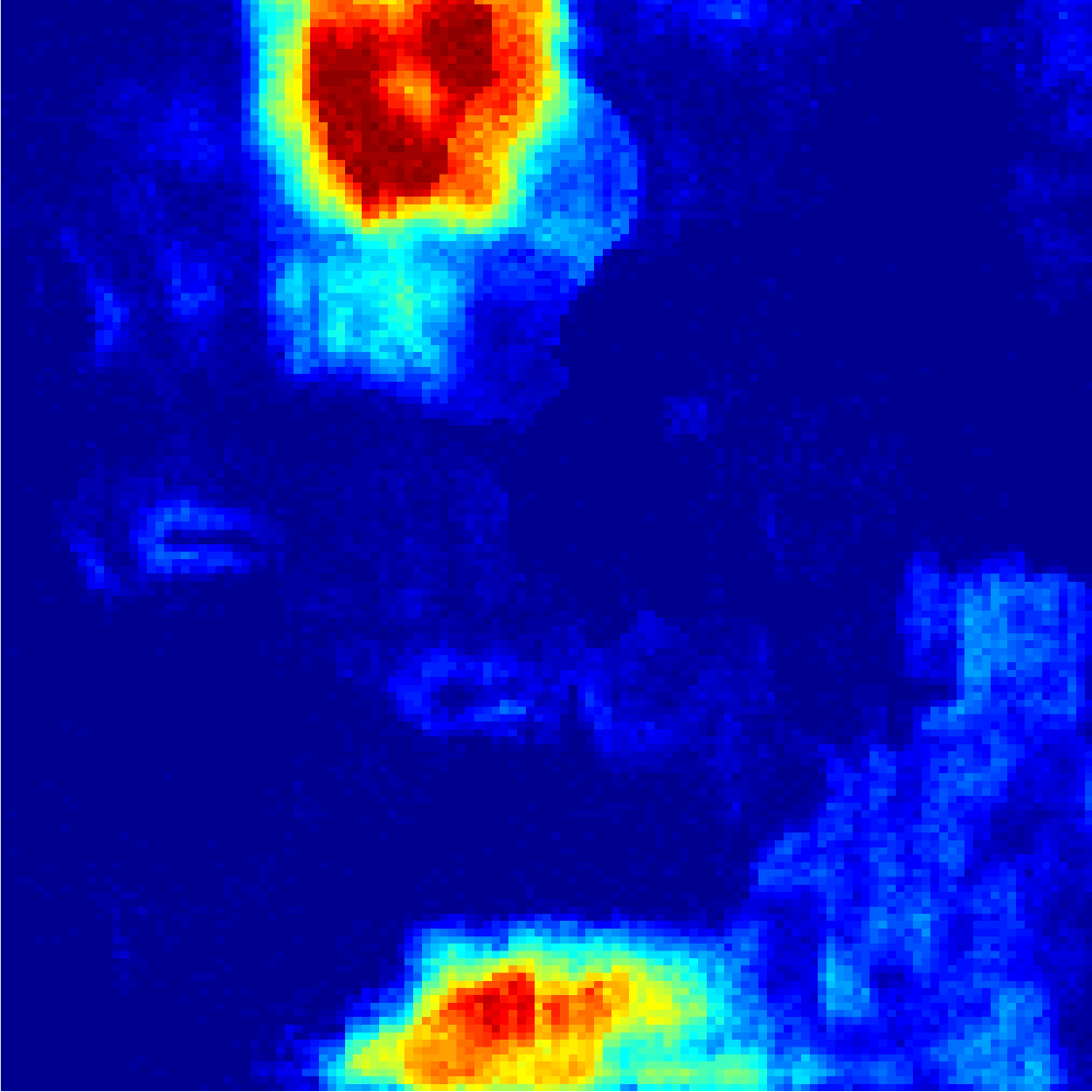}
		\captionsetup{labelformat=empty}
		\caption{}
	\end{subfigure}	
	\begin{subfigure}[t]{0.15\textwidth}
		\centering
		\includegraphics[height=0.9\linewidth]{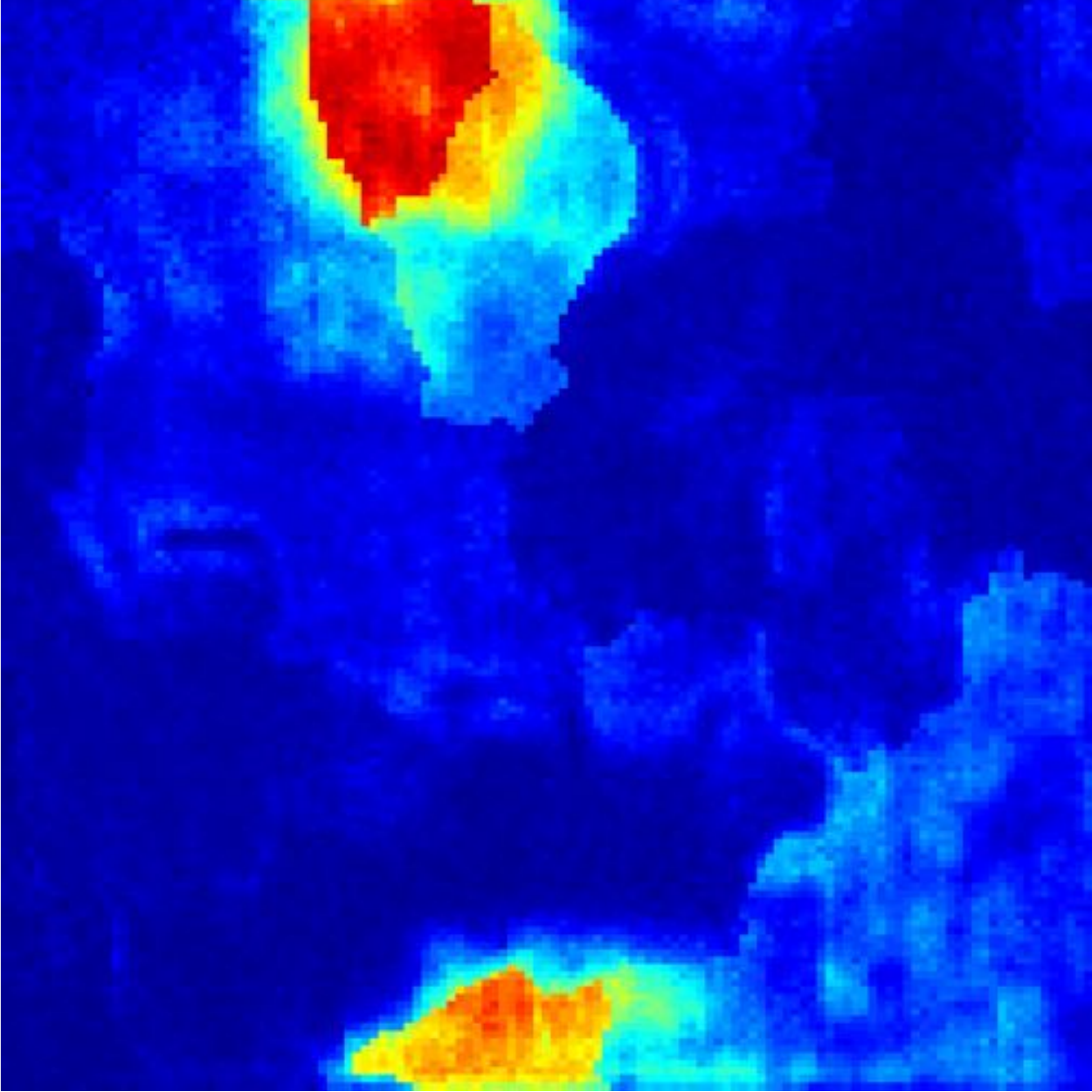}
		\captionsetup{labelformat=empty}
		\caption{}
	\end{subfigure}	
	\begin{subfigure}[t]{0.15\textwidth}
		\centering
		\includegraphics[height=0.9\linewidth]{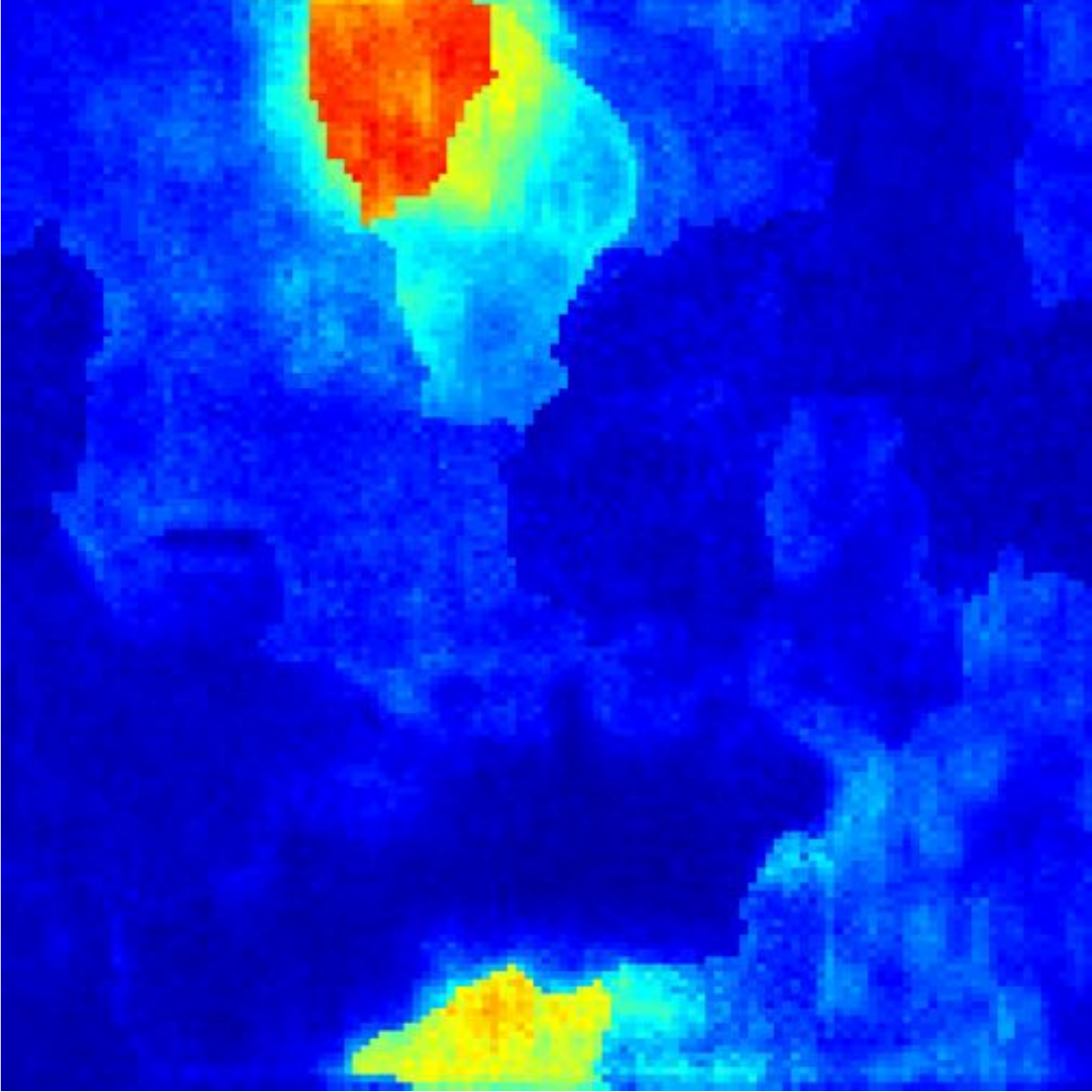}
		\captionsetup{labelformat=empty}
		\caption{}
	\end{subfigure}
	
	\begin{subfigure}[t]{0.15\textwidth}
		\centering
		\includegraphics[height=0.9\linewidth]{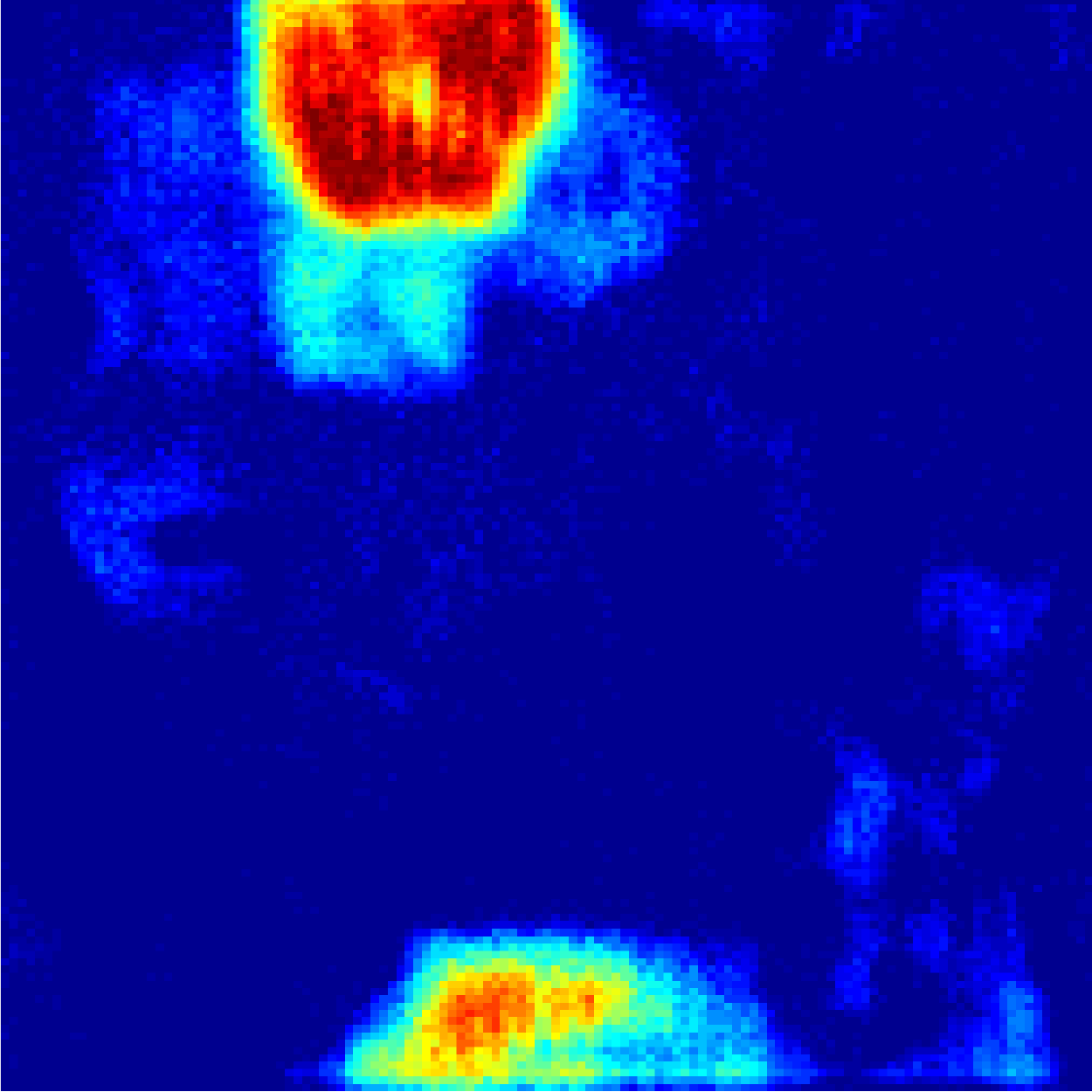}
		\captionsetup{labelformat=empty}
		\caption{48 superpixels}
	\end{subfigure}	
	\begin{subfigure}[t]{0.15\textwidth}
		\centering
		\includegraphics[height=0.9\linewidth]{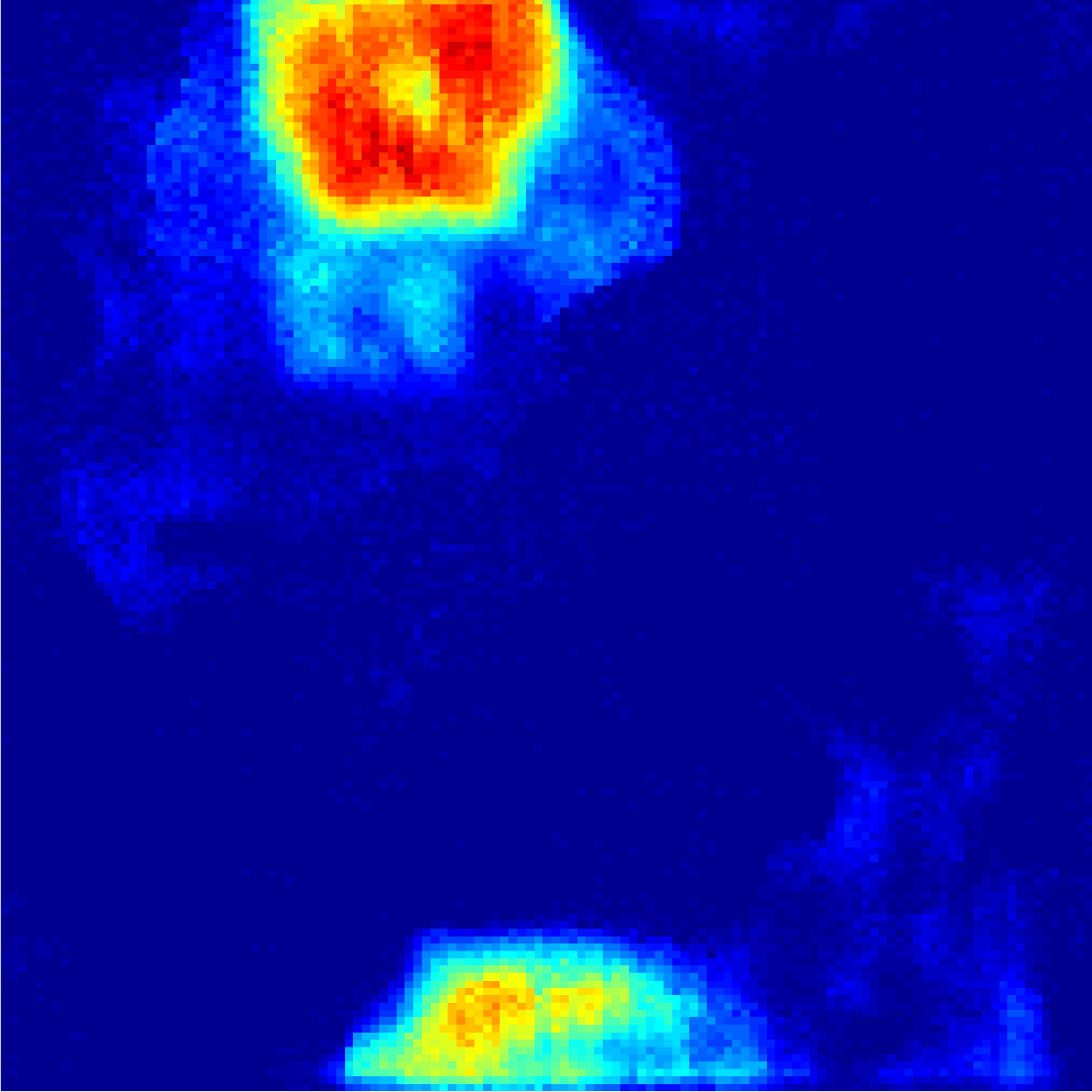}
		\captionsetup{labelformat=empty}
		\caption{}
	\end{subfigure}	
	\begin{subfigure}[t]{0.15\textwidth}
		\centering
		\includegraphics[height=0.9\linewidth]{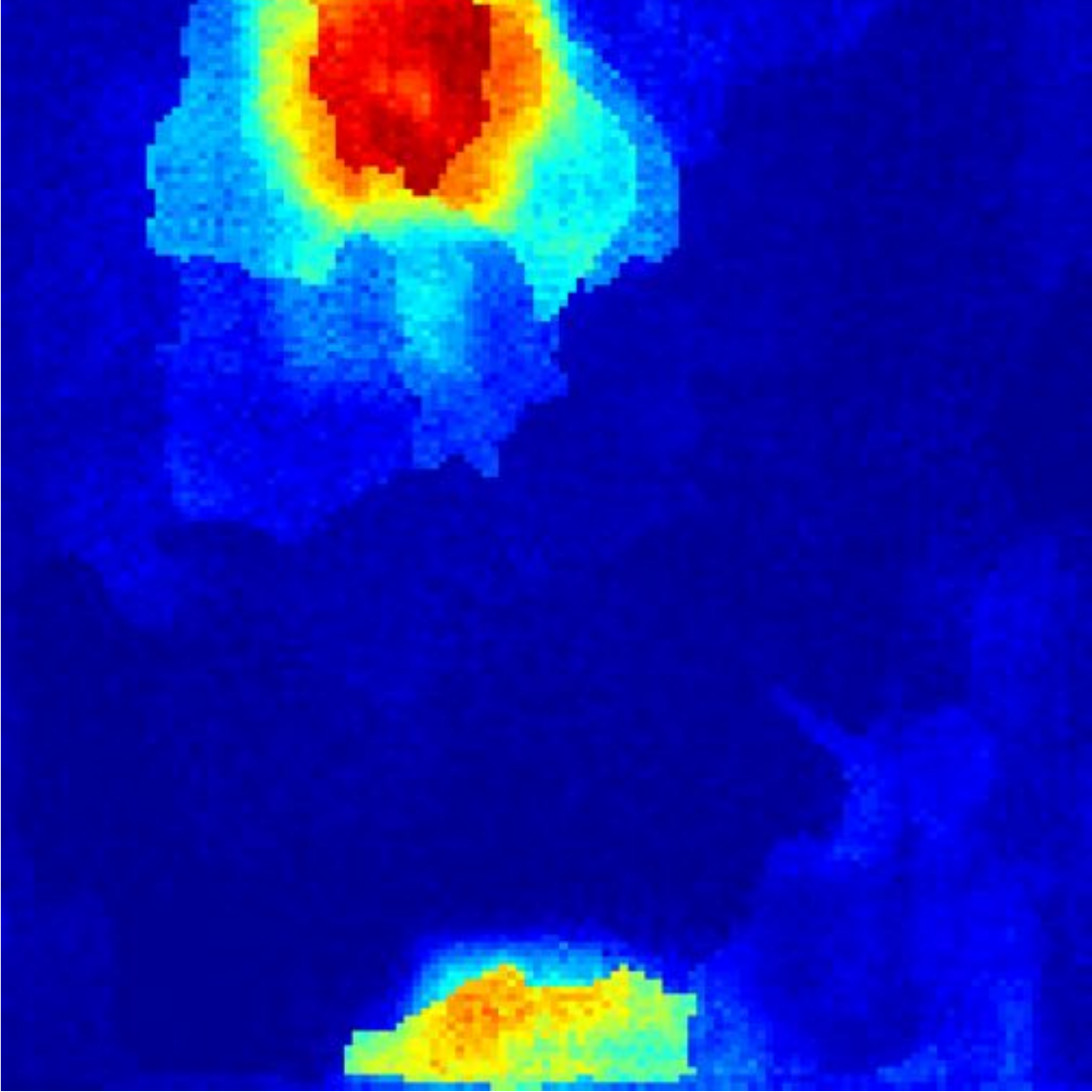}
		\captionsetup{labelformat=empty}
		\caption{}
	\end{subfigure}	
	\begin{subfigure}[t]{0.15\textwidth}
		\centering
		\includegraphics[height=0.9\linewidth]{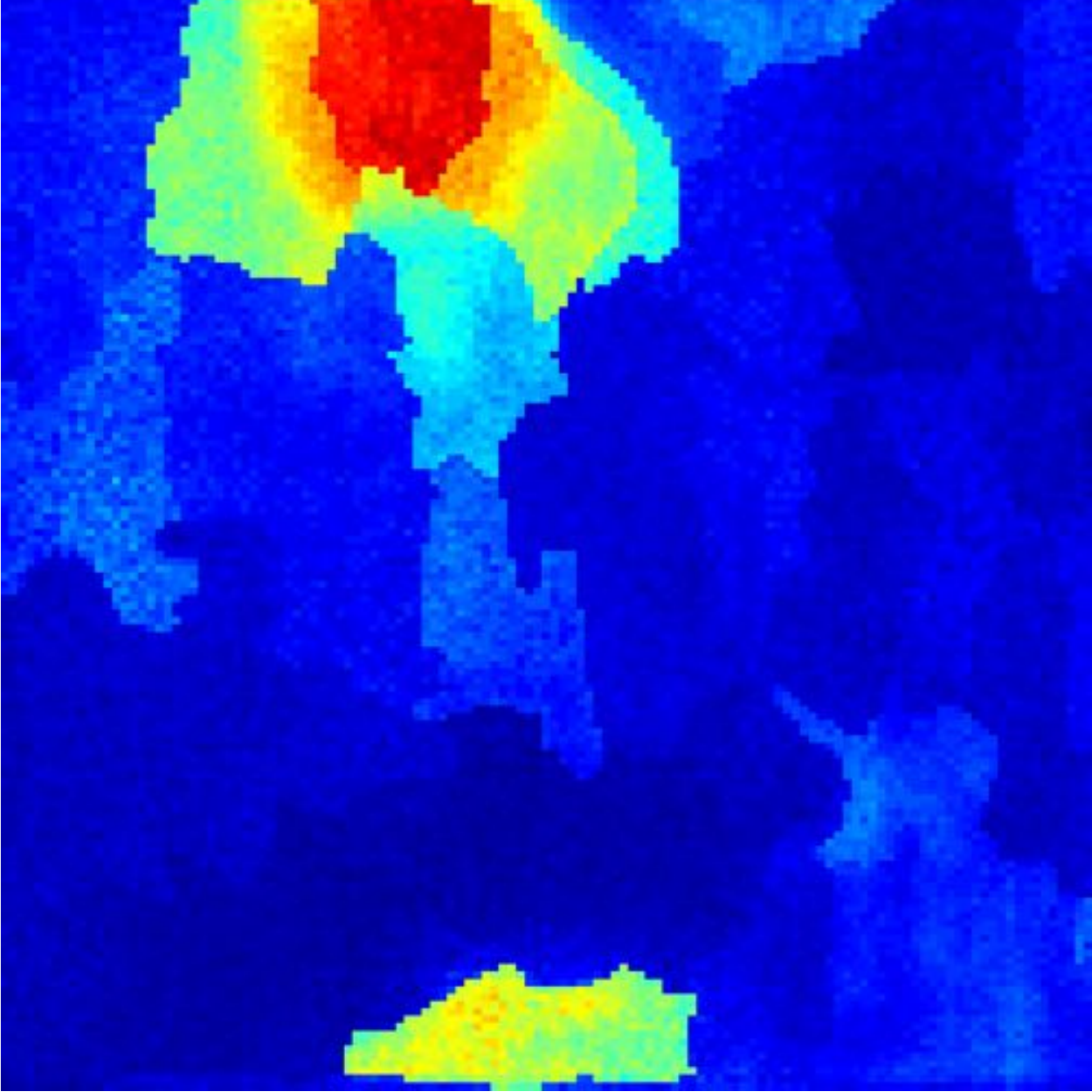}
		\captionsetup{labelformat=empty}
		\caption{}
	\end{subfigure}
	
	
	\begin{subfigure}[t]{0.15\textwidth}
		\centering
		\includegraphics[height=0.9\linewidth]{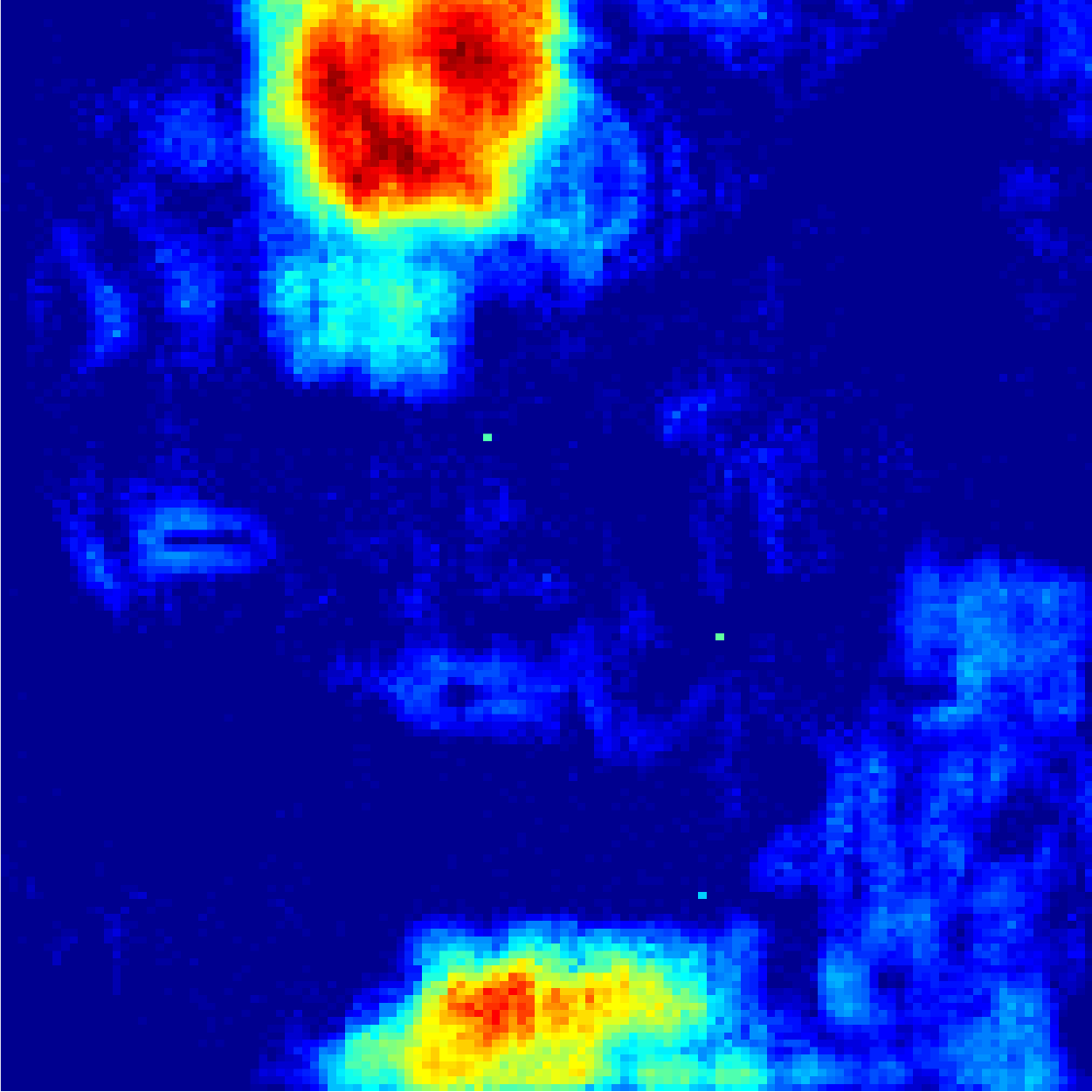}
		\captionsetup{labelformat=empty}
		\caption{ 200 superpixels\\s=1}
	\end{subfigure}	
	\begin{subfigure}[t]{0.15\textwidth}
		\centering
		\includegraphics[height=0.9\linewidth]{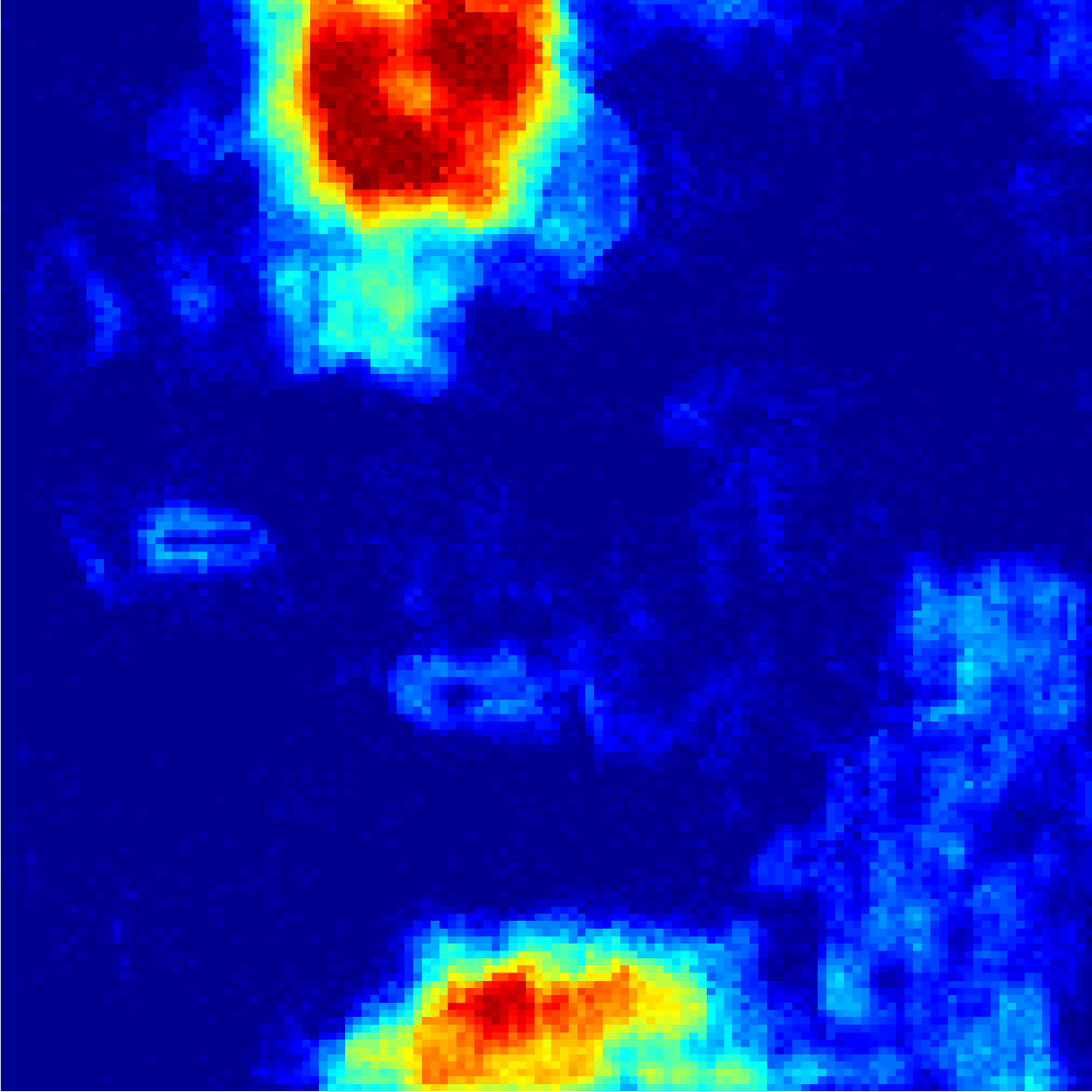}
		\captionsetup{labelformat=empty}
		\caption{\\s=100}
	\end{subfigure}	
	\begin{subfigure}[t]{0.15\textwidth}
		\centering
		\includegraphics[height=0.9\linewidth]{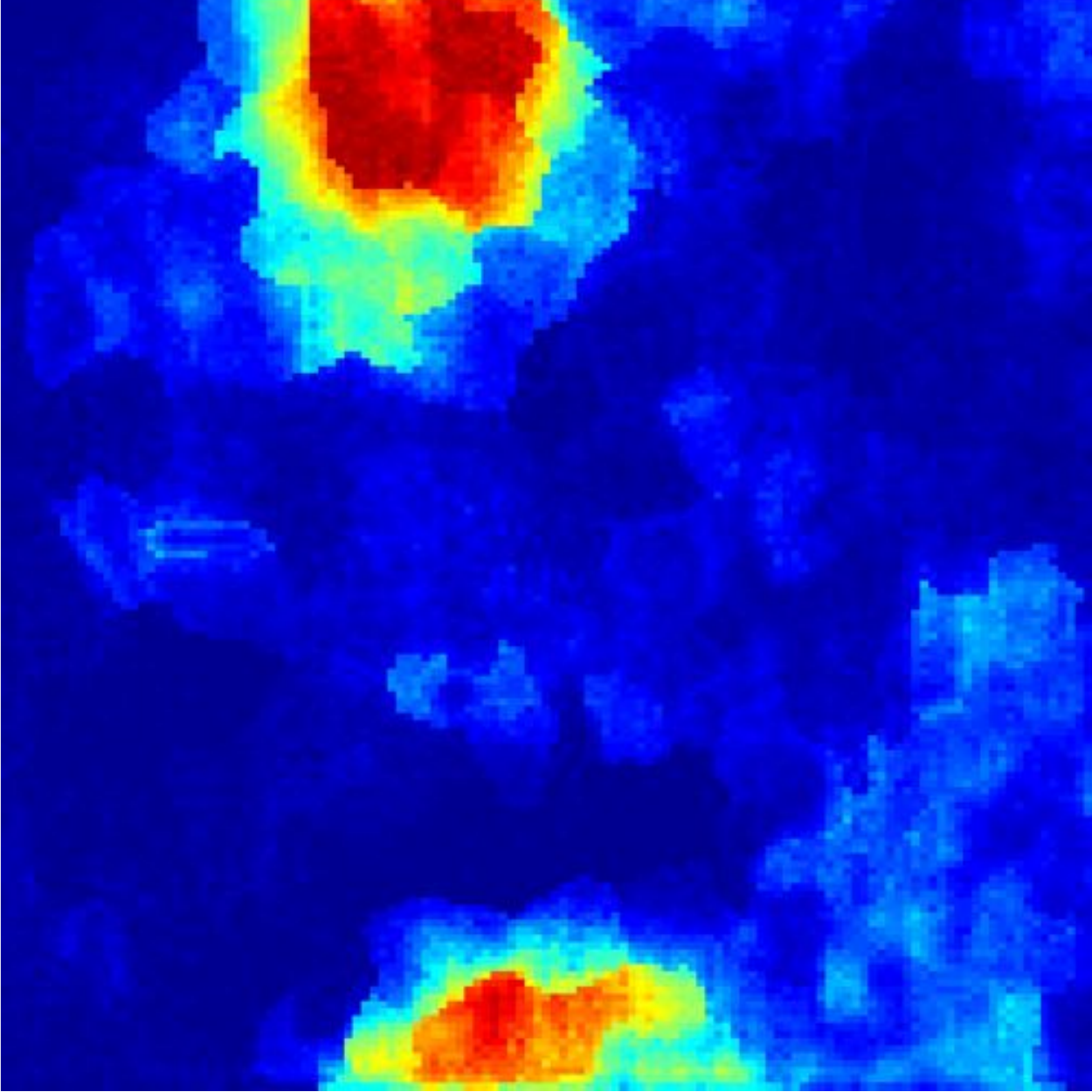}
		\captionsetup{labelformat=empty}
		\caption{\\s=500}
	\end{subfigure}	
	\begin{subfigure}[t]{0.15\textwidth}
		\centering
		\includegraphics[height=0.9\linewidth]{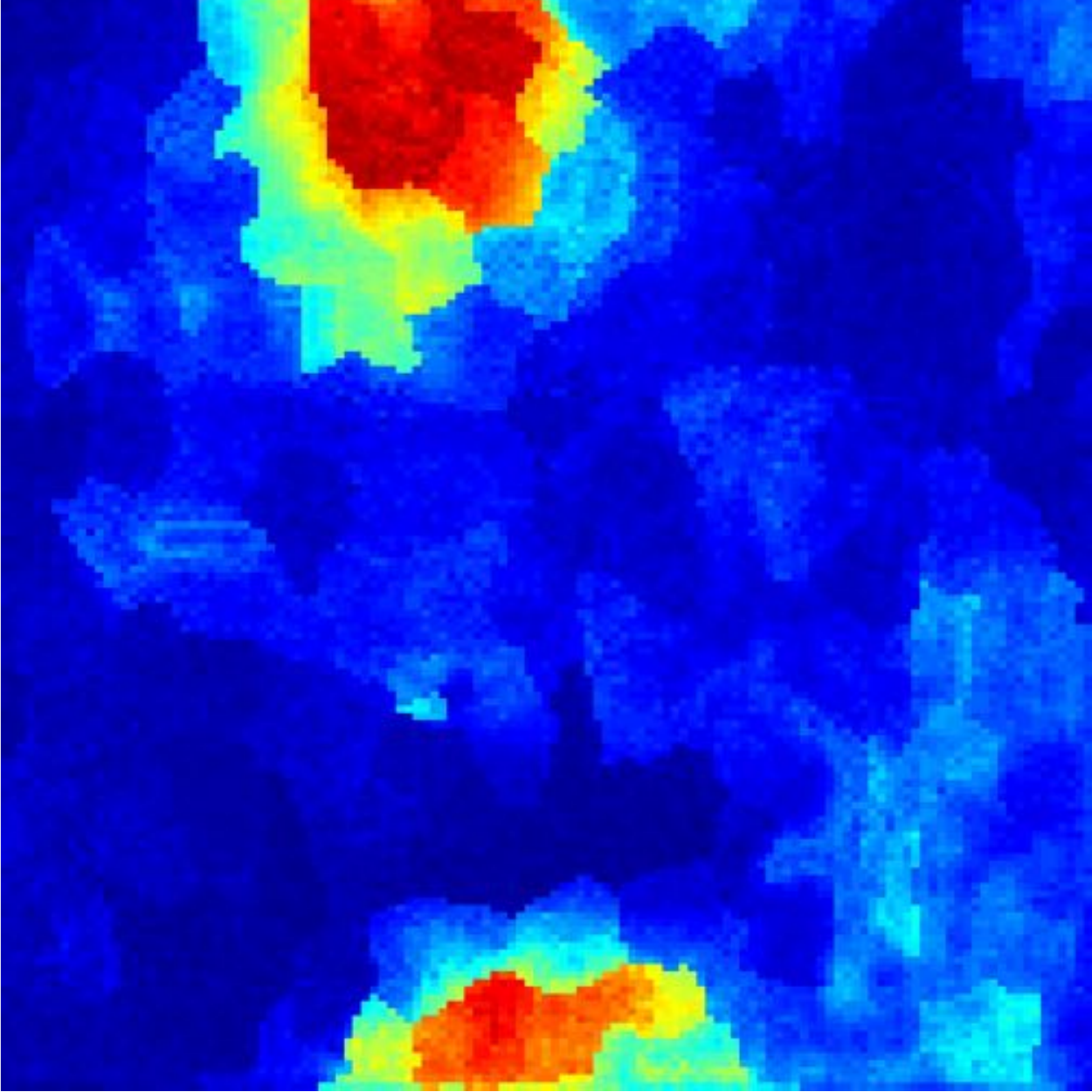}
		\captionsetup{labelformat=empty}
		\caption{\\s=1000}
	\end{subfigure}
	\caption{Partial membership maps of HF-00 with varying number of superpixels and varying scaling factor $s$ for topic 3 - sand ripple. Column 1 to 5 correspond to scaling factor $s=1, 100, 500, 1000$, respectively.}
	\label{fig:hf00_3}
\end{figure*}

In the following experiment, we study the effect of number of superpixels. SONAR Imagery HF-00 is selected for this experiment. The number of superpixel is varied to be $20$, $48$, and $200$. The scaling factor $s$ is varied to be $1, 100, 500,$ and $1000$. In contrast to the previous experiment where the topic proportion is fixed, the topic proportion is learned in this experiment. The experimental results are shown in Fig. \ref{fig:hf00_1} - \ref{fig:hf00_3}. The three figures show the partial membership maps in ``sea-grass'', ``dark flat sand'', and ''sand ripple'' topic, respectively, with varying number of superpixels and varying scaling factor $s$,.

In Fig. \ref{fig:hf00_1} - \ref{fig:hf00_2}, at each row, when the scaling factor $s$ increases, especially when $s$ changes from $500$ to $1000$, a few superpixels within the red region (as shown in Column $s=1$) change their color from red to yellow, thus the red region (when $s=1$) becomes less and less continuous. Based on the discussion in above experiment, with superpixels as documents, a large scaling factor $s$ should enhance the spatial continuity. However, this requires a strict prerequisite that the topic proportions are accurately estimated and consistent across superpixels that belong to the same topic. Otherwise, matching a large scaling factor with a superpixel will damage the membership estimation by pushing the membership vector towards an incorrect topic proportion vector. As illustrated, increasing the number of superpixels and matching a large scaling factor does not guarantee an improved membership estimation.

\subsubsection{Varying fuzzifier}
\begin{figure*}[htb!]
	\centering
		\begin{subfigure}[t]{0.25\textwidth}
			\centering
			\includegraphics[height=0.6\linewidth,width=0.9\linewidth]{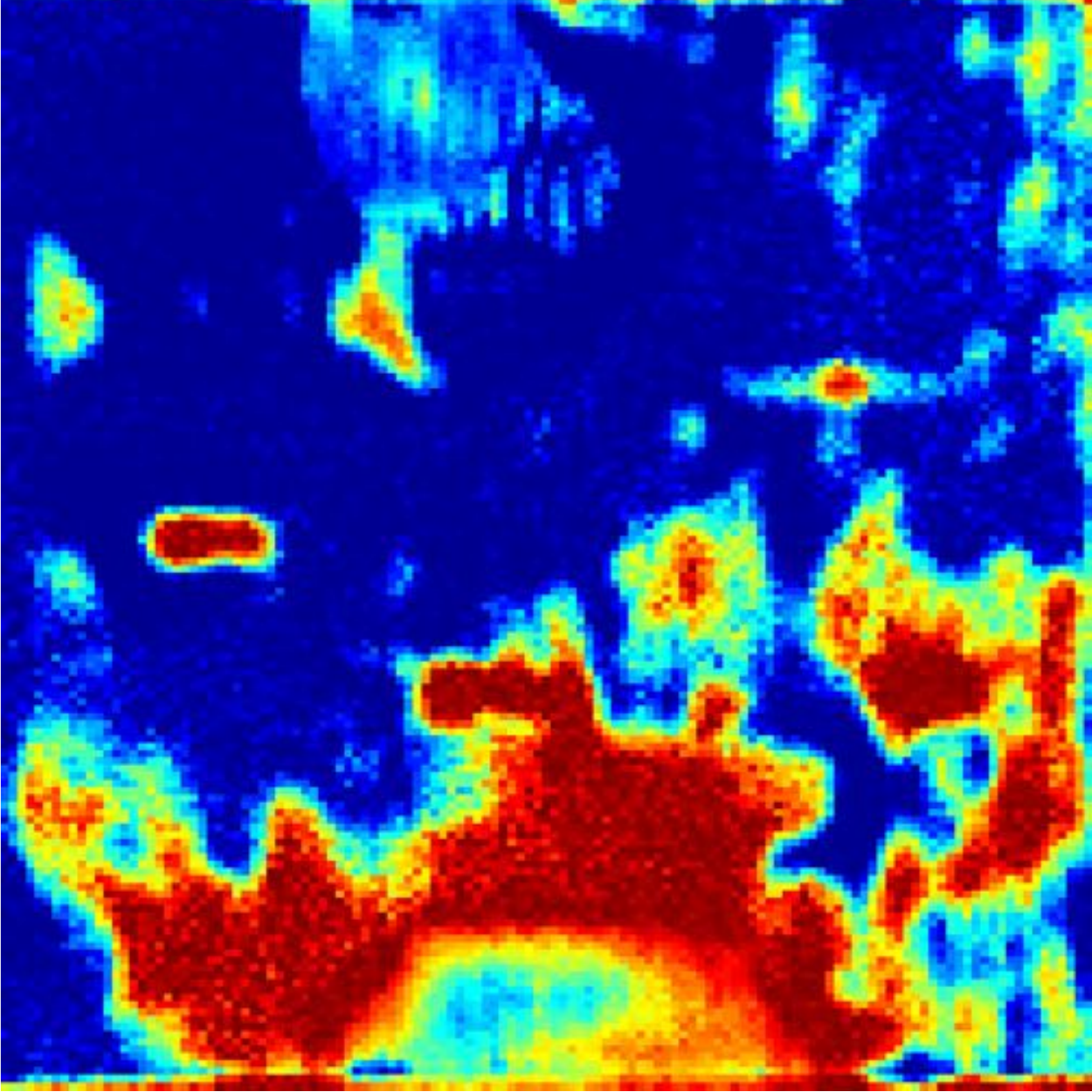}
			\captionsetup{labelformat=empty}
			\caption{(a)$m$ = 1}
		\end{subfigure}	
		\begin{subfigure}[t]{0.25\textwidth}
			\centering
			\includegraphics[height=0.6\linewidth,,width=0.9\linewidth]{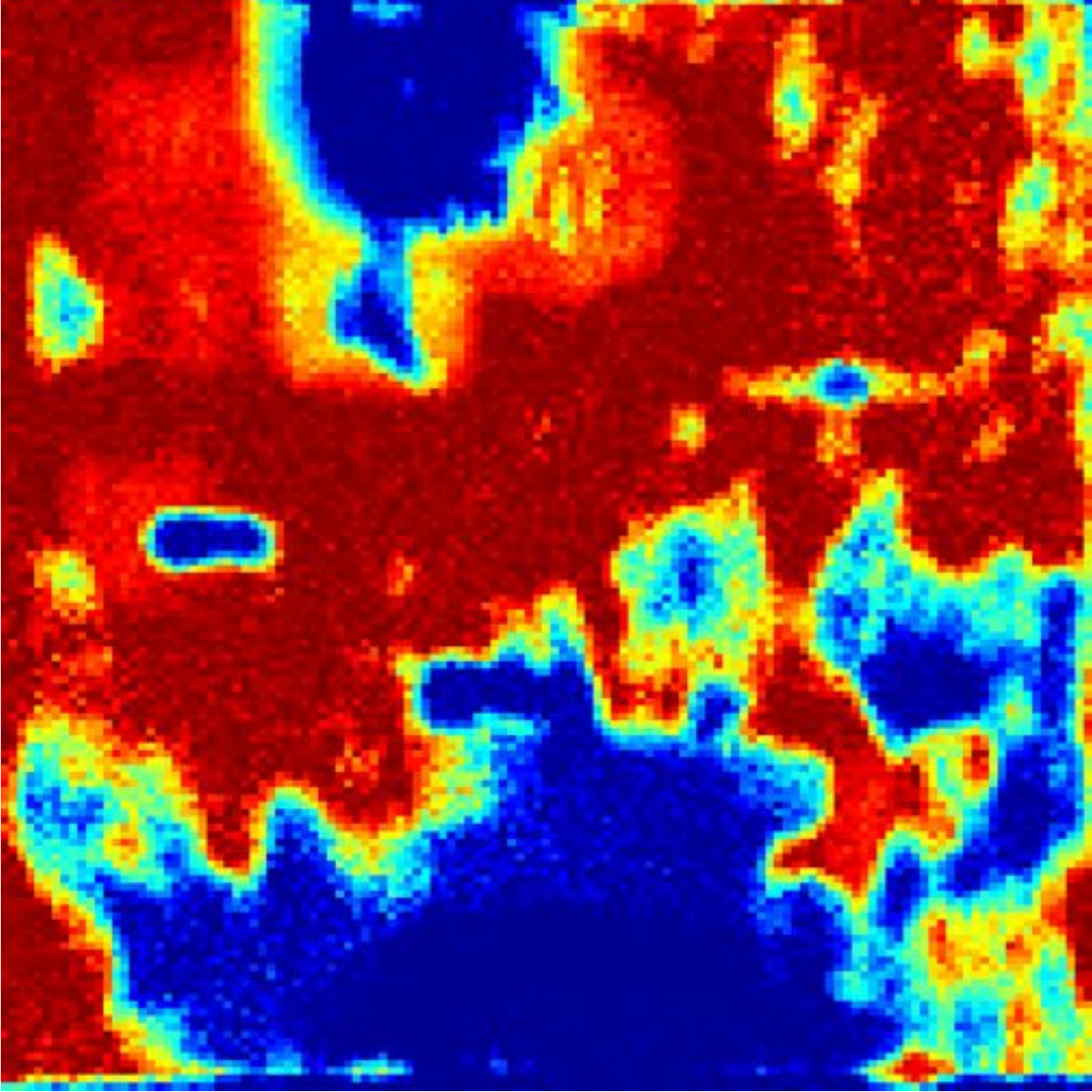}
			\captionsetup{labelformat=empty}
			\caption{(b)}
		\end{subfigure}	
		\begin{subfigure}[t]{0.25\textwidth}
			\centering
			\includegraphics[height=0.6\linewidth,,width=0.9\linewidth]{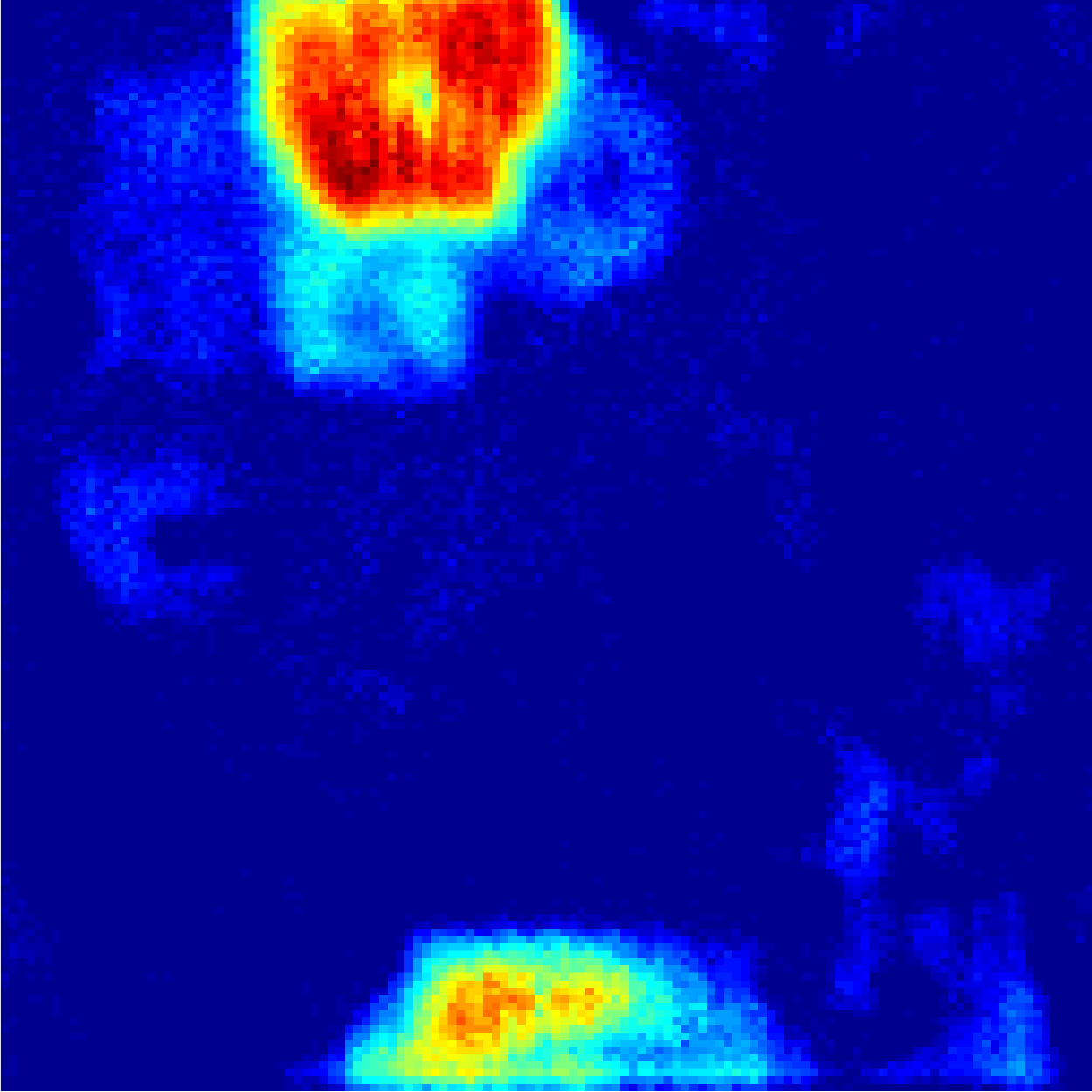}
			\captionsetup{labelformat=empty}
			\caption{(c)}
		\end{subfigure}	
		
		\begin{subfigure}[t]{0.25\textwidth}
			\centering
			\includegraphics[height=0.6\linewidth,,width=0.9\linewidth]{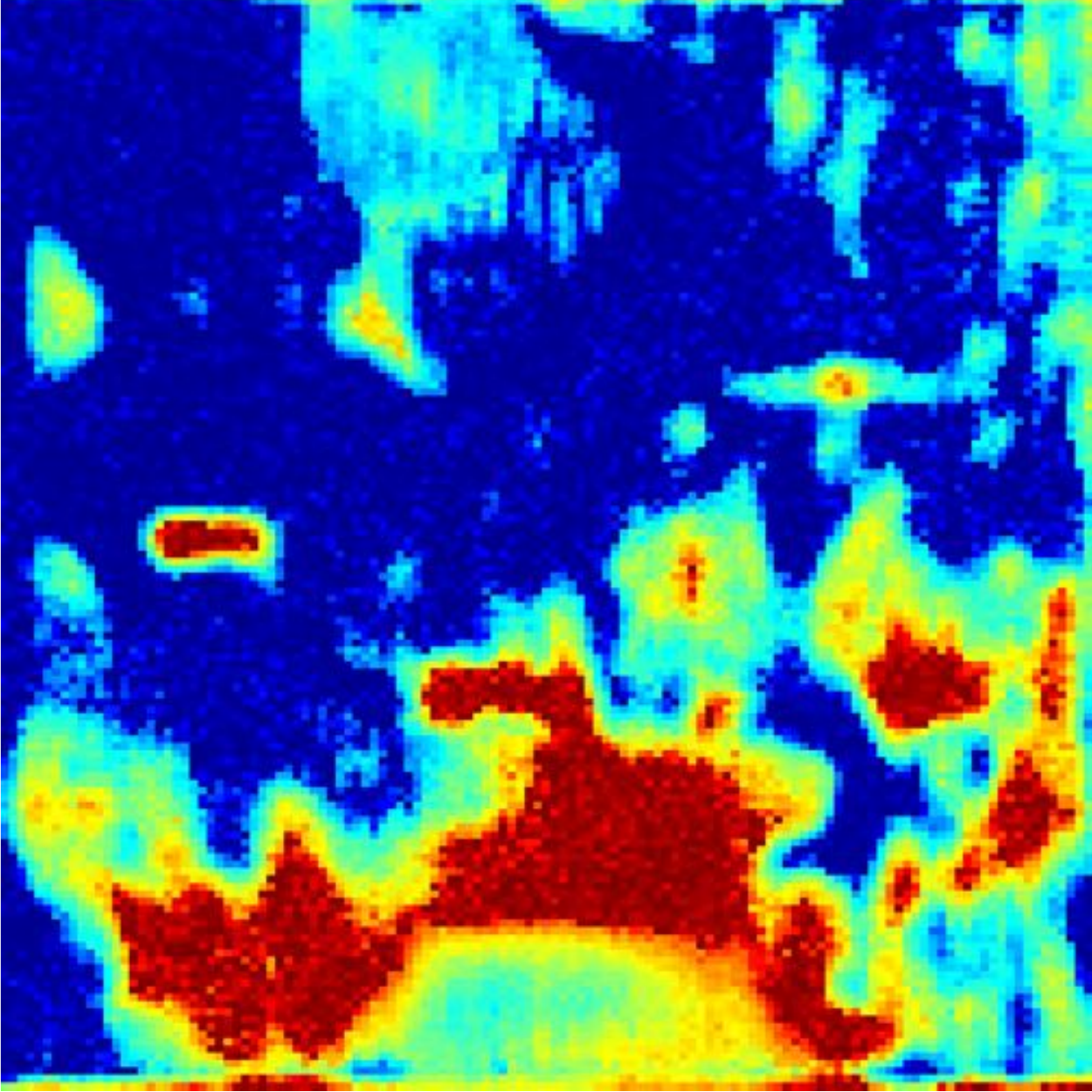}
			\captionsetup{labelformat=empty}
		    \caption{(a)$m$  = 2}
		\end{subfigure}	
		\begin{subfigure}[t]{0.25\textwidth}
			\centering
			\includegraphics[height=0.6\linewidth,,width=0.9\linewidth]{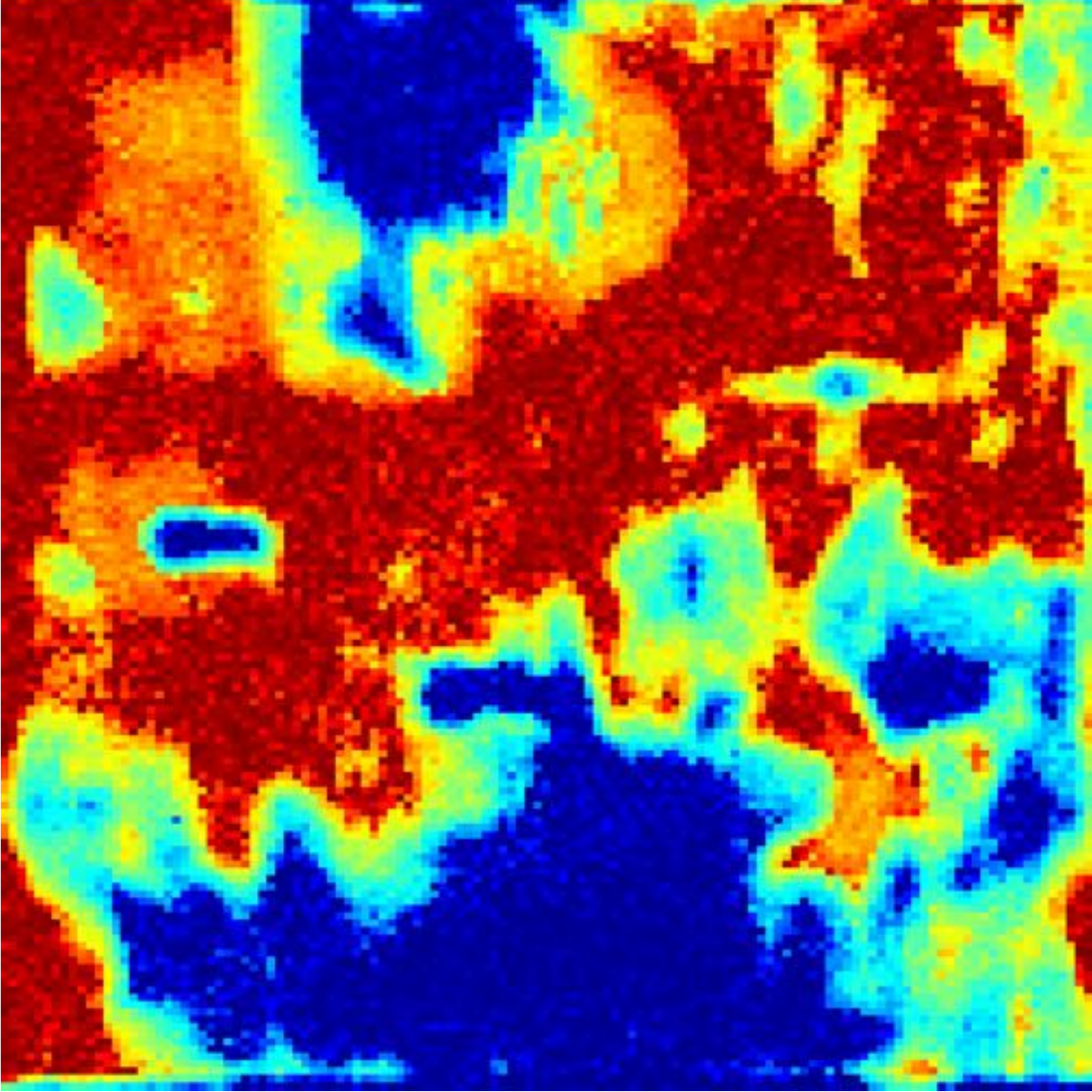}
			\captionsetup{labelformat=empty}
			\caption{(b)}
		\end{subfigure}	
		\begin{subfigure}[t]{0.25\textwidth}
			\centering
			\includegraphics[height=0.6\linewidth,,width=0.9\linewidth]{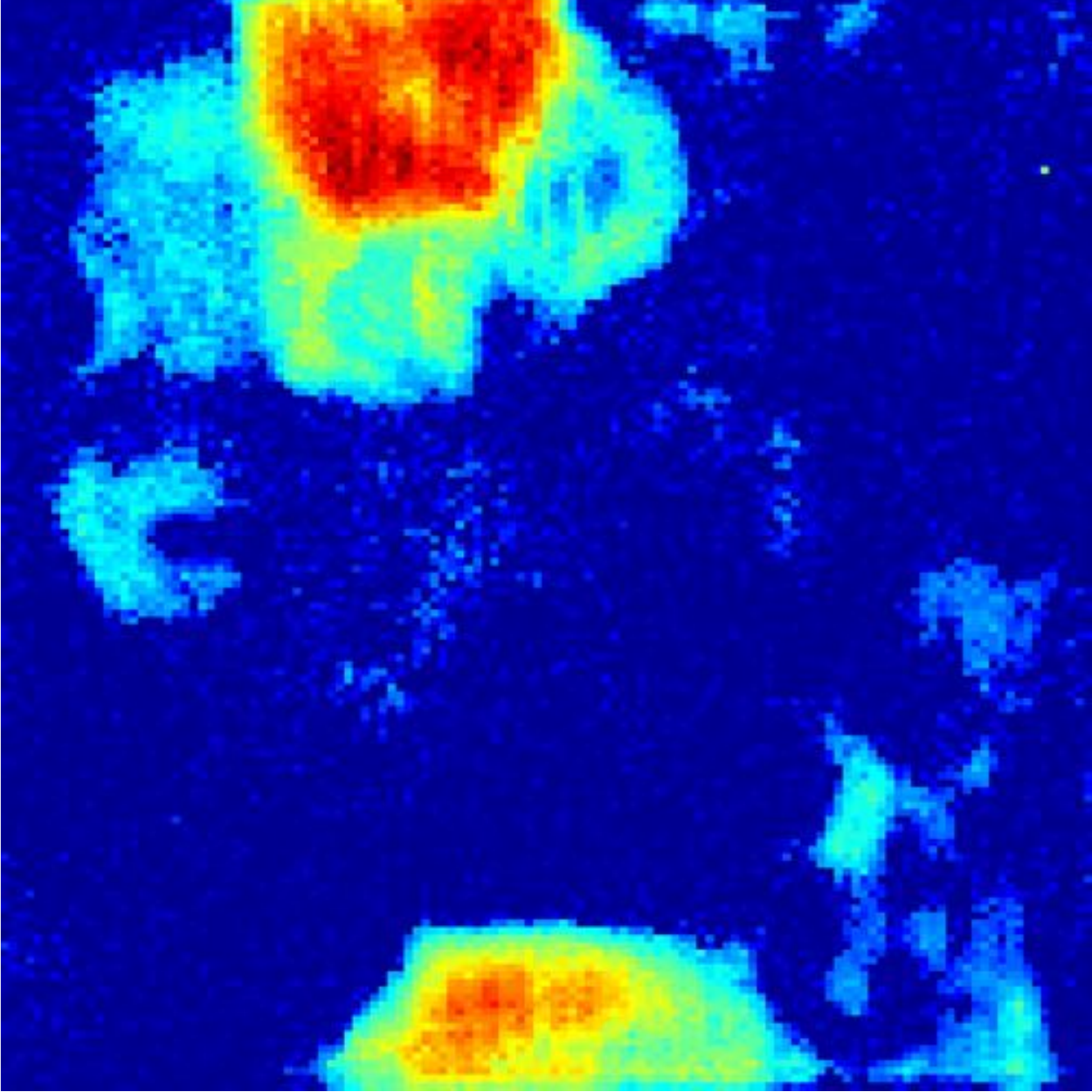}
		    \captionsetup{labelformat=empty}
			\caption{(c)}
		\end{subfigure}	
		
		\begin{subfigure}[t]{0.25\textwidth}
			\centering
			\includegraphics[height=0.6\linewidth,,width=0.9\linewidth]{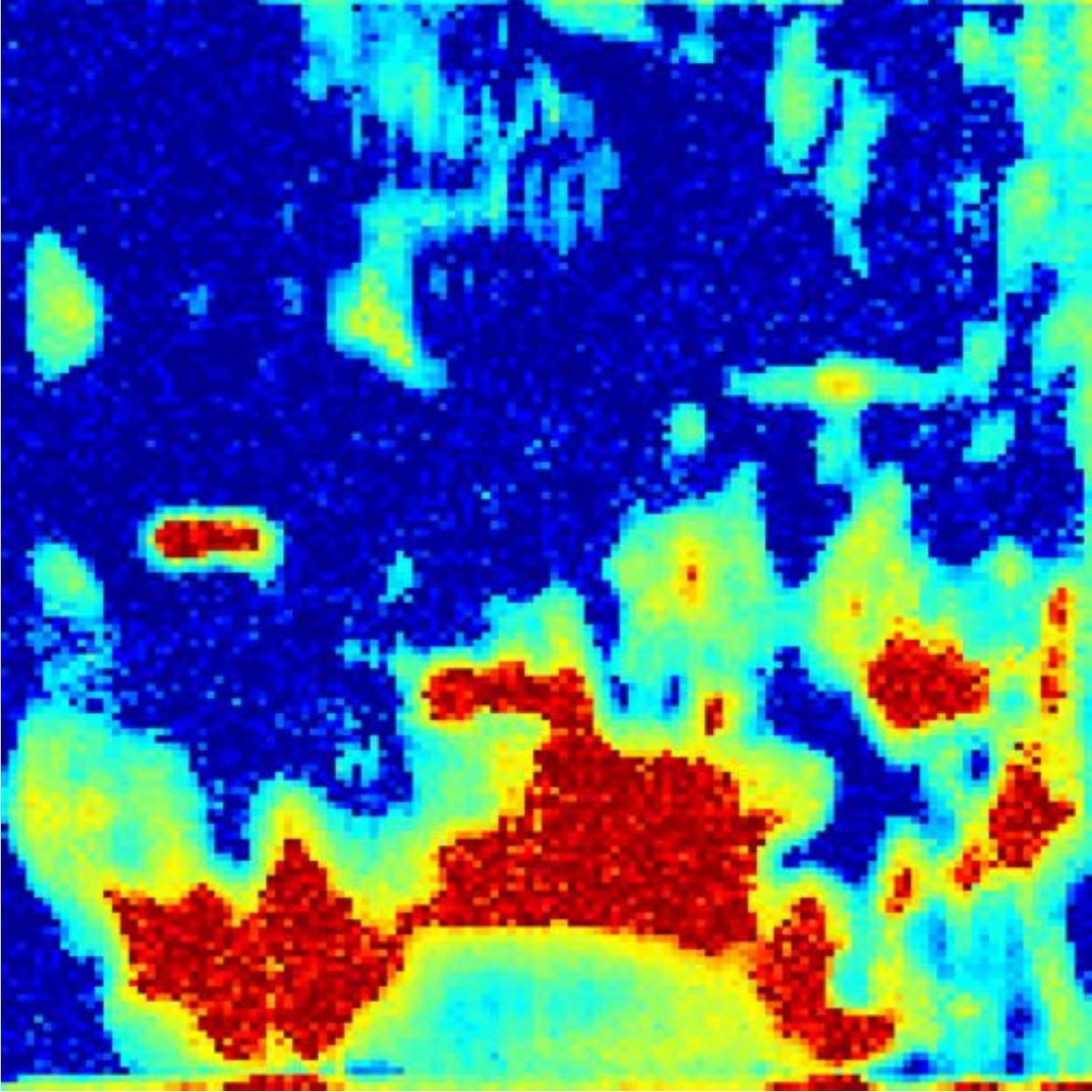}
			\captionsetup{labelformat=empty}
			\caption{(a)$m$  = 3}
		\end{subfigure}	
		\begin{subfigure}[t]{0.25\textwidth}
			\centering
			\includegraphics[height=0.6\linewidth,,width=0.9\linewidth]{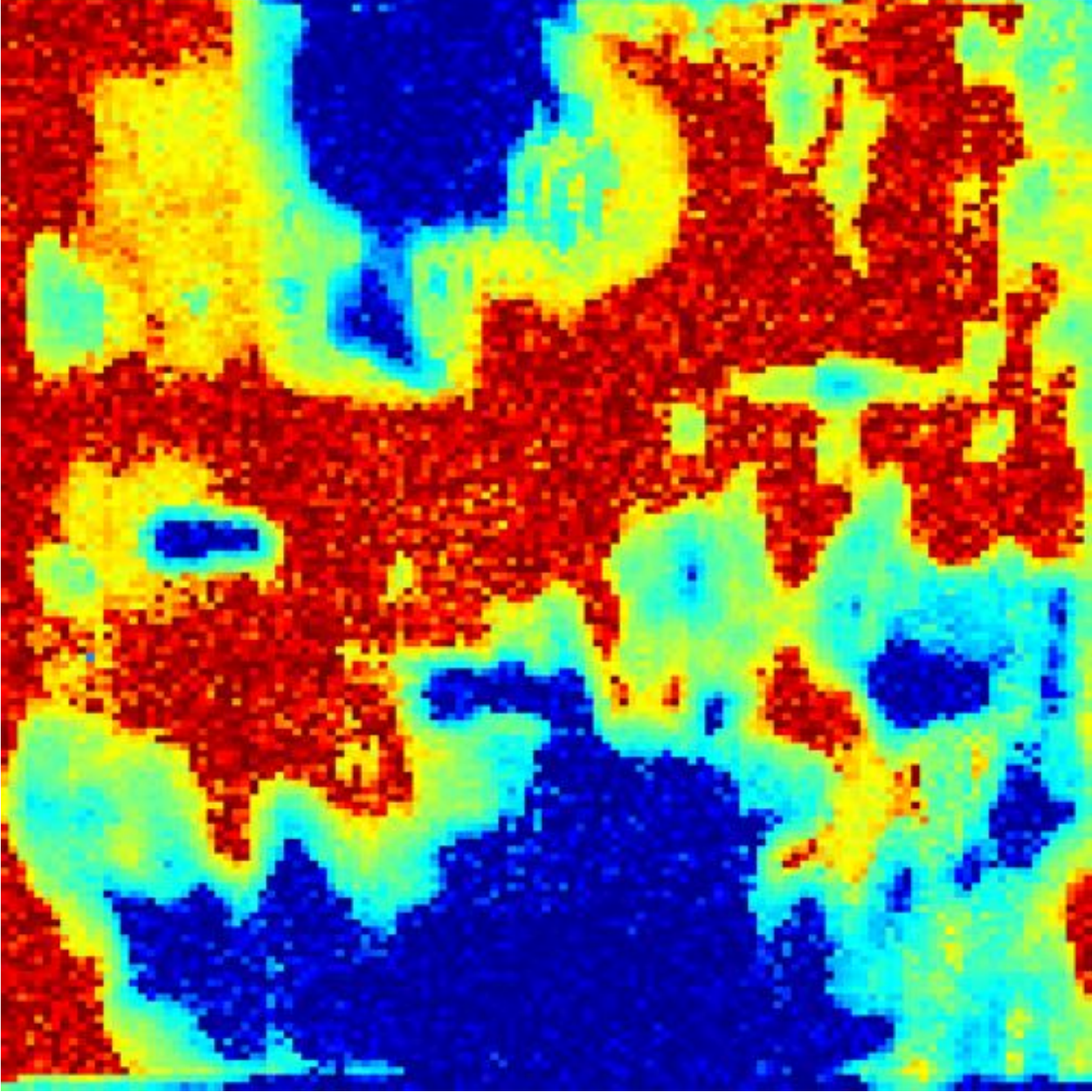}
			\captionsetup{labelformat=empty}
			\caption{(b)}
		\end{subfigure}		
		\begin{subfigure}[t]{0.25\textwidth}
			\centering
			\includegraphics[height=0.6\linewidth,,width=0.9\linewidth]{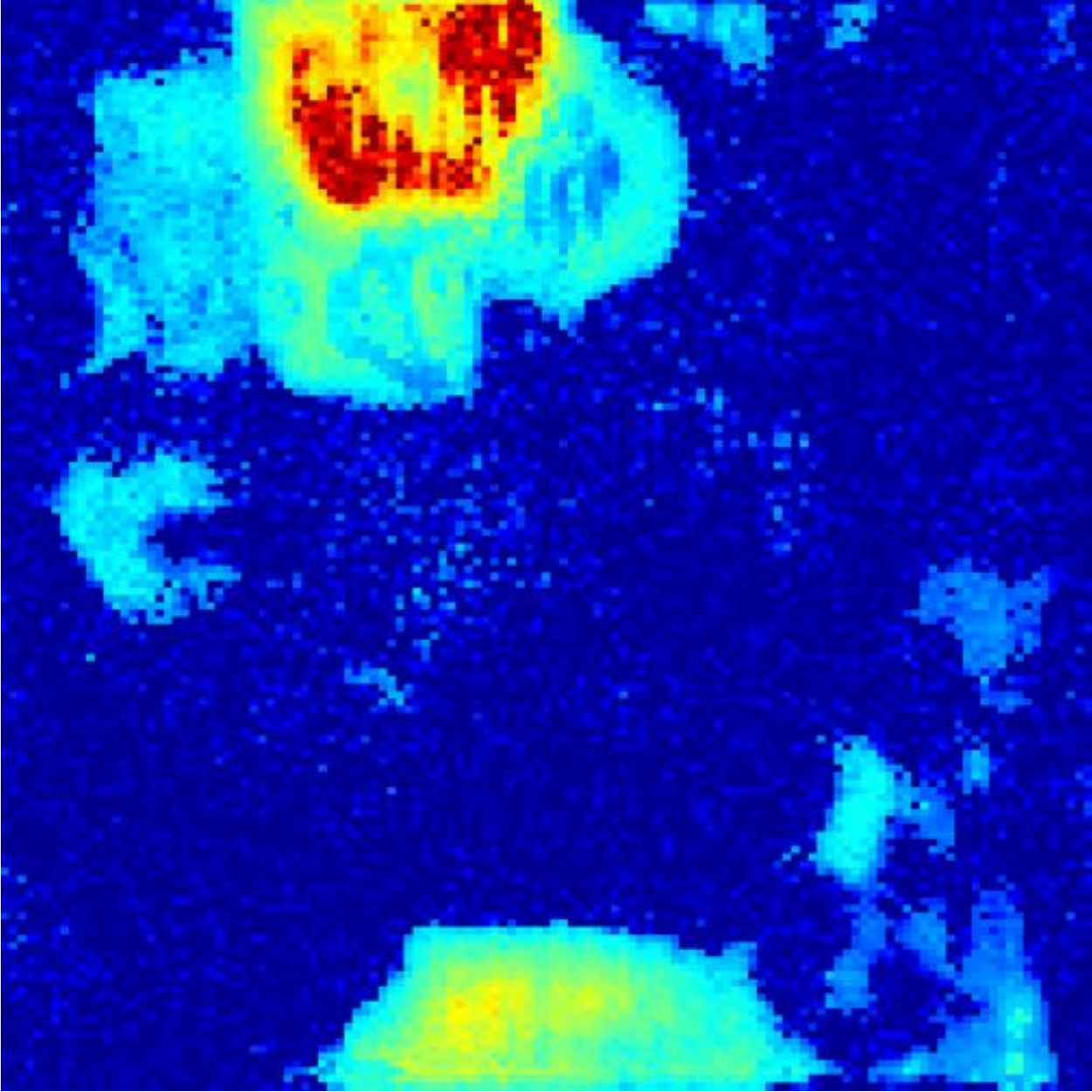}
			\captionsetup{labelformat=empty}
			\caption{(c)}
		\end{subfigure}		

			\caption{The partial membership maps of HF-00 with varying \emph{fuzzifier}. Row 1-3 represent the partial membership maps of HF-00 with \emph{fuzzifier} = 1, 2, and 3, respectively. The superpixel number is 48.}
			\label{fig:varyingF_48}
	
\end{figure*}

In this experiment, we investigate the effect of the {fuzzifier} $m$ using SAS image HF-00 with 48 superpixels. $m$ is varied to be $1, 2,$ and $3$, respectively. Experimental results are shown in Figure \ref{fig:varyingF_48}. In each column, as $m$ increases, some more highly mixed partial membership values are present around boundary areas. 
/

\subsection{Sunset Dataset} 

\begin{figure*}[!htb]
	\centering
	\begin{subfigure}[b]{0.18\textwidth}
		\centering
		\includegraphics[width=1\linewidth]{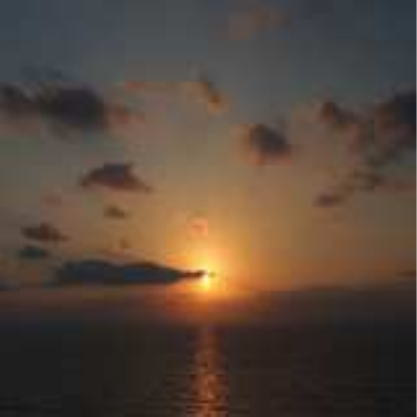}
		\caption{}
	\end{subfigure}
	\begin{subfigure}[b]{0.18\textwidth}
		\centering
		\includegraphics[width=1\linewidth]{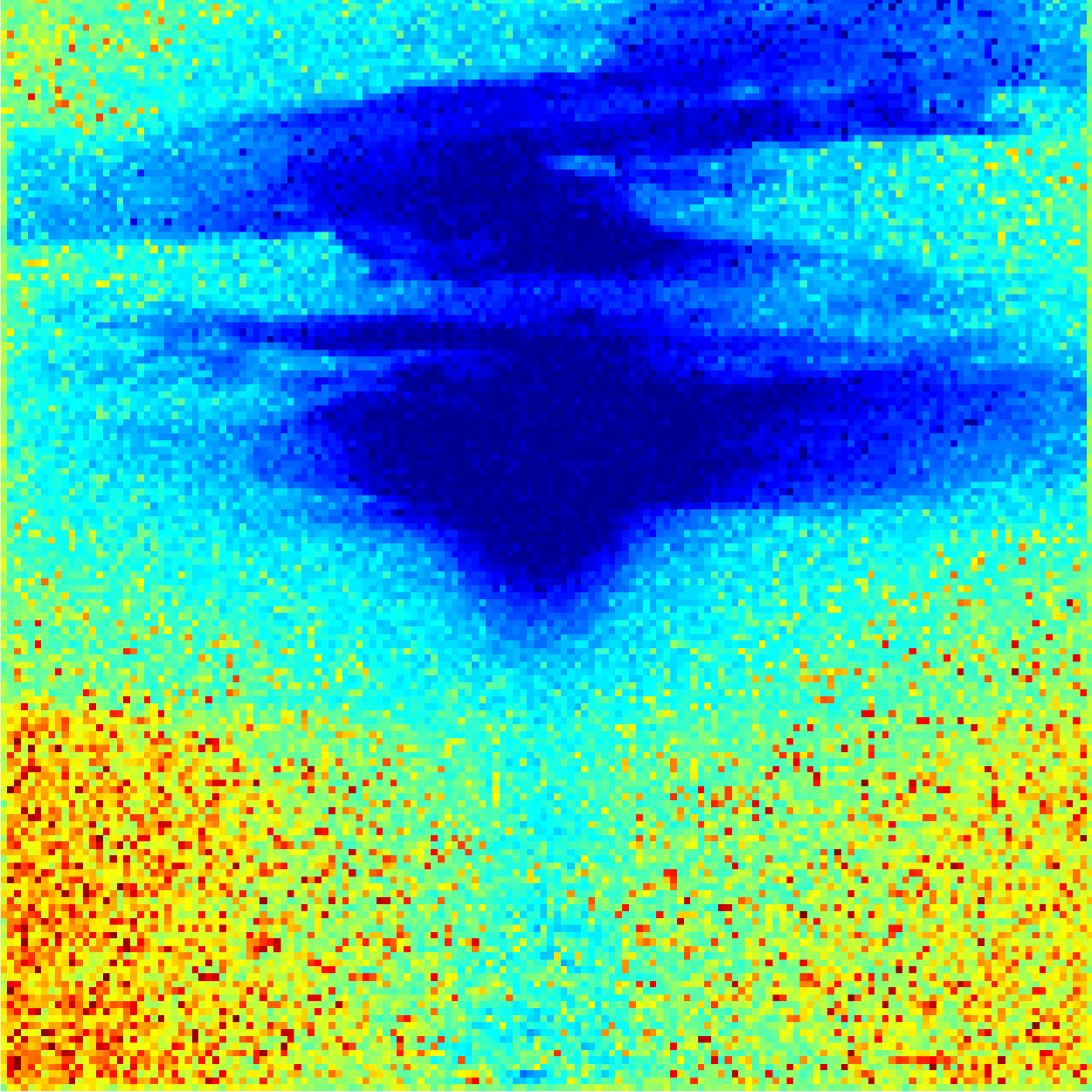}
		\caption{}
	\end{subfigure}
	\begin{subfigure}[b]{0.18\textwidth}
		\centering
		\includegraphics[width=1\linewidth]{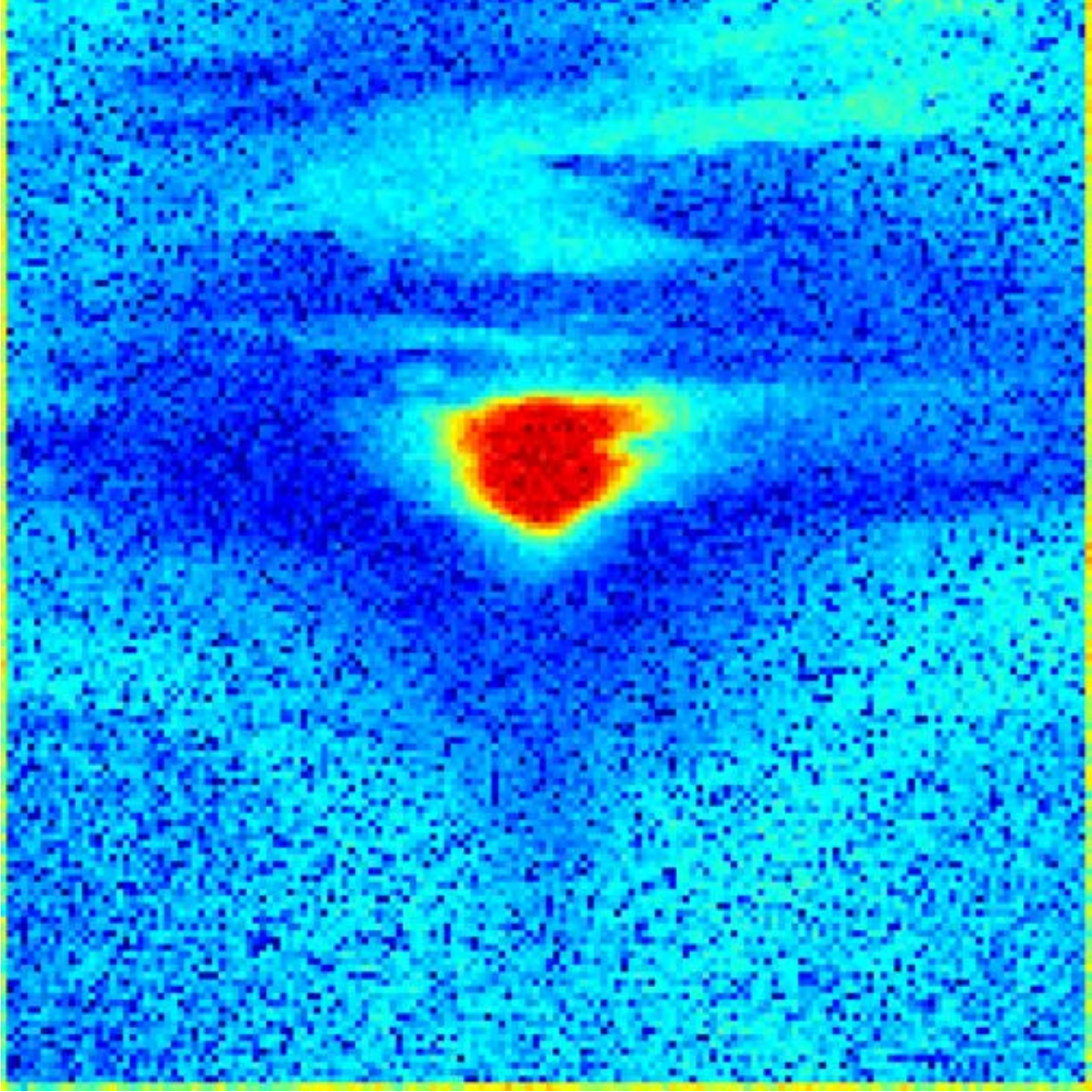}
		\caption{}
	\end{subfigure}
	\begin{subfigure}[b]{0.18\textwidth}
		\centering
		\includegraphics[width=1\linewidth]{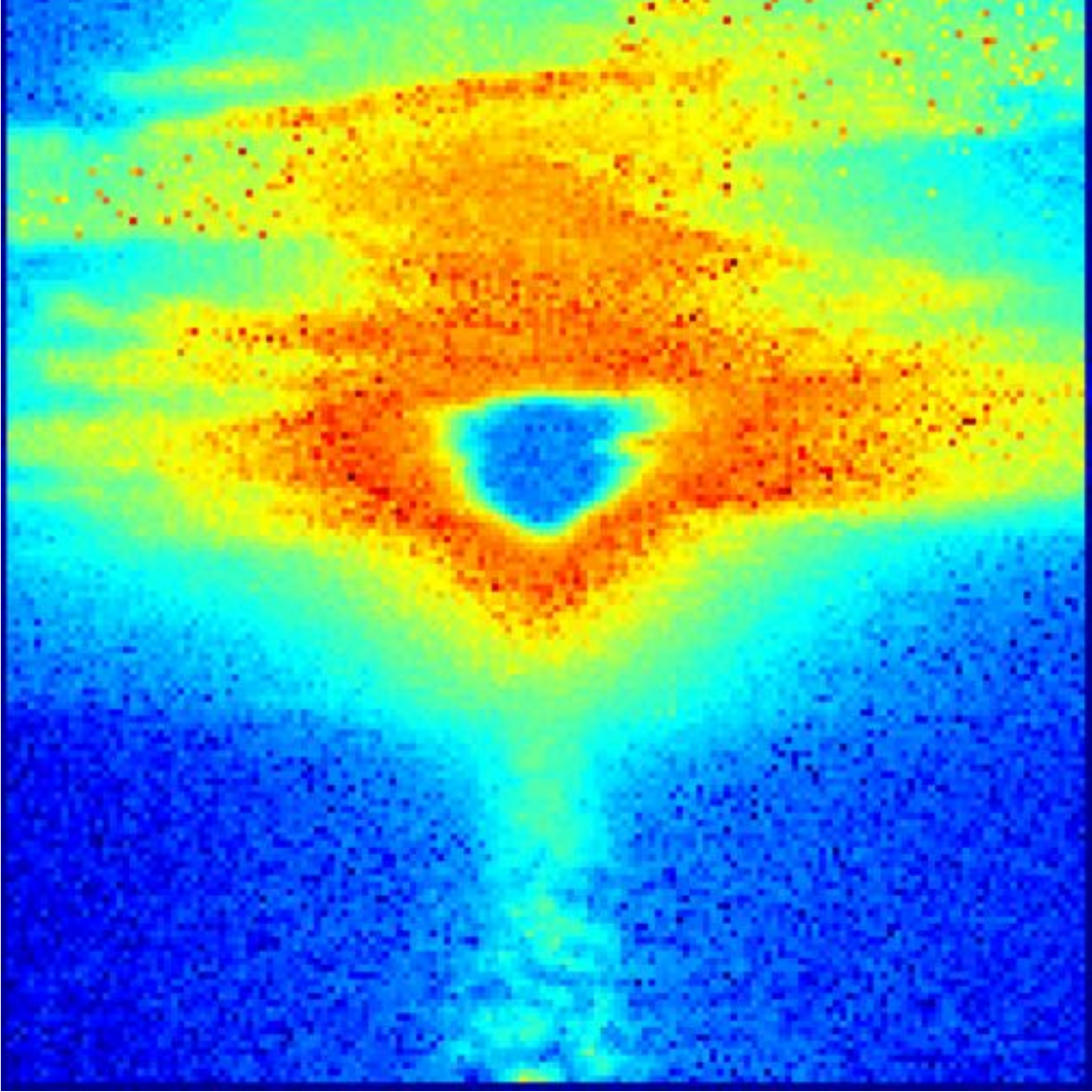}
		\caption{}
	\end{subfigure}
	\begin{subfigure}[b]{0.18\textwidth}
		\centering
		\includegraphics[width=1\linewidth]{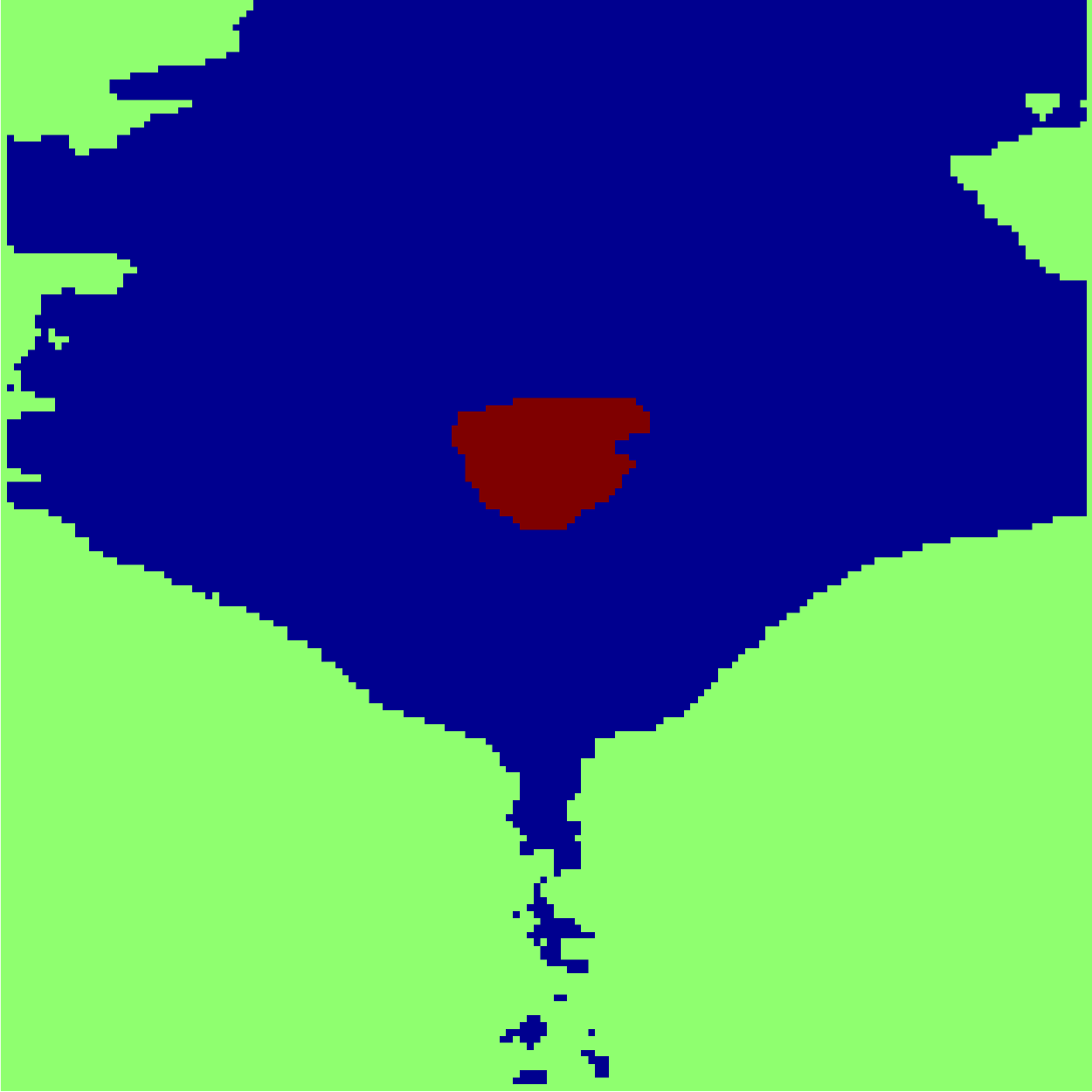}
		\caption{}
	\end{subfigure}
	
	\begin{subfigure}[b]{0.18\textwidth}
		\centering
		\includegraphics[width=1\linewidth]{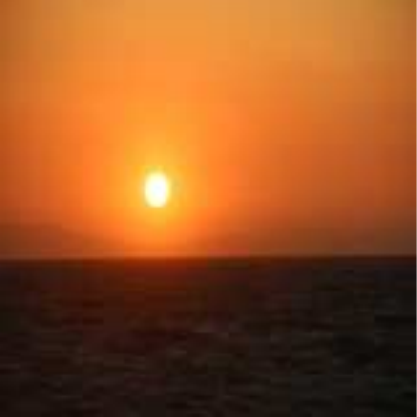}
		\caption{}
	\end{subfigure}
	\begin{subfigure}[b]{0.18\textwidth}
		\centering
		\includegraphics[width=1\linewidth]{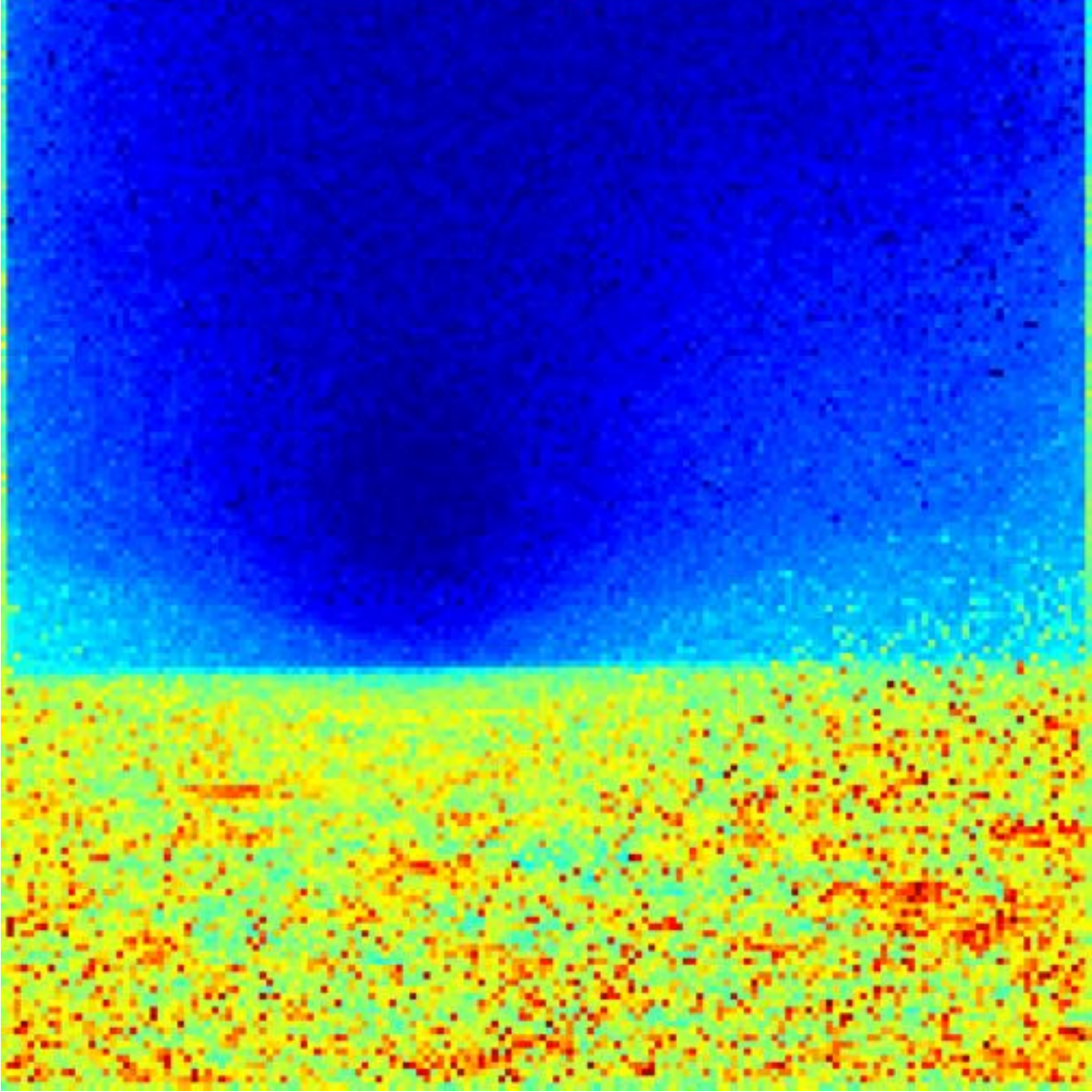}
		\caption{}
	\end{subfigure}
	\begin{subfigure}[b]{0.18\textwidth}
		\centering
		\includegraphics[width=1\linewidth]{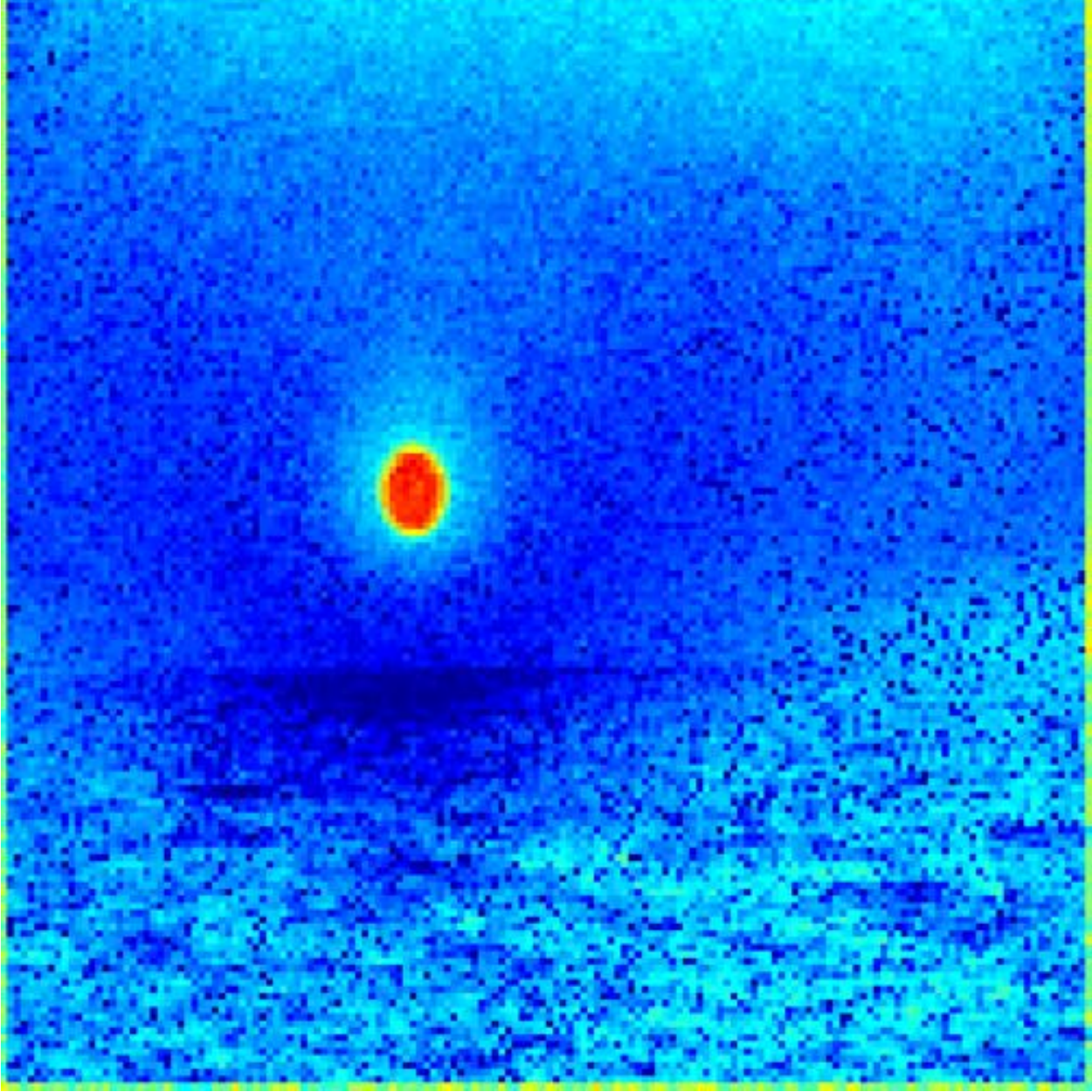}
		\caption{}
	\end{subfigure}
	\begin{subfigure}[b]{0.18\textwidth}
		\centering
		\includegraphics[width=1\linewidth]{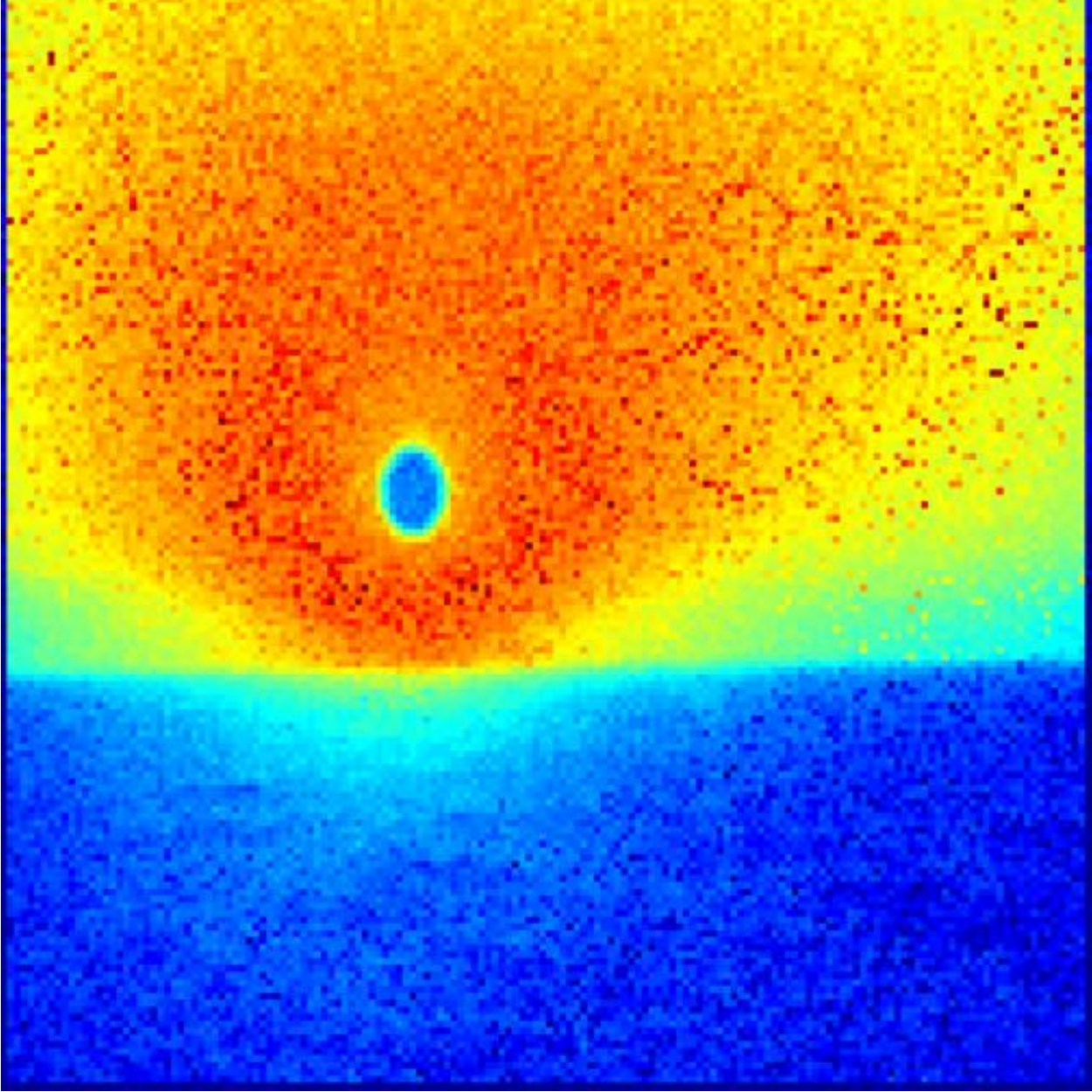}
		\caption{}
	\end{subfigure}
	\begin{subfigure}[b]{0.18\textwidth}
		\centering
		\includegraphics[width=1\linewidth]{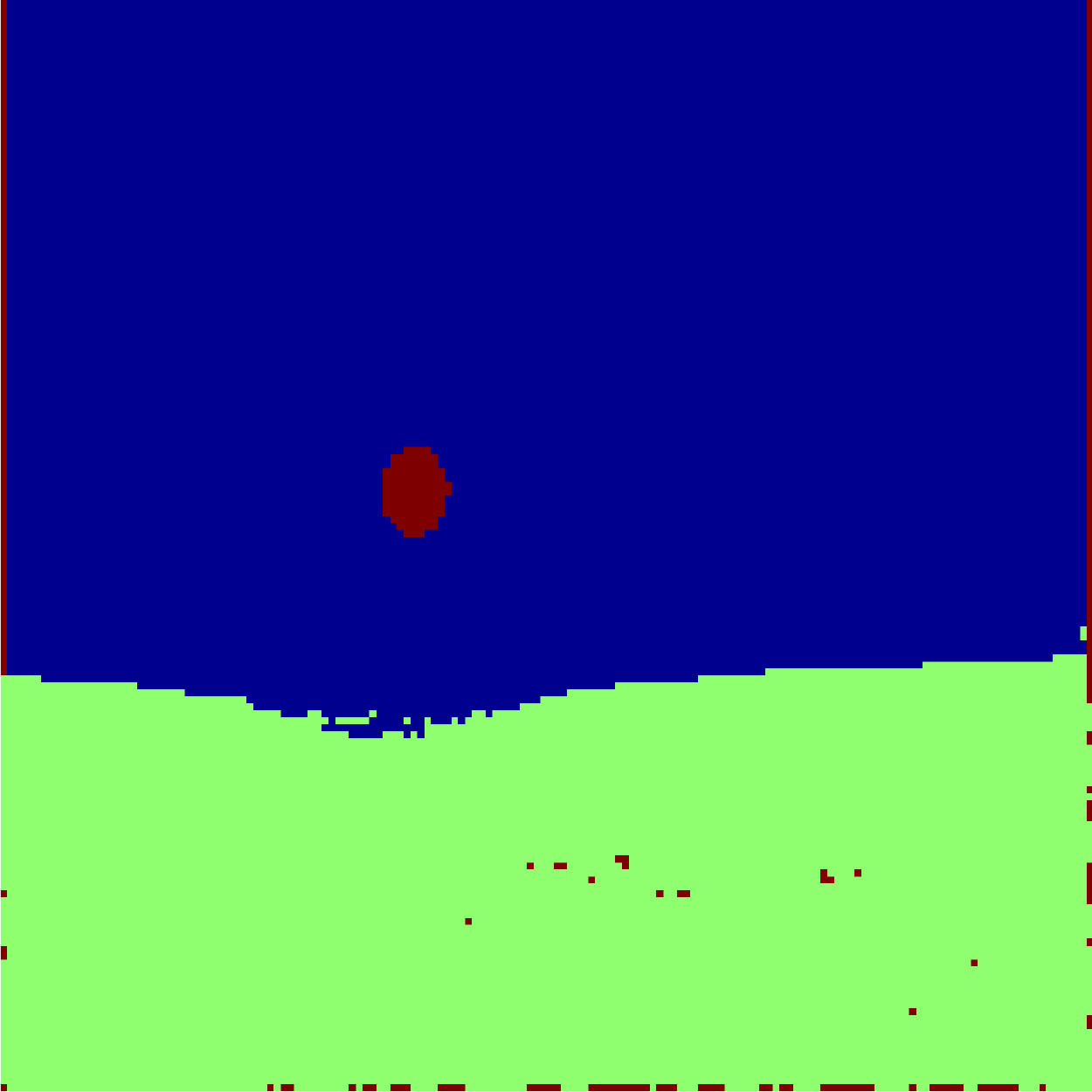}
		\caption{}
	\end{subfigure}
	
	\begin{subfigure}[b]{0.18\textwidth}
		\centering
		\includegraphics[width=1\linewidth]{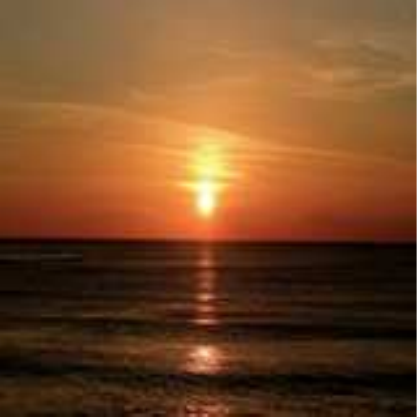}
		\caption{}
	\end{subfigure}
	\begin{subfigure}[b]{0.18\textwidth}
		\centering
		\includegraphics[width=1\linewidth]{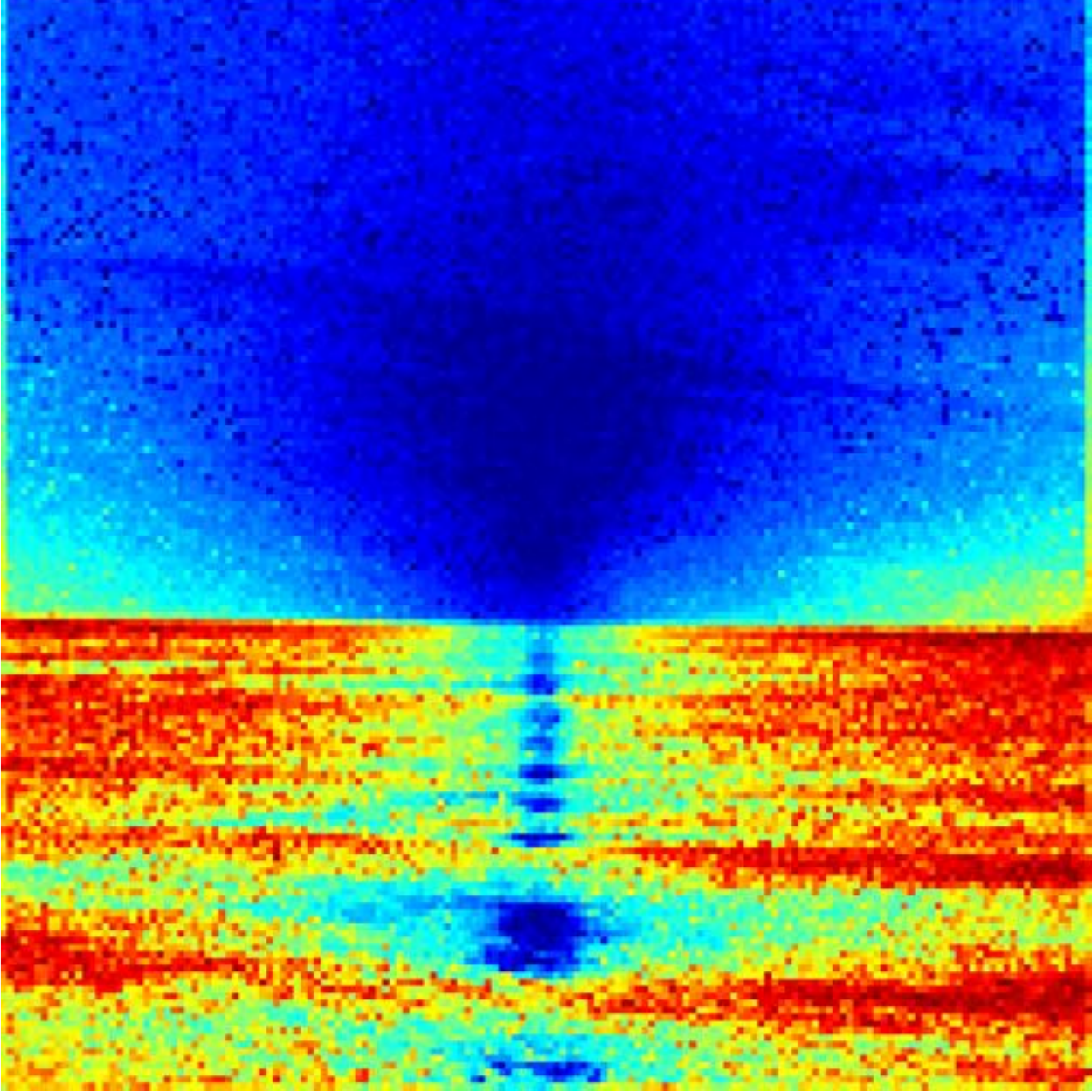}
		\caption{}
	\end{subfigure}
	\begin{subfigure}[b]{0.18\textwidth}
		\centering
		\includegraphics[width=1\linewidth]{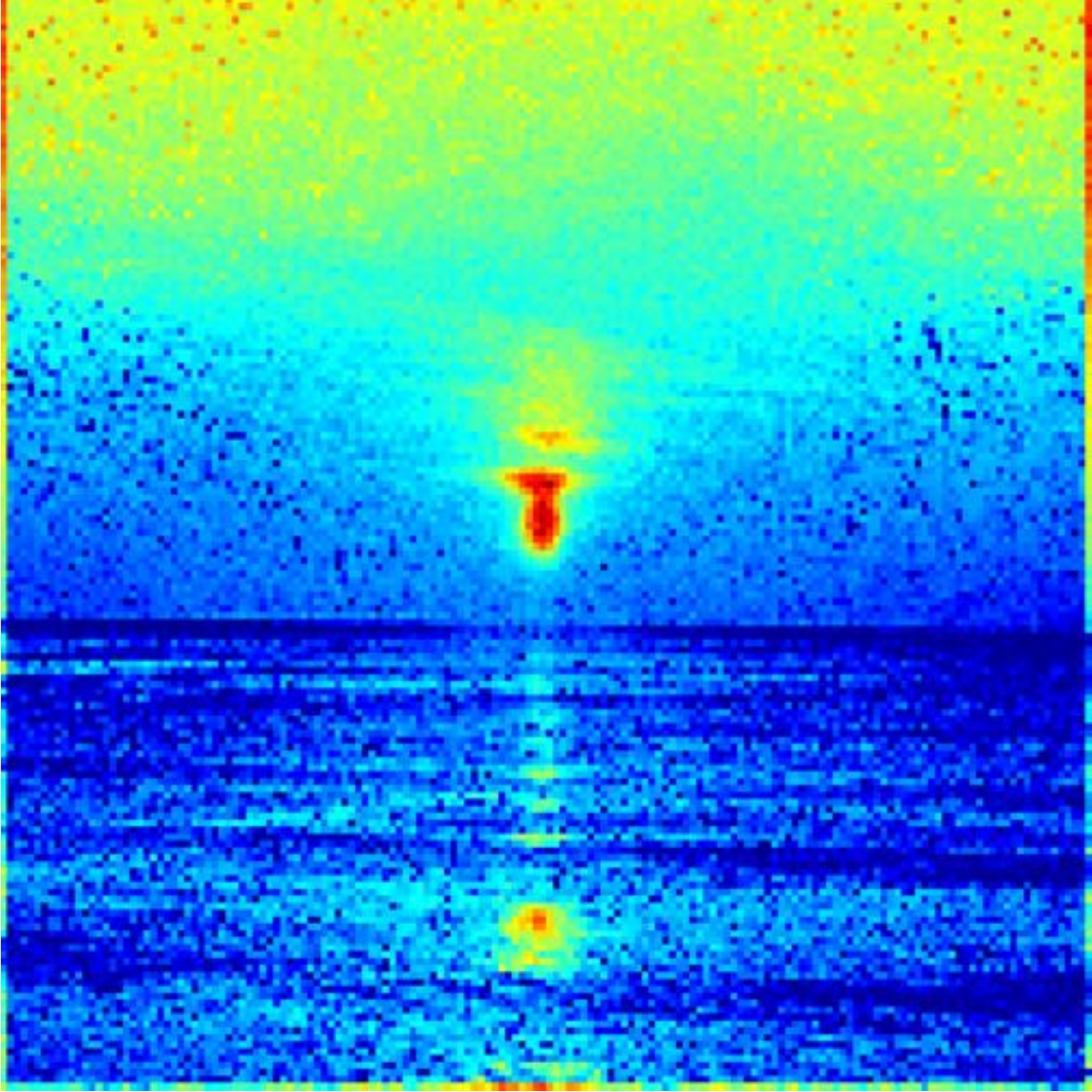}
		\caption{}
	\end{subfigure}
	\begin{subfigure}[b]{0.18\textwidth}
		\centering
		\includegraphics[width=1\linewidth]{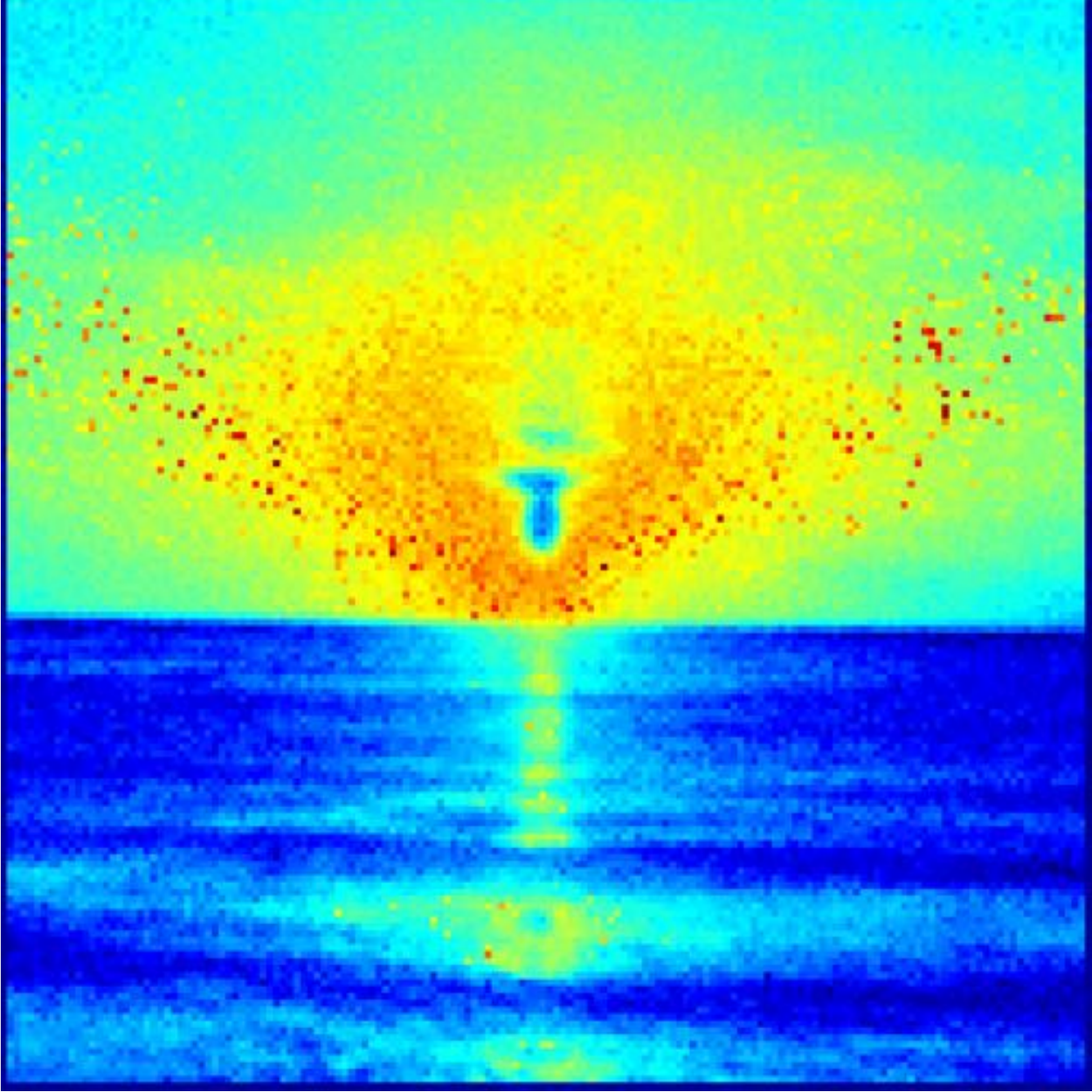}
		\caption{}
	\end{subfigure}
	\begin{subfigure}[b]{0.18\textwidth}
		\centering
		\includegraphics[width=1\linewidth]{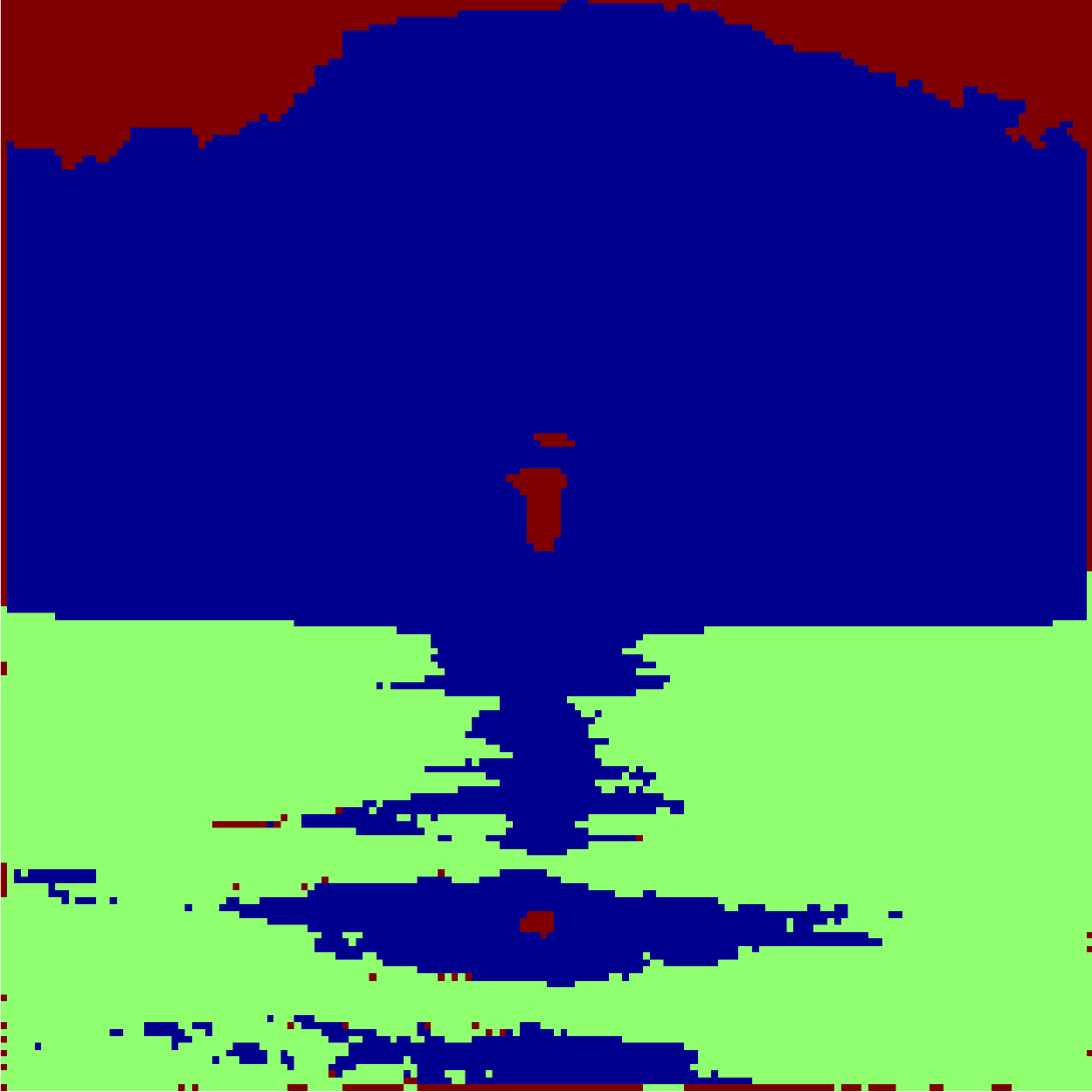}
		\caption{}
	\end{subfigure}
	\caption{Examples of segmentation result on Sunset dataset. (a): Sunset Image 1. (f): Sunset Image 2. (k): Sunset Image 3. (b)-(d), (g)-(i), and (l)-(n) are the PM-LDA partial membership maps in the estimated three topics for Sunset Image 1, Sunset Image 2, and Sunset Image 3, respectively. The color indicates the degree of membership of a visual word in a topic or cluster. (e), (j), and (o) are the LDA results where color indicates the topic.}
	\label{fig:sunset}
\end{figure*}

Finally, experimental results on Sunset dataset show the ability of PM-LDA to perform partial membership segmentation given visual natural imagery. 27 sunset themed images from Flickr (with the necessary permissions)\footnote[2]{\scriptsize{Photo can be found at: https://www.flickr.com/photos/aoa-/6104409480/}}  \footnote[3]{\scriptsize{Photo can be found at: https://www.flickr.com/photos/frenchdave/8482336933/}} were used. Features used in this experiment are $3 \times 3$ Gaussian filter ($\sigma=1$) responses on LAB channels, $3 \times 3$ Gaussian filter response ($\sigma=2$) on blue channel, first order derivative of Gaussian along y-axis on L channel and the log transform of blue channel. Each image is segmented into 15 superpixels using normalized cuts. The number of topics is set to be $3$. Experiments are run on 27 images simultaneously and some of the experimental results are shown in Fig. \ref{fig:sunset}. Columns 2-4 show the segmentation results of PM-LDA.  Column 5 is the LDA results with $3$ topics.  Comparing Column 3 and Column 5 in Fig. \ref{fig:sunset}, we can see that PM-LDA can generate continuous partial membership according to the extent to which the sky is colored by sunlight. The partial membership map illustrates how the topic gradually shift from one to the other. In contrast, LDA can only produce $0$-$1$ segmentation.

\section{Conclusion}
In this paper the PM-LDA model is introduced for soft image segmentation.  PM-LDA improves upon the LDA model by introducing a partial membership rather than requiring a single topic label for each word. Experimental results on three image datasets demonstrate the capacity of PM-LDA model in both soft and crisp image segmentation. Future work will include developing a more efficient sampling approach,  \eg,  a collapsed Gibbs sampler, to accelerate the parameter estimation procedure.  

\section{Acknowledgements}
The authors graciously thank the Office of Naval Research, Code 321, for funding this research. Any opinions, findings, and conclusions or recommendations expressed in this material are those of the author(s) and do not necessarily reflect the views of the Office of Naval Research.

{
	\bibliographystyle{IEEEtran}
	\bibliography{egbib}

\begin{thebibliography}{10}
\providecommand{\url}[1]{#1}
\csname url@samestyle\endcsname
\providecommand{\newblock}{\relax}
\providecommand{\bibinfo}[2]{#2}
\providecommand{\BIBentrySTDinterwordspacing}{\spaceskip=0pt\relax}
\providecommand{\BIBentryALTinterwordstretchfactor}{4}
\providecommand{\BIBentryALTinterwordspacing}{\spaceskip=\fontdimen2\font plus
\BIBentryALTinterwordstretchfactor\fontdimen3\font minus
  \fontdimen4\font\relax}
\providecommand{\BIBforeignlanguage}[2]{{%
\expandafter\ifx\csname l@#1\endcsname\relax
\typeout{** WARNING: IEEEtran.bst: No hyphenation pattern has been}%
\typeout{** loaded for the language `#1'. Using the pattern for}%
\typeout{** the default language instead.}%
\else
\language=\csname l@#1\endcsname
\fi
#2}}
\providecommand{\BIBdecl}{\relax}
\BIBdecl

\bibitem{blei:2003}
D.~M. Blei, A.~Y. Ng, and M.~I. Jordan, ``Latent dirichlet allocation,''
  \emph{Journal of Machine Learning Research}, vol.~3, pp. 993--1022, 2003.

\bibitem{russell:2006}
B.~C. Russell, W.~T. Freeman, A.~Efros, J.~Sivic, and A.~Zisserman, ``Using
  multiple segmentations to discover objects and their extent in image
  collections,'' in \emph{IEEE Conference on Computer Vision and Pattern
  Recognition}, 2006, pp. 1605--1614.

\bibitem{cao:2007}
L.~Cao and L.~Fei-Fei, ``Spatially coherent latent topic model for concurrent
  segmentation and classification of objects and scenes,'' in \emph{IEEE
  International Conference on Computer Vision}, 2007, pp. 1--8.

\bibitem{wang:2008}
X.~Wang and E.~Grimson, ``Spatial latent dirichlet allocation,'' in
  \emph{Advances in Neural Information Processing Systems}, 2008, pp.
  1577--1584.

\bibitem{zhao:2010}
B.~Zhao, L.~Fei-Fei, and E.~P. Xing, ``Image segmentation with topic random
  field,'' in \emph{European Conference on Computer Vision}, 2010, pp.
  785--798.

\bibitem{andreetto:2012}
M.~Andreetto, L.~Zelnik-Manor, and P.~Perona, ``Unsupervised learning of
  categorical segments in image collections,'' \emph{IEEE Transactions on
  Pattern Analysis and Machine Intelligence}, vol.~34, no.~9, pp. 1842--1855,
  2012.

\bibitem{shi:2000}
J.~Shi and J.~Malik, ``Normalized cuts and image segmentation,'' \emph{IEEE
  Transactions on Pattern Analysis and Machine Intelligence}, vol.~22, no.~8,
  pp. 888--905, 2000.

\bibitem{comaniciu:2002}
D.~Comaniciu and P.~Meer, ``Mean shift: A robust approach toward feature space
  analysis,'' \emph{IEEE Transactions on Pattern Analysis and Machine
  Intelligence}, vol.~24, no.~5, pp. 603--619, 2002.

\bibitem{felzenszwalb:2004}
P.~F. Felzenszwalb and D.~P. Huttenlocher, ``Efficient graph-based image
  segmentation,'' \emph{International Journal of Computer Vision}, vol.~59,
  no.~2, pp. 167--181, 2004.

\bibitem{bezdek:1984}
J.~C. Bezdek, R.~Ehrlich, and W.~Full, ``Fcm: The fuzzy c-means clustering
  algorithm,'' \emph{Computers \& Geosciences}, vol.~10, no.~2, pp. 191--203,
  1984.

\bibitem{naz:2010}
S.~Naz, H.~Majeed, and H.~Irshad, ``Image segmentation using fuzzy clustering:
  A survey,'' in \emph{International Conference on Emerging Technologies
  (ICET)}, Oct 2010, pp. 181--186.

\bibitem{Krinidis:2010}
S.~Krinidis and V.~Chatzis, ``A robust fuzzy local information c-means
  clustering algorithm,'' \emph{IEEE Transactions on Image Processing},
  vol.~19, no.~5, pp. 1328--1337, May 2010.

\bibitem{Krinidis:2012}
S.~Krinidis and M.~Krinidis, ``Generalised fuzzy local information c-means
  clustering algorithm,'' \emph{Electronics Letters}, vol.~48, no.~23, pp.
  1468--1470, November 2012.

\bibitem{heller:2008}
K.~A. Heller, S.~Williamson, and Z.~Ghahramani, ``Statistical models for
  partial membership,'' in \emph{International Conference on Machine Learning},
  2008, pp. 392--399.

\bibitem{GlennZare:2014}
T.~Glenn, A.~Zare, and P.~Gader, ``Bayesian fuzzy clustering,'' \emph{IEEE
  Transaction on Fuzzy Systems}, no.~8, pp. 1545--1561, 2015.

\bibitem{Glenn:2013}
T.~C. Glenn, ``Context-dependent detection in hyperspectral imagery,'' 2013.

\bibitem{robert:2013}
C.~Robert and G.~Casella, \emph{Monte Carlo statistical methods}.\hskip 1em
  plus 0.5em minus 0.4em\relax Springer Science \& Business Media, 2013.

\bibitem{cobb2014boundary}
J.~T. Cobb and A.~Zare, ``Boundary detection and superpixel formation in
  synthetic aperture sonar imagery,'' in \emph{International Conference on SAS
  and SAR}, Sept. 2014.

\bibitem{chen:2015}
C.~Chen, A.~Zare, and J.~T. Cobb, ``Sand ripple characterization using an
  extended synthetic aperture sonar model and parallel sampling method,''
  \emph{IEEE Transaction on Geoscience and Remote Sensing.}, vol.~53, no.~10,
  pp. 5547--5559, 2015.

\bibitem{cobb:2013multi}
J.~T. Cobb and A.~Zare, ``Multi-image texton selection for sonar image seabed
  co-segmentation,'' in \emph{SPIE}, vol. 8709, no. 87090H, June 2013.

\end{thebibliography}
	}

\begin{IEEEbiography}{Chao Chen}
	received the B.S. and M.S. degree in control theory both from Xidian University, Xi'an, China, in 2007 and 2010, respectively, and the Ph.D. degree from University of Missouri - Columbia, in 2016. Her research interests include sparse coding, Bayesian inference, and synthetic aperture sonar imagery analysis.
\end{IEEEbiography}
\vspace*{-1cm}

\begin{IEEEbiography}{Alina Zare (S'07--M'08--SM'13)}
	received the Ph.D. degree from the University of Florida, Gainesville, in 2008. She is currently an Associate Professor with the Department of Electrical and Computer Engineering, University of Florida.   Her research interests include machine learning, computational intelligence, Bayesian methods, sparsity promotion, image analysis, pattern recognition, hyperspectral image analysis, and remote sensing. Alina Zare is a recipient of the 2014 National Science Foundation CAREER award and the 2014 National Geospatial-Intelligence Agency New Investigator Program Award. Alina Zare is an Associate Editor of the IEEE Transactions on Geoscience and Remote Sensing.  
\end{IEEEbiography}

\begin{IEEEbiography}{Huy N. Trinh}
	received the B.S. degree in Computer Science at University of Missouri, Columbia, USA, in 2015. He is currently a Graduate Research Assistant working toward the M.S degree in the Department of Computer Science, University of Missouri, Columbia, USA. His research interests include machine learning, image segmentation, computational intelligent, cloud computing and networks.
\end{IEEEbiography}

\begin{IEEEbiography}{Gbenga Omotara}
	is currently pursuing an undergraduate degree in Electrical and Computer Engineering at the University of Missouri, Columbia, MO, USA. His interests include machine learning and computational intelligence.
\end{IEEEbiography}

\begin{IEEEbiography}{J. Tory Cobb (S'99--M'01--SM'13)}
	received the B.S. degree in electrical engineering from the United States Coast Guard Academy, New London, CT, USA in 1994, the M.S. degree in electrical engineering from Auburn University, Auburn, AL, USA, in 2001, and the Ph.D. degree from the University of Florida, Gainesville, FL, USA, in 2011.  From 1994 to 1999 he was an active duty officer in the Coast Guard.  Since 2001 he has been employed as a Research Engineer at the Naval Surface Warfare Center, Panama City, FL, USA. He has served as Principal Investigator or Co-Principal Investigator for various automatic target recognition and sensor fusion projects funded by the Office of Naval Research. His research interests include statistical modeling of sonar signals with applications to automatic target recognition, automated environmental characterization of seabeds in side-look sonar images, and sonar image segmentation algorithm development. Dr. Cobb is an Associate Editor of the IEEE Journal of Oceanic Engineering.
\end{IEEEbiography}

\begin{IEEEbiography}{Timotius A.~Lagaunne}
	is currently pursuing an undergraduate degree in Mathematics at the University of Missouri Columbia, MO, USA. His interests include machine learning, computational intelligence, data science and natural language processing 
\end{IEEEbiography}




\end{document}